\documentclass[twoside,11pt]{article}

\usepackage{jmlr2e}
\usepackage{graphicx,epsf,empheq}
\usepackage{natbib}%,dsfont,cases}
\usepackage{verbatim,epsfig}
\usepackage{multirow}
\usepackage{epsfig,latexsym,enumerate,paralist}
\usepackage{mathtools,color,etoolbox,hyperref,float,mathrsfs}

\usepackage{subcaption,array}

\usepackage{adjustbox,enumitem, paralist,rotating,subcaption}

\newcommand{\bd}{\boldsymbol}
\newcommand{\mb}{\mathbf}

\newcommand{\be}{\begin{equation}}
\newcommand{\ee}{\end{equation}}

\DeclareMathOperator*{\argmin}{arg\,min}
\DeclareMathOperator*{\argmax}{arg\,max}

\DeclareMathOperator{\diag}{diag}

\DeclareMathOperator{\eig}{\lambda}
\DeclareMathOperator{\rank}{rank}
\DeclareMathOperator{\cov}{cov}
\DeclareMathOperator{\corr}{corr}
\DeclareMathOperator{\var}{var}

\DeclareMathOperator{\tr}{tr}
\DeclareMathOperator{\lspan}{span}

\DeclareMathOperator{\PVE}{PVE}
\DeclareMathOperator{\SNR}{SNR}

\DeclareMathOperator{\sign}{sign}

\newtheorem{thm}{Theorem}

\newtheorem{cor}{Corollary}
\newtheorem{assump}{Assumption}

\makeatletter
\DeclareFontFamily{U}{MnSymbolA}{}
\DeclareFontShape{U}{MnSymbolA}{m}{n}{
	<-6>  MnSymbolA5
	<6-7>  MnSymbolA6
	<7-8>  MnSymbolA7
	<8-9>  MnSymbolA8
	<9-10> MnSymbolA9
	<10-12> MnSymbolA10
	<12->   MnSymbolA12}{}
\DeclareFontShape{U}{MnSymbolA}{b}{n}{
	<-6>  MnSymbolA-Bold5
	<6-7>  MnSymbolA-Bold6
	<7-8>  MnSymbolA-Bold7
	<8-9>  MnSymbolA-Bold8
	<9-10> MnSymbolA-Bold9
	<10-12> MnSymbolA-Bold10
	<12->   MnSymbolA-Bold12}{}
\DeclareSymbolFont{MnSyA}{U}{MnSymbolA}{m}{n}
\SetSymbolFont{MnSyA}{bold}{U}{MnSymbolA}{b}{n}

\DeclareRobustCommand{\overleftharpoon}{\mathpalette{\overarrow@\leftharpoonfill@}}
\DeclareRobustCommand{\overrightharpoon}{\mathpalette{\overarrow@\rightharpoonfill@}}
\def\leftharpoonfill@{\arrowfill@\leftharpoondown\mn@relbar\mn@relbar}
\def\rightharpoonfill@{\arrowfill@\mn@relbar\mn@relbar\rightharpoonup}

\DeclareMathSymbol{\leftharpoondown}{\mathrel}{MnSyA}{'112}
\DeclareMathSymbol{\rightharpoonup}{\mathrel}{MnSyA}{'100}
\DeclareMathSymbol{\mn@relbar}{\mathrel}{MnSyA}{'320}
\makeatother

\DeclareFontFamily{U}{mathx}{\hyphenchar\font45}
\DeclareFontShape{U}{mathx}{m}{n}{
	<5> <6> <7> <8> <9> <10>
	<10.95> <12> <14.4> <17.28> <20.74> <24.88>
	mathx10
}{}
\DeclareSymbolFont{mathx}{U}{mathx}{m}{n}
\DeclareMathAccent{\widecheck}{\mathalpha}{mathx}{"71}

\makeatletter
\def\thickhline{%
	\noalign{\ifnum0=`}\fi\hrule \@height \thickarrayrulewidth \futurelet
	\reserved@a\@xthickhline}
\def\@xthickhline{\ifx\reserved@a\thickhline
	\vskip\doublerulesep
	\vskip-\thickarrayrulewidth
	\fi
	\ifnum0=`{\fi}}
\makeatother

\newlength{\thickarrayrulewidth}
\usepackage{lastpage}
\ShortHeadings{Decomposition-based Generalized Canonical Correlation Analysis}{Shu, Qu, and Zhu}

\firstpageno{1}

\begin{document}

\title{D-GCCA: Decomposition-based 
	Generalized Canonical Correlation Analysis 
	for Multi-view High-dimensional Data}

\author{\name Hai Shu\thanks{Corresponding author. The article is published in {\it Journal of Machine Learning Research}, 23(169):1--64, 2022. The publisher’s version is available at \url{https://www.jmlr.org/papers/v23/20-021.html}.}
	\email hs120@nyu.edu \\
       \addr Department of Biostatistics\\
        School of Global Public Health\\
       New York University\\
       New York, NY 10003, USA
       \AND
       \name Zhe Qu \email zqu2@tulane.edu \\
       	\addr Department of Mathematics\\
       	Tulane University\\
       	New Orleans, LA 70118, USA       
       \AND
       \name Hongtu Zhu
        \email htzhu@email.unc.edu \\
       \addr Departments of Biostatistics, Statistics, Computer Science, and Genetics\\
       The University of North Carolina at Chapel Hill\\
       Chapel Hill, NC 27599, USA
       }

\editor{}%Qiaozhu Mei}

\maketitle

\begin{abstract}%   <- trailing '%' for backward compatibility of .sty file
Modern biomedical studies often collect
multi-view data, that is, multiple  
types of data measured on the same set of objects.	
A popular model in high-dimensional multi-view data analysis is to decompose each view's data matrix 
into 
a low-rank common-source matrix generated by latent factors 
common across all data views, 
a low-rank distinctive-source matrix corresponding to each view, 
and an additive noise matrix. 
We propose a novel decomposition method for this model, called decomposition-based generalized canonical correlation analysis (D-GCCA).  The D-GCCA rigorously 
defines the decomposition on the $\mathcal{L}^2$ space of random variables in contrast to
  the Euclidean dot product space used by most existing methods, 
{\color{black} thereby being able to provide the estimation consistency for the low-rank matrix recovery}.
%In contrast,  most existing
%methods are developed on   the Euclidean dot product space.
Moreover,  to well calibrate common latent factors,  we impose a desirable orthogonality constraint on distinctive latent factors. 
{\color{black} Existing methods, however,  inadequately consider
such orthogonality and may thus suffer from substantial loss of undetected common-source variation.} 
{\color{black}Our D-GCCA takes one step further than generalized canonical correlation analysis by separating common and distinctive components among canonical variables, while enjoying an appealing interpretation from the perspective of principal component analysis.
{\color{black} Furthermore, we propose to use the variable-level proportion of signal variance explained by common or distinctive latent factors for selecting the variables most influenced.}
Consistent estimators of our D-GCCA method are established with good finite-sample numerical performance, 
and have closed-form expressions leading to efficient computation especially for large-scale data. 
The superiority of D-GCCA over state-of-the-art methods is also corroborated in simulations and real-world data examples.}

\end{abstract}

\begin{keywords}
Canonical variable, common and distinctive variation structures, data integration, high-dimensional data, multi-view data.
\end{keywords}

\section{Introduction}\label{sec: intro}
Data integration 
is widely used in biomedical studies to combine multi-view data, which are
multiple types (i.e., views) of data obtained from the same set of objects,
into meaningful and valuable information. 
Such studies  include The Cancer Genome Atlas \citep[TCGA;][]{Hoad18}
with multi-platform genomic data for tumor samples,
and Human Connectome Project \citep[HCP;][]{Van13} with multi-modal brain images of healthy adults, among many others \citep{Craw16,Jens17}. 
The use of  multi-view~data can allow us to  enhance   understanding the etiology of 
many complex diseases, such as cancers \citep{Ciri15,Camp18} and neurodegenerative diseases \citep{Wein13,Saee17}.  
Researchers hence have become highly interested in  studying the shared  and individual information  across multi-view data
through separating their common and distinctive variation structures \citep{Kloe16, Smil17}.

Let $\mb{Y}_k\in \mathbb{R}^{p_k\times n}$ ($k=1, \ldots, K$) be the 
row-mean centered data matrix of the $k$th view  of $K$-view data obtained on a common set of $n$ objects, where $p_k$ is the number of variables.
One popular approach for disentangling their common and distinctive variation structures
is to decompose each data matrix into 
\be\label{decomp in mat}
\mb{Y}_k=\mb{X}_k+\mb{E}_k=\mb{C}_k+\mb{D}_k+\mb{E}_k\quad~~~\mbox{for}~~~k=1, \ldots, K, 
\ee 
where $\mb{X}_k$ is a low-rank signal matrix with an additive noise matrix $\mb{E}_k$,  
$\mb{C}_k$ is a low-rank {\it common-{\color{black}source} matrix} 
that represents the part of $\mb{X}_k$ coming from the {\color{black} underlying source of variation (a.k.a. latent factors) common across all  views,
and $\mb{D}_k$ is a low-rank {\it distinctive-{\color{black}source} matrix}
from distinctive latent factors of the corresponding view.
 In other words,
the common-source and distinctive-source matrices 
contain the variation information in each view, respectively, explained by the common and distinctive latent factors of the 
$K$ views.
}

%Both common and distinctive mechanisms, also known as latent factors, denote the underlying causes of variation in the data \citep{Scho13}.

{\color{black}There is a growing literature on developing decomposition
methods for 
model~\eqref{decomp in mat}. Throughout this paper, we will consider 
six state-of-the-art methods, including} orthogonal n-block partial least squares
\citep[OnPLS;][]{Lofs11}, distinctive and common components with simultaneous component analysis \citep[DISCO-SCA;][]{Scho13},
common orthogonal basis extraction \citep[COBE;][]{Zhou16},
joint and individual variation explained \citep[JIVE;][]{Lock13} and its variant R.JIVE \citep{OCon16}, and the angle-based JIVE \citep[AJIVE;][]{Feng18}.
{\color{black}
The decomposition differs per method.
\mbox{OnPLS} is developed from
a multi-block partial least squares (PLS) method
that is equivalent to the generalized canonical correlation analysis (GCCA)
using the sum of covariances criterion \citep{tenenhaus2011regularized}. 
Both
DISCO-SCA and JIVE are  based on the simultaneous
component analysis \citep[SCA;][]{smilde2003framework} that applies the principal component analysis (PCA)
to the concatenation of all observed data matrices, but DISCO-SCA
imposes more orthogonality constraints. 
R.JIVE is a JIVE variant with an additional orthogonality constraint. 
Both 
AJIVE and COBE can be regarded as 
extensions of
the maximum-variance based GCCA \citep{Kett71}, 
but with different denoising strategies. 
Although PLS, SCA, and GCCA 
are widely-used data integration methods,
they solve problems different from \eqref{decomp in mat} and are only used as one step of the above methods.
Problem~\eqref{decomp in mat} belongs to
the scope of multi-block or multi-view data analysis that covers
a wide spectrum of topics,  on which we refer readers to 
\citet{zhao2017multi}, \citet{Li18} and \citet{mishra2021recent} for reviews.}

{\color{black}The   six state-of-the-art methods for model \eqref{decomp in mat}} can be applied to data with $K\ge 2$ views,
but suffer from two major issues.   
(i) They are built on the inappropriate Euclidean dot product space $(\mathbb{R}^n,\cdot)$, which simply approximates the $\mathcal{L}^2$ space of random variables. 
(ii) They 
inadequately consider orthogonality {\color{black} (i.e., uncorrelatedness)} constraints among 
distinctive-source matrices $\{\mb{D}_k\}_{k=1}^K$, 
{\color{black}
so there is no guarantee against the risk that
$\{\mb{D}_k\}_{k=1}^K$
are all pairwise correlated 
and thus retain some undiscovered common latent factors
and their explained variation.}
To address these issues,
a nice decomposition, called decomposition-based canonical correlation analysis (D-CCA), 
is recently proposed in \citet{Shu17}
based on
the canonical correlation analysis \citep[CCA;][]{Hote36}, but unfortunately, it is limited to
two data views, $K=2$. %We will show   that extending from $K=2$ to $K>2$ represents major challenges.

The aim of this paper is to address   issues (i) and (ii) for data with $K\geq 2$ views. 
We assume 
that the columns of each matrix in \eqref{decomp in mat} are $n$ independent copies of 
the corresponding random vector in
\be\label{decomp in vari}
\bd{y}_k=\bd{x}_k+\bd{e}_k=\bd{c}_k+\bd{d}_k+\bd{e}_k\in \mathbb{R}^{p_k},
\ee
with entries of $\bd{c}_k,$ $\bd{d}_k$ and $\bd{e}_k$ belonging to
$\mathcal{L}_0^2$, where $\bd{c}_k$ and $\bd{d}_k$ are {\color{black} called the {\it common-source random vector} and 
the {\it distinctive-source random vector}}, respectively,   
generated by common and distinctive latent factors. 
Here, $\mathcal{L}_0^2$ is the vector space composed of all real-valued random variables with zero mean and finite variance. 
We denote $(\mathcal{L}_0^2,\cov)$ as 
the inner product space of $\mathcal{L}_0^2$ that is endowed with the
covariance operator as the inner product.

{\color{black}
A major drawback of  the six existing methods  
is that 
their decompositions are defined
with respect to the orthogonality of $(\mathbb{R}^n,\cdot)$ 
rather than the more precise orthogonality
of $(\mathcal{L}_0^2,\cov)$.
Obviously, the orthogonality of $(\mathbb{R}^n,\cdot)$ (i.e., zero sample covariance) 
is not equivalent to that of $(\mathcal{L}_0^2,\cov)$  (i.e., zero covariance), and
on the contrary, the former excludes any jointly continuous, uncorrelated random variables.
Specifically, if $v_1,v_2\in\mathcal{L}_0^2$ are jointly continuous with $\cov(v_1,v_2)=0$, then
their $n$ independent paired observations $\bd{v}_1,\bd{v}_2\in\mathbb{R}^n$ 
have
$P(\bd{v}_1\cdot\bd{v}_2 =\bd{v}_1^\top \bd{v}_2\ne 0)=1$
 \citep[][p.\,134]{Roha15}.
Hence, $(\mathbb{R}^n,\cdot)$ is not a correct
space to define a decomposition for model \eqref{decomp in mat}.
Moreover, our decomposition defined from $(\mathcal{L}_0^2, \cov)$ enables us to investigate the asymptotic consistency of estimating unobservable $\{\mb{C}_k,\mb{D}_k\}_{k=1}^K$ and their explained proportions of signal variance. In contrast, the six existing methods are unable to establish 
the estimation consistency.  
%because their estimands defined on $(\mathbb{R}^n,\cdot)$
%are still approximations,
%with unclear high-dimensional convergence,
%to the counterparts rigorously defined on $(\mathcal{L}_0^2, \cov)$.

%For example, for the dataset-level proportion of signal variation explained by the common latent factors, their estimand $\|\mb{C}_k\|_F^2/\|\mb{X}_k\|_F^2$ is still an estimator of the statistical quantity $\tr\{\cov(\bd{c}_k)\}/\tr\{\cov(\bd{x}_k)\}.$

}

{\color{black}
Furthermore, based on $(\mathcal{L}_0^2,\cov)$,  we
can naturally use the variable-level proportion of signal variance explained by either
common or distinctive latent factors  in order
to quantify their influence on each variable 
for the purpose of variable selection.
In contrast,
the existing decomposition methods based on $(\mathbb{R}^n,\cdot)$
only consider the proportion of explained variation at {\color{black}the view level} and barely discuss it at the variable level \citep{Smil17}.
At the view level, 
they measure the variation of data by
the sum of squares of data points;
thus, their proportion of signal variation explained, for example, by
common latent factors 
is $\|\mb{C}_k\|_F^2/\|\mb{X}_k\|_F^2$,
which  essentially approximates the statistical quantity $\tr\{\cov(\bd{c}_k)\}/\tr\{\cov(\bd{x}_k)\}$ in $(\mathcal{L}_0^2,\cov)$.
More clearly seen at the variable level, variance
is superior over the Euclidean sum of squares
to measure the variation of a random variable, but
with the inevitable, challenging question on the
uniform consistency in estimation under high-dimensional settings \citep{fan2018eigenvector}.
This might be a reason that hinders the use  of the variable-level proportion of explained variation in the existing decomposition methods.

}

Even translated into $(\mathcal{L}_0^2, \cov)$,  
the six competing methods focus on the orthogonality (i.e., uncorrelatedness) between $\bd{c}_k$ and $\bd{d}_k$, but they inadequately consider orthogonality constraints among $\{\bd{d}_k\}_{k=1}^K$.  
Specifically, OnPLS, COBE, JIVE, and AJIVE do not impose any orthogonality on $\{\bd{d}_k\}_{k=1}^K$.  R.JIVE enforces such orthogonality at the price of relegating
its unexplained portion of signal $\bd{x}_k$ into noise $\bd{e}_k$. 
DISCO-SCA often only approximates, but not exactly achieves its target orthogonality for $\{\bd{d}_k\}_{k=1}^K$ \citep{Kloe16}.
When $K=2$, the orthogonality between $\bd{d}_1$ and $\bd{d}_2$ desirably assures no common latent factors
retained between them.
For $K>2$, 
with the same aim to well capture the common latent factors,
a similar desirable orthogonality constraint on $\{\bd{d}_k\}_{k=1}^K$ is 
that at least one pair among them are uncorrelated.
However, it is  unclear how to build a decomposition for all $K\ge 2$ that
can ensure both
the above desirable orthogonality among $\{\bd{d}_k\}_{k=1}^K$ and the interpretability of associated $\{\bd{c}_k\}_{k=1}^K$.
%After all, the former alone is insufficient to guarantee the latter.

We propose a novel method, called decomposition-based generalized canonical correlation analysis (D-GCCA), to handle  model   \eqref{decomp in mat}-\eqref{decomp in vari} with $K\ge 2$ views.  
Our method is equivalent to
D-CCA
when $K=2$. The key idea of D-GCCA is to divide the decomposition problem \eqref{decomp in vari} into multiple sub-problems 
via Carroll's GCCA \citep{Carr68}.
We slightly relax the aforementioned desirable orthogonality of $\{\bd{d}_k\}_{k=1}^K$ by enforcing it for each sub-problem. 
This in turn leads to a geometrically interpretable definition of $\{\bd{c}_k\}_{k=1}^K$
on space $(\mathcal{L}_0^2,\cov)$ by connecting Carroll's GCCA with PCA. %\citep{Hote33}
In particular, our defined common latent factors of $\{\bd{x}_k\}_{k=1}^K$
represent the same contribution made by the principal basis
of the entire signal space $\sum_{k=1}^K\lspan(\bd{x}_k^\top)$  
in generating each of the $K$ signal subspaces $\{\lspan(\bd{x}_k^\top)\}_{k=1}^K$. Here,
{\color{black}for any random vectors $\bd{v}_1$ and $\bd{v}_2$ with entries in $(\mathcal{L}_0^2, \cov)$,
$\lspan(\bd{v}_1^\top)$ denotes the subspace of $(\mathcal{L}_0^2, \cov)$ that is spanned by entries of $\bd{v}_1$, 
and $\lspan(\bd{v}_1^\top)+\lspan(\bd{v}_2^\top)=\lspan((\bd{v}_1^\top,\bd{v}_2^\top))$.}

Estimating matrices  $\{\mb{C}_k,\mb{D}_k\}_{k=1}^K$ {\color{black}and their proportions of explained signal variance
poses
theoretical and computational difficulties 
for high-dimensional data.
The observed high-dimensional matrices
$\{\mb{Y}_k\}_{k=1}^K$ are often 
high-rank in practice.
If the high-rank $\mb{Y}_k$ is directly treated as the signal,}
its associated high-rank covariance matrix   can be inconsistently
estimated by the traditional sample covariance matrix
due to the curse of ``intrinsic" high dimensionality
\citep{Yin88,Vers12}. 
Low-rank signal $\mb{X}_k$ or equivalently
low-rank $\cov(\bd{x}_k)$ is thus often assumed to
facilitate the construction of consistent estimates \citep{Shu17}.
Fortunately, big data matrices are often approximately low-rank 
in many real-world applications \citep{Udel19},
and their  low-rank approximations render feasible or more efficient computation, while retaining the major portion of  information \citep{Kish17}.
We consider the low-rank plus noise structure given in \eqref{decomp in mat}-\eqref{decomp in vari} under
the widely used high-dimensional spiked covariance model 
\citep{Fan13,Wang17,Shu17}. Subsequently, we  propose soft-thresholding based estimators for   $\{\mb{C}_k,\mb{D}_k\}_{k=1}^K$
{\color{black}
and therefrom derive estimators for
the proportions of signal variance explained by
either common or distinctive latent factors}.
Convergence properties of our estimators are established 
with reasonably good finite-sample performance shown by simulations.
The proposed estimators have closed-form expressions and
thus are more computationally efficient than
most existing methods that use time-expensive 
iterative optimization algorithms. 
For example, to decompose three $91{,}282{\times} 1080$ data matrices in our HCP application, 
our approach can complete in 18 seconds on a single computing node, whereas 
some state-of-the-art methods cannot converge within 5 hours.

	The contributions of this paper are summarized below:

\begin{itemize}
	\item We propose a novel decomposition method, called D-GCCA, for tackling $K\ge2$ data views under model \eqref{decomp in mat}, based on $(\mathcal{L}_0^2,\cov)$ instead of  $(\mathbb{R}^n,\cdot)$. Our distinctive-source matrices are especially imposed with an orthogonality constraint to avoid substantial loss of undetected common-source variation. The proposed common-source matrices exhibit a geometric interpretation from the perspective of PCA. Our D-GCCA reduces to D-CCA when $K=2$.

	\item We establish consistent estimators for our defined common-source and distinctive-source matrices under high-dimensional settings with convergence rates in both the Frobenius norm and the spectral norm. 
The proposed estimators have closed-form expressions and thus are computationally efficient. {\color{black} To the best of our knowledge, this is the first work that
establishes the high-dimensional estimation consistency under model~\eqref{decomp in mat} with $K\ge2$.}	

	{\color{black}\item

	We propose to use the variable-level proportion of signal variance explained by either common or distinctive latent factors for
selecting the most influenced variables. Consistent estimators are theoretically established and numerically verified. 

}

	\item We compare our D-GCCA with the six competing methods on both simulated and real-world data to show the superiority of proposed method for separating the common-source and distinctive-source variations across multi-view data. 
	
	\item As a byproduct, we reformulate Carroll's GCCA from the traditional $(\mathbb{R}^n,\cdot)$ to the more precise $(\mathcal{L}_0^2,\cov)$ 
	and provide some useful properties, which may facilitate the use of GCCA in statistical data integration.
\end{itemize}

The rest of this paper is organized as follows. We introduce our random-variable version of Carroll's GCCA and propose our D-GCCA method in Section~\ref{Sec: Method}. We propose our estimation approach of high-dimensional D-GCCA and establish its asymptotic properties in Section~\ref{sec: estimation}.
Section~\ref{Sec: simulation} evaluates the finite-sample performance of proposed estimators via simulations. 
{\color{black}We also compare D-GCCA with the six competing methods 
through simulated data in Section~\ref{Sec: simulation}
and through two 
real-world data examples from
TCGA
and HCP in Section~\ref{Sec: real data}.
 Concluding remarks are made in Section~\ref{Sec: conclusion}.
All theoretical proofs and additional simulation results are presented in Appendices.
{\color{black}A Python package for the proposed D-GCCA
method is available at \url{https://github.com/shu-hai/D-GCCA}.}

}

%Consider a toy example where $p_1=\ldots=p_K=1$ and noises $\{e_k\}_{k=1}^K$ have been excluded.
%Let $z_1,\ldots, z_K$ be $K$ standardized random variables.
%We now simplify the problem $x_k=c_k+d_k\in\mathbb{R}^{p_k}$ to $z_k=c+d_k\in \mathbb{R}$ for $k=1,\ldots,K$. 
%Specially, denote $c_{ij}$ to be the common variable $c$ for the pairwise decomposition between $z_i$ and $z_j$.
%The attempt to derive $c$ for $K>2$ from $\{c_{ij}\}_{1\le i<j\le K}$ 
%is equivalent to developing
%a similar but more complicated decomposition $c_{ij}=c+d_{ijk}$ ($1\le i,j\le K$)
%for $(K^2-K)/2$ variables.
%Thus, directly developing a decomposition in the form of \eqref{decomp in vari} for $K>2$ is more convenient than detouring from the pairwise decompositions.  Back to the toy example,
%This indicates the necessity to develop a decomposition approach for $K>2$.
%Regarding the geometric interpretability of $c$, 
%both its direction and magnitude count.
%The direction of $c$ can be naturally determined by
%maximizing the sum of squared correlations $\sum_{k=1}^K\corr^2(c,z_k)$.
%Then, the ideal structure of
%at least one orthogonal pair among $\{d_k\}_{k=1}^K$ 
%is used to calibrate the 
%magnitude of $c$, but its existence remains to be established.

We now introduce some notation.
For a real matrix $\mb{M}=(M_{ij})_{1\le i\le p,1\le j\le n}$, 
the $\ell$th largest singular value is denoted by $\sigma_\ell(\mb{M})$,
the $\ell$th largest eigenvalue when $p=n$
is $\lambda_{\ell}(\mb{M})$, 
the spectral norm is $\| \mb{M} \|_2=\sigma_1(\mb{M} )$, %\{\lambda_1(\mb{M}^\top\mb{M})\}^{1/2}$,
the Frobenius norm  is  $\| \mb{M} \|_F=(\sum_{i=1}^p\sum_{j=1}^n M_{ij}^2)^{1/2}$,
the matrix $\mathcal{L}^{\infty}$ norm  is  $\|\mb{M} \|_\infty=\max_{1\le i\le p}\sum_{j=1}^n |M_{ij}| $, 
the max norm  is  $\| \mb{M}\|_{\max}=\max_{1\le i\le p,1\le j\le n}|M_{ij}|$, 
and the Moore-Penrose pseudoinverse is $\mb{M}^\dag$.
Denote $\mb{M}^{[s:t,u:v]}$, $\mb{M}^{[s:t,:]}$,  and $\mb{M}^{[:,u:v]}$
as the submatrices 
$(M_{ij})_{s\le i\le t, u\le j\le v}$, $(M_{ij})_{s\le i\le t,1\le j\le n}$, 
and $(M_{ij})_{1\le i\le p, u\le j\le v}$
of $\mb{M}$, respectively.
{\color{black}Let $[\mb{M}_1;\dots;\mb{M}_N]=(\mb{M}_1^\top, \dots,\mb{M}_N^\top)^\top$ be the row-wise concatenation of matrices $\mb{M}_1,\dots,\mb{M}_N$
that have the same number of columns.}
We write the $j$th entry of a vector $\bd{v}$ by $\bd{v}^{[j]}$, and $\bd{v}^{[s:t]}=(\bd{v}^{[s]},\bd{v}^{[s+1]},\dots,\bd{v}^{[t]})^\top$.
{\color{black}For any random vectors $\bd{v}_1$ and $\bd{v}_2$,
denote $\cov(\bd{v}_1,\bd{v}_2)$ as the covariance matrix of $\bd{v}_1$ and $\bd{v}_2$
whose $(i,j)$th entry is $\cov(\bd{v}_1^{[i]},\bd{v}_2^{[j]})$,
and write $\cov(\bd{v}_1)=\cov(\bd{v}_1,\bd{v}_1)$.}
Define
$(v_i)_{i\in \mathcal{I}}$ by $(v_{i_1},\ldots,v_{i_q})$ with 
$\mathcal{I}=\{i_1,\ldots,i_q\}$ and $i_1<\ldots<i_q$.
%The equal symbol ``=" used between two random variables sometimes means the almost surely equality. 
The angle between any $x,y\in (\mathcal{L}_0^2, \cov)$ 
is denoted by $\theta(x,y)$,
and the norm of $x$ is 
$\| x\|=\sqrt{\var(x)}$.
We use $\cos\{\theta(x,y)\}$ and $\corr(x,y)$
exchangeably, and 
define $\corr(x,0)=0$.
The symbol $\perp$ used between 
two subspaces, sets, and/or random variables in $(\mathcal{L}_0^2, \cov)$ means their
orthogonality, i.e., uncorrelatedness. 
Define $r_0=0$, $r_k=\rank\{\cov(\bd{x}_k)\}$,
%${\color{black}r_{\min}=\min_{1\le k\le K}r_k}, {\color{black}r_{\max}=\max_{1\le k\le K}r_k},$
 and $r_f=\rank\{\cov([\bd{x}_1;\dots;\bd{x}_K]  )\}$. 
Note that
$r_k=\dim\{\lspan(\bd{x}_k^\top)\}$ and
$r_f=\dim\{\lspan([\bd{x}_1;\dots;\bd{x}_K]^\top )\}$.
{\color{black}For two sequences, write $a_n\asymp b_n$ iff $a_n=O(b_n)$ and $b_n=O(a_n)$, and 	
$a_n\lesssim_P b_n$ iff $a_n=O_P (b_n)$.
}
Throughout the paper, the asymptotic arguments  are by default under $n\to \infty$.

\section{Methodology}\label{Sec: Method}
We first develop the random-variable version of Carroll's GCCA {\color{black}in $(\mathcal{L}_0^2, \cov)$} and then use it to derive our D-GCCA decomposition.

\subsection{Generalized canonical correlation analysis}
{\color{black}In the literature, many GCCA methods   extend CCA to more than two data views based on different optimization criteria, such as the sum of correlations, the maximum variance (MAXVAR), and the minimum variance (MINVAR) \citep{Hors61, Carr68, Kett71}. 
We derive our D-GCCA for  model \eqref{decomp in mat}-\eqref{decomp in vari}
by using Carroll's GCCA~\mbox{\citep{Carr68}}.}

We first translate Carroll's GCCA into the space $(\mathcal{L}_0^2, \cov)$.
Carroll's GCCA was originally proposed and is often studied
in $(\mathbb{R}^n,\cdot)$ using data samples
\citep[e.g.,][]{Carr68,van11,Drap14}.
\citet{Kett71} briefly mentioned that
the random-variable version of Carroll's GCCA
is a mixture of the MAXVAR and MINVAR methods.  
We provide the solution to the optimization problem 
of Carroll's GCCA
in $(\mathcal{L}_0^2,\cov)$ 
as well as  some important properties.

For subspaces
$\{\lspan(\bd{x}_k^\top)\}_{k=1}^K$,
the Carroll's GCCA in $(\mathcal{L}_0^2,\cov)$ sequentially finds the closest elements among the $K$ subspaces.
The method has $r_f$ stages. The $\ell$th stage
finds the closest elements, denoted as $z_1^{(\ell)},\ldots,z_K^{(\ell)}$, among the $K$ subspaces,
which are called the $\ell$th-stage canonical variables,
along with an auxiliary variable 
$w^{(\ell)}$ as follows:
\be\label{GCCA}
\begin{split}
	&\{z_1^{(\ell)},\ldots,z_K^{(\ell)},w^{(\ell)} \}=\argmax_{\{z_1,\ldots,z_K,w\}}	\sum_{k=1}^K  \cos^2\{\theta(z_k, w)\}\\
	&\text{subject to}\quad	
	\begin{cases}
		z_k\in \lspan(\bd{x}_k^\top),\ \| z_k\|=1,
		\\
		w \perp \{w^{(j)}\}_{j=0}^{\ell-1}, \ w\in \mathcal{L}_0^2,\  \|w \|=1, {\color{black} w^{(0)}=0}.
	\end{cases}
\end{split}
\ee
{\color{black}
In $(\mathcal{L}_0^2,\cov)$, the cosine similarity
$\cos\{\theta(\cdot,\cdot)\}$ is equal to $\corr(\cdot,\cdot)$.
The auxiliary variable $w^{(\ell)}$ is the  variable closest
to all $\{z_k^{(\ell)}\}_{k=1}^K$,
and the sum of its squared cosine similarities  
with $\{z_k^{(\ell)}\}_{k=1}^K$ is used 
to measure the closeness of $\{z_k^{(\ell)}\}_{k=1}^K$.
The variable $w^{(\ell)}$ is also called the consensus variable of $\{z_k^{(\ell)}\}_{k=1}^K$ in  the literature \citep{kiers1994generalized,dahl2006bridge}.
Figure~\ref{Fig: D-GCCA steps}\,(a) illustrates the Carroll’s GCCA.
}

Let $\bd{f}_k^\top$ be an arbitrary orthonormal basis of $\lspan(\bd{x}_k^\top)$, $\bd{f}=[\bd{f}_1;\dots;\bd{f}_K]$, and
$\{\bd{\eta}^{(\ell)}\}_{1\le \ell \le r_f}$ be any $r_f$ orthonormal eigenvectors of $\cov(\bd{f})$,
where $\bd{\eta}^{(\ell)}=[\bd{\eta}_1^{(\ell)};\dots;\bd{\eta}_K^{(\ell)}]$ corresponds to   eigenvalue  $\lambda_\ell(\cov(\bd{f}))$ with  $\bd{\eta}_k^{(\ell)}\in \mathbb{R}^{r_k}$. We have $r_f=\rank\{\cov(\bd{f})\}$.
The following theorem presents the solution to \eqref{GCCA}
as well as  some  useful properties for  
our decomposition method.

\begin{thm}\label{GCCA thm}
	%\vspace{-0.1cm}
	The following results hold.
	\begin{enumerate}[label=(\roman*),font=\upshape]
		\item\label{GCCA thm(i)}  For $\ell\le r_f$ and $k\le K$, the solution of \eqref{GCCA}
		is given by 
		\begin{align}
		z_k^{(\ell)}
		&= \begin{cases}
		\text{any standardized variable in}~\lspan(\bd{x}_k^\top) ,&~~\text{if}~ \bd{\eta}_k^{(\ell)}= \bd{0},\\
		\pm (\bd{\eta}_k^{(\ell)}/\|\bd{\eta}_k^{(\ell)}\|_F)^\top\bd{f}_k, &~~\text{if}~ \bd{\eta}_k^{(\ell)}\ne \bd{0},
		\end{cases}
		\label{z_k formula}\\
		w^{(\ell)}&=[\lambda_{\ell}(\cov(\bd{f}))]^{-1/2}(\bd{\eta}^{(\ell)})^\top\bd{f}.
		\label{w formula}
		\end{align}
		Moreover, we have 
		\begin{align}%\label{cos angle w,z_k}
		&	\cos\{\theta(z_k^{(\ell)},w^{(\ell)})\}
		=\pm   [\lambda_{\ell}(\cov(\bd{f}))]^{1/2}\| \bd{\eta}_k^{(\ell)} \|_F,	
		\nonumber\\
		\label{ccorr=eig}
		&	\sum_{k=1}^K\cos^2\{\theta(z_k^{(\ell)},w^{(\ell)}) \}= \lambda_\ell(\cov(\bd{f})),
		\\
		&\sum_{k=1}^K\lspan(\bd{x}_k^\top)=\lspan(\{w^{(\ell)}\}_{\ell=1}^{r_f}).\label{w basis}
		\end{align}

		\item\label{GCCA thm(ii)} For $\ell\le r_f$, re-define $z_k^{(\ell)}$ in \eqref{z_k formula} to be
		\be\label{z_k revised}
		z_k^{(\ell)}
		= \begin{cases}
			0,  &~~\text{if}~ \bd{\eta}_k^{(\ell)}= \bd{0}, \text{i.e.}, w^{(\ell)}\perp \lspan(\bd{x}_k^\top),\\
			(\bd{\eta}_k^{(\ell)}/\|\bd{\eta}_k^{(\ell)}\|_F)^\top\bd{f}_k, &~~\text{otherwise}.
		\end{cases}
		\ee
		Then, we have $\theta(z_k^{(\ell)},w^{(\ell)})\in [0,\pi/2]$ \ and \
		$
		\lspan(\{z_k^{(\ell)}\}_{\ell=1}^{r_f})=\lspan(\bd{x}_k^\top).
		$
		
		\item\label{GCCA thm(iii)} 
		For $z_k^{(\ell)}$ in either \eqref{z_k formula} or \eqref{z_k revised},
		if 	
		$\lambda_{\ell}(\cov(\bd{f}))\le 1$ 
		and $\lspan(\{z_{k}^{(m)}\}_{m=1}^{\ell-1})\ne\lspan(\bd{x}_{k}^\top) $
		for some $\ell$ and $k$, 
		then there exists 
		a $w^{(\ell)}\in \lspan(\bd{x}_{k}^\top)$ such that $w^{(\ell)}\perp \sum_{1\le j\ne k\le K} \lspan(\bd{x}_j^\top)$.
		%	If there does not exists a $w^{(L_1)}\in \lspan(\bd{x}_k^\top)$ for any $k$, then $\lspan(\{z_k^{(\ell)}\}_{\ell\le L_1-1})=\lspan(\bd{x}_k^\top)$ for all $k$, then $L_1-1=r_f$?????????
	\end{enumerate}
\end{thm}
In the following text, if without further clarification,
we refer $z_k^{(\ell)}$ to the one defined in~\eqref{z_k revised} so that
$\theta(z_k^{(\ell)},w^{(\ell)})$ falls into $[0,\pi/2]$.

{\color{black}
Unlike our D-GCCA  for
disentangling the common and distinctive latent factors
and their explained variations among multiple data views, 
the existing GCCA methods \citep[e.g.,][]{Hors61,Carr68, Kett71, tenenhaus2011regularized, Tene14, cai2020sparse}
often focus on finding the canonical variables $\{z_k^{(\ell)}\}_{k=1}^K$, which are merely the most correlated components among the multiple views, and studying the coefficients in their linear expressions
formed by corresponding signal variables $\{\bd{x}_k^{[i]}\}_{i=1}^{p_k}$.
The auxiliary variables $w^{(\ell)}$s 
of Carroll's GCCA or its variant MAXVAR GCCA %(instead with sum of cosines in \eqref{GCCA})
are  also called 
as a consensus or common latent representation of multi-view data in the literature  \citep{kiers1994generalized,dahl2006bridge,fu2017scalable, Bent19},
but they do not solve our problem in~\eqref{decomp in mat}-\eqref{decomp in vari}, which also involves distinctive latent factors. %that help better determine and interpret the common latent factors and their explained variation \citep{westerhuis2019data}.
As extensions of MAXVAR GCCA for \eqref{decomp in mat}-\eqref{decomp in vari}, 
AJIVE and COBE 
treat the consensus variables $w^{(\ell)}$s 
as the common latent factors
and define $\bd{c}_k$ as the projection of $\bd{x}_k$ onto the space spanned by $w^{(\ell)}$s.
This simple approach is undesirable even for $K = 2$ views,
where both Carroll's GCCA and MAXVAR GCCA reduce to CCA.
For example,  if $\bd{x}_k=z_k^{(1)}$ for $k\le K=2$, then 
one only needs to consider  
the first-stage consensus variable $w^{(1)}$. 
Let $\bd{c}_k$ be the projection of $\bd{x}_k=z_k^{(1)}$ 
onto $w^{(1)}$, then $\bd{c}_k=(z_1^{(1)}+z_2^{(1)})/2$
and $\bd{d}_1=-\bd{d}_2=(z_1^{(1)}-z_2^{(1)})/2$,
but the distinctive-source random vectors $\bd{d}_1$ and $\bd{d}_2$ now share the same latent factor $(z_1^{(1)}-z_2^{(1)})/2$, contradicting their definition that they are generated from distinctive latent factors.
%However, $\{w^{(\ell)}\}_{\ell=1}^{r_f}$ fail to characterize the magnitudes of the common latent factors because $\{w^{(\ell)}\}_{\ell=1}^{r_f}$ are standardized variables with unit norm. One may alternatively use the smallest length of the projections of $\{z_k^{(\ell)}\}_{k=1}^K$ onto $w^{(\ell)}$ as the magnitude of the corresponding common latent factor $c^{(\ell)}$. This simple approach is undesirable even for $K=2$ data types, because its distinctive latent factors $d_1^{(\ell)}=-d_2^{(\ell)}=\frac{1}{2}(z_1^{(\ell)}-z_2^{(\ell)})$ with a perfect absolute correlation $|\corr(d_1^{(\ell)},d_2^{(\ell)})|=1$ if $z_1^{(\ell)}\ne z_2^{(\ell)}$. 
%In contrast, our D-GCCA first calibrates each common latent factor $c^{(\ell)}=\alpha^{(\ell)}w^{(\ell)}$ by imposing the desirable orthogonality \ref{(O.2)} on distinctive latent factors $\{d_k^{(\ell)}\}_{k=1}^K$ together with constraints \ref{(C.1)}  and \ref{(C.2)}, and then lets each $c^{(\ell)}$ use the same coefficient of $z_k^{(\ell)}$ in $\bd{x}_k$ to form the $\bd{c}_k$ in \eqref{c_k vec}. In the above example, our D-GCCA yields $\bd{d}_1\perp\bd{d}_2$ with $\bd{c}_k=\frac{1}{2}[1-\tan\theta(z_1^{(1)},z_2^{(1)})](z_1^{(1)}+z_2^{(1)})$.
In contrast, our D-GCCA, detailed in the next subsection, 
yields $\bd{d}_1\perp\bd{d}_2$ with $\bd{c}_k=[1-\tan\{ {\theta(z_1^{(1)},z_2^{(1)})}/2\}](z_1^{(1)}+z_2^{(1)})/2$.
See \citet{guo2019canonical} and \citet{wong2021deep}
for  an overview and recent progress
in GCCA-based multi-view data analysis.

}

{\color{black}\subsection{Decomposition-based generalized canonical correlation analysis}}
\subsubsection{Common-source and distinctive-source matrices and random vectors}\label{subsec: mat def}
In the model given by \eqref{decomp in mat}-\eqref{decomp in vari},
the columns of each common-source matrix $\mb{C}_k$ or distinctive-source matrix $\mb{D}_k$ are assumed to be $n$ independent copies of its corresponding random vector $\bd{c}_k$ or $\bd{d}_k$.
We thus consider the following decomposition with noise excluded: 
\be\label{x=c+d}
\bd{x}_k=\bd{c}_k+\bd{d}_k \quad \mbox{for}~~k=1,\ldots, K.
\ee
The estimation of $\{\mb{C}_k,\mb{D}_k\}_{k=1}^K$ from noisy data $\{\mb{Y}_k\}_{k=1}^K$ will be given in Section~\ref{sec: estimation}.

Like the divide-and-conquer strategy of D-CCA,
our D-GCCA first breaks down  decomposition problem \eqref{x=c+d} into multiple sub-problems. Each $\ell$th sub-problem 
is solved by finding a common variable $c^{(\ell)}$
and $K$ distinctive variables $\{d_k^{(\ell)}\}_{k=1}^K$
for the $\ell$th-stage canonical variables $\{z_{k}^{(\ell)}\}_{k=1}^K$
such that
\be\label{z=c+d}
z_{k}^{(\ell)}=c^{(\ell)}+d_k^{(\ell)} \quad \mbox{for}~~k=1,\ldots, K.
\ee
{\color{black}
The ideal orthogonality among $\{\bd{d}_k\}_{k=1}^K$ and its reduced version on $\{d_k^{(\ell)}\}_{k=1}^K$ are given below.
\begin{enumerate}[label=(O.\arabic*)]
\item\label{(O.1)} At least one pair among $\{\lspan(\bd{d}_k^\top)\}_{k=1}^K$ is orthogonal.
\item\label{(O.2)} At least one pair among $\{d_k^{(\ell)}\}_{k=1}^K$ is orthogonal.
\end{enumerate} 
}
The auxiliary variable $w^{(\ell)}$ in \eqref{GCCA} naturally serves as the direction variable of our common variable $c^{(\ell)}$ of $\{z_{k}^{(\ell)}\}_{k=1}^K$.
We define $c^{(\ell)}$ by
\be\label{c for K>=2}
c^{(\ell)}=\alpha^{(\ell)}w^{(\ell)},
\ee
where $\alpha^{(\ell)}$ satisfies   
\begin{enumerate}[label=(C.\arabic*)]
	\item\label{(C.1)} $|\alpha^{(\ell)}|$ is the smallest value such that \ref{(O.2)} holds;
	\item\label{(C.2)} $\alpha^{(\ell)}<0$ if \ref{(C.1)} has two solutions with respect to $\alpha^{(\ell)}$.
\end{enumerate} 

\begin{figure}[b!]
	\begin{subfigure}[b]{0.32\textwidth}
		%	\centering
		\includegraphics[width=1\textwidth]{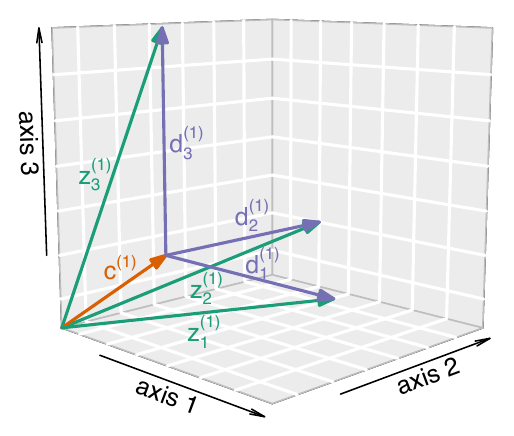}
		\caption{$c^{(1)}=\alpha_1^{(1)}w^{(1)}$}
	\end{subfigure}
%	\hspace{0.3cm}
	\begin{subfigure}[b]{0.32\textwidth}
		%	\centering
		\includegraphics[width=1\textwidth]{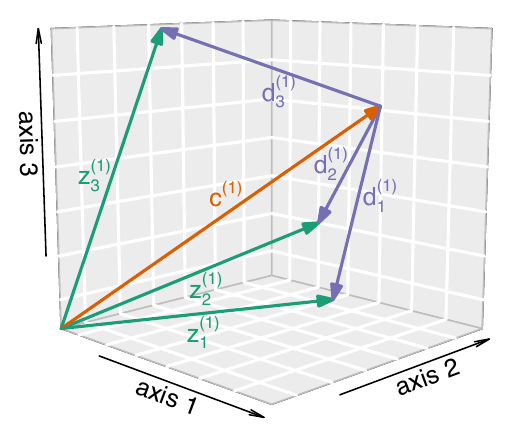}
		\caption{$c^{(1)}=\alpha_2^{(1)}w^{(1)}$}
	\end{subfigure}
%	\hspace{0.3cm}
	\begin{subfigure}[b]{0.32\textwidth}
		%	\centering
		\includegraphics[width=1\textwidth]{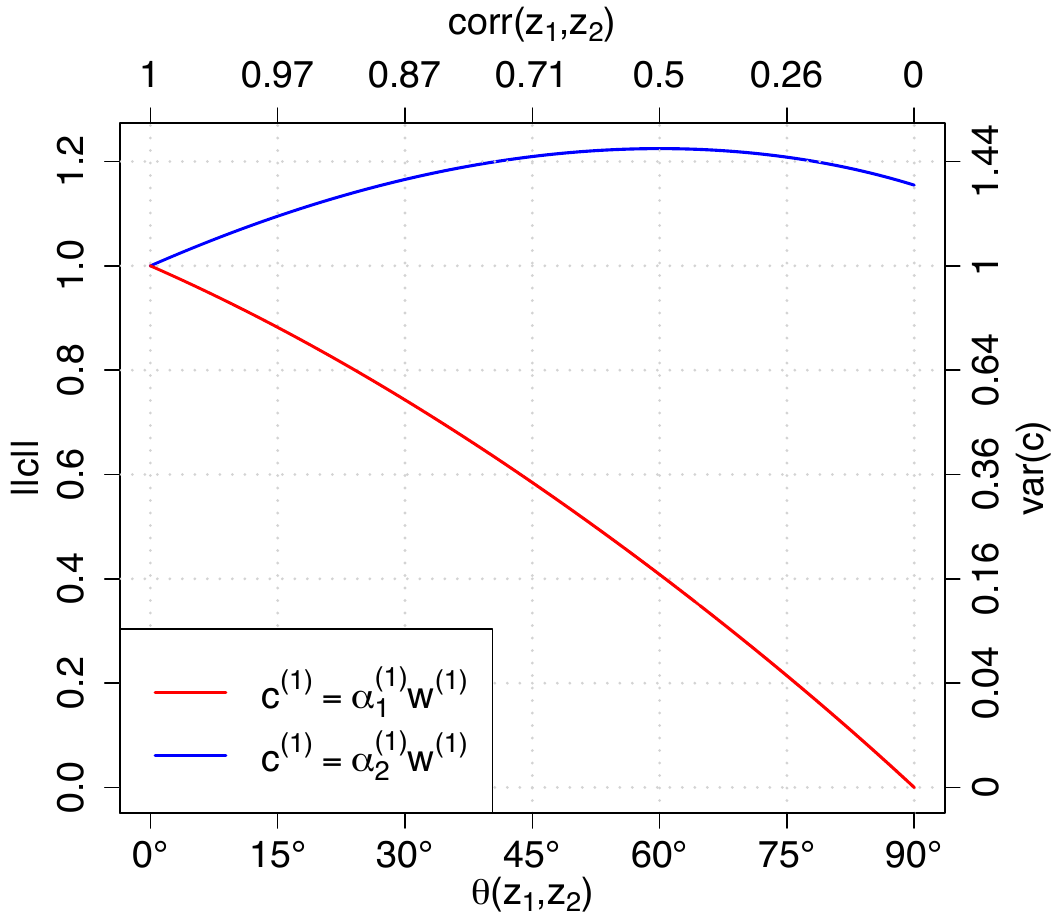}
		\caption{$\|c^{(1)}\|$ vs. $\theta(z_1^{(1)},z_2^{(1)})$}
	\end{subfigure}
	\caption{{\color{black}The geometry of D-GCCA for $K=3$ data views with $\theta(z_1^{(1)},z_2^{(1)})=\theta(z_1^{(1)},z_3^{(1)})=\theta(z_2^{(1)},z_3^{(1)})\in (0^\circ,90^\circ)$.
There are only two possible choices of $\alpha^{(1)}$ for the common variable $c^{(1)}=\alpha^{(1)}w^{(1)}$ such that at least one pair among $\{d_k^{(1)}\}_{k=1}^3$ is orthogonal: 
$c^{(1)}=\alpha_1^{(1)}w^{(1)}$ in panel (a) and $c^{(1)}=\alpha_2^{(1)}w^{(1)}$ in panel (b), where $\alpha_1^{(1)}<\alpha_2^{(1)}$, and $d_1^{(1)},d_2^{(1)}$ and $d_3^{(1)}$ are  mutually orthogonal.
Panel (c) shows that
as $\theta(z_1^{(1)},z_2^{(1)})$ increases
or equivalently as $\corr(z_1^{(1)},z_2^{(1)})=\cos\{\theta(z_1^{(1)},z_2^{(1)})\}$ decreases, 
$\|c^{(1)}\|=\sqrt{\var(c^{(1)})}$ decreases 
if $c^{(1)}=\alpha_1^{(1)}w^{(1)}$, but is not monotonic if $c^{(1)}=\alpha_2^{(1)}w^{(1)}$. D-GCCA chooses $c^{(1)}=\alpha_1^{(1)}w^{(1)}$.}
}
	\label{Fig: c=alpha*w}
\end{figure}

The rationale of  setting  constraints \ref{(C.1)} and \ref{(C.2)} is given  as follows.
Let $\alpha_1^{(\ell)}$ and $\alpha_2^{(\ell)}$ be two candidate values of $\alpha^{(\ell)}$, each of which leads to
the required orthogonality \ref{(O.2)}. 
If $|\alpha_1^{(\ell)}|<|\alpha_2^{(\ell)}|$,  then 
the extra variance $(|\alpha_2^{(\ell)}|^2-|\alpha_1^{(\ell)}|^2)$
for the variable $c^{(\ell)}$ of $\alpha_2^{(\ell)}$
can be alternatively explained by the variables $\{d_{k}^{(\ell)}\}_{k=1}^K$ of $\alpha_1^{(\ell)}$.
{\color{black} 
Figure~\ref{Fig: c=alpha*w} shows 
a motivating example 
with $K=3$ and equal angles among $\{z_k^{(1)}\}_{k=1}^3$; $\alpha^{(1)}=\alpha_1^{(1)}$ is more sensible, because as $\corr(z_1^{(1)},z_2^{(1)})=\cos\{\theta(z_1^{(1)},z_2^{(1)})\}$
 increases from~0 to~1, 
 the variance of $c^{(1)}=\alpha_1^{(1)}w^{(1)}$ also increases from 0 to 1, reflecting the strength of the correlation, 
whereas the variance of $c^{(1)}=\alpha_2^{(1)}w^{(1)}$
is not monotonic.}
If $\alpha_1^{(\ell)}<0<\alpha_2^{(\ell)}$ and $|\alpha_1^{(\ell)}|=|\alpha_2^{(\ell)}|$,  then the $d_{k}^{(\ell)}$ corresponding to 
$\alpha_1^{(\ell)}$, 
for $k=1,\dots,K$,
has a larger variance than that to $\alpha_2^{(\ell)}$.

We provide
the existence and explicit formula of $\alpha^{(\ell)}$  in the theorem below.

\begin{thm}\label{alpha thm}
	For $\ell\le r_f$, $w^{(\ell)}$ in \eqref{w formula}, and $\{z_k^{(\ell)}\}_{k=1}^K$ in \eqref{z_k revised},  we have that $\alpha^{(\ell)}$ in \eqref{c for K>=2} exists and satisfies
	\begin{align*}
	\alpha^{(\ell)}\in\argmin_{\alpha_{jk}^{(\ell)}}\bigg\{
	|\alpha_{jk}^{(\ell)} | : 
	\alpha_{jk}^{(\ell)} =\frac{1}{2}
	\left[
	\cos\{\theta(w^{(\ell)},z_j^{(\ell)})\}
	+
	\cos\{\theta(w^{(\ell)},z_k^{(\ell)})\}
-
	(
	\Delta_{jk}^{(\ell)}
	)^{1/2}
	\right]&
	\nonumber\\
	\text{for}		
	\quad \Delta_{jk}^{(\ell)}\ge 0
	\quad \text{and}\quad
	1\le j< k \le K
	&\bigg\}
	\end{align*}
	with
	$
	\Delta_{jk}^{(\ell)}=
	[\cos\{\theta(w^{(\ell)},z_j^{(\ell)})\}
	+
	\cos\{\theta(w^{(\ell)},z_k^{(\ell)})\}]^2	
	-4	\cos\{\theta(z_j^{(\ell)},z_k^{(\ell)})\}.
	$
\end{thm}

\begin{remark}\label{remark1}
We interpret the decomposition given in \eqref{z=c+d}-\eqref{c for K>=2} 
via analyzing the relationship
between the entire signal space  $\sum_{k=1}^K\lspan(\bd{x}_k^\top)$
and its subspaces $\{\lspan(\bd{x}_k^\top)\}_{k=1}^K$. 
First, from the perspective of PCA, we consider how the $K$ signal subspaces $\{\lspan(\bd{x}_k^\top)\}_{k=1}^K$ contribute to forming the whole signal space $\sum_{k=1}^K\lspan(\bd{x}_k^\top)$. We use an arbitrary orthonormal basis $\bd{f}_k^\top$ of $\lspan(\bd{x}_k^\top)$ to represent its contribution to $\sum_{k=1}^K\lspan(\bd{x}_k^\top)$, because $\bd{f}_k^\top$ fully characterizes $\lspan(\bd{x}_k^\top)$ due to $\lspan(\bd{x}_k^\top)=\{\bd{f}_k^\top \bd{b} : \forall \bd{b}\in \mathbb{R}^{r_k}\}$, and its entries, all of which are standardized variables, provide a fair comparison among subspaces $\{\lspan(\bd{x}_k^\top)\}_{k=1}^K$. 
By \eqref{w formula} and \eqref{w basis}, %we see that
$\{w^{(\ell)}\sqrt{\lambda_{\ell}(\cov(\bd{f}))}\}_{\ell=1}^{r_f}$ are the $r_f$ principal components of $\bd{f}^\top=(\bd{f}_1^\top,\ldots,\bd{f}_K^\top)$, which fully capture the variance of $\bd{f}$, that is, the accumulated contribution to  $\sum_{k=1}^K\lspan(\bd{x}_k^\top)$ from all subspaces $\{\lspan(\bd{x}_k^\top)\}_{k=1}^K$.
They also constitute an orthogonal basis of $\sum_{k=1}^K\lspan(\bd{x}_k^\top)$ that is the closest to these subspaces 
in the sense of~\eqref{GCCA}.
{\color{black}We thus call standardized variables $\{w^{(\ell)}\}_{\ell=1}^{r_f}$ as 
the {\it principal basis} of $\sum_{k=1}^K\lspan(\bd{x}_k^\top)$ with respect to $\{\lspan(\bd{x}_k^\top)\}_{k=1}^K$.}
Next, from the perspective of the principal basis $\{w^{(\ell)}\}_{\ell=1}^{r_f}$, we conversely deduce how
the entire signal space $\sum_{k=1}^K\lspan(\bd{x}_k^\top)$ generates its subspaces
$\{\lspan(\bd{x}_k^\top)\}_{k=1}^K$.
With $0/0:=0$,
$z_k^{(\ell)}$ is the normalized projection of $w^{(\ell)}$ onto $\lspan(\bd{x}_k^\top)$.
Theorem~\ref{GCCA thm}~\ref{GCCA thm(ii)} shows
that the normalized projections $\{z_k^{(\ell)}\}_{\ell=1}^{r_f}$
of $\{w^{(\ell)}\}_{\ell=1}^{r_f}$
span the subspace $\lspan(\bd{x}_k^\top)$ for each $k\le K$.
Hence, the decomposition in \eqref{z=c+d}-\eqref{c for K>=2} essentially measures the same contribution of the principal-basis component $w^{(\ell)}$  
%of the entire signal space $\sum_{k=1}^K\lspan(\bd{x}_k^\top)$ 
in generating each of the $K$ signal subspaces $\{\lspan(\bd{x}_k^\top)\}_{k=1}^K$.
%Our decomposition is illustrated in Figure~\ref{decompose struc}.
\end{remark}

\iffalse
This leads to the following definition.

\begin{definition}
	Standardized variables $\{w^{(\ell)}\}_{\ell=1}^{r_f}$ given in \eqref{w formula} are defined  to be
	the  principal basis of $\sum_{k=1}^K\lspan(\bd{x}_k^\top)$ with respect to
	$\{\lspan(\bd{x}_k^\top)\}_{k=1}^K$.
\end{definition}
\fi

\begin{remark}\label{remark2}
	Let $L=\max\{\ell\in \{1,\ldots,r_f\}:\lambda_\ell(\cov(\bd{f}))> 1\}$. 
%{\color{black} To capture the uniform contribution to all the subspaces $\{\lspan(\bd{x}_k^\top)\}_{k=1}^K$ from the principal basis $\{w^{(\ell)}\}_{\ell=1}^{r_f}$,}
We only need to consider the first $L$ principal-basis components $\{w^{(\ell)}\}_{\ell=1}^L$ due to the following reasons.
	%When $K=2$, Lemma~2 in \citet{Kett71} shows that $\lambda_{r_{12}}(\cov(\bd{f}))>1$ with $r_{12}=\cov(\bd{f}_1,\bd{f}_2)$, and $\lambda_\ell(\cov(\bd{f}))=1$ for $r_{12}+1\le \ell\le r_1+r_2-r_{12}$.
	For $\ell>L$, 
	by Theorem~\ref{GCCA thm}~\ref{GCCA thm(iii)},
	either there exists a $w^{(\ell)}\in \lspan(\bd{x}_k^\top)$ for some $k$ that is orthogonal to
	all the other signal subspaces $\{\lspan(\bd{x}_j^\top)\}_{j\ne k}$, or otherwise, 
	$\{z_k^{(m)}\}_{m=1}^{\ell-1}$   has  spanned the subspace $\lspan(\bd{x}_k^\top)$ for all $k=1,\dots,K$.
	The first scenario results in $c^{(\ell)}=0$, and 
	 the second one indicates that the contribution of $w^{(\ell)}$
	to each signal subspace has already been accomplished by 
	the preceding components $\{w^{(m)}\}_{m=1}^{\ell-1}$.
{\color{black}
Our stopping rule $\ell\le L$ for Carroll's GCCA when $K\ge2$
is an extension from the stopping rule $\ell\le r_{12}$ of the CCA with $K=2$, where $r_{12}=\max\{\ell\in\{1,\dots, r_f\}: \corr(z_1^{(\ell)},z_2^{(\ell)})>0\}$ is the number of positive canonical correlations.
The number $L=r_{12}$ when $K=2$, because
$\lambda_\ell(\cov(\bd{f}))=1+\corr(z_1^{(\ell)},z_2^{(\ell)})>1$ if $\ell\le r_{12}$, and otherwise $\lambda_\ell(\cov(\bd{f}))\le 1$ \citep[][Lemma~2]{Kett71}.

%{\color{black} 
%Consistent with Theorem~\ref{GCCA thm}~\ref{GCCA thm(iii)},   the canonical correlation $\corr(z_1^{(\ell)},z_2^{(\ell)})>0$ if and only if $\ell=1, \dots, r_{12}$.  Therefore, it follows from Theorem~1 in \citet{Shu17} that if $r_{12}<r_k$, then $w^{(L+1)}=w^{(r_{12}+1)}$ can be any variable in $\lspan(\bd{x}_k^\top)\setminus \lspan(\{z_k^{(\ell)}\}_{\ell=1}^{r_{12}})$ and consequently $w^{(L+1)}\perp \lspan(\bd{x}_j^\top)$ for $1\le j\ne k\le 2$, and otherwise $r_{12}=r_1=r_2$ and $\lspan(\bd{x}_k^\top)=\lspan(\{z_k^{(\ell)}\}_{\ell=1}^{r_{12}})$ for $k=1,2$. } 
}	
	
\end{remark}

We now combine the decompositions for all $\ell=1,\ldots,L$  in \eqref{z=c+d} to form the original 
decomposition~\eqref{x=c+d}.
%Theorem~\ref{alpha thm} shows that $\alpha^{(\ell)}=0$ if and only if $z_k^{(\ell)}=0$ for some $k$ or $z_j^{(\ell)}\perp z_k^{(\ell)}$ for some $j\ne k$.
Define the index set of nonzero $c^{(\ell)}$s  by 
$\mathcal{I}_0=\{\ell\in \{1,\ldots,L\}: c^{(\ell)}\ne 0, \ \text{i.e.},\ \alpha^{(\ell)}\ne 0\}$. 
We set $\bd{c}_k=\bd{0}_{p_k\times 1}$ and
$\mb{C}_k=\mb{0}_{p_k\times n}$ for all $k$ when $\mathcal{I}_0= \emptyset$, so  
we only consider $\mathcal{I}_0\ne \emptyset$ as follows.
Let $\bd{z}_k^{\mathcal{I}_0}=(z_k^{(\ell)})_{\ell\in \mathcal{I}_0}^\top$.
The portion of $\bd{x}_k$ generated from latent factors $\bd{z}_k^{\mathcal{I}_0}$ is equivalent to
the  projection of $\bd{x}_k$ onto $\lspan\{(\bd{z}_k^{\mathcal{I}_0})^\top\}$ given by
\be\label{linear rep of x_k on z_k}
\cov(\bd{x}_k,\bd{z}_k^{\mathcal{I}_0})\{\cov(\bd{z}_k^{\mathcal{I}_0})\}^\dag \bd{z}_k^{\mathcal{I}_0}
=\cov(\bd{x}_k,\bd{z}_k^{\mathcal{I}_0})\{\cov(\bd{z}_k^{\mathcal{I}_0})\}^\dag (c^{(\ell)}+d_k^{(\ell)})_{\ell\in\mathcal{I}_0}^\top.
\ee
Here, $\cov(\bd{x}_k,\bd{z}_k^{\mathcal{I}_0})\{\cov(\bd{z}_k^{\mathcal{I}_0})\}^\dag$ is a deterministic coefficient matrix.
We define the common-source vector $\bd{c}_k$ of $\bd{x}_k$
as
\be\label{c_k vec}
\bd{c}_k=\cov(\bd{x}_k,\bd{z}_k^{\mathcal{I}_0})\{\cov(\bd{z}_k^{\mathcal{I}_0})\}^\dag\bd{c}^{\mathcal{I}_0},
\ee
which is the portion of \eqref{linear rep of x_k on z_k}
comes from the common latent factors $(\bd{c}^{\mathcal{I}_0})^\top:=(c^{(\ell)})_{\ell\in \mathcal{I}_0}$.
%Here, $\bd{c}^{\mathcal{I}_0}$ preserves the same coefficient matrix $\cov(\bd{x}_k,\bd{z}_k^{\mathcal{I}_0})[\cov(\bd{z}_k^{\mathcal{I}_0})]^\dag$ 
%of $\bd{z}_k^{\mathcal{I}_0}$ for $\bd{x}_k$.

\begin{definition}
{\color{black}For D-GCCA,}
we define the common-source random vector $\bd{c}_k$ of $\bd{x}_k$ as \eqref{c_k vec} and the distinctive-source random vector $\bd{d}_k=\bd{x}_k-\bd{c}_k$. The common-source matrix $\mb{C}_k$ and distinctive-source matrix $\mb{D}_k$ are the corresponding sample matrices of $\bd{c}_k$ and $\bd{d}_k$, respectively. 
{\color{black}
The  $\{c^{(\ell)}\}_{\ell\in \mathcal{I}_0}$ in \eqref{c for K>=2}
are called the common latent factors of $\{\bd{x}_k\}_{k=1}^K$,
and $\{d_k^{(\ell)}\}_{\ell=1}^{r_f}$ in \eqref{z=c+d} are called the distinctive latent factors
of $\bd{x}_k$.
}

\end{definition}

\begin{figure}[b!]
	\begin{subfigure}[b]{0.6\textwidth}
			\centering
		\includegraphics[width=1\textwidth]{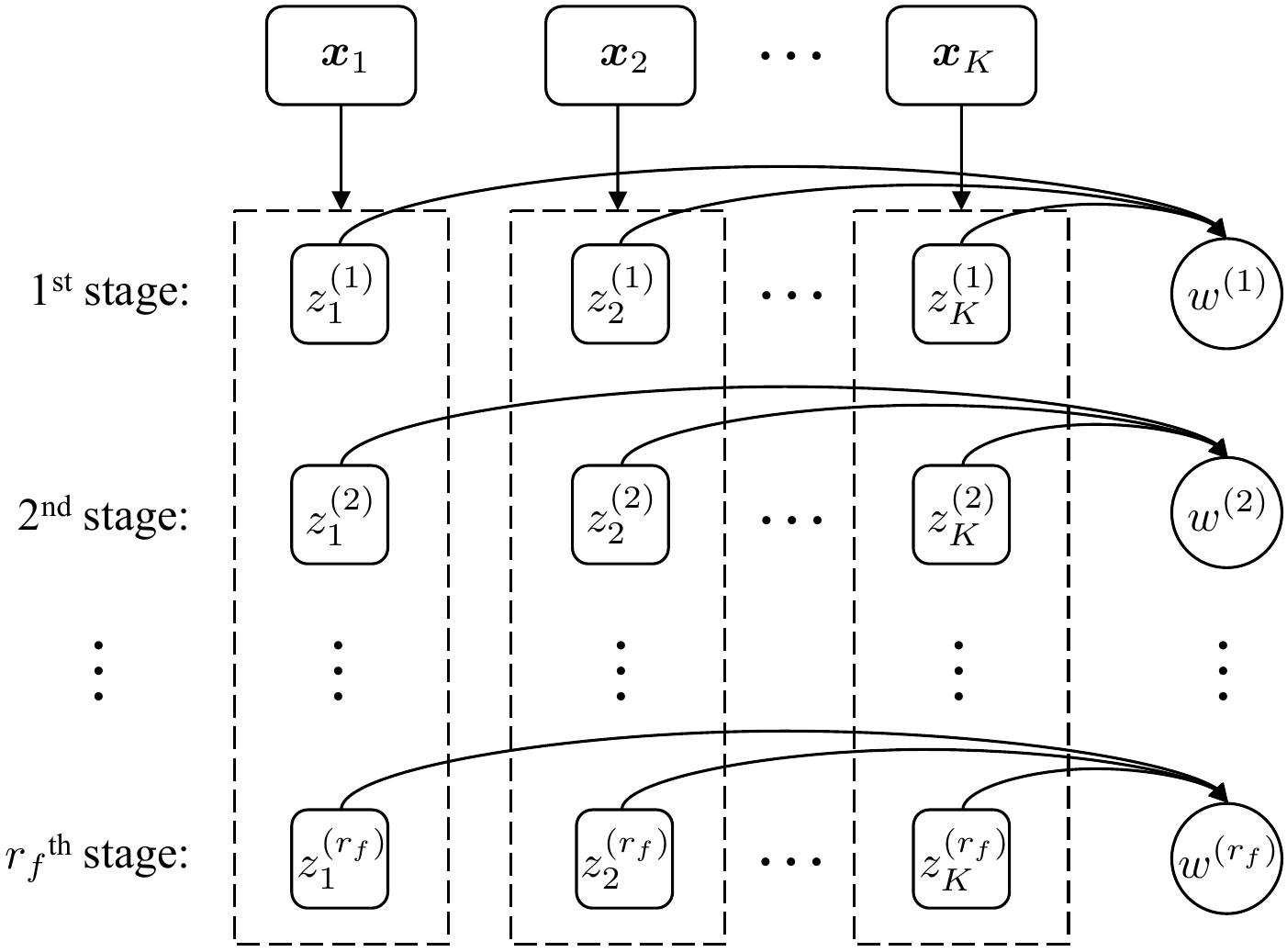}
			\smallskip
		\caption{Step 1: Eqn.\,\eqref{GCCA} (i.e., Carroll's GCCA)}
	\end{subfigure}
	\bigskip
	\hspace{0.05\textwidth}
	\begin{subfigure}[b]{0.35\textwidth}
			\centering
		\includegraphics[width=1\textwidth]{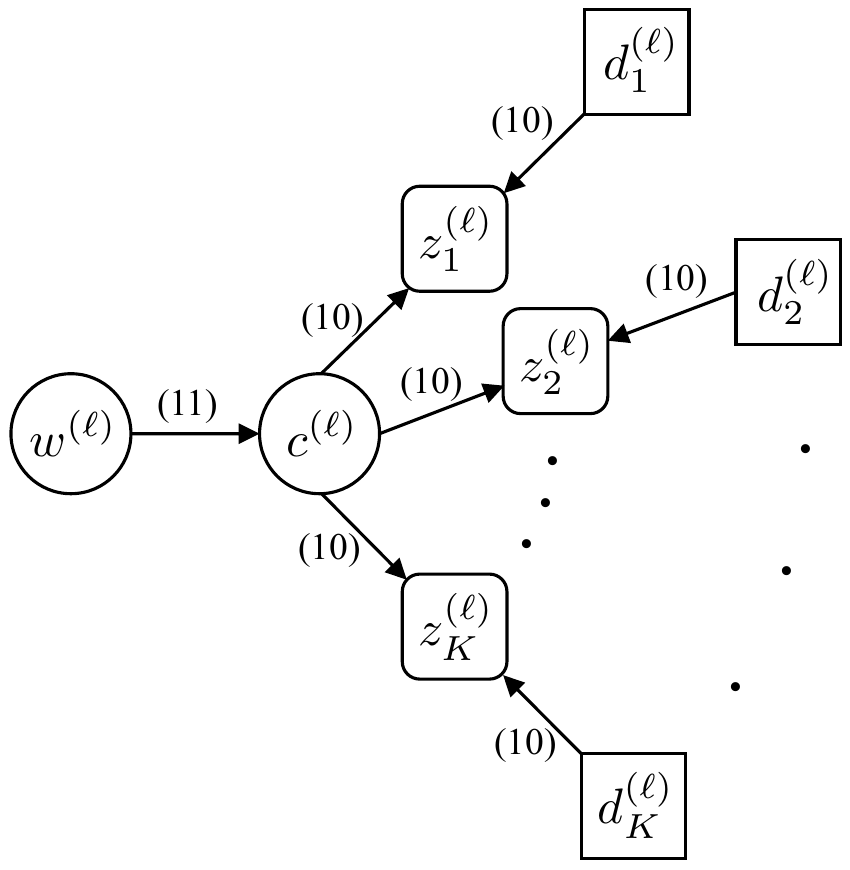}
		\smallskip
		\caption{Step 2: Eqns. \eqref{z=c+d} and \eqref{c for K>=2}}
	\end{subfigure}
	\begin{subfigure}[b]{1\textwidth}
	\bigskip
			\centering
		\includegraphics[width=0.7\textwidth]{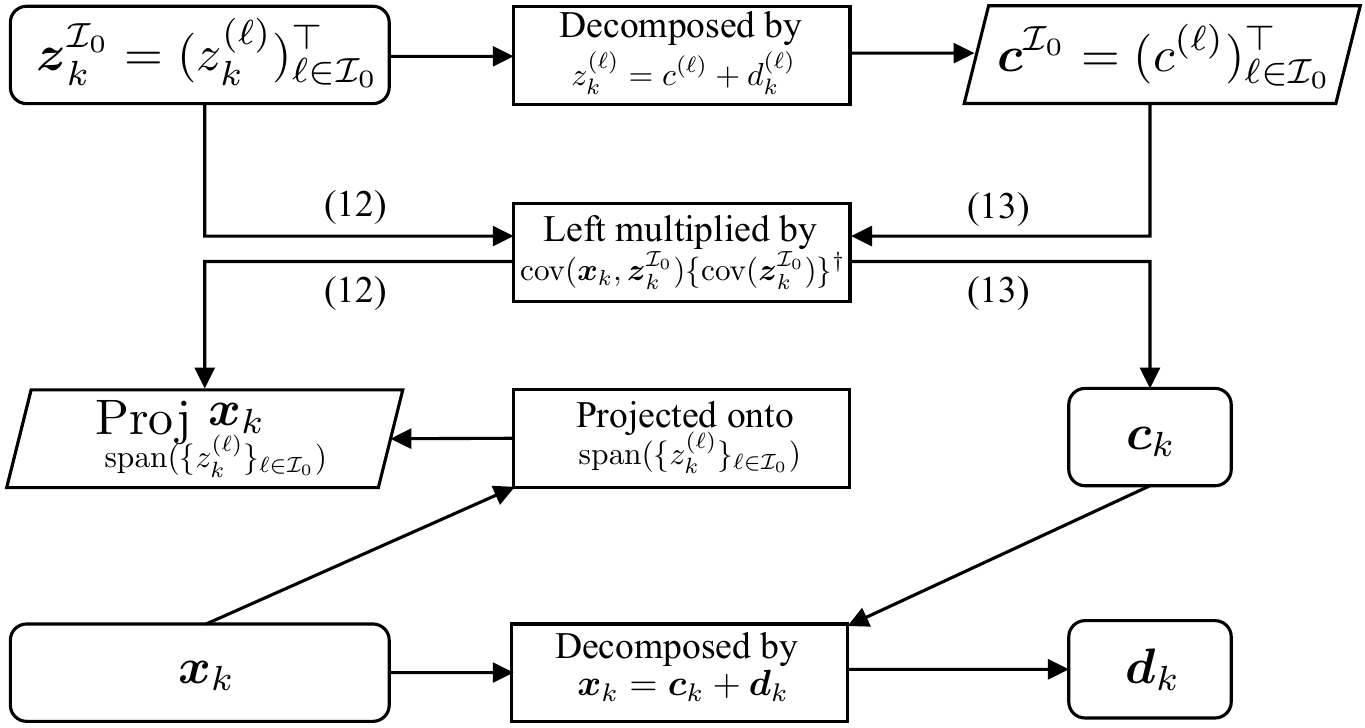}
		\bigskip
		\caption{Step 3: Eqns. \eqref{linear rep of x_k on z_k} and \eqref{c_k vec}}
	\end{subfigure}
	\caption{{\color{black}Illustration of D-GCCA steps.}
}
	\label{Fig: D-GCCA steps}
\end{figure}

{\color{black} Figure~\ref{Fig: D-GCCA steps} illustrates the steps
 of the proposed D-GCCA.}
When $K=2$, 
the following theorem shows that
our D-GCCA is equivalent to D-CCA.

\begin{thm}\label{equivalence to K=2}
	When $K=2$, $\{\bd{c}_k\}_{k=1}^K$ in \eqref{c_k vec} are the same as those of D-CCA in \textup{(16)} of \citet{Shu17}.
\end{thm}

We further investigate the uniqueness of $\{\bd{c}_k\}_{k=1}^K$.

\begin{thm}\label{model uniqueness}
	For $L\ge 1$, if $\lambda_1(\cov(\bd{f})),\ldots,\lambda_L(\cov(\bd{f}))$ are distinct, then  $\{\bd{c}_k\}_{k=1}^K$ are uniquely defined by~\eqref{c_k vec} regardless of the non-unique choice of $\bd{f}$ and $\{\bd{\eta}^{(\ell)}\}_{1\le \ell \le L}$.
\end{thm}

The largest $L$ eigenvalues of $\cov(\bd{f})$ are invariant to the choice of $\bd{f}$. 
For a given $\bd{f}$, the distinctness of these $L$ eigenvalues ensures the identifiability of $\{\bd{\eta}^{(\ell)}\}_{1\le \ell \le L}$ up to a sign change and thus simplifies the analysis. 
Analogous assumptions are often made in 
the literature \citep{Zhou08,Birn13,Wang17}.
{\color{black}
	If the joint distribution of the $n$ $(\ge L)$ samples of $\bd{f}$ is absolutely continuous or elliptically contoured, then the largest $L$ eigenvalues of its sample covariance matrix are distinct with probability one \citep{Okam73,Gupt91}. Hence, our distinct eigenvalues assumption is plausible in practice.
}

{\color{black}
\subsubsection{Proportion of signal variance explained}

The contribution of common latent factors $\{c^{(\ell)}\}_{\ell\in \mathcal{I}_0}$ in generating the signal vector $\bd{x}_k$ of the $k$th data view, 
or the influence of $\{c^{(\ell)}\}_{\ell\in \mathcal{I}_0}$ on $\bd{x}_k$, can be measured by 
\be\label{PVEc set-level}
\PVE_c(\bd{x}_k)=\frac{\tr\{\cov(\bd{c}_k)\}}{\tr\{\cov(\bd{x}_k)\}},
\ee
which is the proportion of $\bd{x}_k$'s variance explained by
their generated common-source vector $\bd{c}_k$.
The influence of distinctive latent factors $\{d_k^{(\ell)}\}_{\ell=1}^{r_f}$ on $\bd{x}_k$ can be quantified 
by 
\be\label{PVEd set-level}
\PVE_d(\bd{x}_k)=1-\PVE_c(\bd{x}_k),
\ee
which is interpreted as the extra proportion
of $\bd{x}_k$'s variance that is explained by adding their generated distinctive-source vector $\bd{d}_k$ \citep{Smil17}.
The above two quantities are the  view-level proportions of signal
variance explained in the $k$th data view.

Similarly, the influences
of $\{c^{(\ell)}\}_{\ell\in \mathcal{I}_0}$ and $\{d_k^{(\ell)}\}_{\ell=1}^{r_f}$ on the signal variable $\bd{x}_k^{[i]}$ can be
assessed by the explained proportions 
\be\label{PVE variable-level}
\PVE_c(\bd{x}_k^{[i]})=\frac{\var(\bd{c}_k^{[i]})}{\var(\bd{x}_k^{[i]})}
\quad\text{and}\quad \PVE_d(\bd{x}_k^{[i]})=1-\PVE_c(\bd{x}_k^{[i]}),
\ee
respectively.
The variable-level proportions of explained signal variance are useful in selecting
variables within each data view that are highly influenced 
by $\{c^{(\ell)}\}_{\ell\in \mathcal{I}_0}$ and $\{d_k^{(\ell)}\}_{\ell=1}^{r_f}$, respectively.
With more easily interpretable feature definitions (e.g., name or location) from original data views, these selected variables are concrete representatives of the common and distinctive latent factors. In contrast, the variables selected by
the sparse GCCA \citep{tenenhaus2014variable,cai2020sparse}
are only linked to the canonical variables, which are merely the most correlated components between the multiple data views,
not linked to their common or distinctive latent factors.
}

{\color{black}
\subsubsection{Additional remarks}\label{sec: t-th level decomp}
%\begin{remark}\label{deficit remark}
Unlike the $K=2$ case,
for $K\ge3$ data views,
it is highly difficult to build a decomposition in the form of~\eqref{decomp in vari} that simultaneously enjoys
both the ideal orthogonality \ref{(O.1)} of distinctive-source vectors $\{\bd{d}_k\}_{k=1}^K$ and a sensible interpretation of the associated 
common-source vectors
$\{\bd{c}_k\}_{k=1}^K$.
We thus relax~\ref{(O.1)} to~\ref{(O.2)}. That is, we  impose~\ref{(O.1)}  on distinctive latent factors $\{d_k^{(\ell)}\}_{k=1}^K$ 
for each $\ell$th stage of GCCA.
This leads to a nice 
interpretation of 
common latent factor $c^{(\ell)}$ given in Remark~\ref{remark1} as the contribution of the principal-basis component $w^{(\ell)}$ made uniformly to generating all signal subspaces $\{\lspan(\bd{x}_k^\top)\}_{k=1}^K$.
Our $\bd{c}_k$ defined
in~\eqref{c_k vec}
is the part of $\bd{x}_k$ that is generated by
$\{c^{(\ell)}\}_{\ell\in \mathcal{I}_0}$.

Due to the relaxation from~\ref{(O.1)} to~\ref{(O.2)},
it is possible that our $\bd{d}_1, \dots, \bd{d}_K$
are all pairwise correlated and thus retain some common underlying source of variation.
In other words, one may continue 
to apply D-GCCA to $\{\bd{d}_k\}_{k=1}^K$ to obtain
their common latent factors.
One solution to fix this issue is to 
uncover these remaining common latent factors
sequentially and treat them hierarchically. 
Specifically, denote $\{(\bd{d}_k^{(0)},\bd{c}_k^{(1)},\bd{d}_k^{(1)})\}_{k=1}^K=\{(\bd{x}_k,\bd{c}_k,\bd{d}_k)\}_{k=1}^K$
and $\{c^{(\ell,1)}\}{\scriptstyle \ell\in \mathcal{I}_0^{(1)}}=\{c^{(\ell)}\}_{\ell\in \mathcal{I}_0}$.
One may iteratively apply D-GCCA to
$\{\bd{d}_k^{(t)}:=\bd{d}_k^{(t-1)}-\bd{c}_k^{(t)}\}_{k=1}^K$
to obtain their common latent factors 
$\{c^{(\ell,t+1)}\}{\scriptstyle \ell\in \mathcal{I}_0^{(t+1)}}$
and common-source random vectors $\{\bd{c}_k^{(t+1)}\}_{k=1}^K$
from $t=1$ up to a given number $T\ge 1$,
or until $\{c^{(\ell,t+1)}\}{\scriptstyle \ell\in \mathcal{I}_0^{(t+1)}}= \emptyset$
or $\PVE_c(\bd{d}_k^{(t)})\prod_{i=0}^{t-1}\PVE_d(\bd{d}_k^{(i)})\le \varepsilon$
for a given tolerance $\varepsilon>0$.
%Here, the dataset-level proportion of $\bd{d}_k^{(t)}$'s variance $\PVE_c(\bd{d}_k^{(t)})=1-\PVE_d(\bd{d}_k^{(t)}):=\tr\{\cov(\bd{c}_k^{(t+1)})\}/\tr\{\cov(\bd{d}_k^{(t)})\}$, and similarly the variable-level proportion of variance $\PVE_c([\bd{d}_k^{(t)}]^{[i]})=1-\PVE_d([\bd{d}_k^{(t)}]^{[i]}):=\var([\bd{c}_k^{(t+1)}]^{[i]})/\var([\bd{d}_k^{(t)}]^{[i]})$.
This iterative procedure yields
a hierarchical decomposition structure for each $\bd{x}_k$,
where $\bd{c}_k^{(t)}$ and $\bd{d}_k^{(t)}$
can be called
the $t$th-level common-source 
and distinctive-source random vectors of $\bd{x}_k$,
and the common and distinctive latent factors of $\{\bd{d}_k^{(t-1)}\}_{k=1}^K$ can be called the $t$th-level common and distinctive latent factors of $\{\bd{x}_k\}_{k=1}^K$.
The importance of the $t$th-level common latent factors
$\{c^{(\ell,t)}\}{\scriptstyle \ell\in \mathcal{I}_0^{(t)}}$ to $\{\bd{x}_k\}_{k=1}^K$ decreases as $t$ increases, because
the more important common latent factors 
are supposed to be uncovered earlier due to Remark~\ref{remark1}.
We thus focus on the first-level decomposition in this paper.
More details about the hierarchical structure are given in 
Appendix~\ref{supp: hierarchical structure}.
%The hierarchical structure is illustrated in Figure~\ref{Fig: D-GCCA high levels}.
Note that when $K=2$ or each $\bd{x}_k$ follows a single-factor model (i.e., $r_k=1$), the first-level distinctive-source vectors $\{\bd{d}_k\}_{k=1}^K$ satisfy~\ref{(O.1)}.

{\color{black}

The difficulties of  imposing \ref{(O.1)} on GCCA for $K\ge3$
views are as follows.
First, the inter-stage orthogonality of canonical variables, the key to realizing \ref{(O.1)} in CCA by D-CCA for $K=2$, may not exist in GCCA for $K\ge3$. Specifically, let $r_1\le \dots\le r_K$,
and augment canonical variables $\{z_k^{(\ell)}\}_{\ell=1}^{r_1}$ of CCA/GCCA with standardized variables $\{z_k^{(\ell)}\}_{r_1<\ell\le r_k}$ and zeros $\{z_k^{(\ell)}\}_{r_k<\ell\le r_K}$
so that $\lspan(\bd{x}_k^\top)=\lspan(\{z_k^{(\ell)}\}_{\ell=1}^{r_K})$.
When $K=2$, CCA satisfies  
the inter-stage orthogonality that
$\lspan(\{z_k^{(i)}\}_{k=1}^K)\perp\lspan(\{z_k^{(j)}\}_{k=1}^K)$ for $1\le i\ne j \le r_K$.
This enables D-CCA to divide
the problem $\bd{x}_k=\bd{c}_k+\bd{d}_k$ $(k\le K)$
into the $r_K$ mutually uncorrelated sub-problems $z_k^{(\ell)}=c^{(\ell)}+d_k^{(\ell)}$ $(k\le K)$, $\ell=1,\dots,r_K$,
and conquer \ref{(O.1)} by imposing \ref{(O.2)}.
Nevertheless, when $K\ge 3$, the inter-stage orthogonality is not guaranteed for GCCA. For example, 
it does not hold for $K=3$ when $\lspan(\bd{x}_k^\top)=\lspan(z_k)$ with $k=1,2$ and $\lspan(\bd{x}_3^\top)=\lspan(\{z_1,z_2\})$
with $z_1,z_2\in \mathcal{L}_0^2$ and $\theta(z_1,z_2)\in (0,\pi/2)$,
even if $\{z_k^{(\ell)}\}_{\ell=1}^{r_K}$ are not canonical variables.
Second,
even under the inter-stage orthogonality, 
\ref{(O.2)} for all $\ell$ does not ensure \ref{(O.1)}. 
For instance, \ref{(O.1)} fails for $K=3$ when $\lspan(\bd{d}_k^\top)=\lspan(\{d_k^{(\ell)}\}_{\ell=1}^2)$ for $k\le 3$,
$\{d_k^{(1)}\}_{k=1}^3\perp \{d_k^{(2)}\}_{k=1}^3$,
$d_1^{(1)}\perp d_2^{(1)} \not\perp d_3^{(1)}  \not\perp d_1^{(1)}$,
and $d_1^{(2)}\not\perp d_2^{(2)}\perp d_3^{(2)}$.
Third, to satisfy \ref{(O.1)},
one may alternatively attempt to design an inner product space
with the subspaces of $\sum_{k=1}^K\lspan(\bd{x}_K^\top)$ as elements and then apply
the Carroll's GCCA~\eqref{GCCA} and our decomposition~\eqref{z=c+d} directly to it for $\ell=1$. 
However, the existence of such an inner product space, particularly with a meaningful geometric interpretation, is unknown.
A close example is the Grassmann algebra of $\sum_{k=1}^K\lspan(\bd{x}_K^\top)$,
which is spanned by the blades algebraically
representing the subspaces of $\sum_{k=1}^K\lspan(\bd{x}_K^\top)$ \citep{MR759340, dorst2010geometric},
but a sum of blades may be a multivector 
not equivalent to a vector space and thus the resulting $\{d_k^{(1)}\}_{k=1}^K$ may not represent $\{\lspan(\bd{d}_k^\top)\}_{k=1}^K$.

}

%Our common-source vectors $\{\bd{c}_k\}_{k=1}^K$ in \eqref{c_k vec} are all generated by the same latent factors $(\bd{c}^{\mathcal{I}_0})^\top=(c^{(\ell)})_{\ell\in \mathcal{I}_0}$.
%As explained in the paragraph before Remark~\ref{remark2}, for $\ell\le L$, $c^{(\ell)}$ is the contribution of the principal-basis component $w^{(\ell)}$ made uniformly to {\color{black}generating} all signal subspaces $\{\lspan(\bd{x}_k^\top)\}_{k=1}^K$.
%Vector $\bd{c}^{\mathcal{I}_0}$ contains these $c^{(\ell)}$s that are nonzero.
%This generative nature of $(\bd{c}^{\mathcal{I}_0})^\top$ indicates that even some part of the common mechanism is possibly retained among $\{\bd{d}_k\}_{k=1}^K$ due to relaxing the latter's desirable orthogonality {\color{black}into each sub-problem~\eqref{z=c+d}},
%it is less important than $(\bd{c}^{\mathcal{I}_0})^\top$ and may be further explored by recursively applying our proposed decomposition. When $K=2$, by Theorem~\ref{equivalence to K=2} below, our D-GCCA decomposition is equivalent to D-CCA, and thus ensures $\lspan(\bd{d}_1^\top)\perp\lspan(\bd{d}_2^\top)$.

%\end{remark}

}

\section{Estimation}\label{sec: estimation}

\subsection{Estimators}\label{subsec: mat est}

We derive the estimators of common-source and distinctive-source matrices $\{\mb{C}_k,\mb{D}_k\}_{k=1}^K$ by starting with the estimation of signal matrices $\{\mb{X}_k\}_{k=1}^K$ from the observed data $\{\mb{Y}_k\}_{k=1}^K$.

Suppose that the low-rank plus noise structure in \eqref{decomp in mat}-\eqref{decomp in vari} follows 
the factor model:
\be\label{factor model}
\mb{Y}_k=\mb{X}_k+\mb{E}_k=\mb{B}_k\mb{F}_k+\mb{E}_k,
\qquad
\bd{y}_k=\bd{x}_k+\bd{e}_k=\mb{B}_k\bd{f}_k+\bd{e}_k,
\ee
where $\mb{B}_k\in \mathbb{R}^{p_k\times r_k}$ is a real deterministic matrix, the columns of $\mb{F}_k$ and $\mb{E}_k$ are, respectively,  the $n$ independent copies of $\bd{f}_k$ and $\bd{e}_k$,
$\bd{f}_k^\top$ is an orthonormal 
basis of $\lspan(\bd{x}_k^\top)$ with $\cov(\bd{f}_k,\bd{e}_k)=\mb{0}_{r_k\times p_k}$,
$\lspan(\bd{x}_k^\top)$ is a fixed space that is independent of $\{p_k\}_{k=1}^K$ and $n$, {\color{black}and $\mb{F}:=[\mb{F}_1;\dots;\mb{F}_K]$ has independent columns}.
We assume
that $\cov(\bd{y}_k)$ is a spiked covariance matrix,  for which
the largest $r_k$ eigenvalues are significantly larger than the rest, namely, signals are distinguishably stronger than noises.
The $r_k$ spiked eigenvalues are majorly contributed by signal $\bd{x}_k$, 
%with a rank-$r_k$ covariance matrix, 
whereas the rest small eigenvalues are induced by noise $\bd{e}_k$.
The spiked covariance model has been widely used in various fields, such as signal processing \citep{Nada10}, machine learning \citep{Huang17}, and economics \citep{Cham83}.

For simplicity, we define the estimators of $\{\mb{X}_k,\mb{C}_k,\mb{D}_k\}_{k=1}^K$
using the true $\{r_k\}_{k=1}^K$, $\mathcal{I}_0$,
$r_k^*=\rank\{\cov(\bd{z}_k^{\mathcal{I}_0})\}$, 
{\color{black}$\mathcal{I}_{\Delta_+}^{(\ell)}=\{(j,k):\Delta_{jk}^{(\ell)}> 0,1\le j<k\le K\}$, 
$\mathcal{I}_{\Delta_0}^{(\ell)}=\{(j,k):\Delta_{jk}^{(\ell)}=0,1\le j<k\le K\}$}, 
and $\sign(\alpha^{(\ell)})$ for all $\ell \in \mathcal{I}_0$.
The practical selection of these {\color{black}nuisance} parameters is discussed in Section~\ref{subsec: rank and set selection}.

We use the following soft-thresholding estimator of $\mb{X}_k$ proposed in \citet{Shu17}. This estimator is originally inspired by the method of \citet{Wang17}
for spiked covariance matrix estimation:
\be\label{X estimator}
\widehat{\mb{X}}_k=\mb{U}_{k1} \diag\{\widehat{\sigma}_1^S(\mb{Y}_k),\ldots, \widehat{\sigma}_{r_k}^S(\mb{Y}_k)\}\mb{U}_{k2}^\top,
\ee
where 
$
\widehat{\sigma}_\ell^S(\mb{Y}_k)=[\max\{\sigma_\ell^2(\mb{Y}_k)- \tau_kp_k,0\}]^{1/2},
$
$
\tau_k=\sum_{\ell=r_k+1}^{p_k} \sigma_\ell^2(\mb{Y}_k)/(np_k-nr_k-p_kr_k),
$ 
and $\mb{U}_{k1}\diag\{\sigma_1(\mb{Y}_k),\dots,\sigma_{r_k}(\mb{Y}_k)\}\mb{U}_{k2}^\top$ is 
the top-$r_k$ singular value decomposition (SVD)~of~$\mb{Y}_k$. 

%Under Assumption~\ref{assump1} given later, $\widehat{\mb{X}}_k$ is rank-$r_k$ with probability tending to 1.

We next use $\widehat{\mb{X}}_k$ to develop estimators for $\mb{C}_k$ and $\mb{D}_k=\mb{X}_k-\mb{C}_k$.
{\color{black}Recall from \eqref{c_k vec} that we have 
the random variable $\bd{c}_k=\cov(\bd{x}_k,\bd{z}_k^{\mathcal{I}_0})\{\cov(\bd{z}_k^{\mathcal{I}_0})\}^\dag\bd{c}^{\mathcal{I}_0}$.

We begin with the estimation of $\cov(\bd{x}_k,\bd{z}_k^{\mathcal{I}_0})$}.
Define an estimator of $\cov(\bd{x}_k)$ by
$\widehat{\cov}(\bd{x}_k)=\widehat{\mb{X}}_k\widehat{\mb{X}}_k^\top/n$ whose SVD is denoted as
$\widehat{\cov}(\bd{x}_k)=\widehat{\mb{V}}_{xk}\widehat{\mb{\Lambda}}_{xk}\widehat{\mb{V}}_{xk}^\top$, where 
$\widehat{\mb{\Lambda}}_{xk}=\diag\{
\lambda_1(\widehat{\cov}(\bd{x}_k)),\dots,
\lambda_{r_k}(\widehat{\cov}(\bd{x}_k))\}$ 
and $\widehat{\mb{V}}_{xk}$ has $r_k$ orthonormal columns. {\color{black}We can obtain
$\lambda_\ell(\widehat{\cov}(\bd{x}_k))=[\widehat{\sigma}_\ell^S(\mb{Y}_k)]^2/n$
and $\widehat{\mb{V}}_{xk}=\mb{U}_{k1}$.} 
Define the estimators of $\mb{F}_k$ and $\mb{F}$ by
$\widehat{\mb{F}}_k=(\widehat{\mb{\Lambda}}_{xk}^{1/2})^\dag\widehat{\mb{V}}_{xk}^\top\widehat{\mb{X}}_k$ and
$\widehat{\mb{F}}=[\widehat{\mb{F}}_1;\dots;\widehat{\mb{F}}_K]$, respectively.
We estimate $\cov(\bd{f})$ by
$\widehat{\cov}(\bd{f})=\widehat{\mb{F}}\widehat{\mb{F}}^\top/n$.
Let $\widehat{\bd{\eta}}^{(\ell)}=[\widehat{\bd{\eta}}_1^{(\ell)};\dots;\widehat{\bd{\eta}}_K^{(\ell)}]$, with $\widehat{\bd{\eta}}_k^{(\ell)}\in \mathbb{R}^{r_k}$, be a normalized eigenvector of $\widehat{\cov}(\bd{f})$
corresponding to $\lambda_\ell(\widehat{\cov}(\bd{f}))$.
We also let different $\widehat{\bd{\eta}}^{(\ell)}$s be orthogonal.
Our estimated sample vector of the variable $w^{(\ell)}$
{\color{black}in \eqref{w formula}} is defined by
\be\label{w estimator}
{\color{black}
(\widehat{\bd{w}}^{(\ell)})^\top
= \begin{cases}
[\lambda_{\ell}(\widehat{\cov}(\bd{f}))]^{-1/2}(\widehat{\bd{\eta}}^{(\ell)})^\top\widehat{\mb{F}}, & ~~\text{if}~ \lambda_{\ell}(\widehat{\cov}(\bd{f}))\ne 0,\\
\bd{0}_{1\times n},&~~\text{otherwise},
\end{cases}}
\ee
and that of the variable $z_k^{(\ell)}$ {\color{black}in \eqref{z_k revised}} is $(\widehat{\bd{z}}_k^{(\ell)})^\top=(\widehat{\bd{\eta}}_k^{(\ell)}/\|\widehat{\bd{\eta}}_k^{(\ell)}\|_F)^\top\widehat{\mb{F}}_k$ if $\|\widehat{\bd{\eta}}_k^{(\ell)}\|_F\ne 0$
and otherwise $\widehat{\bd{z}}_k^{(\ell)}=\bd{0}_{n\times 1}$.
{\color{black}
Then, $\cov(\bd{x}_k,\bd{z}_k^{\mathcal{I}_0})$ is estimated by
\be\label{C 1st comp}
\widehat{\cov}(\bd{x}_k,\bd{z}_k^{\mathcal{I}_0})
=n^{-1}\widehat{\mb{X}}_k(\widehat{\bd{z}}_k^{(\ell)})_{\ell\in \mathcal{I}_0}=
\widehat{\mb{V}}_{xk}\widehat{\mb{\Lambda}}_{xk}^{1/2}{\widehat{\mb{H}}_k}^\top,
\ee
where $\widehat{\mb{H}}_k=(\widehat{\bd{\eta}}_k^{(\ell)}/\|\widehat{\bd{\eta}}_k^{(\ell)}\|_F)_{\ell\in \mathcal{I}_0}^\top$ with $\bd{0}/0:=\bd{0}$.
}

The matrix $\cov(\bd{z}_k^{\mathcal{I}_0})$ is initially estimated by
%an initial covariance matrix estimator of $\bd{z}_k^{\mathcal{I}_0}=(z_k^{(\ell)})_{\ell\in \mathcal{I}_0}^\top$ by 
$\widetilde{\cov}(\bd{z}_k^{\mathcal{I}_0})=\widehat{\mb{H}}_k\widehat{\mb{H}}_k^\top$.
Let $\widetilde{\cov}(\bd{z}_k^{\mathcal{I}_0})=\widehat{\mb{V}}_{zk}\widehat{\mb{\Lambda}}_{zk}  \widehat{\mb{V}}_{zk}^\top $ be
its compact SVD, where $\widehat{\mb{\Lambda}}_{zk}$ has nonincreasing diagonal elements. %of $\widetilde{\cov}(\bd{z}_k^{\mathcal{I}_0})$.
With $\widecheck{r}_k:=\min(r_k^*,\rank\{\widetilde{\cov}(\bd{z}_k^{\mathcal{I}_0})\})$,
our %final 
estimator of $\cov(\bd{z}_k^{\mathcal{I}_0})$
is defined by the top-$\widecheck{r}_k$ SVD of $\widetilde{\cov}(\bd{z}_k^{\mathcal{I}_0})$ as
\be\label{est cov z}
\widehat{\cov}(\bd{z}_k^{\mathcal{I}_0})=\widehat{\mb{V}}_{zk}^{[:,1:\widecheck{r}_k]} \widehat{\mb{\Lambda}}_{zk}^{[1:\widecheck{r}_k,1:\widecheck{r}_k]}  (\widehat{\mb{V}}_{zk}^{[:,1:\widecheck{r}_k]})^\top.
\ee

{\color{black} To approximate the sample matrix $\mb{C}^{\mathcal{I}_0}$ of latent factors $\bd{c}^{\mathcal{I}_0}=(c^{(\ell)})_{\ell\in \mathcal{I}_0}^\top=(\alpha^{(\ell)}w^{(\ell)})_{\ell\in \mathcal{I}_0}^\top$, the key is the estimation 
of the value $\alpha^{(\ell)}$ given in Theorem~\ref{alpha thm}.}
Replacing $\cos\{\theta(w^{(\ell)},z_k^{(\ell)})\}$ 
and $\cos\{\theta(z_j^{(\ell)},z_k^{(\ell)})\}$ by 
$\widehat{\cos}\{\theta(w^{(\ell)},z_k^{(\ell)})\}= (\widehat{\bd{w}}^{(\ell)})^\top \widehat{\bd{z}}_k^{(\ell)}/n$
and  
$\widehat{\cos}\{\theta(z_j^{(\ell)},z_k^{(\ell)})\}=
(\widehat{\bd{z}}_j^{(\ell)})^\top \widehat{\bd{z}}_k^{(\ell)}/n
$
in $\Delta_{jk}^{(\ell)}$
yields its initial estimator $\widetilde{\Delta}_{jk}^{(\ell)}$.
For $(j,k)\in {\color{black}\mathcal{I}_{\Delta_+}^{(\ell)}\cup \mathcal{I}_{\Delta_0}^{(\ell)}}$, define 
\be\label{alpha estimator}
\widehat{\alpha}_{jk}^{(\ell)} 
=\frac{1}{2}\left[
\widehat{\cos}\{\theta(w^{(\ell)},z_j^{(\ell)})\}
+
\widehat{\cos}\{\theta(w^{(\ell)},z_k^{(\ell)})\}
-
(
\widehat{\Delta}_{jk}^{(\ell)}
)^{1/2}     \right]
\ee
where $\widehat{\Delta}_{jk}^{(\ell)}=\max(\widetilde{\Delta}_{jk}^{(\ell)},0){\color{black}[(j,k)\in \mathcal{I}_{\Delta_+}^{(\ell)}]}     $ {\color{black} with $[\cdot]$ being the Iverson bracket}.
For $\ell\in \mathcal{I}_0$, we define
\[
\widehat{\alpha}^{(\ell)}=\argmin_{\widehat{\alpha}_{jk}^{(\ell)}}\left\{ |\widehat{\alpha}_{jk}^{(\ell)}|: \widehat{\alpha}_{jk}^{(\ell)}\sign(\alpha^{(\ell)})>0, (j,k)\in {\color{black}\mathcal{I}_{\Delta_+}^{(\ell)}\cup \mathcal{I}_{\Delta_0}^{(\ell)}} \right\}.
\]
{\color{black} Then, $\mb{C}^{\mathcal{I}_0}$ is estimated
with $\widehat{\bd{w}}^{(\ell)}$ in \eqref{w estimator}
by
\be\label{latent C sample est}
\widehat{\mb{C}}^{\mathcal{I}_0}=(\widehat{\alpha}^{(\ell)}\widehat{\bd{w}}^{(\ell)})_{\ell\in\mathcal{I}_0}^\top.
\ee

Combining \eqref{C 1st comp}, \eqref{est cov z} and \eqref{latent C sample est} yields our estimator of  the common-source matrix~$\mb{C}_k$:
} 
\be\label{C estimator}
\widehat{\mb{C}}_k=\widehat{\cov}(\bd{x}_k,\bd{z}_k^{\mathcal{I}_0})\{\widehat{\cov}(\bd{z}_k^{\mathcal{I}_0})\}^\dag \widehat{\mb{C}}^{\mathcal{I}_0}.
\ee
Our estimator of the distinctive-source matrix $\mb{D}_k$ is defined by 
\be\label{D estimator}
\widehat{\mb{D}}_k=\widehat{\mb{X}}_k-\widehat{\mb{C}}_k.
\ee

{\color{black}The major time cost of proposed matrix estimators
comes from the SVD of each $\mb{Y}_k$ with complexity $O(\min\{np_k^2,n^2p_k\})$.
}

{\color{black} 
We define the estimators
for the view-level and the variable-level
proportions of explained signal variance
$\PVE_c(\bd{x}_k)=1-\PVE_d(\bd{x}_k)$ and $\PVE_c(\bd{x}_k^{[i]}) =1-\PVE_d(\bd{x}_k^{[i]})$
by
\begin{align}
\widehat{\PVE}_c(\bd{x}_k)&=1-\widehat{\PVE}_d(\bd{x}_k)=
\|\widehat{\mb{C}}_k \|_F^2/\|\widehat{\mb{X}}_k \|_F^2,\\
\widehat{\PVE}_c(\bd{x}_k^{[i]})&=1-\widehat{\PVE}_d(\bd{x}_k^{[i]})=
\|\widehat{\mb{C}}_k^{[i,:]} \|_F^2/\|\widehat{\mb{X}}_k^{[i,:]}  \|_F^2.
\label{PVEvar est}
\end{align}

}

\subsection{Asymptotic properties}
We introduce an assumption used in \citet{Wang17} and \citet{Shu17}.

\begin{assump}\label{assump1}
	We assume the following conditions for  model \eqref{factor model}.
	\vspace{-0.1cm}
	\begin{enumerate}[label=(\roman*),font=\upshape]
		\item\label{assump1(i)}	Let $\lambda_{k,1}>\cdots>\lambda_{k,r_k}>\lambda_{k,r_k+1}\ge \cdots\ge \lambda_{k,p_k}>0$ be
		the eigenvalues of $\cov(\bd{y}_k)$. There exist positive constants $\kappa_1,\kappa_2$ and $\delta_0$ such that $\kappa_1\le \lambda_{k,\ell}\le \kappa_2$ for $\ell>r_k$ and  $\min_{\ell\le r_k} (\lambda_{k,\ell}-\lambda_{k,\ell+1})/\lambda_{k,\ell}\ge \delta_0$.
		
		\item\label{assump1(ii)} Assume that $p_k>\kappa_0 n$ with a constant $\kappa_0>0$. When $n\to \infty$, assume $\lambda_{k,r_k}\to \infty$,
		$p_k/(n\lambda_{k,\ell})$ is upper bounded for $\ell\le r_k$, $\lambda_{k,1}/\lambda_{k,r_k}$ is bounded from above and below,
		and $p_k^{1/2}(\log n)^{1/\gamma_{k2}}=o(\lambda_{k,r_k})$ with $\gamma_{k2}$ given in \ref{assump1(v)}.
		
		\item\label{assump1(iii)} The columns of $\mb{Z}_{y_k}= \mb{\Lambda}_{y_k}^{-1/2}\mb{V}_{y_k}^\top \mb{Y}_k$ are independent copies of 
		$\bd{z}_{y_k}= \mb{\Lambda}_{y_k}^{-1/2}\mb{V}_{y_k}^\top \bd{y}_k$, where $\mb{V}_{y_k}\mb{\Lambda}_{y_k}\mb{V}_{y_k}^\top$ is the full SVD of $\cov(\bd{y}_k)$ with
		$\mb{\Lambda}_{y_k}=\diag(\lambda_{k,1},\ldots,\lambda_{k,p_k})$.
		Vector $\bd{z}_{y_k}$'s entries $\{z_{y_k}^{[i]}\}_{i=1}^{p_k}$ are independent with $E(z_{y_k}^{[i]})=0$, $\var(z_{y_k}^{[i]})=1$, and the sub-Gaussian norm $\sup_{q\ge 1}q^{-1/2}[E(|z_{y_k}^{[i]}|^q)]^{1/q}\le \kappa_s$ 
		with a constant $\kappa_s>0$ for all $i\le p_k$.
		
		\item\label{assump1(iv)} Matrix $\mb{B}_k^\top\mb{B}_k$ is a diagonal matrix.
		For all $i\le p_k$ and $\ell\le r_k$,
		$| \mb{B}_k^{[i,\ell]}  |\le \kappa_B(\lambda_{k,\ell}/p_k)^{1/2}$ with a constant $\kappa_B>0$.
		
		\item\label{assump1(v)} Denote $\bd{e}_k=(e_{k,1},\ldots,e_{k,p_k})^\top$ and $\bd{f}_k=(f_{k,1},\ldots, f_{k,r_k})^\top$. Let
		$\| \cov(\bd{e}_k) \|_\infty<s_0$ with a constant $s_0>0$.
		For all $i\le p_k$ and $\ell\le r_k$,  there exist positive constants $\gamma_{k1},\gamma_{k2},b_{k1}$ and $b_{k2}$ such that for $t>0$,
		$
		P(|e_{k,i}|>t)\le \exp\{-(t/b_{k1})^{\gamma_{k1}}\}
		$
		and
		$
		P(|f_{k,\ell}|>t)\le \exp\{-(t/b_{k2})^{\gamma_{k2}}\}.
		$
	\end{enumerate}	
\end{assump}

%Assumption~\ref{assump2} holds for the general case when $\bd{f}$ is independent of $n$ and $\{p_k\}_{k=1}^K$.

{\color{black} Assumption~\ref{assump1} follows Assumptions 2.1-2.3 and 4.1-4.2 of \citet{Wang17} which guarantee the consistency of
each signal estimator $\widehat{\mb{X}}_k$ given in \eqref{X estimator}. 
The diverging leading eigenvalues and bounded nonspiked eigenvalues of $\cov(\bd{y}_k)$
in conditions~\ref{assump1(i)} and~\ref{assump1(ii)}, together with the approximate sparsity constraint $\|\cov(\bd{e}_k) \|_\infty<s_0$ in condition \ref{assump1(v)}, ensure sufficiently strong signals for thresholding. 
These conditions are common in the literature of high-dimensional factor models \citep{bai2003inferential, bai2008large, Fan13}.
The sub-Gaussian constraint in~\ref{assump1(iii)} and the exponential-type tails in~\ref{assump1(v)} generalize
 Gaussian distributions, while  allowing the use of  
 the large deviation theory to establish concentration bounds.
For condition~\ref{assump1(iv)}, letting $\bd{f}_k=\mb{\Lambda}_{xk}^{-1/2}\mb{V}_{xk}^\top\bd{x}_k$
with the compact SVD  $\cov(\bd{x}_k)=\mb{V}_{xk}\mb{\Lambda}_{xk}\mb{V}_{xk}^\top$, we have
  $\mb{B}_k=\cov(\bd{x}_k,\bd{f}_k)=\mb{V}_{xk}\mb{\Lambda}_{xk}^{1/2}$.
Hence, $\mb{B}_k^\top\mb{B}_k=\mb{\Lambda}_{xk}$ is a diagonal matrix. Then, it follows from Weyl's inequality that  $\max_{\ell \le r_k}\| \mb{B}_k^{[:,\ell]}\|_F^2/\lambda_{k,\ell}\le 1+\|\cov(\bd{e}_k)\|_2/\lambda_{k,r_k}=1+o(1)$.
It is thus reasonable to assume $|\mb{B}_k^{[i,\ell]}|=O(\sqrt{\lambda_{k,\ell}/p_k})$. See \citet{Wang17} and \citet{Shu17} for more discussions on Assumption~\ref{assump1}.
}

We have the following asymptotic properties for estimators defined
in \eqref{X estimator} and \eqref{C estimator}-\eqref{PVEvar est}. 

\begin{thm}\label{consistency thm}
	Suppose that Assumption~1 holds and true $\{r_k\}_{k=1}^K$ are given.
	Then for each $k\le K$,  we have 
	\[
	\frac{ \|\widehat{\mb{X}}_k-\mb{X}_k\|_{\star}^2}{\|\mb{X}_k \|_{\star}^2}
	=O_P\Big(
	\min\Big\{
	\frac{1}{n^2}+\frac{\log p_k}{n\SNR_k},1
	\Big\}
	\Big),
	\]
	where $\| \cdot  \|_{\star}$ denotes either the Frobenius norm or the spectral norm and {\color{black}$\SNR_k=\frac{\tr\{\cov(\bd{x}_k)\}}{\tr\{\cov(\bd{e}_k)\}}$} is the signal-to-noise ratio of $\bd{y}_k$.
	Additionally assume that $K$ is a constant, $\mathcal{I}_0\ne \emptyset$,  $\{\lambda_\ell(\cov(\bd{f}))\}_{\ell=1}^L$ are distinct,
	and  true
	$\big\{\mathcal{I}_0,\{r_k^*\}_{k=1}^K, \{{\color{black}\mathcal{I}_{\Delta_+}^{(\ell)},\mathcal{I}_{\Delta_0}^{(\ell)}},\sign(\alpha^{(\ell)})\}_{\ell\in \mathcal{I}_0}\big\}$ are given.
	If
	$
	\delta_\eta:=\frac{1}{\sqrt{n}}
	+	\sum_{k=1}^K \sqrt{\frac{\log p_k}{n\SNR_k}}=o(1) 
	$,
	then 
	\be\label{RNE of C and D}
	\max\left\{ \frac{\|\widehat{\mb{C}}_k-\mb{C}_k\|_{\star}^2}{\|\mb{X}_k \|_{\star}^2}
	,\frac{\|\widehat{\mb{D}}_k-\mb{D}_k\|_{\star}^2}{\|\mb{X}_k \|_{\star}^2} \right\}
	=O_P({\color{black}\delta_\eta^2})
	\ee
and
	\be\label{eq: dataset-level PVE}
{\color{black}\left|\widehat{\PVE}_c(\bd{x}_k)-\PVE_c(\bd{x}_k)\right|
	=O_P(\delta_\eta).}
	\ee
{\color{black} Furthermore, if $\delta_k:=(1+\frac{1}{\SNR_k})\sqrt{\frac{\log p_k}{n}}=o(1)$
and 
$\min_{ i\le p_k}\var(\bd{x}_k^{[i]})\ge M_k
\lambda_{r_k}(\cov(\bd{x}_k))/p_k$ 
with a constant $M_k>0$,
then  we have 
\be\label{eq: variable-level PVE}
\max_{1\le i\le p_k}\left|\widehat{\PVE}_c(\bd{x}_k^{[i]})-\PVE_c(\bd{x}_k^{[i]})\right|=O_P(\delta_\eta+\delta_k).
\ee
}
\end{thm}

%When $K=2$,  
%by Lemma~2 of \citet{Kett71}, the error bounds of $\widehat{\mb{C}}_k$ and $\widehat{\mb{D}}_k$ in~\eqref{RNE of C and D} are equivalent to those in Theorem~3 of the D-CCA paper \citep{Shu17}.
{\color{black} Under Assumption~\ref{assump1}, the signal-to-noise ratio $\SNR_k\asymp \lambda_{k,1}/p_k$. 
For pervasive factor models that have
leading eigenvalues $\lambda_{k,\ell}\asymp p_k$ for $\ell\le r_k$
\citep{Fan13, Wang17}, we have $\SNR_k\asymp 1$, and thus
 $\delta_\eta\asymp \sum_{k=1}^K\sqrt{(\log p_k)/n}$
and $\delta_k\asymp \sqrt{(\log p_k)/n}$.
It is commonly assumed that $\sqrt{(\log p_k)/n}=o(1)$ in the literature of high-dimensional statistics \citep{bickel2008covariance, rothman2009generalized}. 
Hence,
$\delta_\eta=o(1)=\delta_k$ 
holds at least for pervasive factor models.
Note that $\sum_{i=1}^{p_k}\var(\bd{x}_k^{[i]})=\tr(\cov(\bd{x}_k))
=\sum_{\ell=1}^{r_k}\lambda_{\ell}(\cov(\bd{x}_k))$.
Thus, it is reasonable to 
assume
$\min_{i\le p_k}\var(\bd{x}_k^{[i]})\ge M_k
\lambda_{r_k}(\cov(\bd{x}_k))/p_k$.

When $K=2$, the convergence rates of $\widehat{\mb{C}}_k$ and $\widehat{\mb{D}}_k$ in \eqref{RNE of C and D} are
faster than those in Theorem~3 of the D-CCA paper \citep{Shu17}.
This benefits from the predetermination of the nuisance parameters $\{\mathcal{I}_{\Delta_+}^{(\ell)},\mathcal{I}_{\Delta_0}^{(\ell)}\}_{\ell\in\mathcal{I}_0}$ (e.g., by the approach in the next subsection). 
The same convergence rates can be obtained in the proof of \citet{Shu17} if $\max\{\ell:\sigma_\ell(\cov(\bd{x}_1,\bd{x}_2))=1\}
=r_1+r_2-\rank\{\cov([\bd{x}_1;\bd{x}_2])\}
$ is predetermined (e.g., by the two-step test of \citet{Chen19}).
To the best of our knowledge, 
the results 
in \eqref{eq: dataset-level PVE}-\eqref{eq: variable-level PVE}
are the first work to show the high-dimensional estimation consistency
of the view-level and variable-level
proportions of explained signal variance
for the decomposition model in \eqref{decomp in mat}-\eqref{decomp in vari} for $K\ge 2$, which are not seen in \citet{Shu17} even when $K=2$.

The convergence rates of our estimators in \eqref{RNE of C and D}--\eqref{eq: variable-level PVE} depend on the information of all the $K$ data views and decrease when $K$ increases.
The estimation involves denoising all signals $\{\mb{X}_k\}_{k=1}^K$ and accumulates the estimation errors of $\{\mb{X}_k\}_{k=1}^K$,
since the common latent factors generally change as $K$ increases and thus cannot be determined by fewer data views.
For example, consider the case where $\bd{x}_k=z_k^{(1)}$ ($k=1,\dots,K$) with 
$\corr(z_i^{(1)},z_j^{(1)})=\rho\in(0,1)$ for $1\le i\ne j\le K$.
As used in AJIVE, COBE and our D-GCCA,
the desirable direction variable of the common latent factor
is the consensus variable  $w^{(1)}$ of GCCA, which is
$w^{(1)}=\sum_{k=1}^2z_k^{(1)}/\sqrt{2+2\rho}$ when $K=2$, and
$w^{(1)}=(\sum_{k=1}^3z_k^{(1)})/\sqrt{3+6\rho}$ when $K=3$,
thereby resulting in different common latent factors for $K=2$ and $K=3$.

}

\subsection{Selection of {\color{black}nuisance} parameters}\label{subsec: rank and set selection}
We discuss how to practically select the parameters $\{r_k\}_{k=1}^K$, $\mathcal{I}_0$, $\{r_k^*\}_{k=1}^K$, $\{\mathcal{I}_{\Delta_+}^{(\ell)},\mathcal{I}_{\Delta_0}^{(\ell)}\}_{\ell\in\mathcal{I}_0}$, and $\{\sign(\alpha^{(\ell)})\}_{\ell\in \mathcal{I}_0}$.
%{\color{black}The nuisance parameters of higher-level decomposition are selected in the same fashion.}
Denote $\widehat{r}_k, \widehat{L}$, $\widehat{\mathcal{I}}_0$, $\widehat{r}_k^*$, $\widehat{\mathcal{I}}_{\Delta_+}^{(\ell)}$, $\widehat{\mathcal{I}}_{\Delta_0}^{(\ell)}$, and $\widehat{\sign}(\alpha^{(\ell)})$ to be estimators of their true counterparts. 

We select $\{\widehat{r}_k\}_{k=1}^K$ 
by using the edge distribution method of \citet{Onat10}
that consistently estimates the rank for the factor model in \eqref{factor model} under mild conditions.
To determine the other parameters,
we use hypothesis tests based on the denoised data $\{\widehat{\mb{X}}_k\}_{k=1}^K$.
Testing procedures have been widely used in the literature of CCA \citep{Bart41,Lawl59,Cali05,Song16} to select similar parameters.

Consider the selection of $L=\max\{\ell\in \{1,\dots,r_f\}:\lambda_\ell(\cov(\bd{f}))> 1\}$.
Left-multiplying  the both sides of $[\cov(\bd{f})\bd{\eta}^{(\ell)}]^{[\sum_{i=0}^{k-1}r_i:\sum_{i=1}^{k}r_i]}=[\lambda_\ell(\cov(\bd{f}))\bd{\eta}^{(\ell)}]^{[\sum_{i=0}^{k-1}r_i:\sum_{i=1}^{k}r_i]}$ by $\bd{\eta}_k^{(\ell)}$
can obtain $\cov\big((\bd{\eta}_k^{(\ell)})^\top\bd{f}_k,\sum_{j\ne k}(\bd{\eta}_j^{(\ell)})^\top\bd{f}_j\big)=\left[\lambda_\ell(\cov(\bd{f}))-1\right]\|\bd{\eta}_k^{(\ell)}\|_F^2$ for all $k\le K$.
Let $\widehat{L}$ be the largest $\ell\in [0,\rank\{\widehat{\cov}(\bd{f})\}]$ such that 
for at least one $k$, both $\corr(w^{(\ell)}, z_k^{(\ell)})=0$ and $\corr\big((\bd{\eta}_k^{(\ell)})^\top\bd{f}_k,\sum_{j\ne k}(\bd{\eta}_j^{(\ell)})^\top\bd{f}_j\big)=0$ are rejected by a right-tailed test for zero correlation.
The two tests indicate $\|\bd{\eta}_k^{(\ell)}\|_F\ne 0$
and $\cov\big((\bd{\eta}_k^{(\ell)})^\top\bd{f}_k,\sum_{j\ne k}(\bd{\eta}_j^{(\ell)})^\top\bd{f}_j\big)>0$, respectively, thereby implying 
$\lambda_\ell(\cov(\bd{f}))-1=\cov\big((\bd{\eta}_k^{(\ell)})^\top\bd{f}_k,\sum_{j\ne k}(\bd{\eta}_j^{(\ell)})^\top\bd{f}_j\big)\big/\|\bd{\eta}_k^{(\ell)}\|_F^2>0$.
We use
the normal approximation test of \citet{DiCi17} for testing zero correlation.  

To determine $\mathcal{I}_0=\{\ell\in\{1,\dots,L\}:\alpha^{(\ell)}\ne 0\}$,
we retain index $\ell\le \widehat{L}$ in $\widehat{\mathcal{I}}_0$ if 
$\corr(w^{(\ell)}, z_k^{(\ell)})$
$=0$ and $\corr(z_j^{(\ell)}, z_k^{(\ell)})=0$ are rejected  respectively by the right-tailed and the two-tailed zero-correlation tests for all $k\le K$ and all $j\ne k$.

{\color{black}
The rank estimate $\widehat{r}_k^*$ of $r_k^*=\rank\{\cov(z_k^{\mathcal{I}_0})\}$ is obtained by the two-step test of \citet{Chen19} for the rank of matrix $\cov(z_k^{\widehat{\mathcal{I}}_0})$.
}

{\color{black} We next select $\mathcal{I}_{\Delta_+}^{(\ell)}=\{(j,k):\Delta_{jk}^{(\ell)}> 0,1\le j<k\le K\}$
and $\mathcal{I}_{\Delta_0}^{(\ell)}=\{(j,k):\Delta_{jk}^{(\ell)}= 0,1\le j<k\le K\}$. Note that
$\Delta_{jk}^{(\ell)}=-4\cov(z_{j,k}^{(\ell)},z_{k,j}^{(\ell)})$
with $z_{j,k}^{(\ell)}=z_j^{(\ell)}-\frac{1}{2}[\cos\{\theta(w^{(\ell)},z_j^{(\ell)})\}
+
\cos\{\theta(w^{(\ell)},z_k^{(\ell)})\}]w^{(\ell)}$.
For $\ell\in \widehat{\mathcal{I}}_0$,
we include $(j,k)$ into $\widehat{\mathcal{I}}_{\Delta_+}^{(\ell)}$ if $\corr(z_{j,k}^{(\ell)},z_{k,j}^{(\ell)})=0$ is rejected by the left-tailed zero-correlation test, and then include the remaining $(j,k)$ into $\widehat{\mathcal{I}}_{\Delta_0}^{(\ell)}$ if $\corr(z_{j,k}^{(\ell)},z_{k,j}^{(\ell)})=0$ is not rejected by the right-tailed zero-correlation test.
}

\iffalse
We next select $\mathcal{I}_\Delta^{(\ell)}=\{(j,k):\Delta_{jk}^{(\ell)}\ge 0,1\le j<k\le K\}$.
An equivalent formula of $\Delta_{jk}^{(\ell)}=0$ is $\cos\theta(z_{j,k}^{(\ell)},z_{k,j}^{(\ell)})= 0$ with
$z_{j,k}^{(\ell)}=z_j^{(\ell)}-\frac{1}{2}[\cos\theta(w^{(\ell)},z_j^{(\ell)})
+
\cos\theta(w^{(\ell)},z_k^{(\ell)})]w^{(\ell)}$.
For $\ell\in \widehat{\mathcal{I}}_0$,
we exclude $(j,k)$ from $\widehat{\mathcal{I}}_\Delta^{(\ell)}$ if 
$\widetilde{\Delta}_{jk}^{(\ell)}<0$ and meanwhile $\corr(z_{j,k}^{(\ell)},z_{k,j}^{(\ell)})=0$ is rejected by the two-tailed zero-correlation test.

{\color{black}
\begin{align*}
\cov(z_{j,k}^{(\ell)},z_{k,j}^{(\ell)})
&=\cov(z_j^{(\ell)}-\frac{1}{2}[\cos\theta(w^{(\ell)},z_j^{(\ell)})
+
\cos\theta(w^{(\ell)},z_k^{(\ell)})]w^{(\ell)},\\
&~~~~~~~~~~~ z_k^{(\ell)}-\frac{1}{2}[\cos\theta(w^{(\ell)},z_j^{(\ell)})
+
\cos\theta(w^{(\ell)},z_k^{(\ell)})]w^{(\ell)})\\
&=\cov(z_j^{(\ell)},z_k^{(\ell)})
-\frac{1}{4}[\cos\theta(w^{(\ell)},z_j^{(\ell)})
+
\cos\theta(w^{(\ell)},z_k^{(\ell)})]^2\\
&=\cos\theta(z_j^{(\ell)},z_k^{(\ell)})
-\frac{1}{4}[\cos\theta(w^{(\ell)},z_j^{(\ell)})
+
\cos\theta(w^{(\ell)},z_k^{(\ell)})]^2\\
&=-\frac{1}{4}\Delta_{jk}^{(\ell)}
\end{align*}
}
\fi

Finally, consider to determine the sign of $\alpha^{(\ell)}$.
Define 
$\alpha^{(\ell)}_+=\min \{\alpha_{jk}^{(\ell)}:\alpha_{jk}^{(\ell)}>0,(j,k)\in \widehat{\mathcal{I}}_\Delta^{(\ell)} \}$
and
$\alpha^{(\ell)}_-=\max \{\alpha_{jk}^{(\ell)}:\alpha_{jk}^{(\ell)}<0,(j,k)\in \widehat{\mathcal{I}}_\Delta^{(\ell)}\}$,
and define $\widehat{\alpha}^{(\ell)}_+$ and $\widehat{\alpha}^{(\ell)}_-$ in the same way by using $\widehat{\alpha}_{jk}^{(\ell)}$ instead.
Let $\widehat{\sign}(\alpha^{(\ell)})$ be the sign of the existing one of $\widehat{\alpha}^{(\ell)}_+$ and $\widehat{\alpha}^{(\ell)}_-$ if the other does not exist.
Otherwise, first
test $|\alpha^{(\ell)}_+|-|\alpha^{(\ell)}_-|=0$
by applying the bias-corrected and accelerated bootstrap interval \citep{Efro93}.
Let $\widehat{\sign}(\alpha^{(\ell)})=1$ if zero is outside the bootstrap interval and $|\widehat{\alpha}^{(\ell)}_+|<|\widehat{\alpha}^{(\ell)}_-| $, 
and otherwise let
$\widehat{\sign}(\alpha^{(\ell)})=-1$.

\section{Simulation studies}\label{Sec: simulation}
%\subsection{Simulations}\label{subsec: siml}
{\color{black}
In this section, we evaluate the finite-sample performance of proposed D-GCCA estimation via simulations, comparing with the six competing methods mentioned in Section~\ref{sec: intro}.

\subsection{Simulation setups}
We consider $K=3$ data views with signals $\{\bd{x}_k\}_{k=1}^3$ following the four simulation setups below, and generate signal-independent Gaussian noises $\{e_{k,i}\}_{i=1}^{p_k}\overset{iid}{\sim}N(0,\sigma_{e_k}^2)$ that are independent across data views. Simulations are conducted with sample size $n=300$, variable dimension $p_1$ ranging from 100 to 1500, noise variance $\sigma_{e_1}^2$ from 0.25 to 9, and 1000 independent replications under each setting.

\begin{itemize}[leftmargin=*]
\item Setup 1.1: Let $\bd{x}_1\overset{d}{=}\bd{x}_2\overset{d}{=}\bd{x}_3$ and $r_1=r_2=r_3=1$. Set standardized canonical variables $z_1^{(1)},z_2^{(1)},z_3^{(1)}$ to be jointly Gaussian with $\theta_z:=\theta(z_j^{(1)},z_k^{(1)})$ for all $j\ne k $. 
Let $\mb{\Lambda}_k=500$ for $k=1,2,3$. Randomly generate three unit vectors $\{\mb{V}_{xk}\}_{k=1}^3$ that are equal if with the same size and are fixed for all simulation replications of the same $(p_1,p_2,p_3)$. 
Let $\bd{x}_k=\mb{V}_{xk}\mb{\Lambda}_{xk}^{1/2}z_k^{(1)}$.
We vary $\theta_z$ from $10^\circ$ to $70^\circ$, resulting in
D-GCCA's $\{\PVE_c(\bd{x}_k)\}_{k=1}^3$ all from $0.853$ to $0.079$ invariant to $\{p_k\}_{k=1}^3$; see Appendix~\ref{sec: add simul}. Let $\sigma_{e_1}^2=\sigma_{e_2}^2=\sigma_{e_3}^2$.

\item Setup 1.2: Fix  variable dimensions $(p_2,p_3)=(300,900)$ and noise variances $\sigma_{e_2}^2=\sigma_{e_3}^2=1$. 
The other settings are the same as in Setup~1.1.

\item Setup 2.1: 	
Let $p_1=p_2=p_3$ and $r_1=r_2=r_3=5$. Design $\cov(\bd{f})$ with
eigenvalues %$(3,2.799,2.25,1.5,1,1,1,1,0.635,0.415,0.401,0,0,0,0)$
$(3,2.8,2.25,1.5,1,1,1,1,0.635,0.415,0.4,0,0,0,0)$
such that, respectively for $\ell =1,\ldots, 4$,
$\{\theta(w^{(\ell)},z_k^{(\ell)})\}_{k\le 3}$ are all approximately
$0^\circ,$ $15^\circ,$ $30^\circ$,  and $45^\circ$, and $\{\theta(z_j^{(\ell)},z_k^{(\ell)})\}_{j<k\le 3}$ are all close to
$0^\circ, 25^\circ, 50^\circ$ and $75^\circ$.
Matrix $\cov(\bd{f})$ is given in Appendix~\ref{sec: add simul}.
Set $\bd{f}$ to be multivariate Gaussian with mean zero and covariance matrix $\cov(\bd{f})$.
Let $\mb{\Lambda}_k=\diag(500,400,300,200,100)$ for all $k\le 3$. 
Randomly generate three matrices $\{\mb{V}_{xk}\}_{k=1}^3$,  each with orthonormal columns, which are equal if with the same size and are fixed for all simulation replications of the same $(p_1,p_2,p_3)$.
Let $\bd{x}_k=\mb{V}_{xk}\mb{\Lambda}_{xk}^{1/2}\bd{f}_k$.
D-GCCA has $(\PVE_c(\bd{x}_1),\PVE_c(\bd{x}_2),\PVE_c(\bd{x}_3))=(0.387, 0.324, 0.427)$ invariant to $\{p_k\}_{k=1}^3$.
Let $\sigma_{e_1}^2=\sigma_{e_2}^2=\sigma_{e_3}^2$.

\item Setup 2.2: Fix $(p_2,p_3)=(300,900)$ and $\sigma_{e_2}^2=\sigma_{e_3}^2=1$. The other settings are the same as in Setup~2.1.  

\end{itemize}

\subsection{Finite-sample performance of D-GCCA estimators}\label{Sec: simul finite sample of proposed}

We first evaluate the performance of the D-GCCA estimation that uses true nuisance parameters $\{\{r_k,r_k^*\}_{k=1}^K,\mathcal{I}_0,\{{\color{black}\mathcal{I}_{\Delta_+}^{(\ell)},\mathcal{I}_{\Delta_0}^{(\ell)}},\sign(\alpha^{(\ell)})\}_{\ell\in\mathcal{I}_0}\}$.
The practical selection of these nuisance parameters has been discussed in Section~\ref{subsec: rank and set selection} and its performance is investigated later in this subsection.
It is easily seen that $\SNR_k=\tr(\mb{\Lambda}_k)/(p_k\sigma_{e_k}^2)$ in the above simulation setups.
For simplicity, we hence examine the trend of estimation errors in Theorem~\ref{consistency thm} with respect to $(p_k,\sigma_{e_k}^2)$ instead of $(p_k,\SNR_k)$.

Figure~\ref{Fig: Setup 1.1 and 1.2, theta=50} shows the estimation errors of D-GCCA
under Setups~1.1 and~1.2 with $\theta_z=50^\circ$.
Similar results are observed and provided in Appendix~\ref{sec: add simul} for the other values of $\theta_z$.
For Setup~1.1 in Figure~\ref{Fig: Setup 1.1 and 1.2, theta=50}\,(a), the average estimation errors are almost the same for
the three identically distributed data views, indicating the fair treatment of proposed
estimation to each view.
As expected in Theorem~\ref{consistency thm}, 
the errors generally increase as either dimension $p_1$ or noise variance $\sigma_{e_1}^2$ grows, and the relatively slower error trend of $\widehat{\PVE}_c(\bd{x}_k)$ reflects its slower convergence rate
as compared with those of $\{\widehat{\mb{X}}_k,\widehat{\mb{C}}_k, \widehat{\mb{D}}_k \}$.
The errors are acceptable even for some cases 
when $p_1$ or $\sigma_{e_1}^2$ is large along
with very low $\SNR_k$. For example, the errors are smaller than 0.05 at $(p_1,\sigma_{e_1}^2)=(1500,4)$ with $\SNR_k=0.083$.
In Figure~\ref{Fig: Setup 1.1 and 1.2, theta=50}\,(b) for Setup 1.2, the estimation result of the first data view is similar to that in Figure~\ref{Fig: Setup 1.1 and 1.2, theta=50}\,(a).
As for the second and third data views with fixed variable dimensions and noise variances,
when $(p_1,\sigma_{e_1}^2)$ the parameters of the first data view grow, 
the signal matrix estimation is not affected, while the estimation errors of the other three quantities are observed with slight increasing trends. 
These results are consistent with those shown in Theorem~\ref{consistency thm}.
{\color{black} Because Setups~1.1 and~1.2 are single-factor models, we have 
$\PVE_c(\bd{x}_k^{[i]})=\PVE_c(\bd{x}_k)$ and 
$\widehat{\PVE}_c(\bd{x}_k^{[i]})=\widehat{\PVE}_c(\bd{x}_k)$
for all $i\le p_k$.
The max absolute error of $\{\widehat{\PVE}_c(\bd{x}_k^{[i]})\}_{i=1}^ {p_k}$
hence coincides with
the absolute error of $\widehat{\PVE}_c(\bd{x}_k)$
shown in the seventh row of Figure~\ref{Fig: Setup 1.1 and 1.2, theta=50}.

Now we consider Setups 2.1 and 2.2 that are multi-factor models.
Figure~\ref{Fig: Setup 2.1 and 2.2, Frobenius} presents similar results as in Figure~\ref{Fig: Setup 1.1 and 1.2, theta=50}
for the estimation of $\{\mb{X}_k,\mb{C}_k,\mb{D}_k,\PVE_c(\bd{x}_k)\}_{k=1}^3$.
Figure~\ref{Fig: Setup 2.1 and 2.2, PVE}
shows the performance of $\{\widehat{\PVE}_c(\bd{x}_k^{[i]})\}_{i=1}^{p_k}$.
The first three rows of Figure~\ref{Fig: Setup 2.1 and 2.2, PVE} 
summarize the maximum, the third quartile,
and the median of their absolute errors. 
As in Theorem~\ref{consistency thm},
those errors increase as either dimension $p_1$ or noise
variance $\sigma_{e_1}^2$ grows.
For large $p_1$ or $\sigma_{e_1}^2$,
although the estimated PVE values
have large maximum absolute errors,
the fourth row of Figure~\ref{Fig: Setup 2.1 and 2.2, PVE}
shows strong average correlations ($>0.75$) between the true and  estimated PVEs. 
In terms of variable selection,
the consequent ranking of variables may be more
informative.
We evaluate the ranking quality 
by the Spearman's $\rho$ coefficient \citep{spearman1904}
and the normalized discounted cumulative gain \citep[nDCG;][]{pmlr-v30-Wang13}.
Spearman's $\rho\in [-1,1]$ computes the correlation between
the rank values of the true and estimated PVEs.
The considered nDCG ranges on $[0,1]$ and uses the true PVE as the degree of relevancy with the logarithmic discount.
For both metrics, a higher value indicates better concordance
between the rankings of variables from the true and  estimated PVEs.
The fifth row of Figure~\ref{Fig: Setup 2.1 and 2.2, PVE}
shows high average Spearman's $\rho$ values
mostly above 0.95 for low noise $\sigma_{e_1}^2\le 1$, 
above 0.85 for modest to moderate dimension $p_1\le 600$,
and nearly all above 0.75 for strong noise $\sigma_{e_1}^2\in \{4,9\}$
or large dimension $p_1\in [900,1500]$.
For the ranking of variables
based on either $\{\widehat{\PVE}_c(\bd{x}_k^{[i]})\}_{i=1}^{p_k}$ or $\{\widehat{\PVE}_d(\bd{x}_k^{[i]})\}_{i=1}^{p_k}$,
strong agreement with that based on their estimands is observed 
in the last two rows of Figure~\ref{Fig: Setup 2.1 and 2.2, PVE}
with the average nDCG above 0.97 for considering all ranks
and above $0.86$ for only the top $p_k/10$ ranks. 
}

We also numerically evaluate the selection approach of nuisance parameters that is proposed in Section~\ref{subsec: rank and set selection}.
Figures~\ref{Fig for nuisance in main text}, \ref{Fig1 for nuisance in appendix} and \ref{figA: Setup 1.2 prop nuiparameter plot} show the accuracy of the selection approach for the four simulation setups. 
For simplicity, we apply the same significance level $\alpha$, ranging from 0.5 down to 0.0001, to all hypothesis tests involved in the selection approach.
For Setups 1.1 and 1.2, $\alpha\in [0.0001,0.5]$ and $\alpha\in [0.005,0.5]$ perform the same well for $\theta_z\in [10^\circ,60^\circ]$
and $\theta_z=70^\circ$, respectively, with accuracy values all above 90\% and most around or above 95\%.
As for Setups 2.1 and 2.2, as shown in Figure~\ref{Fig for nuisance in main text} (e) and (f),
when the significance level is 0.1,
the accuracy achieves nearly 90\% for most considered cases.}
There is no dramatic change when the significance level is down from 0.2 to 0.05.
In practice, it is worth trying several significance levels to monitor the change of nuisance parameters,
and also suggested to report the used significance level
along with the obtained decomposition.
One may also expect to potentially improve the accuracy by additionally using the Bagging technique \citep{Hast09}, that is, for each nuisance parameter applying the selection approach to a large number of resampled data sets and then combining the results by majority voting. We leave this to interested readers.

\begin{figure}[p!]
	\begin{subfigure}[b]{0.5\textwidth}
		%	\centering
		\includegraphics[width=1\textwidth]{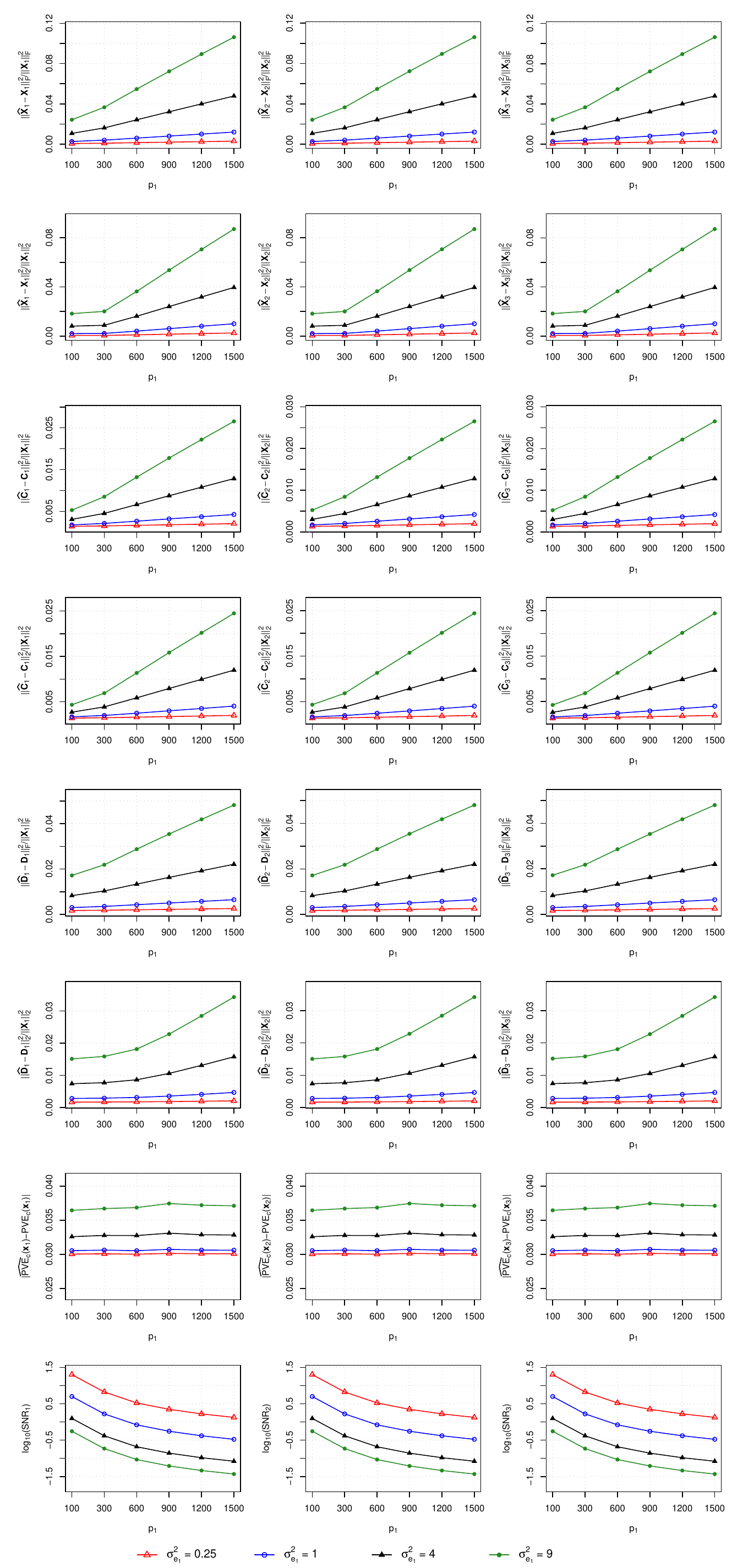}
		\caption{Setup 1.1 with $\theta_z=50^\circ$}
	\end{subfigure}
	\hspace{0.3cm}
	\begin{subfigure}[b]{0.5\textwidth}
		%	\centering
		\includegraphics[width=1\textwidth]{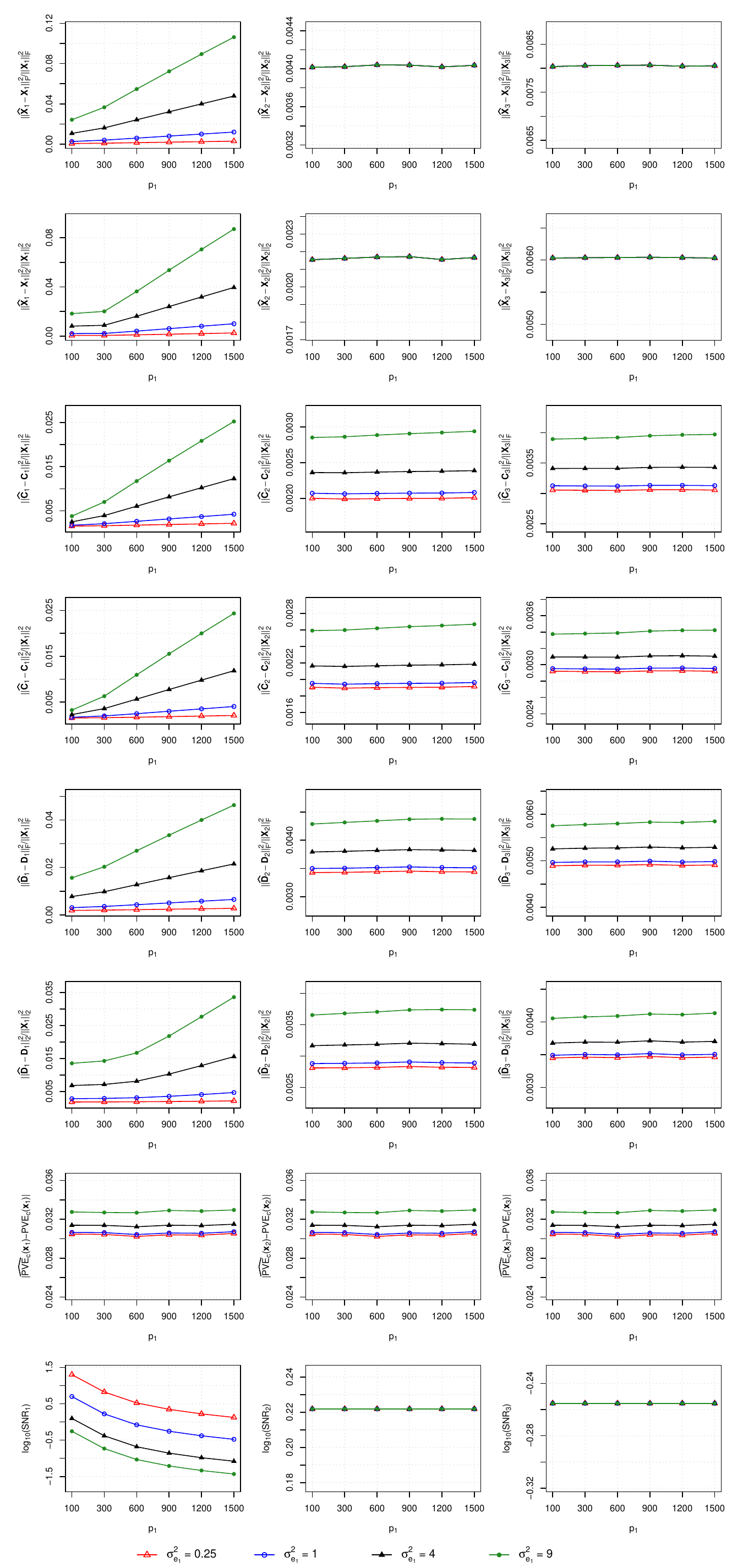}
		\caption{Setup 1.2 with $\theta_z=50^\circ$}
	\end{subfigure}
	\caption{Average errors of D-GCCA estimates over 1000 replications for Setups~1.1 and~1.2.}
	\label{Fig: Setup 1.1 and 1.2, theta=50}
\end{figure}

\begin{figure}[p!]
	\begin{subfigure}[b]{0.5\textwidth}
		%	\centering
		\includegraphics[width=1\textwidth]{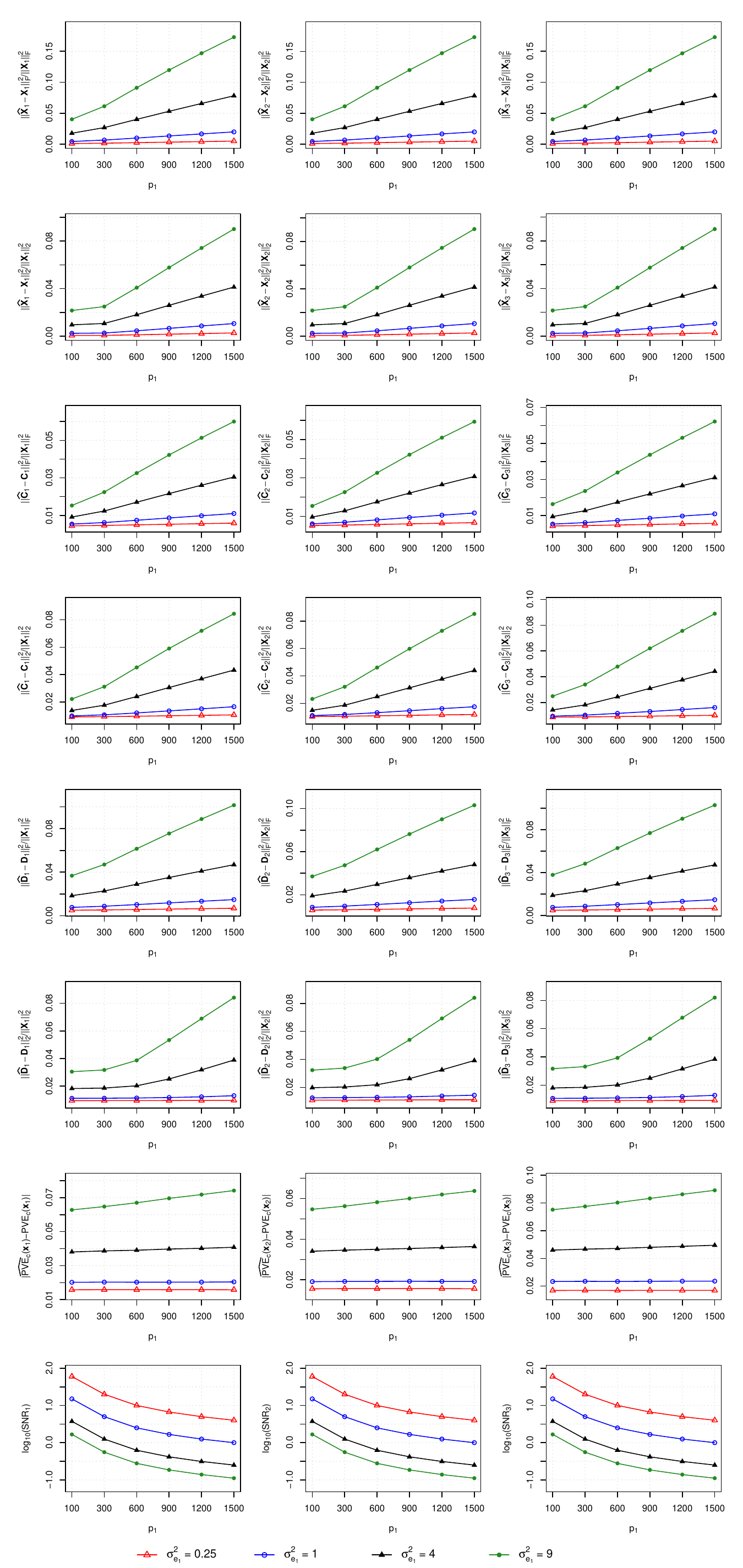}
		\caption{Setup 2.1}
	\end{subfigure}
	\hspace{0.3cm}
	\begin{subfigure}[b]{0.5\textwidth}
		%	\centering
		\includegraphics[width=1\textwidth]{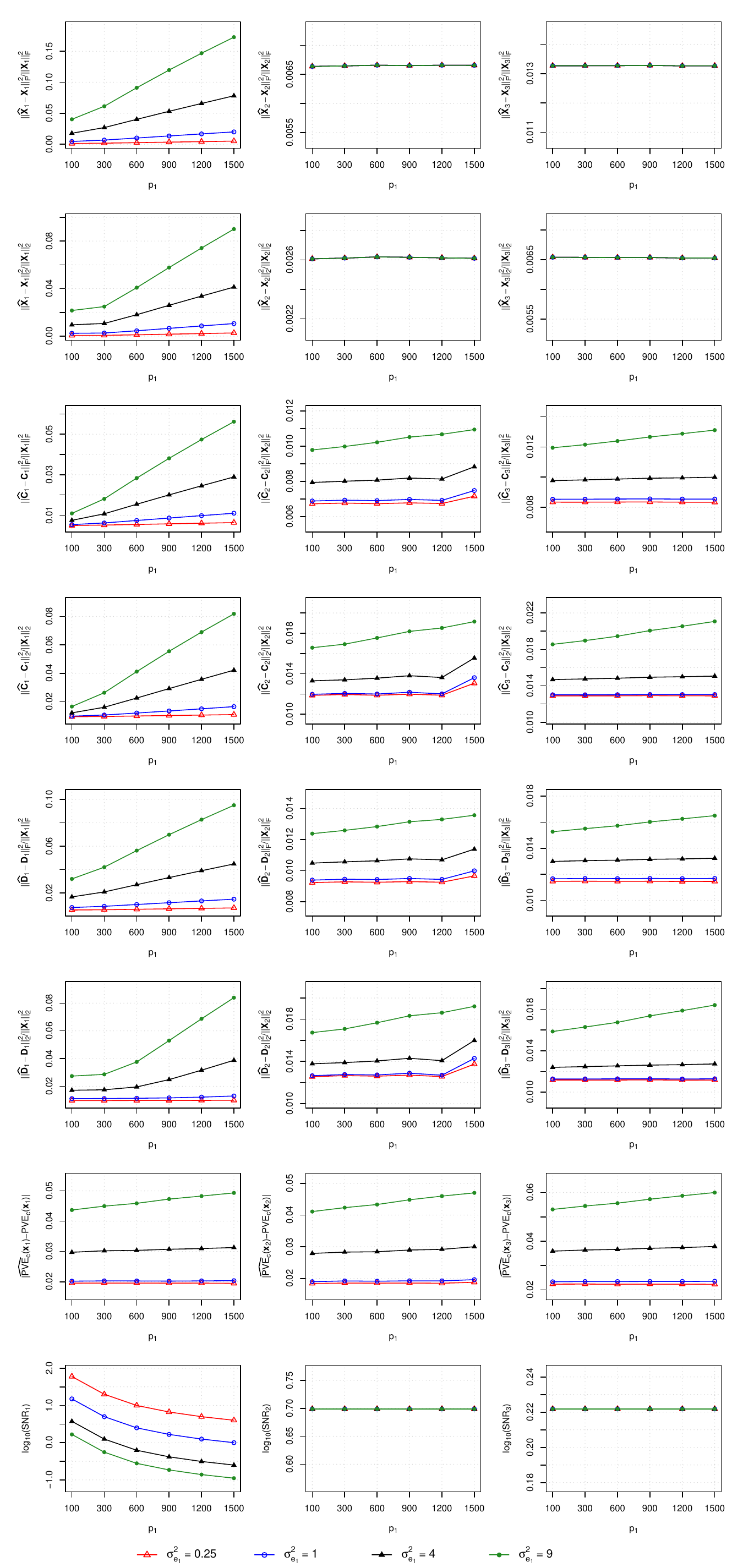}
		\caption{Setup 2.2}
	\end{subfigure}
	\caption{Average errors of D-GCCA estimates over 1000 replications for Setups~2.1 and~2.2.}
	\label{Fig: Setup 2.1 and 2.2, Frobenius}
\end{figure}

\begin{figure}[p!]
	\begin{subfigure}[b]{0.5\textwidth}
		%	\centering
		\includegraphics[width=1\textwidth]{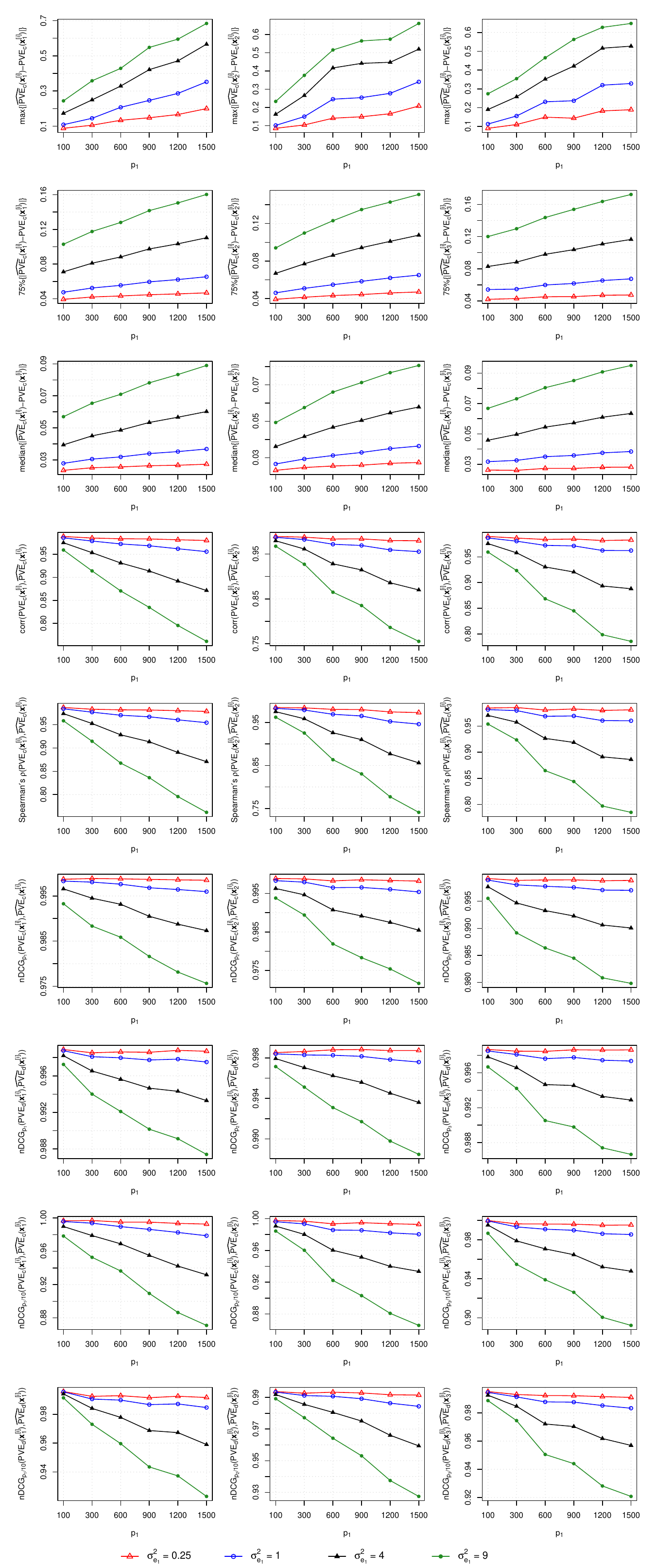}
		\caption{Setup 2.1}
	\end{subfigure}
	\hspace{0.3cm}
	\begin{subfigure}[b]{0.5\textwidth}
		%	\centering
		\includegraphics[width=1\textwidth]{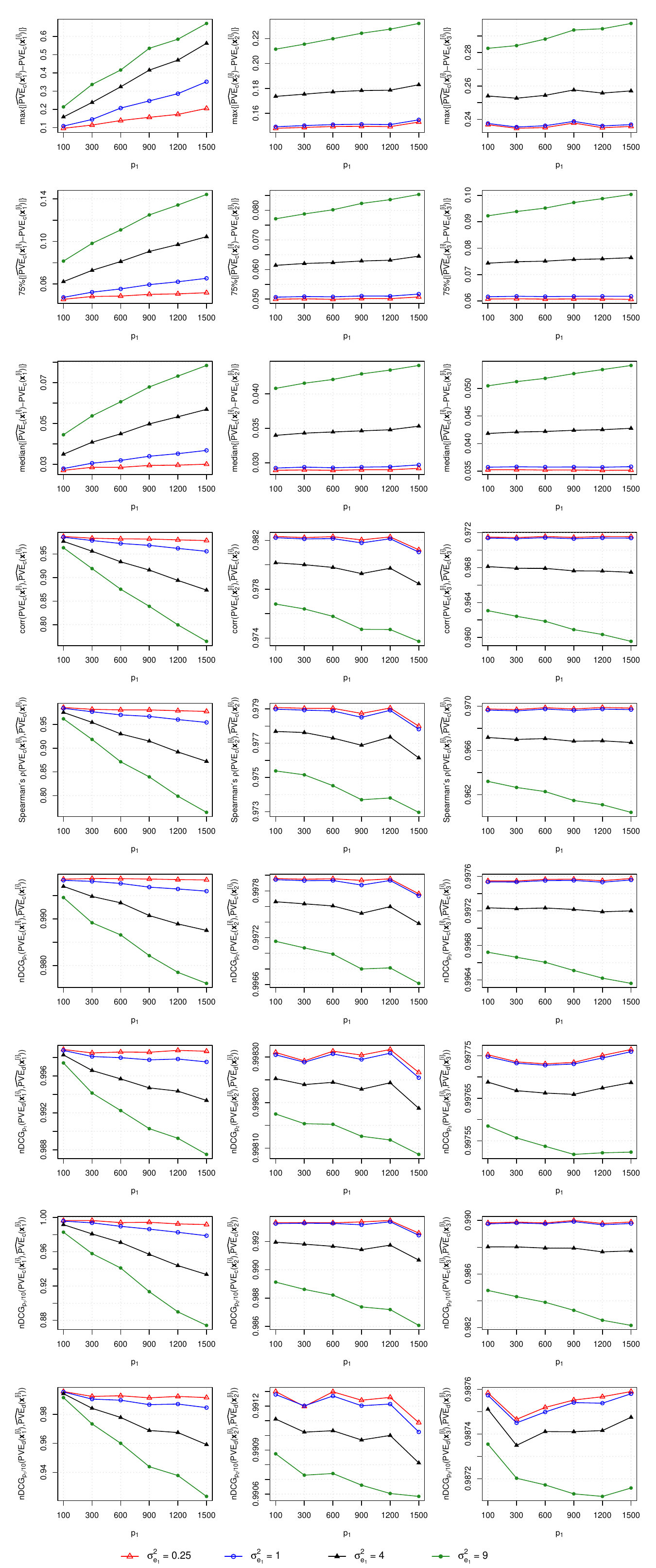}
		\caption{Setup 2.2}
	\end{subfigure}
	\caption{\color{black}Average results of D-GCCA's variable-level PVE estimation over 1000 replications for Setups~2.1 and~2.2.}
	\label{Fig: Setup 2.1 and 2.2, PVE}
\end{figure}

\begin{figure}[p!]
	\begin{subfigure}[b]{1\textwidth}
		\centering
		\includegraphics[width=0.75\textwidth]{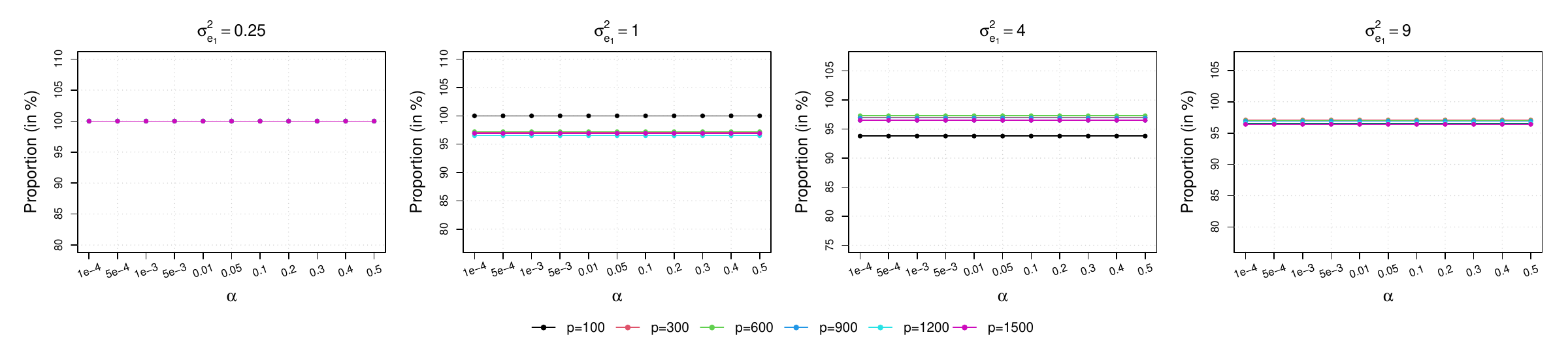}
		\caption{Setup 1.1 with $\theta_z=50^\circ$}
		\label{noise_level}
	\end{subfigure}\hspace{0.3cm}
	\begin{subfigure}[b]{1\textwidth}
		\centering
		\includegraphics[width=0.75\textwidth]{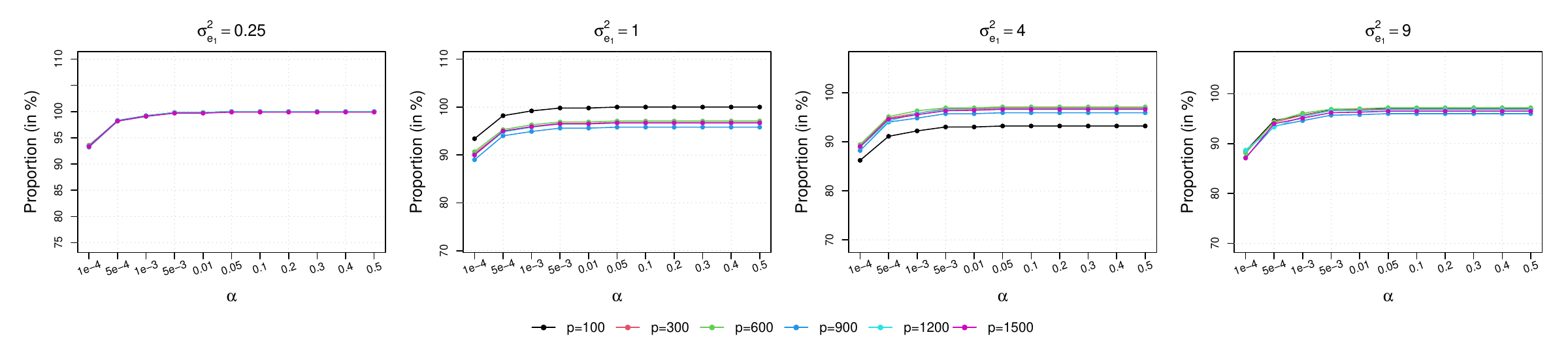}
		\caption{Setup 1.1 with $\theta_z=70^\circ$}
		\label{noise_level}
	\end{subfigure}
	\begin{subfigure}[b]{1\textwidth}
		\centering
		\includegraphics[width=0.75\textwidth]{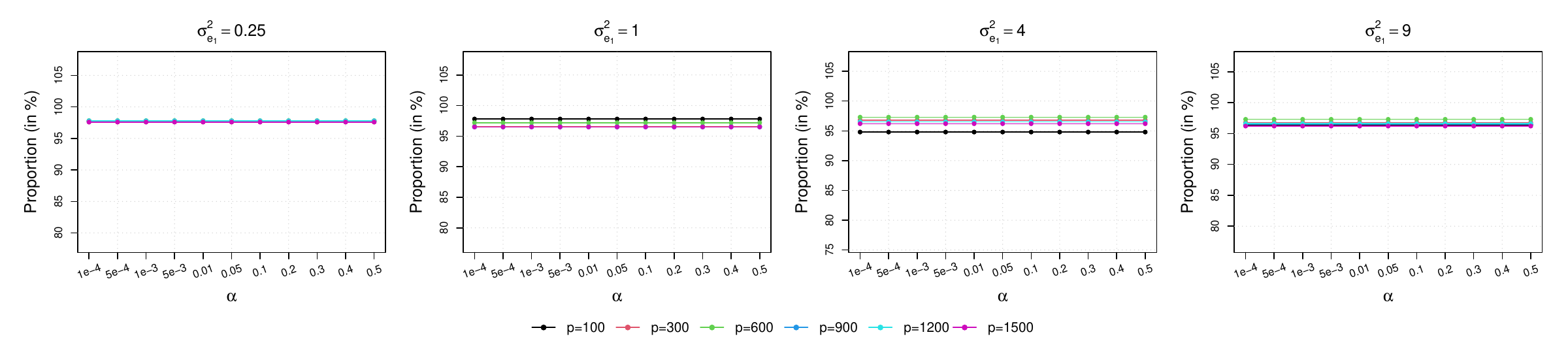}
		\caption{Setup 1.2 with $\theta_z=50^\circ$}
		\label{noise_level}
	\end{subfigure}\hspace{0.3cm}
	\begin{subfigure}[b]{1\textwidth}
		\centering
		\includegraphics[width=0.75\textwidth]{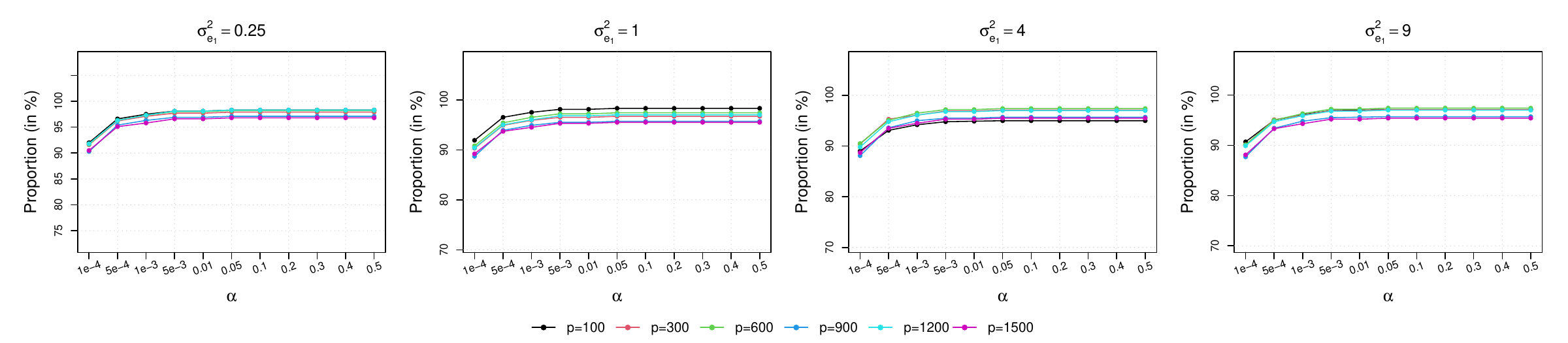}
		\caption{Setup 1.2 with $\theta_z=70^\circ$}
	\end{subfigure}
	\begin{subfigure}[b]{1\textwidth}
		\centering
		\includegraphics[width=0.75\textwidth]{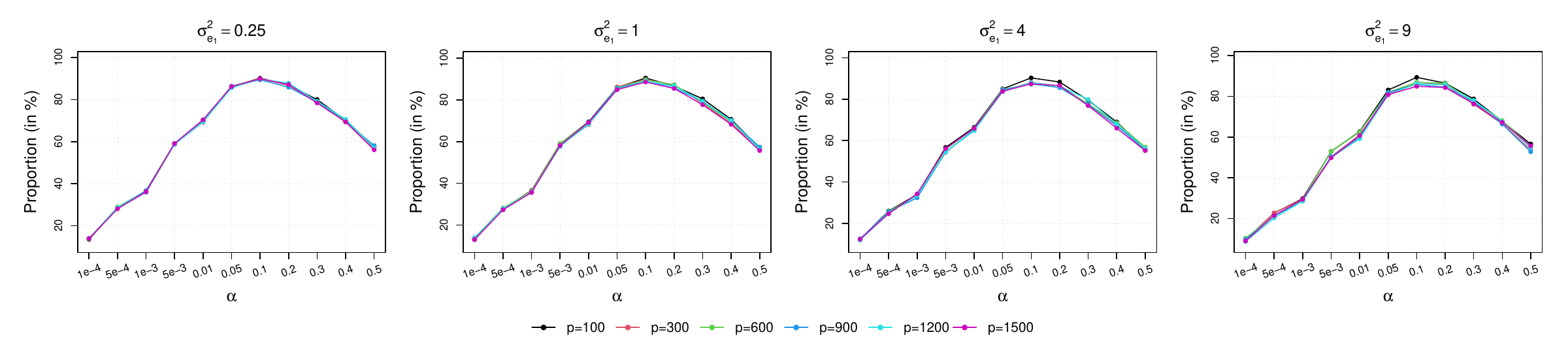}
		\caption{Setup 2.1}
	\end{subfigure}\hspace{0.3cm}
	\begin{subfigure}[b]{1\textwidth}
		\centering
		\includegraphics[width=0.75\textwidth]{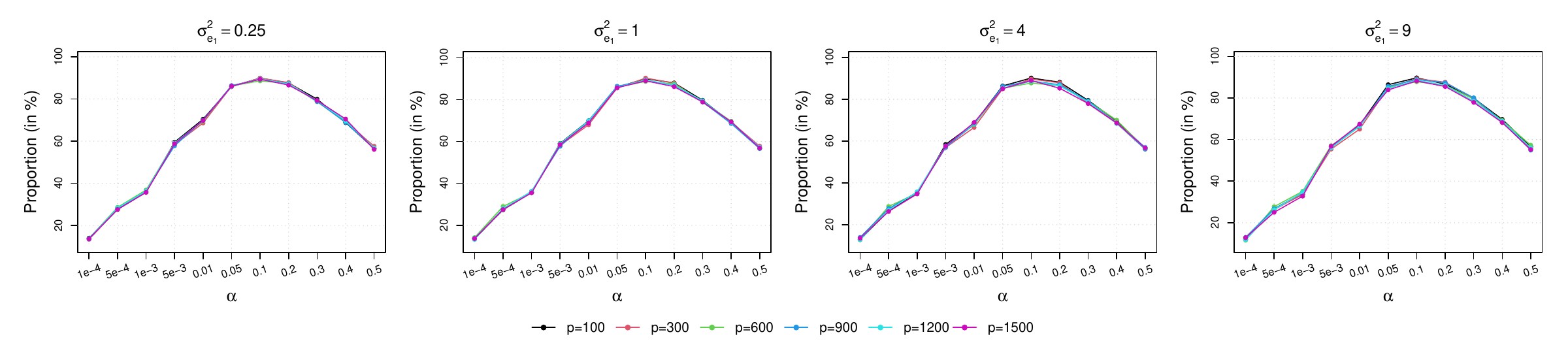}
		\caption{Setup 2.2}
	\end{subfigure}
	\caption{The proportion of 1000 simulation replications where all nuisance parameters of D-GCCA are correctly selected. The nuisance parameters 
		%$\big\{\{r_k,r_k^*\}_{k=1}^K, \mathcal{I}_0, \{\mathcal{I}_\Delta^{(\ell)},\sign(\alpha^{(\ell)})\}_{\ell\in \mathcal{I}_0}\big\}$ 
		are selected using the approach described in Section~\ref{subsec: rank and set selection} with a significance level $\alpha$ uniformly applied to all tests.
	}
	\label{Fig for nuisance in main text}
\end{figure}

\subsection{Comparison with related methods}\label{sec: siml comp}
{\color{black}
We now compare the performance of D-GCCA and the six competing methods (JIVE, R.JIVE, AJIVE, COBE, OnPLS, and DISCO-SCA) under the four simulation setups.

Since the decompositions defined by the seven methods are different, 
it is unfair to compare the errors of their matrix estimates to D-GCCA's true matrices. Alternatively, 
under the general model given in \eqref{decomp in mat} and \eqref{decomp in vari}, for each method we consider 
whether at least one orthogonal pair among $\{\bd{d}_k\}_{k=1}^K$ exists, and otherwise {\color{black}
how severe the common underlying source of variation}
is retained among $\{\bd{d}_k\}_{k=1}^K$.

The orthogonality between
$\bd{d}_j$ and $\bd{d}_k$ is equivalent to
$\sum_{m=1}^{r_{d_j}}\sum_{\ell=1}^{r_{d_k}}[\corr(d_j^{(m)},d_k^{(\ell)})\ne 0]=0$, 
where $\{d_k^{(\ell)}\}_{\ell=1}^{r_{d_k}}$ denote the latent factors of $\bd{d}_k$.  
We detect each
$\corr(d_j^{(m)},d_k^{(\ell)})\ne 0$ using the normal approximation test \citep{DiCi17}, with false discovery rate controlled at 0.05 \citep{Benj95} and the $\ell$th right-singular vector of $\widehat{\mb{D}}_k$ used as the
$n$ samples of $d_k^{(\ell)}$.

Let $\rho_\ell(\{\bd{x}_k\}_{k=1}^K)$ be the maximum of the objective function in \eqref{GCCA}.
If no pairs in $\{\bd{d}_k\}_{k=1}^K$ are orthogonal,
we use 
$\rho_1(\{\bd{d}_k\}_{k=1}^K)\in [1,K]$ to measure
{\color{black} the severity of common underlying source retained by} $\{\bd{d}_k\}_{k=1}^K$.
From equation~\eqref{ccorr=eig},
we estimate $\rho_1(\{\bd{d}_k\}_{k=1}^K)$ by $\widehat{\rho}_1(\{\bd{d}_k\}_{k=1}^K)=\lambda_1(\widehat{\mb{F}}\widehat{\mb{F}}^\top/n)$ with the matrix $\widehat{\mb{F}}$ defined in Section~\ref{subsec: mat est} but uses $\{\widehat{\mb{D}}_k\}_{k=1}^K$ here instead of~$\{\widehat{\mb{X}}_k\}_{k=1}^K$.

\begin{table}[p!]
	\begin{center}
		\scalebox{0.83}{
			\begin{tabular}{cccccc}	
				\thickhline\\[-12pt]
				Setup &		Method	&	 $\ge$ 1 orth. pair  & $\widehat{\rho}_1(\{\bd{d}_k\}_{k=1}^3)$ & $\frac{\|\widehat{\mb{X}}_1-\mb{X}_1\|_F^2}{\|\mb{X}_1\|_F^2}$, $\frac{\|\widehat{\mb{X}}_2-\mb{X}_2\|_F^2}{\|\mb{X}_2\|_F^2}$,
				$\frac{\|\widehat{\mb{X}}_3-\mb{X}_3\|_F^2}{\|\mb{X}_3\|_F^2}$ 
				\\ \hline
				&			D-GCCA1                  &{\bf 100\%}  & 1.10 (0.05) & 0.006 (6.0e-4), 0.006 (5.9e-4), 0.006 (5.6e-4)\\
				&			D-GCCA2                  &{\bf  100\%}  &1.10 (0.05) & 0.006 (1.0e-3), 0.006 (1.1e-3), 0.006 (1.6e-3) \\
				Setup 1.1&			JIVE                          & 0\% & 2.22 (0.06)& 0.014 (1.4e-3), 0.014 (1.4e-3), 0.014 (1.3e-3)\\
				($p_1=600$, &	R.JIVE    			   	& {\bf  100\%}& 1.00 (0.00)& {\bf 0.032(1.0e-2),\,0.021(3.1e-3),\,0.023(7.1e-3)}\\
				~$\theta_z=50^\circ$,&AJIVE   			 &0\% (zero $\widehat{\mb{C}}_k$s) &2.28 (0.05) &0.006 (6.0e-4), 0.006 (6.0e-4), 0.006 (5.6e-4)\\
				$\sigma_{e_1}^2=1$)&COBE  &0\% (zero $\widehat{\mb{C}}_k$s)&2.28 (0.05) &0.006 (6.0e-4), 0.006 (6.0e-4), 0.006 (5.6e-4)\\
				&			OnPLS   				    &1.1\% & 1.87 (0.05) & 0.026 (2.3e-3), 0.026 (2.3e-3), 0.026 (2.2e-3)\\
				&			DISCO-SCA 				&0\% (zero $\widehat{\mb{C}}_k$s)& 3.00 (0.00)& 0.014 (1.3e-3), 0.014 (1.3e-3), 0.014 (1.3e-3)\\
				\hline
				&			D-GCCA1                  &{\bf 100\%} &1.10 (0.05)& 0.006 (6.0e-4), 0.004 (4.1e-4), 0.008 (7.3e-4)\\
				&			D-GCCA2                  &{\bf 100\%} &1.10 (0.05)& 0.006 (1.0e-3), 0.004 (7.9e-4), 0.008 (1.7e-3)\\
				Setup 1.2 &			JIVE                          &0\%  &2.20 (0.06) & 0.014 (1.4e-3), 0.009 (1.0e-3), 0.018 (1.6e-3)\\
				($p_1=600$,&	R.JIVE    			   	&{\bf  100\%} &1.00 (0.00)&{\bf 0.033(1.0e-2),\,0.014(2.3e-3),\,0.029(6.7e-3)}\\
				~$\theta_z=50^\circ$,&AJIVE   			 &0\% (zero $\widehat{\mb{C}}_k$s)&2.28 (0.05)& 0.006 (6.0e-4), 0.004 (4.1e-4), 0.008 (7.3e-4)\\
				$\sigma_{e_1}^2=1$)&COBE  &0\% (zero $\widehat{\mb{C}}_k$s)&2.28 (0.05) & 0.006 (6.0e-4), 0.004 (4.1e-4), 0.008 (7.3e-4)\\
				&			OnPLS   				    & 0.9\%& 1.83 (0.05) & 0.026 (2.4e-3), 0.018 (1.6e-3), 0.026 (2.2e-3)\\
				&			DISCO-SCA 				&0\% (zero $\widehat{\mb{C}}_k$s)& 3.00 (0.00)& 0.014 (1.3e-3), 0.008 (7.6e-4), 0.020 (1.8e-3)\\
				\hline
				&			D-GCCA1                  & 0\%&{\bf2.13 (0.05)}& 0.010 (4.5e-4), 0.010 (4.5e-4), 0.010 (4.8e-4)\\
				&			D-GCCA2                  &0\% &{\bf2.14 (0.06)} & 0.010 (4.5e-4), 0.010 (4.5e-4), 0.010 (4.8e-4) \\
				Setup 2.1 &			JIVE                          & 0\% &2.52 (0.21) & 0.016 (2.0e-3), 0.016 (2.2e-3), 0.016 (2.1e-3)\\
				($p_1=600$,&	R.JIVE    			   	&{\bf 100\%} &1.00 (0.00)& {\bf 0.076(4.3e-2),\,0.080(4.9e-2),\,0.065(3.4e-2)}\\
				$\sigma_{e_1}^2=1$)&AJIVE   			 &0\% &2.80 (0.02)& 0.010 (4.4e-4), 0.010 (4.3e-4), 0.010 (4.7e-4)\\
				&COBE  &0\% &2.80 (0.02) & 0.010 (4.6e-4), 0.010 (4.6e-4), 0.010 (4.8e-4)\\
				&			OnPLS   				    &0.1\% & 2.65 (0.18)& 0.014 (1.7e-3), 0.014 (3.1e-3), 0.015 (1.8e-3)\\
				&			DISCO-SCA 				&NA & NA  & NA\\
				\hline
				&			D-GCCA1                  & 0\%&{\bf2.13 (0.05)} & 0.010 (4.5e-4), 0.007 (3.2e-4), 0.013 (6.1e-4)\\
				&			D-GCCA2                  &0\% &{\bf2.14 (0.06)} & 0.010 (4.5e-4), 0.007 (3.2e-4), 0.013 (6.1e-4)\\
				Setup 2.2&			JIVE                          & 0\% &2.41 (0.26) & 0.016 (2.3e-3), 0.010 (1.4e-3), 0.021 (3.1e-3) \\
				($p_1=600$, &	R.JIVE    			   	&{\bf 100\%} &1.00 (0.00)& {\bf 0.064(4.0e-2),\,0.063(5.0e-2),\,0.079(4.3e-2)}\\
				$\sigma_{e_1}^2=1$)&AJIVE   			 &0\% &2.80 (0.02) & 0.010 (4.4e-4), 0.006 (3.0e-4), 0.013 (6.1e-4)\\
				&COBE  &0\% &2.80 (0.02) & 0.010 (4.6e-4), 0.007 (3.2e-4), 0.013 (6.2e-4)\\
				&			OnPLS   				    &0.5\% & 2.51 (0.18) & 0.015 (6.5e-3), 0.009 (1.3e-3), 0.020 (2.6e-3)\\
				&			DISCO-SCA 				& NA & NA &  NA\\
				\thickhline
		\end{tabular}}
	\end{center}
	\vspace{-0.5cm}
	\caption{
		{\color{black}
			The proportions of replications with at least one orthogonal pair among $\{\bd{d}_k\}_{k=1}^3$, averages (standard deviations) of $\widehat{\rho}_1(\{\bd{d}_k\}_{k=1}^3)$, and averages (standard deviations) of scaled squared errors of signal matrix estimates over 1000 simulation replications.
			D-GCCA1: the D-GCCA using true nuisance parameters. 	
			D-GCCA2: the D-GCCA using nuisance parameters selected by the approach in Section~\ref{subsec: rank and set selection}. 	
			NA: not available due to out of the 24-hour time limit on a CPU core (up to 3.0GHz) per simulation replication.
			By the paired t-test, both D-GCCA1 and D-GCCA2 have significantly different mean $\widehat{\rho}_1(\{\bd{d}_k\}_{k=1}^3)$ values from those of all the other methods with p-values$<$1e-10.}
	}
	\label{Siml table}
\end{table}

Table~\ref{Siml table} reports the comparison results for Setups 1.1 and 1.2
with $(p_1,\theta_z, \sigma_{e_1}^2)=(600,50^\circ,1)$ and 
Setups 2.1 and 2.2 with $(p_1, \sigma_{e_1}^2)=(600,1)$.
We first observe that all simulation replications of R.JIVE for the four setups have at least one orthogonal pair among $\{\bd{d}_k\}_{k=1}^3$, but its scaled squared errors of signal matrix estimates
are much larger than those of JIVE (its original version with no orthogonality constraint on $\{\bd{d}_k\}_{k=1}^K$) and our D-GCCA. 
This agrees with the design of R.JIVE, which can discard some signal as noise
to ensure the orthogonality of $\{\bd{d}_k\}_{k=1}^K$. 
For Setups 1.1 and 1.2 with three one-dimensional signal subspaces $\{\lspan(\bd{x}_k^\top)\}_{k=1}^3$, our D-GCCA  also has all its simulation replications satisfying the desirable orthogonality among $\{\bd{d}_k\}_{k=1}^3$, which is consistent with its decomposition in \eqref{z=c+d} for canonical variables.
In contrast, the other five methods do not show the desirable orthogonality 
for nearly all replications under the four setups. 
For Setups 2.1 and 2.2 with higher-dimensional signal subspaces,
neither does D-GCCA own the desirable orthogonality,
as explained in Section~\ref{sec: t-th level decomp} due to its relaxation into each sub-problem~\eqref{z=c+d}, 
but D-GCCA still has significantly smaller
mean $\widehat{\rho}_1(\{\bd{d}_k\}_{k=1}^K)$ values than those available for the other five methods.
}

\section{Real-world Data Examples}\label{Sec: real data}

\subsection{Application to TCGA breast cancer genomic data}\label{real data analysis}

We compare our D-GCCA with the six state-of-the-art  methods in analyzing the TCGA breast cancer genomic data \citep{Kobo12}.
We consider  three types of genomic data
on a common set of 664 tumor samples
that contain
mRNA expression data for the top 2642 variably expressed genes, miRNA expression data for 437 highly variant miRNAs, and DNA methylation data for 3298 most variable probes. 
The data have been preprocessed following the procedure of \citet{Lock13b}. 
The tumor samples are categorized by the classic PAM50 model \citep{Park09} into
four intrinsic subtypes that are relevant with clinical outcomes, including
111 Basal-like, 56 HER2-enriched, 346 Luminal A, and
151 Luminal B tumors. 
The PAM50 intrinsic subtypes are defined by mRNA expression only.
We investigate whether these intrinsic subtypes are also characterized by other genomic data types such as DNA methylation and miRNA expression that represent different biological components. In particular, we
study the relationship
between the PAM50 intrinsic subtypes 
and the common and distinctive underlying mechanisms of the three genomic data types by evaluating the ability of their corresponding matrices in model~\eqref{decomp in mat}
to separate the four intrinsic subtypes.

\begin{table}[b!]
	\begin{center}
		\begin{tabular}{ccccc}	
			\thickhline
			\\[-12pt]
			Method		& $\widehat{\mb{X}}_k$ 
			& $\widehat{\mb{C}}_k$ 
			& $\widehat{\mb{D}}_k$ 
			& $\widehat{\mb{E}}_k$ \\
			\hline
			D-GCCA    & 48.0, 62.7, 73.6&{\bf21.5}$^\ddag$,\,{\bf21.2},\,{\bf26.8}$^\sharp$  &74.2, 84.9, 93.2  &99.0, 98.4, 98.3  \\
			JIVE      &74.0, 80.0, 82.5    & 65.6, 65.3, 58.9&86.1, 87.9, 92.1  &  99.8, 99.6, 99.7\\
			R.JIVE    &74.5, 74.7, 80.8  &41.7, 38.1, 64.6  &93.3, 99.7, 99.6 & 99.8, 97.0, 97.6   \\
			AJIVE     &48.2, 62.7, 73.6   &NA, NA, NA  & 48.2, 62.7, 73.6& 99.0, 98.4, 98.3 \\
			COBE      &48.2, 62.7, 73.6    &NA, NA, NA  &48.2, 62.7, 73.6 &99.0, 98.4, 98.3    \\
			OnPLS     &60.0, 70.8, 78.1    &36.4, 34.1, 36.4   &89.6, 95.8, 98.6 &99.5, 98.9, 99.6    \\
			DISCO-SCA & 56.7, 67.4, 75.0  &52.6, 53.0, 48.5  &99.0, 97.7, 99.3 &99.4, 99.5, 99.6  \\[0.3ex]
			JIVE* &48.0, 62.7, 73.6   &35.0, 33.0, 50.8   &89.0, 93.8, 97.3 &NA, NA, NA  \\
			R.JIVE* &47.6, 60.2, 72.2  &34.0, 28.5, 61.4    &84.7, 98.7, 99.4   &99.3, 84.7, 83.5   \\
			AJIVE* &48.0, 62.7, 73.6   & NA, NA, NA &48.0, 62.7, 73.6  &NA, NA, NA  \\
			COBE* & 48.0, 62.7, 73.6   & NA, NA, NA &48.0, 62.7, 73.6  &NA, NA, NA  \\
			OnPLS* &48.0, 62.7, 73.6   &22.6$^\ddag$, 26.4, 30.5   &75.1, 87.8, 94.0   & NA, NA, NA \\
			DISCO-SCA* &48.0, 62.7, 73.6  &28.0, 28.0, 28.0$^\sharp$  & 77.9, 82.7, 94.9& NA, NA, NA   \\	
			\hline
			$\mb{Y}_k$ & 84.8, 87.8, 90.0\\
			\thickhline
		\end{tabular}
	\end{center}
	\vspace{-0.5cm}
	\caption{
		{\color{black}
			SWISS scores (in \%) for TCGA breast cancer genomic data types ($k=$ mRNA, miRNA, DNA).
			Lower SWISS scores indicate better subtype separation.
			Methods suffixed with * use D-GCCA's $\widehat{\mb{X}}_k$s instead of  $\mb{Y}_k$s as the input data. NA: not available due to a zero matrix estimate. 
			All methods have	$\text{SWISS}(\widehat{\mb{X}}_k)<\text{SWISS}({\mb{Y}}_k)$ for each $k$.
			Except AJIVE and COBE with $\widehat{\mb{C}}_k=\mb{0}$, all the other methods have $\text{SWISS}(\widehat{\mb{C}}_k)< \text{SWISS}(\widehat{\mb{X}}_k)< \text{SWISS}(\widehat{\mb{D}}_k)$
			for each $k$. Our D-GCCA has the lowest $\text{SWISS}(\widehat{\mb{C}}_k)$ for all $k$.
			By the test of \citet{Caba10},
			all above comparisons of SWISS scores are
			significantly different with p-values$<$0.001, except for the two annotated respectively by $\ddag$ and $\sharp$ with p-values$>$0.05.}
	}	
	\label{SWISS table}
\end{table}

Each observed data matrix is row-centered by subtracting the average within each row.
The nuisance parameters of our D-GCCA method are selected by using 
the approach described in Section~\ref{subsec: rank and set selection}.
The selection approach yields the same decomposition by the choices 0.2 and 0.0001 for the significance level uniformly applied to all involved hypothesis tests.
The values $(\rank(\widehat{\mb{X}}_k), \rank(\widehat{\mb{C}}_k),\rank(\widehat{\mb{D}}_k),\widehat{\PVE}_c(\bd{x}_k))$ from the D-GCCA method
are $(4, 2, 4, 0.239)$, $(3, 2, 3, 0.184)$ and $(3, 2, 3, 0.147)$ for the mRNA, miRNA, and DNA data types, respectively.
%The common-source matrices of  $\{\widehat{\mb{D}}_k\}_{k=1}^3$ only account for less than 0.01 of the variation of their corresponding $\{\widehat{\mb{X}}_k \}_{k=1}^3$, which appear to be ign
To quantify the subtype separation in a matrix, we adopt the SWISS score of \citet{Caba10} that calculates
the standardized within-subtype sum of squares: 
{\color{black}For a matrix $\mb{M}=(M_{ij})_{p\times n}$,
\[
\text{SWISS}(\mb{M})=\frac{\sum_{i=1}^p\sum_{j=1}^n(M_{ij}-\bar{M}_{i,s(j)})^2}{\sum_{i=1}^p\sum_{j=1}^n(M_{ij}-\bar{M}_{i,\cdot})^2},
\]
where $\bar{M}_{i,s(j)}$ is the average of the $j$th sample's subtype on the $i$th row, and $\bar{M}_{i,\cdot}$ is the average of the $i$th row's elements.}
The lower score indicates better subtype separation.

Table~\ref{SWISS table} shows the SWISS scores computed for the D-GCCA method and also the six competing methods mentioned in Section~\ref{sec: intro}. 
The denoised signal matrix $\widehat{\mb{X}}_k$ from all methods 
gains an improved ability on subtype separation with a %much 
smaller score,
comparing to the noisy data matrix ${\mb{Y}}_k$.
All methods, except AJIVE and COBE, discover nonzero common-source matrices, and show a clear pattern of decreasing SWISS scores from 
$\widehat{\mb{D}}_k$ to $\widehat{\mb{X}}_k$ and then to $\widehat{\mb{C}}_k$.
This pattern indicates that the four PAM50 intrinsic subtypes
are more likely to be an inherent feature of the common mechanism underlying the three different genomic data types.
Moreover, our D-GCCA method has the lowest scores for estimated common-source matrices when compared with the other methods.
{\color{black}
The result analysis remains the same even when our D-GCCA's $\widehat{\mb{X}}_k$s, which have the smallest SWISS scores among all signal estimates, are used as the input data for the other six methods.

\begin{table}[b!]
	\begin{center}
			\begin{tabular}{cccc}	
				\thickhline
				Method		& $\bd{d}_{\text{mRNA}}$ \&  $\bd{d}_{\text{miRNA}}$
				&  $\bd{d}_{\text{mRNA}}$ \&  $\bd{d}_{\text{DNA}}$
				& $\bd{d}_{\text{miRNA}}$ \&  $\bd{d}_{\text{DNA}}$ 
				\\ \hline
				D-GCCA    &58.3\% &58.3\% & 0\%  \\
				JIVE      & 15.9\%&21.0\% &17.9\%  \\
				R.JIVE    & 0\%& 0\%&0\% \\
				AJIVE     &75.0\% & 75.0\%&77.8\%\\
				COBE     &75.0\% & 75.0\%&77.8\% \\
				OnPLS     &41.3\% & 60.0\%&36.1\%  \\
				DISCO-SCA &62.5\% & 68.8\%& 56.3\% \\[0.3ex]
				JIVE* &83.3\% & 75.0\%&66.7\%  \\
				R.JIVE* &0\% & 0\%&  0\%\\
				AJIVE* &75.0\% & 75.0\%&77.8\%  \\
				COBE* &75.0\% & 75.0\%&77.8\% \\
				OnPLS* & 83.3\%& 50.0\%& 25.0\%\\
				DISCO-SCA* & 66.7\% &83.3\%  &55.6\% \\	
				\thickhline
			\end{tabular}
	\end{center}
	\vspace{-0.5cm}
	\caption{
		{\color{black}
			The proportions of significant nonzero correlations between distinctive latent factors across TCGA breast cancer genomic data types.
			The proportion is computed by $\frac{1}{d_jd_k}\sum_{m=1}^{r_{d_j}}\sum_{\ell=1}^{r_{d_k}}[\corr(d_j^{(m)},d_k^{(\ell)})\ne 0]$ for $j\ne k$, where $\{d_k^{(\ell)}\}_{\ell=1}^{r_{d_k}}$ are latent factors of $\bd{d}_k$,
			and $\corr(d_j^{(m)},d_k^{(\ell)})\ne 0$ is detected by the normal approximation test \citep{DiCi17} with false discovery rate controlled at 0.05 \citep{Benj95} and the $\ell$th right-singular vector of $\widehat{\mb{D}}_k$ used as the
			$n$ samples of $d_k^{(\ell)}$.
			Methods suffixed with * use D-GCCA's $\widehat{\mb{X}}_k$s
			instead of $\mb{Y}_k$s as the input data. 
		}	
	}
	\label{d's corr table}
\end{table}

The better SWISS scores of D-GCCA for common-source matrix estimates indicate its enhanced ability to capture the common latent factors than the other methods,
which
benefits from our well designed orthogonality constraint on distinctive latent factors.
Table~\ref{d's corr table} further verifies this conclusion,
and shows that significant nonzero correlations do not exist
between D-GCCA's $\bd{d}_{\text{miRNA}}$ and $\bd{d}_{\text{DNA}}$ but account for over 15\% among all pairs of $\bd{d}_k$s from the other methods except R.JIVE.
However, R.JIVE enforces the orthogonality of $\bd{d}_k$s by sacrificing its unexplained signal to be noise. This can be seen in Table~\ref{SWISS table}, where R.JIVE has slightly lower SWISS scores for $\widehat{\mb{E}}_k$s than JIVE, its original version with no orthogonality constraint on $\bd{d}_k$s, and moreover has nonzero $\widehat{\mb{E}}_k$s when using low-rank D-GCCA's signal estimates as the input data.

}

\begin{table}[b!]
	\begin{center}
		\scalebox{0.82}{
			\begin{tabular}{ccccc|ccccc}	
				\thickhline
				%\\[-12pt]
				&&\multicolumn{3}{c|}{SWISS score}&
				&&\multicolumn{3}{c}{SWISS score}\\
				\cline{3-5}\cline{8-10}
				Name	& $\widehat{\PVE}_c(\bd{x}_k^{[i]})$	& $\widehat{\mb{X}}_k^{[i,:]}$ & $\widehat{\mb{C}}_k^{[i,:]}$ & $\widehat{\mb{D}}_k^{[i,:]}$  &
				Name	& $\widehat{\PVE}_d(\bd{x}_k^{[i]})$	& $\widehat{\mb{X}}_k^{[i,:]}$ & $\widehat{\mb{C}}_k^{[i,:]}$ & $\widehat{\mb{D}}_k^{[i,:]}$ 
				
				\\
				\hline
				&&&\multicolumn{3}{c}{Top 10 genes for $k=\text{mRNA}$}			\\
				AKR7A3 & 0.449& 0.156 & 0.182 &0.318   &
				FGG &0.9999&0.750&0.253&0.750\\	
				RHCG  & 0.449& 0.153 &0.183 & 0.309 &
				NEU4 &0.9995& 0.718&0.210&0.716 \\
				AADAT & 0.448& 0.136 &0.179 & 0.281 &
				TAS1R3 &0.9995&0.738&0.745&0.732\\
				GAL & 0.448&0.145 & 0.180 & 0.300 &
				PCSK1&0.9993&0.832&0.289& 0.833 \\
				SLC26A9 & 0.448&0.175 & 0.185 &0.363 &
				HMGCLL1&0.9993&0.708&0.315&0.710\\
				PLAC1   & 0.447&0.162 &0.186 & 0.324 &
				HNF4G&0.9993&0.774&0.248&0.776\\
				KIAA1257 & 0.447& 0.177 &0.187 &0.362 &
				CRISP3&0.9991&0.725&0.469&0.723\\
				FMO6P & 0.447& 0.133 & 0.176 & 0.281 &
				TYRP1&0.9988&0.783&0.515&0.787\\
				GDF15   & 0.447& 0.166 & 0.184 & 0.345  &
				LHFPL4&0.9987&0.775&0.335&0.778\\
				TNNT2   & 0.445& 0.131 &0.178 &0.287 &
				NTS&0.9985&0.761&0.679&0.766\\
				
				&&&\multicolumn{3}{c}{Top 10 miRNAs for $k=\text{miRNA}$}\\
				hsa-mir-584 & 0.448& 0.322 &0.190 &0.732& 
				hsa-mir-34b& 0.99995&0.924&0.233& 0.923\\
				hsa-mir-1468 & 0.444& 0.276 &0.174 &0.657&
				hsa-mir-26a-2&0.9999&0.927&0.388&0.926\\
				hsa-mir-203 & 0.443&0.346 &0.196 &0.763 &
				hsa-mir-196a-1&0.9998&0.928&0.403&0.927\\
				hsa-mir-135b & 0.433&0.270 &0.169 &0.642 &
				hsa-mir-874& 0.9973&0.893&0.511&0.906\\
				hsa-mir-519a-1 &0.428&0.265 &0.167 &0.632&
				hsa-mir-193a& 0.9953 &0.881&0.539&0.899\\
				hsa-mir-190b &0.420&0.384 &0.210 &0.782&
				hsa-mir-615&0.9947&0.872&0.667& 0.892\\
				hsa-mir-29c &0.415& 0.341 &0.193 &0.747 &
				hsa-mir-326 &0.9944&0.882&0.444&0.901\\
				hsa-mir-526b &0.413&0.371 & 0.182 &0.797&
				hsa-mir-296 & 0.9934&0.943&0.291&0.936\\
				hsa-mir-28 &0.411&0.424 &0.200 &0.859&
				hsa-mir-26b& 0.9912&0.856&0.663& 0.882\\
				hsa-mir-30e &0.409&0.299 &0.166 & 0.681&
				hsa-let-7e&0.9877&0.854 &0.537&0.884\\
				
				&&&\multicolumn{3}{c}{Top 10 probes for $k=\text{DNA}$}\\
				cg04220579&0.438&0.314&0.178&0.726& 
				cg24030449& 0.9999&0.981&0.424&0.980\\
				cg02085507&0.437&0.309&0.190&0.700& 
				cg17296078&0.9998&0.984&0.665&0.982\\
				cg18055007&0.432&0.314 &0.195&0.701& 
				cg14009688&0.9997&0.984&0.684&0.983\\
				cg26668713&0.432&0.319&0.182&0.732& 
				cg00121904&0.9997&0.972&0.722&0.975\\
				cg23178308&0.430&0.337&0.186&0.748& 
				cg02789485&0.9997&0.982&0.281&0.981\\
				cg12406559&0.428&0.329&0.176&0.756& 
				cg07482936&0.9996&0.977&0.200&0.977\\
				cg25167447&0.427&0.351&0.168&0.776& 
				cg01817393&0.9996&0.977&0.197&0.978\\
				cg14385738&0.422&0.337&0.176&0.770& 
				cg10484958&0.9993&0.986&0.497&0.984\\
				cg02433671&0.420&0.333&0.207&0.718& 
				cg17532978&0.9986&0.969&0.383&0.974\\
				cg00916635&0.420&0.346&0.171&0.786&
				cg08291098&0.9985&0.971&0.268&0.974\\
				
				\thickhline
		\end{tabular}}
	\end{center}
	\vspace{-0.5cm}
	\caption{
		{\color{black}
			Variables with top 10 largest $\widehat{\PVE}_c(\bd{x}_k^{[i]})$ (the left half table)
			or $\widehat{\PVE}_d(\bd{x}_k^{[i]})$ (the right half table) for each of the three TCGA breast cancer genomic data types.
			The SWISS score shows the separation of PAM50 subtypes; a lower score indicates a better separation.}
	}	
	\label{PVEtop10 table}
\end{table}

{\color{black}
For each genomic data type,
Table~\ref{PVEtop10 table}
lists the top 10 variables most influenced by
common latent factors
and those by distinctive latent factors
according to their explained variable-level proportions
of signal variance, $\{\widehat{\PVE}_c(\bd{x}_k^{[i]})\}_{i=1}^{p_k}$
or $\{\widehat{\PVE}_d(\bd{x}_k^{[i]})\}_{i=1}^{p_k}$.
Table~\ref{PVEtop10 table} also reports
the SWISS scores
for the data $\widehat{\mb{X}}_k^{[i,:]}$, $\widehat{\mb{C}}_k^{[i,:]}$
and $\widehat{\mb{D}}_k^{[i,:]}$ of each selected variable to
corroborate the influences of those underlying mechanisms,
because the PAM50 subtype separation has been shown above as a good indicator of the common underlying mechanism.
Indeed,
$\text{SWISS}(\widehat{\mb{C}}_k^{[i,:]})$ is significantly smaller than $\text{SWISS}(\widehat{\mb{D}}_k^{[i,:]})$ (p-value $<0.05$)
for all selected variables
except for the gene TAS1R3
that has comparable scores
0.745 and 0.732.
For the top 10 variables
with largest $\widehat{\PVE}_c(\bd{x}_k^{[i]})$ values $>40\%$,
their signal data $\widehat{\mb{X}}_k^{[i,:]}$s well inherit 
from their $\widehat{\mb{C}}_k^{[i,:]}$s
the ability to separate the PAM50 subtypes
with small SWISS scores $\le 0.424$,
confirming the considerable influence
of the common underlying mechanism on these variables.
The top 10 variables with largest $\widehat{\PVE}_d(\bd{x}_k^{[i]})$ values all have much less informative
$\text{SWISS}(\widehat{\mb{X}}_k^{[i,:]})\ge 0.708$
nearly the same as $\text{SWISS}(\widehat{\mb{D}}_k^{[i,:]})$ and therefore are almost immune to the influence from the common underlying mechanism, which is consistent with their negligible $\widehat{\PVE}_c(\bd{x}_k^{[i]})$ values $<1.5\%$.
}

\subsection{Application to HCP motor-task functional MRI}
We consider the motor-task functional MRI data obtained from the HCP \citep{Barc13}.
During the image scanning, each of 1080 participants was asked by visual cues to either tap left or right fingers, or squeeze left or right toes, or move their tongue.
From the acquired brain images,
the HCP generated for every participant the $z$-statistic maps of the individual contrasts of the five tasks and also their average contrast against the fixation baseline.
The average contrast represents the impact of the overall motor task.
All the maps were computed at 91,282 grayordinates including 59,412 cortical surface vertices and 31,870 subcortical gray matter voxels.
For each task, its $z$-statistic maps of all participants constitute a 91,282$\times$1080 data matrix.
We focus on the left-hand, right-hand, and overall motor tasks, and aim to discover the brain regions that are most affected by their common underlying mechanism.

The D-GCCA method 
is applied to the three data matrices of interest
that are row-centered beforehand,
with nuisance parameters selected by
the approach discussed in Section~\ref{subsec: rank and set selection}. 
The selection approach yields the same decomposition by the choices 0.2 and 0.0001 for the significance level uniformly applied to all involved tests.
All signal and common-source matrix estimates are rank-2.
{\color{black}
The distinctive-source random vectors 
of the left-hand and   right-hand tasks 
are tested to be  uncorrelated by the approach in 
Section~\ref{sec: siml comp},
and thus the common-source variation of
the three tasks is fully captured by
their common-source random vectors.}
The estimated view-level proportion of signal variance explained by common latent factors, $\widehat{\PVE}_c(\bd{x}_k)$,
has values 0.122, 0.120 and 0.128, respectively, for the left-hand, right-hand and overall motor tasks.
This quantity reflects the overall influence of the common underlying mechanism on the $k$th considered motor~task.

\begin{figure}[p!]
	\centering
	\includegraphics[width=0.7\textwidth]{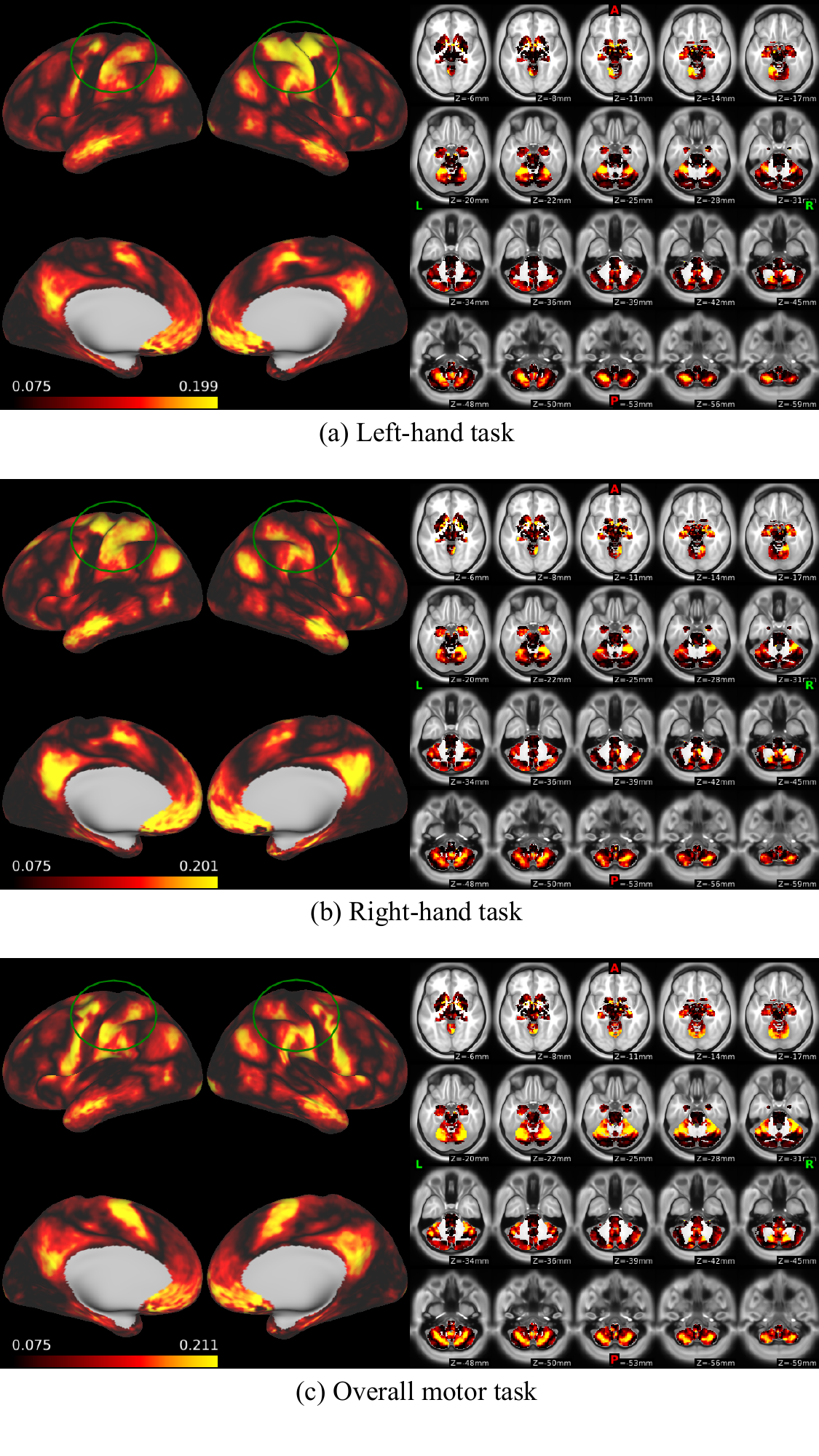}
	\caption{Maps of $\{\widehat{\PVE}_c(\bd{x}_k^{[i]})\}_{i=1}^{91282}$ from D-GCCA for the three HCP motor tasks.
		In each subfigure, the left part displays the cortical surface with the outer side shown in the first row and the inner side in the second row;
		the right part shows the subcortical area on 20 $xy$ slides at the $z$ axis.
		The somatomotor cortex is annotated by green circles. }
	\label{local PVE}
\end{figure}

To assess the local influence of the common underlying mechanism on the $i$th brain 
grayordinate of the $k$th task, we use 
$\widehat{\PVE}_c(\bd{x}_k^{[i]})$
the estimated variable-level proportion of signal variance explained
by common latent factors.
Figure~\ref{local PVE} illustrates
the map of $\{\widehat{\PVE}_c(\bd{x}_k^{[i]})\}_{i=1}^{91282}$ for each task.
In Figure~\ref{local PVE}\,(a) for the left-hand task,
we see that the common underlying mechanism has stronger impacts on the right cortical surface, particularly, the somatomotor cortex in the right green circle,
whereas it affects more on the left subcortical regions such as the cerebellum shown in the first and last rows of the right part of the figure.
The influence pattern is almost opposite for the right-hand task, and is nearly symmetric on the two sides of the brain for the overall motor task.
The contralateral change in the somatomotor cortex and the cerebellum
is consistent with their intrinsic functional connectivity shown in \citet{Buck11}.

On this large-scale data,
we also compare the computational performance 
of our D-GCCA and the six competing methods mentioned in Section~\ref{sec: intro}. 
All methods were implemented separately on a computing node with two 10-core Intel Xeon E5-2690v2 3.0GHz CPUs, total 62GB memory, and 24-hour time limit. 
{\color{black} The three methods, JIVE (with 5.47 hours), R.JIVE (with 17.4 hours) and DISCO-SCA (out of 24 hours), all involving time-expensive iterative optimization,
cannot converge within 5 hours}.
The OnPLS method 
ran out of memory due to
computing the SVD
of each large matrix $\mb{Y}_{j}\mb{Y}_{k}^\top$ for $j\ne k$.
Both D-GCCA and AJIVE have closed-form expressions, and COBE uses a fast alternating optimization strategy. 
The computational time costs of the D-GCCA, AJIVE and COBE methods are {\color{black}18.0, 180.5 and 25.3 seconds}, respectively.
However, the AJIVE and COBE methods were unable to identify nonzero common-source matrices.

{\color{black}
\section{Conclusion}\label{Sec: conclusion}
In this paper, we propose a novel decomposition method, called D-GCCA, to separate the common and distinctive variation structures of two or more data views on the same objects. 
In contrast with existing methods,
we build the decomposition on $(\mathcal{L}_0^2,\cov)$
rather than the traditional $(\mathbb{R}^n,\cdot)$,
 and particularly impose a certain orthogonality constraint on the distinctive latent factors to better capture the common-source variation, along with a geometric interpretation from PCA for the associated common latent factors.  
 Asymptotic result of proposed estimation under high-dimensional settings is established and supported by simulations. Moreover, the D-GCCA decomposition has a closed-form expression and thus is more computationally efficient, especially for large-scale data, than most existing methods with  time-expensive iterative optimization. 
Simulated and real-world data show the advantages of D-GCCA over state-of-the-art methods in capturing the common-source variation and also in the computational time cost. 
}

%Deep GCCA \citep{Bent19}, Kernel GCCA \citep{Tene15}

% Acknowledgements should go at the end, before appendices and references

\acks{{\color{black}
Dr. Shu's work was partially supported by the NIH grant R21AG070303. 
Dr. Zhu's work was partially supported by NIH grants R01MH086633 and R01MH116527.
The  content  is  solely the responsibility of the authors and does not necessarily represent the official views of the NIH.
We
are grateful to the Action Editor Dr. Qiaozhu Mei and three
anonymous reviewers for their constructive comments that greatly reshaped this article.}}

% Manual newpage inserted to improve layout of sample file - not
% needed in general before appendices/bibliography.

\appendix

% Note: in this sample, the section number is hard-coded in. Following
% proper LaTeX conventions, it should properly be coded as a reference:

%In this appendix we prove the following theorem from
%Section~\ref{sec:textree-generalization}:

{\color{black}
\section{A Hierarchical Extension}\label{supp: hierarchical structure}
The hierarchical decomposition structure in Section~\ref{sec: t-th level decomp} is illustrated in Figure~\ref{Fig: D-GCCA high levels}.

\begin{figure}[th!]\bigskip
	\centering
	\includegraphics[width=1\textwidth]{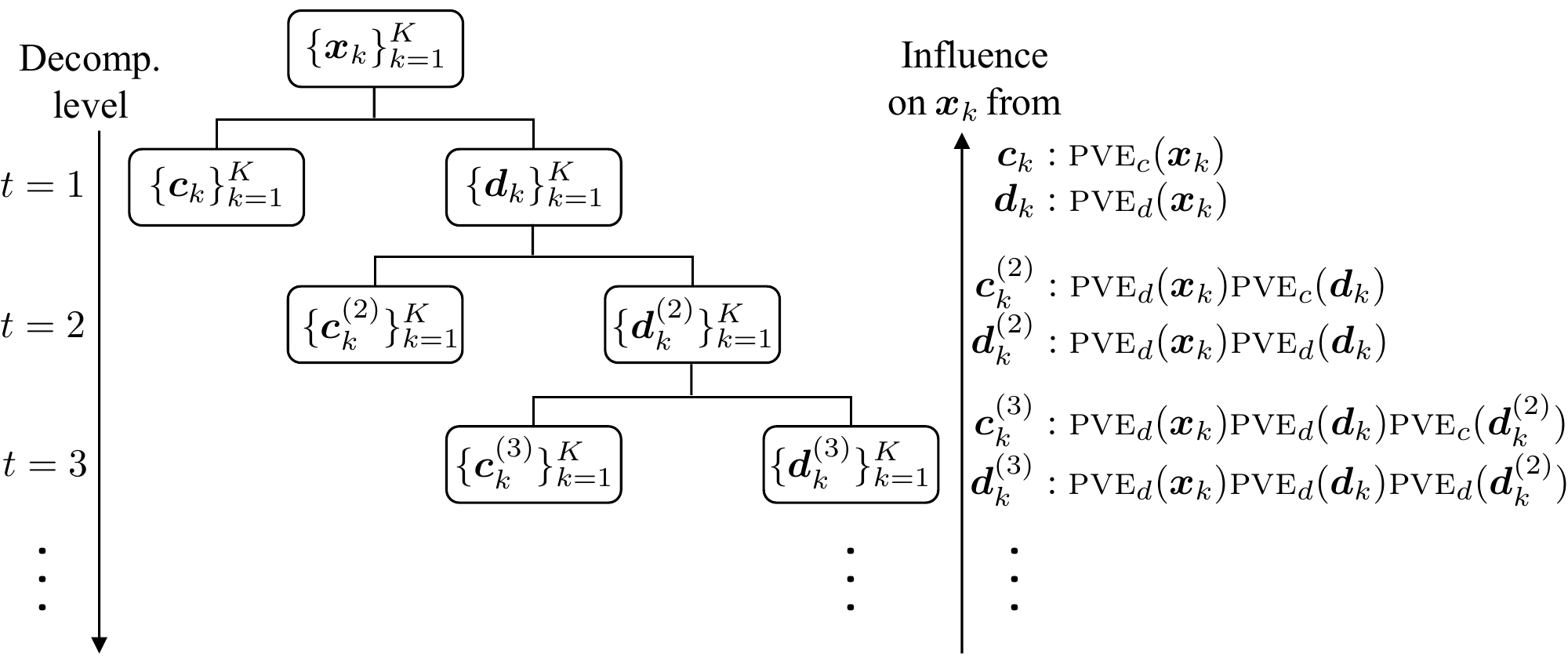}
	\caption{{\color{black}A hierarchical extension of D-GCCA.}}
	\label{Fig: D-GCCA high levels}
\end{figure}

For the $(t+1)$th-level decomposition ($t\ge 1$), 
recall that the view-level explained proportion of $\bd{d}_k^{(t)}$'s variance $\PVE_c(\bd{d}_k^{(t)})=1-\PVE_d(\bd{d}_k^{(t)}):=\tr\{\cov(\bd{c}_k^{(t+1)})\}/\tr\{\cov(\bd{d}_k^{(t)})\}$, and the variable-level explained proportion of variance $\PVE_c([\bd{d}_k^{(t)}]^{[i]})=1-\PVE_d([\bd{d}_k^{(t)}]^{[i]}):=\var([\bd{c}_k^{(t+1)}]^{[i]})/\var([\bd{d}_k^{(t)}]^{[i]})$.
Denote the sample matrices and their estimators
of $(\bd{c}_k^{(t)}, \bd{d}_k^{(t)})$ by 
 $(\mb{C}_k^{(t)},\mb{D}_k^{(t)})$
and $(\widehat{\mb{C}}_k^{(t)},\widehat{\mb{D}}_k^{(t)})$,
and the estimators 
of $(\PVE_c(\bd{d}_k^{(t)}),\PVE_d(\bd{d}_k^{(t)}), \PVE_c([\bd{d}_k^{(t)}]^{[i]}), $
$\PVE_d([\bd{d}_k^{(t)}]^{[i]}))$
by $(\widehat{\PVE}_c(\bd{d}_k^{(t)}),\widehat{\PVE}_d(\bd{d}_k^{(t)}),$
$\widehat{\PVE}_c([\bd{d}_k^{(t)}]^{[i]}), \widehat{\PVE}_d([\bd{d}_k^{(t)}]^{[i]}))$.
We define estimators $(\widehat{\mb{C}}_k^{(t+1)},\widehat{\mb{D}}_k^{(t+1)})$ 
in the same way as $(\widehat{\mb{C}}_k,\widehat{\mb{D}}_k)$ given in Section~\ref{subsec: mat est} by replacing $\widehat{\mb{X}}_k$ with 
$\widehat{\mb{D}}_k^{(t)}$, 
where $\widehat{\mb{D}}_k^{(1)}=\widehat{\mb{D}}_k$,
and define
$\widehat{\PVE}_c(\bd{d}_k^{(t)})=1-\widehat{\PVE}_d(\bd{d}_k^{(t)})=\|\widehat{\mb{C}}_k^{(t+1)} \|_F^2/\|\widehat{\mb{D}}_k^{(t)} \|_F^2$
and 
$\widehat{\PVE}_c([\bd{d}_k^{(t)}]^{[i]})=1-\widehat{\PVE}_d([\bd{d}_k^{(t)}]^{[i]})=\|[\widehat{\mb{C}}_k^{(t+1)} ]^{[i,:]}\|_F^2/\|[\widehat{\mb{D}}_k^{(t)}]^{[i,:]} \|_F^2$.
The corresponding
nuisance parameters are selected in the same fashion
as in Section 3.3.

We have the following asymptotic properties for the above estimators. 

\begin{cor}\label{consistency cor}
Suppose that Assumption~\ref{assump1} holds and
the other conditions on $\{\bd{x}_k\}_{k=1}^K$ for \eqref{RNE of C and D}-\eqref{eq: dataset-level PVE} in Theorem~\ref{consistency thm}
are also satisfied on $\{\bd{d}_k^{(t)}\}_{k=1}^K$ for all $0\le t\le T$
with a fixed number $T\ge 1$.
For all $1\le k\le K$ and $1\le t\le T$,
further assume that
the distinct eigenvalues of $\cov(\bd{d}_k^{(t)})$, denoted
by $\lambda_{k,1}^{(t)}>\dots>\lambda_{k,m_k^{(t)}+1}^{(t)}=0$,
satisfy
$\lambda_{k,1}^{(t)}>\kappa^{(t)}\lambda_{r_k}(\cov(\bd{x}_k))$, $\lambda_{k,1}^{(t)}\asymp\lambda_{k,m_k^{(t)}}^{(t)}$, and 
$(\lambda_{k,\ell}^{(t)}-\lambda_{k,\ell+1}^{(t)})/\lambda_{k,\ell}^{(t)}\ge \delta^{(t)}$ for $1\le \ell\le m_k^{(t)}$
with constants $\kappa^{(t)},\delta^{(t)}>0$.
If $\delta_\eta=o(1)$, then
\be\label{level-t: RNE of C and D}
	\max\left\{ \frac{\|\widehat{\mb{C}}_k^{(T+1)}-\mb{C}_k^{(T+1)}\|_{\star}^2}{\|\mb{D}_k^{(T)} \|_{\star}^2}
	,\frac{\|\widehat{\mb{D}}_k^{(T+1)}-\mb{D}_k^{(T+1)}\|_{\star}^2}{\|\mb{D}_k^{(T)} \|_{\star}^2} \right\}
	=O_P({\color{black}\delta_\eta^2})
\ee
and
	\be\label{level-t: dataset-level PVE}
{\color{black}\left|\widehat{\PVE}_c(\bd{d}_k^{(T)})-\PVE_c(\bd{d}_k^{(T)})\right|
	=O_P(\delta_\eta).}
	\ee
Additionally, if the nonzero eigenvalues of $\cov(\bd{d}_k^{(T)})$ are distinct, a basis of $\lspan([\bd{d}_k^{(T)}]^\top)$ 
has all elements with the sub-Gaussian norm bounded from above,
$\min_{ i\le p_k}\var([\bd{d}_k^{(T)}]^{[i]})\ge M_k^{(T)}
\lambda_{k,m_k^{(T)}}^{(T)}/p_k$ with a constant $M_k^{(T)}>0$,
and $\delta_k=o(1)$, then we have
\be\label{level-t: variable-level PVE}
\max_{1\le i\le p_k}\left|\widehat{\PVE}_c([\bd{d}_k^{(T)}]^{[i]})-\PVE_c([\bd{d}_k^{(T)}]^{[i]})\right|=O_P(\delta_\eta+\delta_k).
\ee
\end{cor}

In Corollary~\ref{consistency cor}, the condition
$\lambda_{k,1}^{(t)}>\kappa^{(t)}\lambda_{r_k}(\cov(\bd{x}_k))$ implies that the variance ratio $\tr\{\cov(\bd{d}_k^{(t)})\}/\tr\{\cov(\bd{x}_k)\}$ is bounded away from zero
and hence is worth the $(t+1)$th-level decomposition.
The other conditions on $\{\bd{d}_k^{(t)}\}_{k=1}^K$
are similar to those in Assumption~\ref{assump1} and Theorem~\ref{consistency thm}.

}

\section{Theoretical Proofs}\label{sec: Proofs}
\begin{proof}{\bf of Theorem~\ref{GCCA thm}.}
%	It is easily seen that $\sum_{k=1}^K \cos^2\theta(z_1,z_k)\ge 1$.
Consider stage $\ell\le r_f$.
	If $w\perp \lspan(\bd{f}^\top)$, then $\sum_{k=1}^K \cos^2\{\theta(w,z_k)\}=0$, and thus this $w$ is not optimal because there always exists
another $w\in \lspan(\bd{f}^\top)$ and $z_k\in \lspan(\bd{x}_k^\top)$, $k=1,\dots,K$, such that $\sum_{k=1}^K \cos^2\{\theta(w,z_k)\}>0$
for stage $\ell\le r_f$.
	When $w\not\perp \lspan(\bd{f}^\top)$,
	since
	$\cos\{\theta(w,z_k)\}=\cos\{\theta(w,w_0)\}\cos\{\theta(w_0,z_k)\}$, where
	$w_0$ denotes the projection of $w$ onto $\lspan(\bd{f}^\top)$,
	we only need to consider $w\in \lspan(\bd{f}^\top)$.
	Then there exists a vector $\bd{b}=(b_1,\dots,b_K)^\top$ such that
	$w= \bd{b}^\top \bd{f}$ 
	and $\cov(w)=\bd{b}^\top\cov(\bd{f})\bd{b}=1$.
	Let $z_k^*$ be the projection of $w$ onto $\lspan(\bd{x}_k^\top)$.
	We only need to consider $z_k$ such that
	\[
	z_k
	\begin{cases}
	= \text{any standardized variable in}~ \lspan(\bd{x}_k^\top),~~&\text{if}~~z_k^*=0,\\
	\propto z_k^*, ~~&\text{if}~~z_k^*\ne0.\\
	\end{cases}
	\]
	Define $\mb{\Phi}_k=(\mb{0}_{r_k\times \sum_{j=1}^{k-1}r_j  },\mb{I}_{r_k\times r_k},\mb{0}_{r_k\times \sum_{j=k+1}^{K}r_j })$.
	Then $\bd{f}_k=\mb{\Phi}_k\bd{f}$ and $\mb{I}_{\sum_{k=1}^K r_k \times \sum_{k=1}^K r_k }= \sum_{k=1}^K   \mb{\Phi}_k^\top\mb{\Phi}_k $.
	Note that the inner product
	$
	\langle w,\bd{f}_k\rangle=\cov(w,\bd{f}_k)=\cov(\bd{b}^\top \bd{f}, \mb{\Phi}_k\bd{f})=\bd{b}^\top \cov(\bd{f})\mb{\Phi}_k^\top,
	$
	which is zero if $z_k^*=0.$
	We have
	\begin{align}
	&z_k^*= \langle w,\bd{f}_k\rangle \bd{f}_k
	= \bd{b}^\top \cov(\bd{f})\mb{\Phi}_k^\top \mb{\Phi}_k\bd{f},
	\label{z* formula}\\
	&\var(z_k^*)=\langle w,\bd{f}_k\rangle\cov(\bd{f}_k)\langle w,\bd{f}_k\rangle^\top
	=\bd{b}^\top \cov(\bd{f})\mb{\Phi}_k^\top\mb{\Phi}_k\cov(\bd{f})\bd{b},
	\nonumber\\
	&\cov(w,z_k^*) =  \bd{b}^\top \cov(\bd{f})  \mb{\Phi}_k^\top \mb{\Phi}_k     \cov(\bd{f}) \bd{b},   
	\label{cov(w,z*)}\nonumber\\
	&\corr^2(w,z_k^*) = \bd{b}^\top \cov(\bd{f})  \mb{\Phi}_k^\top \mb{\Phi}_k     \cov(\bd{f}) \bd{b},  
	\end{align}
	and then
	\be\label{b*cov2*b}
	\sum_{k=1}^K  \cos^2\{\theta(w,z_k) \} =\sum_{k=1}^K\corr^2(w,z_k^*) = \bd{b}^\top \cov^2(\bd{f}) \bd{b} .
	\ee
	Let $w^{(\ell)}=(\bd{b}^{(\ell)})^\top\bd{f}$.
	To maximize \eqref{b*cov2*b} with respect to $\bd{b}$ under the constraints $\bd{b}^\top\cov(\bd{f})\bd{b}=1$ and $\bd{b}^\top\cov(\bd{f})\bd{b}^{(j)}=0$ for $j\le\ell-1$,
	the associated Lagrange function from the method of Lagrange multipliers is 
	\[
	\mathcal{L}(\bd{b},l_1,\dots,l_\ell)=\bd{b}^\top \cov^2(\bd{f}) \bd{b} -
	l_\ell(\bd{b}^\top\cov(\bd{f})\bd{b}-1)
	-\sum_{j=1}^{\ell-1}l_j \bd{b}^\top\cov(\bd{f})\bd{b}^{(j)}.
	\]
	There exist $l_1^{(\ell)},\dots,l_\ell^{(\ell)}$ such that
	$\nabla \mathcal{L}(\bd{b}^{(\ell)},l_1^{(\ell)},\dots,l_\ell^{(\ell)})=\bd{0}$, which yields
\begin{subequations}\label{b*cov2*b Lag}
 \begin{empheq}[left={\empheqlbrace\,}]{align}
&		2\cov^2(\bd{f}) \bd{b}^{(\ell)} = 2 l_\ell^{(\ell)} \cov(\bd{f})\bd{b}^{(\ell)} +   \sum_{j=1}^{\ell-1}l_j^{(\ell)} \cov(\bd{f})\bd{b}^{(j)},  \label{b*cov2*b Lag1}  \\
&		(\bd{b}^{(\ell)})^\top\cov(\bd{f})\bd{b}^{(\ell)}=1,   \label{b*cov2*b Lag2}  \\
&		(\bd{b}^{(\ell)})^\top\cov(\bd{f})\bd{b}^{(j)}=0,~~\text{for}~~j=1,\dots,\ell-1. 
	\end{empheq}
\end{subequations}
	When $\ell=1$, \eqref{b*cov2*b Lag1} becomes
	$
	\cov^2(\bd{f}) \bd{b}^{(1)} =l_1^{(1)} \cov(\bd{f})\bd{b}^{(1)}.
	$
	Then by \eqref{b*cov2*b Lag2}, we have $l_1^{(1)}= (\bd{b}^{(\ell)} )^\top \cov^2(\bd{f}) \bd{b}^{(\ell)} $.
	Thus, the maximum of \eqref{b*cov2*b} when $\ell=1$, i.e., the maximum of $l_1^{(1)}$ is $l_{f,1}\coloneqq \eig_1(\cov(\bd{f}))$.
	We have $ l_{f,1}^{-1/2}\cov(\bd{f}) \bd{b}^{(1)} =\bd{\eta}^{(1)}$.
	Hence, $\bd{b}^{(1)}=l_{f,1}^{1/2}[\cov(\bd{f})]^\dag \bd{\eta}^{(1)}+ \bd{\zeta}$
	for any vector $\bd{\zeta}$ satisfying $\mb{V}_{f}^\top\bd{\zeta}=\bd{0}$, where 
	$\cov(\bd{f})=\mb{V}_{f}\mb{\Lambda}_{f} \mb{V}_{f}^\top $ is
	a compact SVD of $\cov(\bd{f})$,
	and $[\cov(\bd{f})]^\dag=\mb{V}_{f}\mb{\Lambda}_{f}^{-1} \mb{V}_{f}^\top$ is the pseudo-inverse of $\cov(\bd{f})$.
	Let $\bd{u}=  \mb{\Lambda}_{f}^{-1/2} \mb{V}_{f}^\top   \bd{f}$.
	Then $\bd{f}=\cov(\bd{f},\bd{u})\bd{u}=\mb{V}_{f}\mb{\Lambda}_{f}^{1/2}\bd{u}$.
	We have
	\begin{align*}
	w^{(1)}=(\bd{b}^{(1)})^\top \bd{f}&=
	(l_{f,1}^{1/2}(\bd{\eta}^{(1)})^\top[\cov(\bd{f})]^\dag + \bd{\zeta}^\top)\mb{V}_{f}\mb{\Lambda}_{f}^{1/2}\bd{u}\\
	&=l_{f,1}^{1/2}(\bd{\eta}^{(1)})^\top [\cov(\bd{f})]^\dag \mb{V}_{f}\mb{\Lambda}_{f}^{1/2}\bd{u}\\
	&=l_{f,1}^{-1/2} (\bd{\eta}^{(1)})^\top\cov(\bd{f})[\cov(\bd{f})]^\dag\mb{V}_{f}\mb{\Lambda}_{f}^{1/2}\bd{u}\\
	&=l_{f,1}^{-1/2} (\bd{\eta}^{(1)})^\top\mb{V}_{f}\mb{\Lambda}_{f}^{1/2}\bd{u}\\
	&=l_{f,1}^{-1/2} (\bd{\eta}^{(1)})^\top\bd{f}.
	\end{align*}
	Hence, we can simply let $\bd{b}^{(1)}=l_{f,1}^{-1/2}\bd{\eta}^{(1)}$.
	When $\ell=2$, left-multiplying \eqref{b*cov2*b Lag1} by $\bd{b}^{(1)}$ yields $l_1^{(2)}=0$.
	Then \eqref{b*cov2*b Lag} becomes
	\[
	\begin{cases}
	\cov^2(\bd{f}) \bd{b}^{(2)} =  l_2^{(2)} \cov(\bd{f})\bd{b}^{(2)},  \\
	(\bd{b}^{(2)})^\top\cov(\bd{f})\bd{b}^{(2)}=1,   \\
	(\bd{b}^{(2)})^\top\cov(\bd{f})\bd{b}^{(1)}=0.
	\end{cases}
	\]
	Thus, we have $[\eig_2(\cov(\bd{f}))]^{-1/2}\cov(\bd{f})\bd{b}^{(2)}=\bd{\eta}^{(2)}$.
	Then using the same skill for obtaining $\bd{b}^{(1)}$,
	we can simply let 
	$\bd{b}^{(2)}=[\eig_2(\cov(\bd{f}))]^{-1/2} \bd{\eta}^{(2)}$
	and have $\sum_{k=1}^K\cos^2\{\theta(w^{(2)},z_k^{(2)})\} = \eig_2(\cov(\bd{f}))$.
	Similarly, 
	for $2<\ell\le r_f$,
	we can simply let $\bd{b}^{(\ell)}=[\eig_\ell(\cov(\bd{f}))]^{-1/2} \bd{\eta}^{(\ell)}$
	and have $\sum_{k=1}^K\cos^2\{\theta(w^{(\ell)},z_k^{(\ell)})\} = \eig_\ell(\cov(\bd{f}))$.
	%For $\ell>L$, $\sum_{k=1}^K\cos^2(w^{(\ell)},z_k^{(\ell)}) =0$, and thus $w^{(\ell)} \perp \sum_{k=1}^K \lspan(\bd{x}_k^\top)$.
	%which contradicts our beginning assumption that $w^{(\ell)}\in \lspan(\bd{f})$. Hence, if $w^{(\ell)}$ exists for $\ell>L$, then $w^{(\ell)} \perp \sum_{k=1}^K \lspan(\bd{x}_k^\top)$.
	
	For $\ell\le r_f$, by \eqref{z* formula}, the projection of $w^{(\ell)}$ onto space $\lspan(\bd{x}_k^\top)$
	is 
	$
	{z_k^*}^{(\ell)}=
	[\eig_\ell(\cov(\bd{f}))]^{1/2} (\bd{\eta}_k^{(\ell)})^\top\bd{f}_k
	$
	with $\var({z_k^*}^{(\ell)})= \eig_\ell(\cov(\bd{f}))\|\bd{\eta}_k^{(\ell)}\|_F^2$.
	Thus, 
	\[
	z_k^{(\ell)}
	= \begin{cases}
	\text{any standardized variable in}~\lspan(\bd{x}^\top) ,&~~\text{if}~ \bd{\eta}_k^{(\ell)}= \bd{0},\\
	\pm (\bd{\eta}_k^{(\ell)}/\|\bd{\eta}_k^{(\ell)}\|_F)^\top\bd{f}_k, &~~\text{if}~ \bd{\eta}_k^{(\ell)}\ne \bd{0}.
	\end{cases}
	\]
	From equation \eqref{cov(w,z*)}, we have $\cov(w^{(\ell)},{z_k^*}^{(\ell)})= \eig_\ell(\cov(\bd{f}))\| \bd{\eta}_k^{(\ell)} \|_F^2   $.
	Then,
	$
	\cos\{\theta(w^{(\ell)},z_k^{(\ell)})\}
	=\pm   [\eig_{\ell}(\cov(\bd{f}))]^{1/2}\| \bd{\eta}_k^{(\ell)} \|_F.
	$
	
	To prove $\sum_{k=1}^K\lspan(\bd{x}_k^\top)=\lspan(\{w^{(\ell)}\}_{\ell=1}^{r_f})$,	
	since $w^{(\ell)}\in \lspan(\bd{f}^\top)$, we only need to show $\dim(\lspan(\{w^{(\ell)}\}_{\ell=1}^{r_f}))=\dim(\lspan(\bd{f}^\top))=r_f$, which is true because
	the $r_f$ nonzero variables $\{w^{(\ell)}\}_{\ell=1}^{r_f}$ are orthogonal.
	
	Now consider the revised $z_k^{(\ell)}$ in \eqref{z_k revised} for result~\ref{GCCA thm(ii)}.
	By	
	$\cos\{\theta(w^{(\ell)},z_k^{(\ell)})\}
	= [\eig_{\ell}(\cov(\bd{f}))]^{1/2}\| \bd{\eta}_k^{(\ell)} \|_F$
	$\ge 0$, we have 
	$\theta(w^{(\ell)},z_k^{(\ell)})\in [0,\pi/2]$.
	Since $\lspan(\{z_k^{(\ell)}\}_{\ell=1}^{r_f})$ is the projection of $\lspan(\{w^{(\ell)}\}_{\ell=1}^{r_f})$ onto $\lspan(\bd{x}_k^\top)\subseteq \lspan(\{w^{(\ell)}\}_{\ell=1}^{r_f})=\sum_{k=1}^K\lspan(\bd{x}_k^\top)$, we have $\lspan(\{z_k^{(\ell)}\}_{\ell=1}^{r_f})=\lspan(\bd{x}_k^\top)$.
	
	Next, consider result~\ref{GCCA thm(iii)}.
	For some $k$ and $\ell$, since $\lspan(\{z_{k}^{(m)}\}_{m=1}^{\ell-1})\ne\lspan(\bd{x}_{k}^\top) $, 
	there exists a unit-variance variable $v\in \lspan(\bd{x}_k^\top)$ such that 
	$v\perp \lspan(\{z_{k}^{(m)}\}_{m=1}^{\ell-1})$.
	Moreover, $v\perp w^{(m)}$ for all $m\le \ell-1$, because
	$v$ is orthogonal to both the projection of $w^{(m)}$ onto $\lspan(\bd{x}_k^\top)$
	and the rejection of $w^{(m)}$ from $\lspan(\bd{x}_k^\top)$.
	Thus, we just let $w^{(\ell)}=v$. Then, 
	$\cos^2\{\theta(w^{(\ell)},z_k^{(\ell)})\}=1$.
	By $\sum_{k=1}^K\cos^2\{\theta(w^{(\ell)},z_k^{(\ell)})\}=\eig_{\ell}(\cov(\bd{f}))\le1$,
	we have $\sum_{j\ne k}\cos^2\{\theta(w^{(\ell)},z_j^{(\ell)})\}=0$,
	which implies $w^{(\ell)}\perp \sum_{1\le j\ne k\le K} \lspan(\bd{x}_j^\top)$.
\end{proof}

\begin{proof}{\bf of Theorem~\ref{alpha thm}.}
	If $z_k^{(\ell)}=0$ for some $k$, it is easy to see $\alpha^{(\ell)}=0$.
	We only consider that for all $k\le K$, $z_k^{(\ell)}\ne 0$, i.e., $\theta(w^{(\ell)},z_k^{(\ell)})\in [0,\pi/2)$.
	If $d_j^{(\ell)}\perp d_k^{(\ell)}$, 
	then $\|d_j^{(\ell)}\|^2+\| d_k^{(\ell)}\|^2=\| z_j^{(\ell)}-z_k^{(\ell)} \|^2$,
	and consequently by the law of cosines we have	
	\begin{align*}	
	&\left(\|z_j^{(\ell)}\|^2+\| c^{(\ell)} \|^2-2\| z_j^{(\ell)}\|  \|c^{(\ell)}\| \sign(\alpha^{(\ell)})\cos\{\theta(z_j^{(\ell)},w^{(\ell)})\}\right)\\
	&+\left(\|z_k^{(\ell)}\|^2+\| c^{(\ell)} \|^2-2\| z_k^{(\ell)}\| \|c^{(\ell)}\|  \sign(\alpha^{(\ell)}) \cos\{\theta(z_k^{(\ell)},w^{(\ell)})\}\right)
	\\
	&\quad= 	\|z_j^{(\ell)}\|^2+\| z_k^{(\ell)} \|^2-2\| z_j^{(\ell)}\|  \|z_k^{(\ell)}\|  \cos\{\theta(z_j^{(\ell)},z_k^{(\ell)})\}
	\end{align*}
	which gives
	$
	\alpha^{(\ell)}=\frac{1}{2}\left[
	\cos\{\theta(z_j^{(\ell)},w^{(\ell)})\}
	+
	\cos\{\theta(z_k^{(\ell)},w^{(\ell)})\}
	\pm
	(
	\Delta_{jk}^{(\ell)}
	)^{1/2}
	\right].
	$ 	
	Hence, the desired value of $\alpha^{(\ell)}$ is the one
	given in Theorem~\ref{alpha thm}. 
	
	To prove the existence of $\alpha^{(\ell)}$, we only need to show that there exists a $\Delta_{jk}^{(\ell)}\ge 0$ with $j\ne k$. 
	Denote
	$\lambda_{\ell}=\eig_{\ell}(\cov(\bd{f}))$, and $\bd{\nu}_{\ell}=(\nu_{\ell,1},\dots,\nu_{\ell,K})^\top$ with $\nu_{\ell,k}=\| \bd{\eta}_k^{(\ell)} \|_F$.
	We have
	\[
	\cov(\bd{z}^{(\ell)})= \diag\big( \frac{(\bd{\eta}_1^{(\ell)})^\top }{\|\bd{\eta}_1^{(\ell)}\|_F}, \dots, \frac{(\bd{\eta}_K^{(\ell)})^\top}{\|\bd{\eta}_K^{(\ell)}\|_F} \big)  \cov(\bd{f}) 
	\diag\big( \frac{\bd{\eta}_1^{(\ell)} }{\|\bd{\eta}_1^{(\ell)}\|_F}, \dots, \frac{\bd{\eta}_K^{(\ell)}}{\|\bd{\eta}_K^{(\ell)}\|_F} \big),
	\]
{\color{black}	$$\bd{\nu}_\ell^\top\cov(\bd{z}^{(\ell)})\bd{\nu}_\ell=\lambda_\ell,~~~~
	\cos\{\theta(w^{(\ell)},z_k^{(\ell)})\}
	=\lambda_\ell^{1/2}\nu_{\ell,k},
	$$ } 
	and for all $j,k\le K$,
$
	\Delta_{jk}^{(\ell)}=
	\lambda_{\ell}\nu_{\ell,j}^2+\lambda_{\ell}\nu_{\ell,k}^2
	+2\lambda_{\ell}\nu_{\ell,j}\nu_{\ell,k}
	-4\cov(z_j^{(\ell)},z_k^{(\ell)}).
$ 
	Then, we have 
	\begin{align}\label{sum delta}
	\lefteqn{\sum_{j=1}^K\sum_{k=1}^K\cos\{\theta(w^{(\ell)},z_j^{(\ell)})\}\Delta_{jk}^{(\ell)}\cos\{\theta(w^{(\ell)},z_k^{(\ell)})\}}\nonumber\\
	&=\sum_{j=1}^K\sum_{k=1}^K\cov(w^{(\ell)},z_j^{(\ell)})\Delta_{jk}^{(\ell)}\cov(w^{(\ell)},z_k^{(\ell)})\nonumber\\
	&=\sum_{j=1}^K\sum_{k=1}^K \lambda_{\ell}^{1/2}\nu_{\ell,j}    \left( 		\lambda_{\ell}\nu_{\ell,j}^2+\lambda_{\ell}\nu_{\ell,k}^2
	+2\lambda_{\ell}\nu_{\ell,j}\nu_{\ell,k}
	-4\cov(z_j^{(\ell)},z_k^{(\ell)})   \right)\lambda_{\ell}^{1/2}\nu_{\ell,k}\nonumber\\
	&=\left[\sum_{j=1}^K\sum_{k=1}^K \lambda_{\ell}^{1/2}\nu_{\ell,j}  ( 		\lambda_{\ell}\nu_{\ell,j}^2+\lambda_{\ell}\nu_{\ell,k}^2
	+2\lambda_{\ell}\nu_{\ell,j}\nu_{\ell,k}
	)\lambda_{\ell}^{1/2}\nu_{\ell,k}
	\right]
	-4\lambda_{\ell} \bd{\nu}_\ell^\top \cov(\bd{z}^{(\ell)})   \bd{\nu}_\ell\nonumber\\
	&=\left[\sum_{j=1}^K\sum_{k=1}^K \lambda_{\ell}^{1/2}\nu_{\ell,j}  ( 		\lambda_{\ell}\nu_{\ell,j}^2+\lambda_{\ell}\nu_{\ell,k}^2
	+2\lambda_{\ell}\nu_{\ell,j}\nu_{\ell,k}
	)\lambda_{\ell}^{1/2}\nu_{\ell,k}
	\right]
	-4\lambda_{\ell}^2 \bd{\nu}_\ell^\top (\bd{\nu}_\ell\bd{\nu}_\ell^\top) \bd{\nu}_\ell\nonumber\\
	&=\sum_{j=1}^K\sum_{k=1}^K \lambda_{\ell}^{1/2}\nu_{\ell,j}    \left( 		\lambda_{\ell}\nu_{\ell,j}^2+\lambda_{\ell}\nu_{\ell,k}^2
	+2\lambda_{\ell}\nu_{\ell,j}\nu_{\ell,k}
	-4\lambda_{\ell}\nu_{\ell,j}\nu_{\ell,k}  \right)\lambda_{\ell}^{1/2}\nu_{\ell,k}\nonumber\\
	&=\sum_{j=1}^K\sum_{k=1}^K\cos\{\theta(w^{(\ell)},z_j^{(\ell)})\}(\lambda_{\ell}^{1/2}\nu_{\ell,j}-\lambda_{\ell}^{1/2}\nu_{\ell,k})^2\cos\{\theta(w^{(\ell)},z_k^{(\ell)})\}
	\nonumber\\
	&\ge 0.
	\end{align}
	For all $k< K$,
	$\cos\{\theta(w^{(\ell)},z_k^{(\ell)})\}>0$ for
	$\theta(w^{(\ell)},z_k^{(\ell)})\in [0,\pi/2)$,
	and moreover, we have
	$\Delta_{kk}^{(\ell)}=4\cos^2\{\theta(w^{(\ell)},z_k^{(\ell)})\}-4\le 0$. 
	Hence, by~\eqref{sum delta},
	we have at least one $\Delta_{jk}^{(\ell)}\ge 0$ with $j\ne k$.
\end{proof}

\begin{proof}{\bf of Theorem~\ref{equivalence to K=2}.}
	When $K=2$,
	by Lemma~2 in \cite{Kett71}, $L$ is equal to the number of positive canonical correlations between $\bd{x}_1$ and $\bd{x}_2$. 
	Then following the constructions of these two decomposition methods, the proof is easy to complete.
	The details are omitted.
\end{proof}

\begin{proof}{\bf of Theorem~\ref{model uniqueness}.}
	Let $\widetilde{\bd{f}}_k^\top$ be another orthonormal basis of $\lspan(\bd{x}_k^\top)$.
	Then, there exists an orthogonal matrix $\mb{O}_k$ such that $\widetilde{\bd{f}}_k=\mb{O}_k \bd{f}_k$. Define $\widetilde{\bd{f}}=[\widetilde{\bd{f}}_1;\dots;\widetilde{\bd{f}}_K]$.
	We have $\widetilde{\bd{f}}=\mb{O}\bd{f}$ and $\cov(\widetilde{\bd{f}})=\mb{O}\cov(\bd{f})\mb{O}^\top$
	with $\mb{O}=\diag(\mb{O}_1,\dots,\mb{O}_K)$.
	Hence, $\eig_\ell(\cov(\widetilde{\bd{f}}))=\eig_\ell(\cov(\bd{f}))$ for $\ell\le \sum_{k=1}^{K}r_k$.
	Denote $\widetilde{\bd{\eta}}^{(\ell)}=[\widetilde{\bd{\eta}}_1^{(\ell)};\dots;\widetilde{\bd{\eta}}_K^{(\ell)}]$, with $\widetilde{\bd{\eta}}_k^{(\ell)}\in \mathbb{R}^{r_k}$, to be a 
	normalized eigenvector of $\cov(\widetilde{\bd{f}})$ corresponding to $\lambda_{\ell}(\cov(\widetilde{\bd{f}}))$ for $\ell\le L$.
	Now, from the assumption that
	$\eig_1(\cov(\bd{f})),\dots,\eig_L(\cov(\bd{f}))$ are distinct,
	we have
	$\widetilde{\bd{\eta}}^{(\ell)}=\pm \mb{O}\bd{\eta}^{(\ell)}$
	and $\widetilde{\bd{\eta}}_k^{(\ell)}=\pm \mb{O}_k\bd{\eta}_k^{(\ell)}$.
	Denote $\widetilde{w}^{(\ell)}$, $\widetilde{z}_k^{(\ell)}$, $\widetilde{\alpha}^{(\ell)}$ and $\widetilde{c}^{(\ell)}$ to be the counterparts of $w^{(\ell)}$, $z_k^{(\ell)}$, $\alpha^{(\ell)}$ and $c^{(\ell)}$ that are defined in \eqref{w formula}, \eqref{z_k revised} and \eqref{c for K>=2}
	by using $\widetilde{\bd{f}}$ and $\widetilde{\bd{\eta}}^{(\ell)}$ instead of $\bd{f}$ and $\bd{\eta}^{(\ell)}$. 
	We have $\widetilde{w}^{(\ell)}=\pm w^{(\ell)}$, $\widetilde{z}_k^{(\ell)}=\pm z_k^{(\ell)}$,
	$\widetilde{\alpha}^{(\ell)}=\alpha^{(\ell)}$ 
	due to the formula in Theorem~\ref{alpha thm}, and then $\widetilde{c}^{(\ell)}=\pm c^{(\ell)}$.
	Let $\widetilde{\bd{z}}_k^{\mathcal{I}_0}=(\widetilde{z}_k^{(\ell)})_{\ell\in \mathcal{I}_0}^\top$
	and $\widetilde{\bd{c}}^{\mathcal{I}_0}=(\widetilde{c}^{(\ell)})_{\ell\in \mathcal{I}_0}^\top$.
	There exists a diagonal matrix $\mb{D}$ with diagonal entries being either $1$ or $-1$
	such that $\widetilde{\bd{z}}_k^{\mathcal{I}_0}=\mb{D}\bd{z}_k^{\mathcal{I}_0}$ and $\widetilde{\bd{c}}^{\mathcal{I}_0}=\mb{D}\bd{c}^{\mathcal{I}_0}$.
	Then,
	\begin{align*}
	\cov(\bd{x}_k,\widetilde{\bd{z}}_k^{\mathcal{I}_0})[\cov(\widetilde{\bd{z}}_k^{\mathcal{I}_0})]^\dagger\widetilde{\bd{c}}^{\mathcal{I}_0}
	&=\cov(\bd{x}_k,\bd{z}_k^{\mathcal{I}_0})\mb{D}[\mb{D}\cov(\bd{z}_k^{\mathcal{I}_0})\mb{D}]^\dag\mb{D}\bd{c}^{\mathcal{I}_0}\\
	&=\cov(\bd{x}_k,\bd{z}_k^{\mathcal{I}_0})\mb{D}[\mb{D}\mb{V}_{zk} \mb{\Lambda}_{zk}   \mb{V}_{zk}^\top \mb{D}]^\dag\mb{D}\bd{c}^{\mathcal{I}_0}\\
	&=\cov(\bd{x}_k,\bd{z}_k^{\mathcal{I}_0})\mb{D}[\mb{D}\mb{V}_{zk} \mb{\Lambda}_{zk}^{-1}   \mb{V}_{zk}^\top \mb{D}]\mb{D}\bd{c}^{\mathcal{I}_0}\\
	&=\cov(\bd{x}_k,\bd{z}_k^{\mathcal{I}_0})[\cov(\bd{z}_k^{{\color{black}\mathcal{I}_0}})]^\dag\bd{c}^{\mathcal{I}_0}=\bd{c}_k.
	\end{align*}
	The proof is complete. 
\end{proof}

\begin{proof}{\bf of Theorem~\ref{consistency thm}.}
	First of all, it is worth mentioning that
	$\widehat{\mb{X}}_k$ is rank-$r_k$ with probability tending to 1.
	This is because
	we have		
	\[
	\eig_{r_k}(\widehat{\cov}(\bd{x}_k))\ge (1-o_P(1))\eig_{r_k}(\cov(\bd{x}_k))
	\]
	from (S.17) in the supplement of \cite{Shu17}.
	Due to their Lemma~S.1, in the rest of the proof we simply assume that
	$\widehat{\mb{X}}_k$ is rank-$r_k$.
	
{\color{black}
For the convergence results of $\{\widehat{\mb{X}}_k, \widehat{\mb{C}}_k,\widehat{\mb{D}}_k\}$,
we will follow the similar proof techniques of Theorem~3 in \citet{Shu17}.
The key difference is that
our $\mb{C}_k$
and $\widehat{\mb{C}}_k$ are defined
from Carroll's GCCA for $K\ge 2$ 
which are more complex than  
those in \citet{Shu17} from CCA for $K=2$. Hence, our proof needs
extra effort to
establish the error bounds of each component in $\widehat{\mb{C}}_k$
defined in \eqref{C estimator} and then combine them together to yield the final error bound for $\widehat{\mb{C}}_k$.
Moreover, to the best of our knowledge,
the results 
in \eqref{eq: dataset-level PVE}-\eqref{eq: variable-level PVE}
are the first work to show the high-dimensional estimation consistency
of the view-level and variable-level
proportions of explained signal variance
for the decomposition model in \mbox{\eqref{decomp in mat}-\eqref{decomp in vari}} for $K\ge 2$,
which are not seen in \citet{Shu17} even when $K=2$.
In particular, the uniform consistency of the variable-level proportions of explained signal variance given in \eqref{eq: variable-level PVE} will be derived from the $\ell_\infty$ eigenvector perturbation bound recently given in \mbox{\citet{fan2018eigenvector}}.}

\begin{enumerate}[wide, labelwidth=!, labelindent=0pt]
\item We first consider the error bounds of $\widehat{\mb{X}}_k$.	

	By (S.13) and (S.14) in \cite{Shu17},
	there exists a constant $\kappa_x>0$ such that 
	\be\label{scale of norm of X}
	\kappa_x+o_P(1)\le \frac{\|\mb{X}_k \|_2}{ [ n\eig_1(\cov(\bd{x}_k)) ]^{1/2}   }
	\le \frac{\|\mb{X}_k \|_F}{   [n\eig_1(\cov(\bd{x}_k))  ]^{1/2} }
	\le r_k^{1/2}+o_P(1).
	\ee
	From their (S.15), we have
	\begin{align}\label{diff in norm of X}
	\lefteqn{\|\widehat{\mb{X}}_k-\mb{X}_k\|_2\le \|\widehat{\mb{X}}_k-\mb{X}_k\|_F}\nonumber\\
	&\lesssim_P
	\min\left\{
	\Big[\frac{\eig_1(\cov(\bd{x}_k))}{n}\Big]^{1/2}
	+   (p_k\log p_k )^{1/2}  , [ n\eig_1(\cov(\bd{x}_k)) ]^{1/2}  
	\right\}.
	\end{align}
	From (S.7) of \cite{Shu17}, we have $\lambda_1(\cov(\bd{x}_k))\asymp \lambda_{r_k}(\cov(\bd{x}_k))$.
	By Weyl's inequality (see Theorem 3.3.16(a) in \cite{Horn94}) as well as Assumption~\ref{assump1}~(i) and~(v), 
	$\kappa_1\le \lambda_{k,p_k}= \lambda_{k,(r_k+1)+(p_k-r_k)-1}-\lambda_{r_k+1}(\cov(\bd{x}_k))\le \lambda_{p_k-r_k}(\cov(\bd{e}_k))\le \lambda_1(\cov(\bd{e}_k))=\|\cov(\bd{e}_k)\|_2\le\|\cov(\bd{e}_k)\|_\infty\le s_0 $.
	Thus,
	\be\label{lambda_k=R_k}
\frac{	\lambda_1(\cov(\bd{x}_k))}{p_k}\asymp \frac{\tr(\cov(\bd{x}_k))}{\tr(\cov(\bd{e}_k))}= \SNR_k.
	\ee	
	By \eqref{scale of norm of X}, \eqref{diff in norm of X} and \eqref{lambda_k=R_k}, we obtain
	\be\label{diff in X in relative norm}
	\max\left\{
	\frac{\|\widehat{\mb{X}}_k-\mb{X}_k\|_2^2}{\|\mb{X}_k \|_2^2}
	,\frac{\|\widehat{\mb{X}}_k-\mb{X}_k\|_F^2}{\|\mb{X}_k \|_F^2}
	\right\}\lesssim_P
	\min\left\{
	\frac{1}{n^2}+ \frac{\log p_k}{n\SNR_k}   ,1
	\right\}.
	\ee

\item 	
 We next consider the error bounds of $\widehat{\mb{C}}_k$ and $\widehat{\mb{D}}_k$.	
	
	Simply choose $\bd{f}_k=\mb{\Lambda}_{xk}^{-1/2}\mb{V}_{xk}^\top\bd{x}_k$,
	where $\cov(\bd{x}_k)=\mb{V}_{xk}\mb{\Lambda}_{xk}\mb{V}_{xk}^\top$ is
	a compact SVD. % of $\cov(\bd{x}_k)$.
	Then, we have $\bd{z}_k^{\mathcal{I}_0}=\mb{H}_k\bd{f}_k=\mb{H}_k\mb{\Lambda}_{xk}^{-1/2}\mb{V}_{xk}^\top\bd{x}_k$
	with $\mb{H}_k=(\bd{\eta}_k^{(\ell)}/\|\bd{\eta}_k^{(\ell)}\|_F)_{\ell\in \mathcal{I}_0}^\top$. 
	From~\eqref{c_k vec}, it follows that 
	we can write the common-source matrix $\mb{C}_k$ as
	\be\label{C_k mat}
	\mb{C}_k=\cov(\bd{x}_k,\bd{z}_k^{\mathcal{I}_0})\{\cov(\bd{z}_k^{\mathcal{I}_0})\}^\dag \mb{C}^{\mathcal{I}_0},
	\ee
	where the three components are formulated by
	$\cov(\bd{x}_k,\bd{z}_k^{\mathcal{I}_0})=\mb{V}_{xk}\mb{\Lambda}_{xk}^{1/2}\mb{H}_k^\top$,
	$\cov(\bd{z}_k^{\mathcal{I}_0})=\mb{H}_k\mb{H}_k^\top$,
	and
	$
	\mb{C}^{\mathcal{I}_0}=\mb{A}\mb{N}\mb{F}
	$
	with
	$\mb{A}=\diag\{(\alpha^{(\ell)} [\lambda_{\ell}\{\cov(\bd{f})\}]^{-1/2}){}_{\ell\in \mathcal{I}_0} \}$,
	$\mb{N}=(\bd{\eta}^{(\ell)})_{\ell\in \mathcal{I}_0}^\top$, and 
	$
	\mb{F}=[\mb{F}_1;\dots;\mb{F}_K] 
	$
	in which
	$\mb{F}_k=\mb{\Lambda}_{xk}^{-1/2}\mb{V}_{xk}^\top\mb{X}_k$.

	Since $K$ is a constant and each $\lspan(\bd{x}_k^\top)$ is a fixed space
	independent of $n$ and $\{p_k\}_{k=1}^K$, 
	we have that $r_1,\dots,r_K$ are constants
	and there exist positive constants $\kappa_z$, $\kappa_\eta$, {\color{black}$\kappa_\Delta$} and $\kappa_{zz}$ such that 	
	$\min_{k\le K}\lambda_{r_k^*}(\cov(\bd{z}_k^{\mathcal{I}_0}))
	>\kappa_z$,
{\color{black}},	
	$\min_{k\le K, \ell\in \mathcal{I}_0} \| \bd{\eta}_k^{(\ell)} \|_F>\kappa_\eta$, {\color{black}$\min_{(j,k)\in\mathcal{I}_{\Delta_+}^{(\ell)},\ell\in \mathcal{I}_0}\Delta_{jk}^{(\ell)}>\kappa_\Delta$},
	and
	$\min_{(j,k)\in {\color{black}\mathcal{I}_{\Delta_+}^{(\ell)}\cup \mathcal{I}_{\Delta_0}^{(\ell)}},\ell\in \mathcal{I}_0    }\left|\cos\{\theta(z_j^{(\ell)},z_k^{(\ell)})\}\right|
	> \kappa_{zz}$.

	From \cite{Shu17},	
	using their (S.8), (S.30) and the first inequality on page~10 of their supplement, we have that for all $j,k\le K$, 
	\be\label{bound for |cov(f_k)|_2}
	\eig_1(\widehat{\cov}(\bd{x}_k))\lesssim_P \eig_1(\cov(\bd{x}_k)),
	\ee
	\be\label{VLam1/2}
	\|\widehat{\mb{V}}_{xk}\widehat{\mb{\Lambda}}_{xk}^{1/2}-\mb{V}_{xk}\mb{\Lambda}_{xk}^{1/2} \|_2\lesssim_P\eig_1^{1/2}(\cov(\bd{x}_k)) n^{-1/2} ,
	\ee
	and
	\begin{align*}
	&	\| \widehat{\cov}(\bd{f}_j,\bd{f}_k)-\cov(\bd{f}_j,\bd{f}_k) \|_F 
	\le [\max(r_j,r_k)  ]^{1/2}\| \widehat{\cov}(\bd{f}_j,\bd{f}_k)-\cov(\bd{f}_j,\bd{f}_k) \|_2\\
	&\quad \lesssim_P  \min\left\{
	n^{-1/2}
	+ \left(\frac{p_j\log p_j}{n\eig_1(\cov(\bd{x}_j))} \right)^{1/2}  
	+  \left( \frac{p_k\log p_k}{n\eig_1(\cov(\bd{x}_k))}  \right)^{1/2} ,
	1
	\right\},
	\end{align*}
	where $\widehat{\cov}(\bd{f}_j,\bd{f}_k)=n^{-1}\widehat{\mb{F}}_j \widehat{\mb{F}}_k^\top$ is a submatrix of $\widehat{\cov}(\bd{f})$.
	Then,
	\begin{align}\label{diff in cov(f)}
	\| \widehat{\cov}(\bd{f})-\cov(\bd{f}) \|_F &=\Big(\sum_{1\le j,k\le K}\| \widehat{\cov}(\bd{f}_j,\bd{f}_k)-\cov(\bd{f}_j,\bd{f}_k) \|_F^2\Big)^{1/2}
	\nonumber\\
	&\lesssim_P  \min\left\{
	n^{-1/2}
	+
	\sum_{k=1}^K\Big(   \frac{p_k\log p_k}{n\eig_1(\cov(\bd{x}_k))} \Big)^{1/2}   ,
	1
	\right\}\nonumber\\
	&\lesssim_P\delta_\eta.
	\end{align}
	
	By the uniqueness given in Theorem~\ref{model uniqueness}, we let $\bd{\eta}^{(\ell)}$ satisfy $(\bd{\eta}^{(\ell)})^\top\widehat{\bd{\eta}}^{(\ell)}\ge 0$ for all $\ell\in\mathcal{I}_0$.
	By Corollary~1 in \cite{Yu15}, $\delta_\eta=o(1)$, and the condition that $\{\lambda_{\ell}(\cov(f))\}_{\ell=1}^L$ are distinct, we have
	\begin{align}\label{diff in eta}
		\max_{\ell\in\mathcal{I}_0}\| \widehat{\bd{\eta}}^{(\ell)}-\bd{\eta}^{(\ell)} \|_F 
	&\lesssim_P 
	\frac{ \delta_\eta }{
		\min_{\ell\in\mathcal{I}_0}\{ \eig_{\ell-1}(\cov(\bd{f}))-\eig_{\ell}(\cov(\bd{f})),\eig_{\ell}(\cov(\bd{f}))- \eig_{\ell+1}(\cov(\bd{f}))\}
	}\nonumber\\
	&\lesssim_P  \delta_\eta.
	\end{align}
	Since $\delta_\eta=o(1)$ and $\min_{k\le K, \ell\in \mathcal{I}_0} \| \bd{\eta}_k^{(\ell)} \|_F>\kappa_\eta$, then by \eqref{diff in eta} we have 
	\be\label{norm of eta_k hat lower bound}
	\min_{k\le K, \ell\in \mathcal{I}_0}\| \widehat{\bd{\eta}}_k^{(\ell)}  \|_F\ge \kappa_\eta-o_P(1),
	\ee
	and thus	
	\begin{align}\label{diff in H}
	&\|\widehat{\mb{H}}_k- \mb{H}_k \|_2\le \| \widehat{\mb{H}}_k-\mb{H}_k \|_F
	\lesssim_P L^{1/2} \max_{\ell\in \mathcal{I}_0} 
	\left\|\widehat{\bd{\eta}}_k^{(\ell)}/\|\widehat{\bd{\eta}}_k^{(\ell)}\|_F- \bd{\eta}_k^{(\ell)}/\|\bd{\eta}_k^{(\ell)}\|_F  \right \|_F\nonumber\\
	&\qquad\lesssim_P L^{1/2} \max_{\ell\in \mathcal{I}_0} 
	\left\|\widehat{\bd{\eta}}_k^{(\ell)}(\|\bd{\eta}_k^{(\ell)}\|_F-\|\widehat{\bd{\eta}}_k^{(\ell)}\|_F )+ ( \widehat{\bd{\eta}}_k^{(\ell)}-\bd{\eta}_k^{(\ell)} )\|\widehat{\bd{\eta}}_k^{(\ell)}\|_F \right \|_F
	/(\|\widehat{\bd{\eta}}_k^{(\ell)}\|_F\|\bd{\eta}_k^{(\ell)}\|_F)
	\nonumber\\
	&\qquad\lesssim_P 2L^{1/2}\max_{\ell\in \mathcal{I}_0} \|\widehat{\bd{\eta}}_k^{(\ell)}-\bd{\eta}_k^{(\ell)}\|_F/\|\bd{\eta}_k^{(\ell)}\|_F\nonumber\\
	&\qquad\lesssim_P \delta_\eta.
	\end{align}
	We will frequently use the following matrix inequality:
	\begin{align}\label{chain ineq}
\| \widehat{\mb{M}}_1 \widehat{\mb{M}}_2- \mb{M}_1\mb{M}_2 \|_2
\le \begin{cases}
&
\| \widehat{\mb{M}}_1 \|_2\|  \widehat{\mb{M}}_2- \mb{M}_2  \|_2+\| \mb{M}_2 \|_2\|  \widehat{\mb{M}}_1- \mb{M}_1  \|_2,\\
&
\| \widehat{\mb{M}}_2 \|_2\|  \widehat{\mb{M}}_1- \mb{M}_1  \|_2+\| \mb{M}_1 \|_2\|  \widehat{\mb{M}}_2- \mb{M}_2  \|_2.
\end{cases}
\end{align} 
	Then together with $\max_{k\le K}\{\|\mb{H}_k\|_F,\|\widehat{\mb{H}}_k\|_F\}\le L^{1/2}$,
	we have
	\be
	\|\widetilde{\cov}(\bd{z}_k^{\mathcal{I}_0})-\cov(\bd{z}_k^{\mathcal{I}_0})\|_2
	=\|\widehat{\mb{H}}_k\widehat{\mb{H}}_k^\top-\mb{H}_k\mb{H}_k^\top\|_2
	\le (\| \widehat{\mb{H}}_k\|_2+\|  \mb{H}_k  \|_2)\|\widehat{\mb{H}}_k- \mb{H}_k \|_2
	\lesssim_P \delta_\eta.
	\ee
	
	Recall that $\min_{k\le K} \eig_{r_k^*}(\cov(\bd{z}_k^{\mathcal{I}_0}))>\kappa_z$.
	Let $\cov(\bd{z}_k^{\mathcal{I}_0})=\mb{V}_{zk}\mb{\Lambda}_{zk}\mb{V}_{zk}^\top$ be its compact SVD, 
{\color{black}where $\mb{\Lambda}_{zk}$ has nonincreasing diagonal elements}.
	Let $\widehat{\mb{\Lambda}}_{zk}^{[j,j]}=0$ for $j>\widetilde{r}_k:=\rank(\widetilde{\cov}(\bd{z}_k^{\mathcal{I}_0}))$,
	and $\mb{\Lambda}_{zk}^{[j,j]}=0$ for $j>r_k^*$.
	By Weyl's inequality (see Theorem 3.3.16(c) in \cite{Horn94}), for all $j$,
	\[
	|  \widehat{\mb{\Lambda}}_{zk}^{[j,j]}- \mb{\Lambda}_{zk}^{[j,j]}    |
	\le \|\widetilde{\cov}(\bd{z}_k^{\mathcal{I}_0})-\cov(\bd{z}_k^{\mathcal{I}_0})\|_2\lesssim_P \delta_\eta.
	\]
	Hence,
	\be\label{lower bound for eig hat cov(z)}
	\widehat{\mb{\Lambda}}_{zk}^{[\widecheck{r}_k,\widecheck{r}_k]} \ge \mb{\Lambda}_{zk}^{[\widecheck{r}_k,\widecheck{r}_k]} -O_P(\delta_\eta)   
	\ge \kappa_z-o_P(1)
	\ee
	%	(change $\widecheck{r}_k$ to $r_k$? No, in practice $|\widehat{\mathcal{I}}_k|$ may be not r_k. Or think to extend $\widehat{\mathcal{I}}_k$ with 0 (where the corresponding eta set to be zero) so that $|\widehat{\mathcal{I}}_k|=r_k$.)
	and
	\[
	\max_{j>r_k^*} \widehat{\mb{\Lambda}}_{zk}^{[j,j]} \lesssim_P \delta_\eta.
	\]
	Then, 
	\begin{align}
&\|\widehat{\cov}(\bd{z}_k^{\mathcal{I}_0})-\cov(\bd{z}_k^{\mathcal{I}_0})\|_2
	=\left\|
	\sum_{j=1}^{\widecheck{r}_k} \widehat{\mb{\Lambda}}_{zk}^{[j,j]} \widehat{\mb{V}}_{zk}^{[,j]}(\widehat{\mb{V}}_{zk}^{[,j]})^\top 
	-\sum_{j=1}^{r_k^*} \mb{\Lambda}_{zk}^{[j,j]} \mb{V}_{zk}^{[,j]}(\mb{V}_{zk}^{[,j]})^\top
	\right\|_2
	\nonumber\\
	&\qquad\le  \left\|
	\sum_{j=1}^{\widetilde{r}_k} \widehat{\mb{\Lambda}}_{zk}^{[j,j]} \widehat{\mb{V}}_{zk}^{[,j]}(\widehat{\mb{V}}_{zk}^{[,j]})^\top 
	-\sum_{j=1}^{r_k^*} \mb{\Lambda}_{zk}^{[j,j]} \mb{V}_{zk}^{[,j]}(\mb{V}_{zk}^{[,j]})^\top
	\right\|_2
	+
	\sum_{j=\widecheck{r}_k+1}^{\widetilde{r}_k} \|  \widehat{\mb{\Lambda}}_{zk}^{[j,j]} \widehat{\mb{V}}_{zk}^{[,j]}(\widehat{\mb{V}}_{zk}^{[,j]})^\top  \|_2
	\nonumber\\
	&\qquad= \|\widetilde{\cov}(\bd{z}_k^{\mathcal{I}_0})-\cov(\bd{z}_k^{\mathcal{I}_0})\|_2
	+ \sum_{j=\widecheck{r}_k+1}^{\widetilde{r}_k} \widehat{\mb{\Lambda}}_{zk}^{[j,j]}\nonumber\\
	&\qquad\lesssim_P \delta_\eta+ \max(\widetilde{r}_k-r_k^*,0)\delta_\eta\nonumber\\
	&\qquad\lesssim_P \delta_\eta.
	\end{align}
	By Theorem 2.1 in \cite{Meng10}, \eqref{lower bound for eig hat cov(z)}, and $\min_{k\le K} \eig_{r_k^*}(\cov(\bd{z}_k^{\mathcal{I}_0}))>\kappa_z$, 
	\begin{align}\label{diff in inv cov(z)}
&\left\|[\widehat{\cov}(\bd{z}_k^{\mathcal{I}_0})]^\dag-[\cov(\bd{z}_k^{\mathcal{I}_0})]^\dag\right\|_2
	\le\left\|[\widehat{\cov}(\bd{z}_k^{\mathcal{I}_0})]^\dag-[\cov(\bd{z}_k^{\mathcal{I}_0})]^\dag\right\|_F\nonumber\\
	&\qquad\le \max\left\{
	\left\| [ \widehat{\cov}(\bd{z}_k^{\mathcal{I}_0})]^\dag\right\|_2^2, \left\| [\cov(\bd{z}_k^{\mathcal{I}_0})]^\dag \right\|_2^2
	\right\} \left\| \widehat{\cov}(\bd{z}_k^{\mathcal{I}_0})-\cov(\bd{z}_k^{\mathcal{I}_0})  \right\|_F\nonumber\\
	&\qquad\le \max\left\{
	\left\| [ \widehat{\cov}(\bd{z}_k^{\mathcal{I}_0})]^\dag\right\|_2^2, \left\| [\cov(\bd{z}_k^{\mathcal{I}_0})]^\dag \right\|_2^2
	\right\} L^{1/2}  \left\| \widehat{\cov}(\bd{z}_k^{\mathcal{I}_0})-\cov(\bd{z}_k^{\mathcal{I}_0})  \right\|_2\nonumber\\
	&\qquad\lesssim_P  \delta_\eta.
	\end{align}
	By \eqref{chain ineq}, \eqref{diff in H}, and \eqref{VLam1/2}, we have
	\begin{align*}
\| \widehat{\mb{V}}_{xk}\widehat{\mb{\Lambda}}_{xk}^{1/2}\widehat{\mb{H}}_k^\top   -\mb{V}_{xk}\mb{\Lambda}_{xk}^{1/2} \mb{H}_k^\top   \|_2
	&\le \| \mb{V}_{xk}\mb{\Lambda}_{xk}^{1/2} \|_2\|\widehat{\mb{H}}_k- \mb{H}_k  \|_2
	+\| \widehat{\mb{H}}_k \|_2\| \widehat{\mb{V}}_{xk}\widehat{\mb{\Lambda}}_{xk}^{1/2}- \mb{V}_{xk}\mb{\Lambda}_{xk}^{1/2} \|_2
	\nonumber\\
	&\lesssim_P \eig_1^{1/2}(\cov(\bd{x}_k))\delta_\eta+\eig_1^{1/2}(\cov(\bd{x}_k)) n^{-1/2}\nonumber\\
	&\lesssim_P \eig_1^{1/2}(\cov(\bd{x}_k))\delta_\eta.
	\end{align*}
	Using \eqref{chain ineq} again together with the above inequality, \eqref{diff in inv cov(z)}, and \eqref{lower bound for eig hat cov(z)} yields
	\begin{align}\label{VLHcov(z)}
	\lefteqn{\left\| \widehat{\mb{V}}_{xk}\widehat{\mb{\Lambda}}_{xk}^{1/2}\widehat{\mb{H}}_k^\top[\widehat{\cov}(\bd{z}_k^{\mathcal{I}_0})]^\dag
		-\mb{V}_{xk}\mb{\Lambda}_{xk}^{1/2} \mb{H}_k^\top[\cov(\bd{z}_k^{\mathcal{I}_0})]^\dag \right\|_2}\nonumber\\
	&\le \left\| \mb{V}_{xk}\mb{\Lambda}_{xk}^{1/2} \mb{H}_k \right\|_2\left\| [\widehat{\cov}(\bd{z}_k^{\mathcal{I}_0})]^\dag  -  [\cov(\bd{z}_k^{\mathcal{I}_0})]^\dag\right\|_2\nonumber\\
	&\quad+\left\|  [\widehat{\cov}(\bd{z}_k^{\mathcal{I}_0})]^\dag \right\|_2
	\left\| \widehat{\mb{V}}_{xk}\widehat{\mb{\Lambda}}_{xk}^{1/2}\widehat{\mb{H}}_k^\top   -\mb{V}_{xk}\mb{\Lambda}_{xk}^{1/2} \mb{H}_k^\top  \right \|_2
	\nonumber\\
	&\lesssim_P 
	\eig_1^{1/2}(\cov(\bd{x}_k))\delta_\eta.
	\end{align}
	
	By Weyl's inequality (see Theorem 3.3.16(c) in \cite{Horn94}) and \eqref{diff in cov(f)}, for all $\ell\in\mathcal{I}_0$ we have
	\[
	\left|\eig_\ell(\widehat{\cov}(\bd{f}))-\eig_{\ell}(\cov(\bd{f}))\right|
	\le \left\| \widehat{\cov}(\bd{f})-\cov(\bd{f}) \right\|_2\lesssim_P \delta_\eta.
	\]
	Then by $\delta_\eta=o(1)$ and $\eig_{L}(\cov(\bd{f}))> 1$,
	for all $\ell\in\mathcal{I}_0$ we have
	\begin{align}\label{diff in (eig f )^1/2}
	\left|\eig_\ell^{1/2}(\widehat{\cov}(\bd{f}))-\eig_{\ell}^{1/2}(\cov(\bd{f}))\right|
	&=\left|\eig_\ell^{1/2}(\widehat{\cov}(\bd{f}))+\eig_{\ell}^{1/2}(\cov(\bd{f}))\right|^{-1}
	\left|\eig_\ell(\widehat{\cov}(\bd{f}))-\eig_{\ell}(\cov(\bd{f}))\right|
	\nonumber\\
	&\le  \eig_{\ell}^{-1/2}(\cov(\bd{f}))\left\| \widehat{\cov}(\bd{f})-\cov(\bd{f}) \right\|_2
	\nonumber\\
	&\lesssim_P  \delta_\eta=o(1).
	\end{align}
	Thus, for all $\ell\in\mathcal{I}_0$,
	\[
	\eig_\ell^{1/2}(\widehat{\cov}(\bd{f}))\ge \eig_{\ell}^{1/2}(\cov(\bd{f}))-\left|\eig_\ell^{1/2}(\widehat{\cov}(\bd{f}))-\eig_{\ell}^{1/2}(\cov(\bd{f}))\right|\ge 1-o_P(1),
	\]
	and then
	\begin{align}\label{diff in (eig f )^-1/2}
	\left|\eig_\ell^{-1/2}(\widehat{\cov}(\bd{f}))-\eig_{\ell}^{-1/2}(\cov(\bd{f}))\right|
	&=
	\left|\eig_\ell^{1/2}(\widehat{\cov}(\bd{f}))-\eig_{\ell}^{1/2}(\cov(\bd{f}))\right|
	\eig_\ell^{-\frac{1}{2}}(\widehat{\cov}(\bd{f}))\eig_{\ell}^{-\frac{1}{2}}(\cov(\bd{f}))
	\nonumber\\
	&\lesssim_P  \delta_\eta.
	\end{align}
	
	For all $k\le K$ and $\ell\in\mathcal{I}_0$,
	by \eqref{chain ineq}, $\eig_1(\cov(\bd{f}))\le \tr(\cov(\bd{f}))\le \sum_{k=1}^K r_k$, 
	\eqref{diff in eta}, $\| \widehat{\bd{\eta}}_k^{(\ell)} \|_F\le\| \widehat{\bd{\eta}}^{(\ell)} \|_F=1$,
	and \eqref{diff in (eig f )^1/2},
	we obtain
	\begin{align}\label{cos theta w,z}
&	\left|\widehat{\cos}\{\theta(w^{(\ell)},z_k^{(\ell)})\}-\cos\{\theta(w^{(\ell)},z_k^{(\ell)})\}\right|
	=\left|
	\eig_{\ell}^{1/2}(\widehat{\cov}(\bd{f}))\| \widehat{\bd{\eta}}_k^{(\ell)} \|_F-
	\eig_{\ell}^{1/2}(\cov(\bd{f}))\| \bd{\eta}_k^{(\ell)} \|_F
	\right |\nonumber\\
	&\qquad\le |\eig_{\ell}^{1/2}(\cov(\bd{f}))|\left|\| \widehat{\bd{\eta}}_k^{(\ell)} \|_F-\| \bd{\eta}_k^{(\ell)} \|_F\right|
	+\| \widehat{\bd{\eta}}_k^{(\ell)} \|_F\left|\eig_\ell^{1/2}(\widehat{\cov}(\bd{f}))-\eig_{\ell}^{1/2}(\cov(\bd{f}))\right|
	\nonumber\\
	%&\lesssim_P \delta_\eta+\delta_f\nonumber\\
	&\qquad\lesssim_P \delta_\eta.
	\end{align}
	For all $\ell\in \mathcal{I}_0$ and $j,k\le K$,
	\[
	\cos\{\theta(z_j^{(\ell)},z_k^{(\ell)})\}
	=\frac{(\bd{\eta}_j^{(\ell)})^\top\cov(\bd{f}_j,\bd{f}_k)\bd{\eta}_k^{(\ell)}}{\|\bd{\eta}_j^{(\ell)}\|_F\| \bd{\eta}_k^{(\ell)} \|_F}.
	\]
	By \eqref{chain ineq}, \eqref{norm of eta_k hat lower bound}, \eqref{diff in cov(f)}, and \eqref{diff in H}, 
	\begin{align*}
	\lefteqn{\left\|(\widehat{\bd{\eta}}_j^{(\ell)})^\top\|\widehat{\bd{\eta}}_j^{(\ell)}\|_F^{-1}\widehat{\cov}(\bd{f}_j,\bd{f}_k)-(\bd{\eta}_j^{(\ell)})^\top\|\bd{\eta}_j^{(\ell)}\|_F^{-1}\cov(\bd{f}_j,\bd{f}_k)\right\|_2}\\
	&\le \left\| (\widehat{\bd{\eta}}_j^{(\ell)})^\top\|\widehat{\bd{\eta}}_j^{(\ell)}\|_F^{-1} \right\|_2
	\left\| \widehat{\cov}(\bd{f}_j,\bd{f}_k)-\cov(\bd{f}_j,\bd{f}_k) \right\|_2 \nonumber\\
	&\quad+\left\| \cov(\bd{f}_j,\bd{f}_k) \right\|_2\left\|(\widehat{\bd{\eta}}_j^{(\ell)})^\top\|\widehat{\bd{\eta}}_j^{(\ell)}\|_F^{-1} -(\bd{\eta}_j^{(\ell)})^\top\|\bd{\eta}_j^{(\ell)}\|_F^{-1}   \right\|_2\nonumber\\
%	&\lesssim_P  \delta_f+\delta_\eta\nonumber\\
	&\lesssim_P \delta_\eta,
	\end{align*}
	and then,
	\begin{align}\label{cos theta z's}
	\lefteqn{\left|\widehat{\cos}\{\theta(z_j^{(\ell)},z_k^{(\ell)})\}-\cos\{\theta(z_j^{(\ell)},z_k^{(\ell)})\}\right|}\nonumber\\
	&\le \left\| \widehat{\bd{\eta}}_k^{(\ell)}\|\widehat{\bd{\eta}}_k^{(\ell)}\|_F^{-1} \right\|_2
	\left\|(\widehat{\bd{\eta}}_j^{(\ell)})^\top\|\widehat{\bd{\eta}}_j^{(\ell)}\|_F^{-1}\widehat{\cov}(\bd{f}_j,\bd{f}_k)-(\bd{\eta}_j^{(\ell)})^\top\|\bd{\eta}_j^{(\ell)}\|_F^{-1}\cov(\bd{f}_j,\bd{f}_k)\right\|_2\nonumber\\
	&\qquad+\left\| (\bd{\eta}_j^{(\ell)})^\top\|\bd{\eta}_j^{(\ell)}\|_F^{-1}\cov(\bd{f}_j,\bd{f}_k)  \right\|_2
	\left\|\widehat{\bd{\eta}}_k^{(\ell)}\|\widehat{\bd{\eta}}_k^{(\ell)}\|_F^{-1} -\bd{\eta}_k^{(\ell)}\|\bd{\eta}_k^{(\ell)}\|_F^{-1}   \right\|_2\nonumber\\
	&\lesssim_P   \delta_\eta.
	\end{align}
	
{\color{black}Consider $\ell \in \mathcal{I}_0$ and $(j,k)\in \mathcal{I}_{\Delta_+}^{(\ell)}$. }
	By \eqref{cos theta w,z} and \eqref{cos theta z's}, 
	\begin{align}\label{diff in Delta}
	\lefteqn{\left|\widetilde{\Delta}_{jk}^{(\ell)}-\Delta_{jk}^{(\ell)}\right|}\nonumber\\
	&\le \left|\big[\widehat{\cos}\{\theta(w^{(\ell)},z_j^{(\ell)})\}
	+
	\widehat{\cos}\{\theta(w^{(\ell)},z_k^{(\ell)})\}\big]^2
	-		\big[\cos\{\theta(w^{(\ell)},z_j^{(\ell)})\}
	+
	\cos\{\theta(w^{(\ell)},z_k^{(\ell)})\}\big]^2
	\right|	\nonumber\\	
	&\quad+4\left|	\widehat{\cos}\{\theta(z_j^{(\ell)},z_k^{(\ell)})\}-		\cos\{\theta(z_j^{(\ell)},z_k^{(\ell)})\}\right|\nonumber\\
	&\le 4\left|\big[\widehat{\cos}\{\theta(w^{(\ell)},z_j^{(\ell)})\}
	+
	\widehat{\cos}\{\theta(w^{(\ell)},z_k^{(\ell)})\}\big]
	-		\big[\cos\{\theta(w^{(\ell)},z_j^{(\ell)})\}
	+
	\cos\{\theta(w^{(\ell)},z_k^{(\ell)})\}\big]
	\right| \nonumber\\
	&\quad+4\left|	\widehat{\cos}\{\theta(z_j^{(\ell)},z_k^{(\ell)})\}-		\cos\{\theta(z_j^{(\ell)},z_k^{(\ell)})\}\right|\nonumber\\
	&\le 8 \max_{1\le k\le K} \left| \widehat{\cos}\{\theta(w^{(\ell)},z_k^{(\ell)})\}-\cos\{\theta(w^{(\ell)},z_k^{(\ell)}) \}   \right|
	+4\left|	\widehat{\cos}\{\theta(z_j^{(\ell)},z_k^{(\ell)})\}-		\cos\{\theta(z_j^{(\ell)},z_k^{(\ell)})\}\right|\nonumber\\
	&\lesssim_P  \delta_\eta.
	\end{align}
	{\color{black}
By \eqref{diff in Delta}, $\Delta_{jk}^{(\ell)}>\kappa_\Delta$ and $\delta_\eta=o(1)$, we have
$\widetilde{\Delta}_{jk}^{(\ell)}>\kappa_\Delta-o_P(1)$.
Then by the mean value theorem, we have 
	\begin{align}\label{diff in sqrt Delta}
	\left|  (\widehat{\Delta}_{jk}^{(\ell)})^{1/2} - ( \Delta_{jk}^{(\ell)} )^{1/2} \right|
	&\le {\color{black}\frac{1}{2} [\min(\widehat{\Delta}_{jk}^{(\ell)},\Delta_{jk}^{(\ell)})]^{-1/2}}
		\left|\widehat{\Delta}_{jk}^{(\ell)}-\Delta_{jk}^{(\ell)}\right|\nonumber\\
&	\le 
\frac{1}{2}[\kappa_\Delta-o_P(1)]^{-1/2}\left|\widetilde{\Delta}_{jk}^{(\ell)}-\Delta_{jk}^{(\ell)}\right|\nonumber\\
&	\lesssim_P \delta_\eta.
	\end{align}

Now consider $\ell \in \mathcal{I}_0$ and $(j,k)\in \mathcal{I}_{\Delta}^{(\ell)}:=\mathcal{I}_{\Delta_+}^{(\ell)}\cup \mathcal{I}_{\Delta_0}^{(\ell)}$.
}
From \eqref{cos theta w,z} and \eqref{diff in sqrt Delta}, 
	\be\label{diff in alpha_jk}
	|\widehat{\alpha}_{jk}^{(\ell)}-\alpha_{jk}^{(\ell)}|
	\lesssim_P {\color{black}\delta_\eta}.
	\ee
	Recall that 	
	$\min_{(j,k)\in \mathcal{I}_\Delta^{(\ell)},\ell\in \mathcal{I}_0     }\left|\cos\{\theta(z_j^{(\ell)},z_k^{(\ell)})\}\right|>\kappa_{zz}$.
	By \eqref{cos theta z's} and $\delta_\eta=o(1)$,
	with probability tending to 1 we have that
	$\widehat{\cos}\{\theta(z_j^{(\ell)},z_k^{(\ell)})\}\cos\{\theta(z_j^{(\ell)},z_k^{(\ell)})\}>0$
	and thus 
	$\widehat{\alpha}_{jk}^{(\ell)}\alpha_{jk}^{(\ell)}>0$.
	Without loss of generality, we assume $\alpha^{(\ell)}>0$.
	Let $\mathcal{I}_+^{(\ell)}=\{(j,k)\in \mathcal{I}_\Delta^{(\ell)}: \alpha_{jk}^{(\ell)}>0  \}$,
	then
	$\alpha^{(\ell)}=\min\{\alpha_{jk}^{(\ell)}: \alpha_{jk}^{(\ell)}>0, (j,k)\in \mathcal{I}_+^{(\ell)} \}$.
	With probability tending to 1,
	$\widehat{\alpha}^{(\ell)}=\min\{\widehat{\alpha}_{jk}^{(\ell)}: \widehat{\alpha}_{jk}^{(\ell)}>0, (j,k)\in \mathcal{I}_+^{(\ell)} \}$.
	Due to Lemma~S.1 in \cite{Shu17}, we simply assume $\widehat{\alpha}^{(\ell)}=\min\{\widehat{\alpha}_{jk}^{(\ell)}: \widehat{\alpha}_{jk}^{(\ell)}>0, (j,k)\in \mathcal{I}_+^{(\ell)} \}$ in the rest of the proof.	
	Without loss of generality, denote $\alpha_{12}^{(\ell)}=\alpha^{(\ell)}$.
	If $\widehat{\alpha}_{12}^{(\ell)}=\widehat{\alpha}^{(\ell)}$, then $|\widehat{\alpha}^{(\ell)}-\alpha^{(\ell)}|\lesssim_P {\color{black}\delta_\eta}$.
	Otherwise, without loss of generality we asume $\widehat{\alpha}^{(\ell)}=\widehat{\alpha}_{23}^{(\ell)}<\widehat{\alpha}_{12}^{(\ell)}$ and $\alpha^{(\ell)}=\alpha_{12}^{(\ell)}<\alpha_{23}^{(\ell)}$.
Then by \eqref{diff in alpha_jk} and $\delta_\eta=o(1)$,
$\widehat{\alpha}_{23}^{(\ell)}-\widehat{\alpha}_{12}^{(\ell)}\ge \alpha_{23}^{(\ell)}-\alpha_{12}^{(\ell)}-o_P(1)$, 
which contradicts $\widehat{\alpha}_{12}^{(\ell)}>\widehat{\alpha}_{23}^{(\ell)}= \widehat{\alpha}^{(\ell)}$.		
	Hence, 
	\be\label{diff in alpha}
	|\widehat{\alpha}^{(\ell)}-\alpha^{(\ell)}|\lesssim_P {\color{black}\delta_\eta}.
	\ee
	
	By \eqref{chain ineq},  \eqref{diff in (eig f )^-1/2} and \eqref{diff in alpha},
	for all $\ell\in \mathcal{I}_0$,
	\begin{align*}
	\lefteqn{\left|\widehat{\alpha}^{(\ell)} \eig_{\ell}^{-1/2}(\widehat{\cov}(\bd{f}))-\alpha^{(\ell)} \eig_{\ell}^{-1/2}(\cov(\bd{f}))\right|}\nonumber\\
	&\le \widehat{\alpha}^{(\ell)}\left|\eig_\ell^{-1/2}(\widehat{\cov}(\bd{f}))-\eig_{\ell}^{-1/2}(\cov(\bd{f}))\right|
	+\eig_{\ell}^{-1/2}(\cov(\bd{f}))|\widehat{\alpha}^{(\ell)}-\alpha^{(\ell)}|
	\nonumber\\
%	&\lesssim_P  \delta_f+{\color{black}\delta_\eta}\nonumber\\
	&\lesssim_P {\color{black}\delta_\eta}.
	\end{align*}
	Then together with \eqref{chain ineq} and \eqref{diff in eta} gives
	\be\label{diff in AN}
	\|\widehat{\mb{A}}\widehat{\mb{N}} -\mb{A}\mb{N}  \|_2
	\le \| \mb{A} \|_2\| \widehat{\mb{N}}- \mb{N} \|_F
	+\| \widehat{\mb{N}} \|_F \|\widehat{\mb{A}}- \mb{A} \|_2
	\lesssim_P  {\color{black}\delta_\eta},
	\ee
where $\widehat{\mb{A}}:=\diag\{(\widehat{\alpha}^{(\ell)} [\lambda_{\ell}(\widehat{\cov}(\bd{f})) ]^{-1/2})_{\ell\in \mathcal{I}_0} \}$ with $0/0:=0$,
and $\widehat{\mb{N}}:=(\widehat{\bd{\eta}}^{(\ell)})_{\ell\in \mathcal{I}_0}^\top$.
	From the inequalities respectively below (S.12) and (S.22) in the supplement of \cite{Shu17}, we obtain
	\[
	n^{-1}\| \mb{F}_k \|_F^2=r_k+O_P(  n^{-1/2} )
	\]
	and
	\[
	\| \widehat{\mb{F}}_k-\mb{F}_k \|_F
	\le r_k^{1/2}   \| \widehat{\mb{F}}_k-\mb{F}_k \|_2
	\lesssim_P 
	\min\left\{
	1+  [p_k\eig_1^{-1}(\cov(\bd{x}_k))\log p_k  ]^{1/2}  ,n^{1/2}
	\right\}=:\delta_{F_k}.
	\]
	Hence,
	\be\label{norm F}
	\| \mb{F}  \|_F=\big(\sum_{k=1}^K\| \mb{F}_k  \|_F^2\big)^{1/2}=O_P(  n^{1/2})
	\ee
	and
	\be\label{diff in F}
	\|\widehat{\mb{F}}-\mb{F}   \|_F=\Big(\sum_{k=1}^K{\| \widehat{\mb{F}}_k-\mb{F}_k \|_F^2}\Big)^{1/2}
	\lesssim_P \sum_{k=1}^K\delta_{F_k}.
	\ee
	
	By \eqref{chain ineq}, \eqref{norm F}, \eqref{diff in AN} and \eqref{diff in F}, we obtain
	\begin{align}\label{diff in C_{0,k}}
	\|\widehat{\mb{C}}^{\mathcal{I}_0} -\mb{C}^{\mathcal{I}_0}\|_2
	&\le \| \mb{F} \|_F \|\widehat{\mb{A}}\widehat{\mb{N}} -\mb{A}\mb{N}  \|_2
	+ \|\widehat{\mb{A}}\|_2\|\widehat{\mb{N}}  \|_F \|\widehat{\mb{F}}-\mb{F}   \|_F 
	\nonumber\\
	&\lesssim_P n^{1/2}  {\color{black}\delta_\eta}
	+\sum_{k=1}^K\delta_{F_k}
	\lesssim_P n^{1/2} {\color{black}\delta_\eta}.
	\end{align}
	
	Using \eqref{chain ineq}, \eqref{norm F}, \eqref{VLHcov(z)}, \eqref{bound for |cov(f_k)|_2}, \eqref{lower bound for eig hat cov(z)} and \eqref{diff in C_{0,k}} yields
	\begin{align}\label{diff in C in spectral norm}
	\lefteqn{\|\widehat{\mb{C}}_k-\mb{C}_k\|_2}\nonumber\\
	&\le\| \mb{A}\mb{N}\mb{F} \|_2\left\| \widehat{\mb{V}}_{xk}\widehat{\mb{\Lambda}}_{xk}^{1/2}\widehat{\mb{H}}_k^\top[\widehat{\cov}(\bd{z}_k^{\mathcal{I}_0})]^\dag
	-\mb{V}_{xk}\mb{\Lambda}_{xk}^{1/2} \mb{H}_k^\top[\cov(\bd{z}_k^{\mathcal{I}_0})]^\dag \right\|_2
	\nonumber\\
	&\qquad  +\left\|  \widehat{\mb{V}}_{xk}\widehat{\mb{\Lambda}}_{xk}^{1/2}\widehat{\mb{H}}_k^\top[\widehat{\cov}(\bd{z}_k^{\mathcal{I}_0})]^\dag \right\|_2
	\|\widehat{\mb{C}}^{\mathcal{I}_0} -\mb{C}^{\mathcal{I}_0} \|_2
	\nonumber\\
	&\lesssim_P n^{1/2}   [\eig_1^{1/2}(\cov(\bd{x}_k))\delta_\eta+\eig_1^{1/2}(\cov(\bd{x}_k)) n^{-1/2}]
+\eig_1^{1/2}(\cov(\bd{x}_k)) n^{1/2}   {\color{black}\delta_\eta}\nonumber\\
	&\lesssim_P \eig_1^{1/2}(\cov(\bd{x}_k))  n^{1/2}   {\color{black}\delta_\eta}.
	\end{align}
	By $\rank(\mb{M}_1\mb{M}_2)\le \min (\rank(\mb{M}_1),\rank(\mb{M}_2))$
	and $\rank(\mb{M}_1-\mb{M}_2)\le \rank(\mb{M}_1)+\rank(\mb{M}_2)$ for any real matrices $\mb{M}_1$ and $\mb{M}_2$ with compatible sizes,
	we have
	$\rank(\widehat{\mb{C}}_k-\mb{C}_k)\le 2L$.
	Thus,
	\be\label{diff in C in Frob norm}
	\| \widehat{\mb{C}}_k-\mb{C}_k  \|_F\le \| \widehat{\mb{C}}_k-\mb{C}_k \|_2
	[ \rank(\widehat{\mb{C}}_k-\mb{C}_k) ]^{1/2} 
	\lesssim_P \eig_1^{1/2}(\cov(\bd{x}_k))  n^{1/2} {\color{black}\delta_\eta}.
	\ee
	By \eqref{diff in C in spectral norm}, \eqref{diff in C in Frob norm}
	and \eqref{scale of norm of X}, we obtain
	\be\label{diff in C in relative norm}
	\max\left\{
	\frac{\|\widehat{\mb{C}}_k-\mb{C}_k\|_2}{\|\mb{X}_k \|_2}
	,\frac{\|\widehat{\mb{C}}_k-\mb{C}_k\|_F}{\|\mb{X}_k \|_F}
	\right\}\lesssim_P  {\color{black}\delta_\eta}.
	\ee
	
	By $\| \widehat{\mb{D}}_k-\mb{D}_k \|\le \| \widehat{\mb{X}}_k-\mb{X}_k  \|+\|\widehat{\mb{C}}_k-\mb{C}_k   \|$ for both  the Frobenius norm and the spectral norm, \eqref{diff in X in relative norm} and \eqref{diff in C in relative norm}, 
	we obtain	
	\be\label{diff in D in relative norm}
	\max\left\{
	\frac{\|\widehat{\mb{D}}_k-\mb{D}_k\|_2}{\|\mb{X}_k \|_2}
	,\frac{\|\widehat{\mb{D}}_k-\mb{D}_k\|_F}{\|\mb{X}_k \|_F}
	\right\}\lesssim_P  {\color{black}\delta_\eta}.
	\ee

\item	Now we consider the estimated view-level
	proportion of explained signal variance.
	
	Note that $\|\widehat{\mb{X}}_k  \|_F^2/n=\tr(\widehat{\mb{X}}_k \widehat{\mb{X}}_k^\top /n)=\tr(\widehat{\cov}(\bd{x}_k))$.
	By inequality (S.16) of \cite{Shu17},
	\begin{align}\label{diff in trace cov_k}
	\left|\|\widehat{\mb{X}}_k  \|_F^2/n-\tr(\cov(\bd{x}_k))\right|
	&=\big|\tr(\widehat{\cov}(\bd{x}_k))-\tr(\cov(\bd{x}_k))\big|\le \sum_{\ell=1}^{r_k} \big|
	\eig_{\ell}( \widehat{\cov}(\bd{x}_k)) 
	- \eig_{\ell}(\cov(\bd{x}_k)) 
	\big|\nonumber\\
	&\lesssim_P     \eig_1(\cov(\bd{x}_k)) n^{-1/2},
	\end{align}
	and by their (S.17),
	\be\label{trace cov_k lower bd}
	\|\widehat{\mb{X}}_k  \|_F^2/n=\tr(\widehat{\cov}(\bd{x}_k))=\sum_{\ell=1}^{r_k}\eig_{\ell}( \widehat{\cov}(\bd{x}_k)) \ge r_k(1-o_P(1)) \eig_{r_k}(\cov(\bd{x}_k)).
	\ee
	Since \eqref{diff in C in Frob norm} and	
	\[
	\big\|\mb{C}_k\big\|_F\le L^{1/2}   \big\|\mb{C}_k\big\|_2 =L^{1/2} \big\|\mb{V}_{xk}\mb{\Lambda}_{xk}^{1/2}\mb{H}_k^\top[\cov(\bd{z}_k^{\mathcal{I}_0})]^\dag \mb{A}\mb{N}\mb{F}\big\|_2
	\lesssim_P \eig_1^{1/2}(\cov(\bd{x}_k)) n^{1/2}  ,
	\]
	we obtain
	\be\label{C_k Frob norm upper bd}
	\big\|\widehat{\mb{C}}_k\big\|_F
	\le \| \widehat{\mb{C}}_k-\mb{C}_k  \|_F+\big\|\mb{C}_k\big\|_F
	\lesssim_P  \eig_1^{1/2}(\cov(\bd{x}_k))  n^{1/2} .
	\ee
	Then,
	\begin{align}\label{diff in C_k frob norm sq}
	\left|
	\|\widehat{\mb{C}}_k  \|_F^2 /n-\|\mb{C}_k  \|_F^2/n
	\right|
	&=n^{-1} \left|\|\widehat{\mb{C}}_k  \|_F- \|\mb{C}_k\|_F\right|  (\|\widehat{\mb{C}}_k  \|_F+ \|\mb{C}_k\|_F)\nonumber\\
	&\le n^{-1}  \|\widehat{\mb{C}}_k - \mb{C}_k\|_F (\|\widehat{\mb{C}}_k  \|_F+ \|\mb{C}_k\|_F) \nonumber\\
	&\lesssim_P \eig_1(\cov(\bd{x}_k))  {\color{black}\delta_\eta}.
	\end{align}	
	From the central limit theorem,
	\[
	\left\|\mb{F}\mb{F}^\top/n -\cov(\bd{f})  \right\|_2\le  \sum_{k=1}^K r_k \left\|  \mb{F}\mb{F}^\top/n -\cov(\bd{f})  \right\|_{\max} \lesssim_P  n^{-1/2}  .
	\]
	Let $\mb{Q}_k=\mb{V}_{xk}\mb{\Lambda}_{xk}^{1/2}\mb{H}_k^\top[\cov(\bd{z}_k^{\mathcal{I}_0})]^\dag \mb{A}\mb{N}$, then $\| \mb{Q}_k \|_2\lesssim_P \eig_1^{1/2}(\cov(\bd{x}_k))$.
	By Weyl's inequality (see Theorem 3.3.16(c) in \cite{Horn94}),
	\begin{align*}
	&\max_{\ell\le L}\left|\eig_\ell ( \mb{C}_k\mb{C}_k^\top/n)-\eig_\ell(\cov(\bd{c}_k))\right|
	\le \left\| \mb{C}_k\mb{C}_k^\top/n -\cov(\bd{c}_k)  \right\|_2
	\nonumber\\
	&\qquad=\left\| \mb{Q}_kn^{-1}  \mb{F} \mb{F}^\top   \mb{Q}_k^\top  
	-
	\mb{Q}_k\cov(\bd{f})\mb{Q}_k^\top
	\right\|_2 
	 \le 
	\| \mb{Q}_k \|_2
	\left\|\mb{F}\mb{F}^\top/n -\cov(\bd{f})  \right\|_2
	\| \mb{Q}_k^\top \|_2 \nonumber\\
	&\qquad\lesssim_P \eig_1(\cov(\bd{x}_k)) n^{-1/2} .
	\end{align*}
	Then applying the same skill used for \eqref{diff in trace cov_k} yields
	\be\label{diff C_k F norm and trace}
	\left|\|\mb{C}_k  \|_F^2/n-\tr(\cov(\bd{c}_k))\right|
	\le \sum_{\ell=1}^L \left|
	\eig_{\ell}(  \mb{C}_k\mb{C}_k^\top/n) 
	- \eig_{\ell}(\cov(\bd{c}_k)) 
	\right|
	\lesssim_P \eig_1(\cov(\bd{x}_k))  n^{-1/2} .
	\ee
	Combining \eqref{diff in C_k frob norm sq} and \eqref{diff C_k F norm and trace}
	with the triangle inequality gives
	\be\label{error for C_k hat and trace}
	\left|\|\widehat{\mb{C}}_k  \|_F^2/n-\tr(\cov(\bd{c}_k))\right|
	\lesssim_P \eig_1(\cov(\bd{x}_k))  {\color{black}\delta_\eta}.
	\ee
	From \eqref{chain ineq}, \eqref{diff in trace cov_k}, \eqref{trace cov_k lower bd},
	\eqref{C_k Frob norm upper bd}, \eqref{error for C_k hat and trace}
	and \eqref{lambda_k=R_k},
	we have
	\begin{align*}
	\lefteqn{
{\color{black}\left|\widehat{\PVE}_c(\bd{x}_k)-\PVE_c(\bd{x}_k)\right|}	
	=\left|
		\frac{\frac{1}{n}\|\widehat{\mb{C}}_k \|_F^2}{\frac{1}{n}\|\widehat{\mb{X}}_k \|_F^2}
		-\frac{\tr(\cov(\bd{c}_k))}{\tr(\cov(\bd{x}_k))}
		\right|}\nonumber\\
	&\le \left| \frac{1}{\frac{1}{n}\|\widehat{\mb{X}}_k \|_F^2}
	-\frac{1}{\tr(\cov(\bd{x}_k))}\right| \cdot \frac{1}{n}\|\widehat{\mb{C}}_k \|_F^2
	+\left|\frac{1}{n}\|\widehat{\mb{C}}_k  \|_F^2-\tr(\cov(\bd{c}_k))\right| \frac{1}{\tr(\cov(\bd{x}_k))}  \nonumber\\
	&\le 
	\frac{\left|\tr(\cov(\bd{x}_k))-\frac{1}{n}\|\widehat{\mb{X}}_k \|_F^2\right|}{\frac{1}{n}\|\widehat{\mb{X}}_k \|_F^2\tr(\cov(\bd{x}_k))}
	\cdot \frac{1}{n}\|\widehat{\mb{C}}_k \|_F^2
	+\left|\frac{1}{n}\|\widehat{\mb{C}}_k  \|_F^2-\tr(\cov(\bd{c}_k))\right| \frac{1}{\tr(\cov(\bd{x}_k))}\\
	&\lesssim_P {\color{black}\delta_\eta}.
	\end{align*}

{\color{black}
\item Next, we consider the estimated variable-level proportion of explained signal variance.

First consider the error of $\widehat{\mb{V}}_{xk}$ in the max norm.
We will use Theorem 3 of \citet{fan2018eigenvector}.
Before applying the theorem,
we need to check the conditions therein. 	
By Assumption~\ref{assump1}~\ref{assump1(iv)} and \ref{assump1(v)}, we have
$\|\cov(\bd{y}_k)\|_{\max}\le \|\cov(\bd{x}_k)\|_{\max}+\|\cov(\bd{e}_k)\|_{\max}\le r_k \kappa_B^2 \lambda_{k,1}/p_k
+s_0$.	Then from the proof of Lemma A2 (i) in \citet{shu2019estimation} and Assumption~\ref{assump1}~\ref{assump1(iv)}, we obtain
\[
\Big\|\frac{1}{n}\mb{Y}\mb{Y}^\top-\cov(\bd{y}_k) \Big\|_{\max}
\lesssim_P\|\cov(\bd{y}_k)\|_{\max}\sqrt{\frac{\log p_k}{n}}\lesssim_P\Big(\frac{\lambda_{k,1}}{p_k}+s_0\Big)\sqrt{\frac{\log p_k}{n}}.
\]
Thus, in our context, their notation $\epsilon=0$, 
$\mu(\mb{V}_{xk})=\frac{p_k}{r_k}\max_{1\le i\le p_k}\sum_{j=1}^{r_k}(\mb{V}_{xk}^{[i,j]})^2=O(
\frac{p_k}{r_k} r_k \frac{\kappa_B^2}{p_k})$
$=O(1)$
(by Assumption~\ref{assump1} (iv) and (ii)),
and 
$\|E\|_\infty=\|\cov(\bd{e}_k)+\frac{1}{n}\mb{Y}\mb{Y}^\top-\cov(\bd{y}_k)\|_\infty\le s_0+p_k \|\frac{1}{n}\mb{Y}\mb{Y}^\top-\cov(\bd{y}_k)\|_{\max}
\lesssim_P 1+ (\lambda_{k,1}+p_k)\sqrt{(\log p_k)/n}$.	
Hence, by Assumption~\ref{assump1}~(i) and letting 
their notation $\delta=\delta_0\lambda_{r_k}(\cov(\bd{x}_k))/2$,
if $\lambda_{r_k}(\cov(\bd{x}_k)) > \widetilde{M}_k (\lambda_{k,r_k}+p_k)\sqrt{(\log p_k)/n}$ with a sufficiently large constant $\widetilde{M}_k>0$, which is satisfied due to $\delta_k=o(1)$ and \eqref{lambda equal}, then
from Theorem 3 of \citet{fan2018eigenvector}, we
have
\be\label{max diff in V}
\|\widehat{\mb{V}}_{xk}-\mb{V}_{xk}\|_{\max}
=O\left(\frac{\|E\|_\infty}{\lambda_{k,r_k}\sqrt{p_k}}\right)
=O_P\left((\frac{1}{\sqrt{p_k}}+\frac{\sqrt{p_k}}{\lambda_{k,r_k}})
\sqrt{\frac{\log p_k}{n}}\right)
:=O_P(\delta_{V_k}).
\ee

From (S.6), (S.16) and (S.18) in \citet{Shu17}, we have
\be\label{lambda equal}
\lambda_{k,\ell}/\lambda_\ell(\cov(\bd{x}_k))\to 1\quad \text{for} \quad 1\le \ell\le r_k,
\ee
\be\label{max diff in Lambda}
\|\widehat{\mb{\Lambda}}_{xk}-\mb{\Lambda}_{xk}\|_{\max}\lesssim_P
\lambda_{k,1}/\sqrt{n},
\ee
and
\be\label{max diff in Lambda sqrt}
\|\widehat{\mb{\Lambda}}_{xk}^{1/2}-\mb{\Lambda}_{xk}^{1/2}\|_{\max}\lesssim_P
\sqrt{\lambda_{k,1}/n}.
\ee
Then by \eqref{max diff in V}, \eqref{max diff in Lambda}, \eqref{max diff in Lambda sqrt}, and Assumption~\ref{assump1} (iv), we obtain 
\begin{align*}
	\|\widehat{\mb{V}}_{xk}\widehat{\mb{\Lambda}}_{xk}- \mb{V}_{xk}\mb{\Lambda}_{xk}\|_{\max}
	&\le \|(\widehat{\mb{V}}_{xk}-\mb{V}_{xk})\widehat{\mb{\Lambda}}_{xk}\|_{\max}
	+\|\mb{V}_{xk}(\widehat{\mb{\Lambda}}_{xk}-\mb{\Lambda}_{xk})\|_{\max}\\
	&\lesssim_P\delta_{V_k}\lambda_{k,1}+  \sqrt{1/p_k} \lambda_{k,1}/\sqrt{n}
	\lesssim_P\delta_{V_k}\lambda_{k,1},
\end{align*}
\begin{align}
\| \widehat{\cov}(\bd{x}_k)-\cov(\bd{x}_k)\|_{\max}
&=	\|\widehat{\mb{V}}_{xk}\widehat{\mb{\Lambda}}_{xk}\widehat{\mb{V}}_{xk}^\top- \mb{V}_{xk}\mb{\Lambda}_{xk}\mb{V}_{xk}^\top\|_{\max}\nonumber\\
	&\le \|(\widehat{\mb{V}}_{xk}\widehat{\mb{\Lambda}}_{xk}-\mb{V}_{xk}\mb{\Lambda}_{xk})\widehat{\mb{V}}_{xk}^\top\|_{\max}
	+\|\mb{V}_{xk}\mb{\Lambda}_{xk}(\widehat{\mb{V}}_{xk}-\mb{V}_{xk})^\top\|_{\max}\nonumber\\
	&\le \|\widehat{\mb{V}}_{xk}\widehat{\mb{\Lambda}}_{xk}- \mb{V}_{xk}\mb{\Lambda}_{xk}\|_{\max}\| \widehat{\mb{V}}_{xk}^\top\|_1
	+\|\widehat{\mb{V}}_{xk}- \mb{V}_{xk}\|_{\max}\| \mb{\Lambda}_{xk}\mb{V}_{xk}^\top  \|_1\nonumber\\
	&\lesssim_P \delta_{V_k}\lambda_{k,1}/\sqrt{p_k},
\label{max diff in cov_xk}	
\end{align}
and
\begin{align*}
\|\widehat{\mb{V}}_{xk}\widehat{\mb{\Lambda}}_{xk}^{1/2}- \mb{V}_{xk}\mb{\Lambda}_{xk}^{1/2}\|_{\max}
&\le \|(\widehat{\mb{V}}_{xk}-\mb{V}_{xk})\widehat{\mb{\Lambda}}_{xk}^{1/2}\|_{\max}
+\|\mb{V}_{xk}(\widehat{\mb{\Lambda}}_{xk}^{1/2}-\mb{\Lambda}_{xk}^{1/2})\|_{\max}\\
&\lesssim_P\delta_{V_k}\lambda_{k,1}^{1/2}+  \sqrt{1/p_k} \sqrt{\lambda_{k,1}/n}
\lesssim_P\delta_{V_k}\lambda_{k,1}^{1/2}.
\end{align*}

By the last inequality, Assumption~\ref{assump1}~(iv), and \eqref{diff in H}, 
\begin{align}
&\|\widehat{\cov}(\bd{x}_k,\bd{z}_k^{\mathcal{I}_0})-\cov(\bd{x}_k,\bd{z}_k^{\mathcal{I}_0})\|_{\max}
=\|\widehat{\mb{V}}_{xk}\widehat{\mb{\Lambda}}_{xk}^{1/2}\widehat{\mb{H}}_k^\top - \mb{V}_{xk}\mb{\Lambda}_{xk}^{1/2}\mb{H}_k^\top\|_{\max}\nonumber\\
&\le \| \mb{V}_{xk}\mb{\Lambda}_{xk}^{1/2}(\widehat{\mb{H}}_k^\top-\mb{H}_k^\top) \|_{\max}
+\|(\widehat{\mb{V}}_{xk}\widehat{\mb{\Lambda}}_{xk}^{1/2}-\mb{V}_{xk}\mb{\Lambda}_{xk}^{1/2}) \widehat{\mb{H}}_k^\top\|_{\max}\nonumber\\
&\le \| \mb{V}_{xk}\mb{\Lambda}_{xk}^{1/2}\|_{\max}\sqrt{r_k}\|\widehat{\mb{H}}_k^\top-\mb{H}_k^\top \|_2
+\|\widehat{\mb{V}}_{xk}\widehat{\mb{\Lambda}}_{xk}^{1/2}-\mb{V}_{xk}\mb{\Lambda}_{xk}^{1/2}\|_{\max}\sqrt{r_k}\|\widehat{\mb{H}}_k^\top\|_2\nonumber\\
&\lesssim_P (\lambda_{k,1}/p_k)^{1/2}\delta_\eta+
\delta_{V_k}\lambda_{k,1}^{1/2}.\nonumber
\end{align}
Then
by \eqref{diff in inv cov(z)}, $\min_{k\le K} \eig_{r_k^*}(\cov(\bd{z}_k^{\mathcal{I}_0}))>\kappa_z$, and $\delta_\eta=o(1)$, we similarly have
\begin{align}
\lefteqn{\delta_{B_k}:=\|\widehat{\cov}(\bd{x}_k,\bd{z}_k^{\mathcal{I}_0})\{\widehat{\cov}(\bd{z}_k^{\mathcal{I}_0})\}^\dag
-\cov(\bd{x}_k,\bd{z}_k^{\mathcal{I}_0})\{\cov(\bd{z}_k^{\mathcal{I}_0})\}^\dag\|_{\max}}\nonumber\\
&\le 
\|\cov(\bd{x}_k,\bd{z}_k^{\mathcal{I}_0})\|_{\max}\sqrt{|\mathcal{I}_0|}\|\{\widehat{\cov}(\bd{z}_k^{\mathcal{I}_0})\}^\dag- \{\cov(\bd{z}_k^{\mathcal{I}_0})\}^\dag\|_2\nonumber\\
&\quad+\|\widehat{\cov}(\bd{x}_k,\bd{z}_k^{\mathcal{I}_0})-\cov(\bd{x}_k,\bd{z}_k^{\mathcal{I}_0})\|_{\max}\sqrt{|\mathcal{I}_0|}\|\{\widehat{\cov}(\bd{z}_k^{\mathcal{I}_0})\}^\dag\|_2\nonumber\\
&\le  \| \mb{V}_{xk}\mb{\Lambda}_{xk}^{1/2}\|_{\max}\sqrt{r_k}\|\mb{H}_k^\top \|_2\sqrt{|\mathcal{I}_0|}\|\{\widehat{\cov}(\bd{z}_k^{\mathcal{I}_0})\}^\dag- \{\cov(\bd{z}_k^{\mathcal{I}_0})\}^\dag\|_2\nonumber\\
&\quad+\|\widehat{\cov}(\bd{x}_k,\bd{z}_k^{\mathcal{I}_0})-\cov(\bd{x}_k,\bd{z}_k^{\mathcal{I}_0})\|_{\max}\sqrt{|\mathcal{I}_0|}
( \|[\cov(\bd{z}_k^{\mathcal{I}_0})]^\dag\|_2+\|[\widehat{\cov}(\bd{z}_k^{\mathcal{I}_0})]^\dag- [\cov(\bd{z}_k^{\mathcal{I}_0})]^\dag\|_2)\nonumber\\
&\lesssim_P (\lambda_{k,1}/p_k)^{1/2}\delta_\eta+
\delta_{V_k}\lambda_{k,1}^{1/2}.
\label{max diff in B_k}
\end{align}

From \eqref{chain ineq} and \eqref{diff in C_{0,k}},
$\| \frac{1}{n}\widehat{\mb{C}}^{\mathcal{I}_0}(\widehat{\mb{C}}^{\mathcal{I}_0})^\top-\frac{1}{n}\mb{C}^{\mathcal{I}_0}(\mb{C}^{\mathcal{I}_0})^\top\|_2
\lesssim_P \delta_\eta$.
Besides, by the central limit theorem, 
$\|\frac{1}{n}\mb{C}^{\mathcal{I}_0}(\mb{C}^{\mathcal{I}_0})^\top-
\cov(\bd{c}^{\mathcal{I}_0})\|_2\le |\mathcal{I}_0|\| \frac{1}{n}\mb{C}^{\mathcal{I}_0}(\mb{C}^{\mathcal{I}_0})^\top-
\cov(\bd{c}^{\mathcal{I}_0}) \|_{\max}\lesssim_Pn^{-1/2}$.
Thus, by the triangle inequality,
\be\label{max diff in cov(c^I)}
 \left\| \widehat{\mb{C}}^{\mathcal{I}_0}(\widehat{\mb{C}}^{\mathcal{I}_0})^\top/n-\cov(\bd{c}^{\mathcal{I}_0})\right\|_2
\lesssim_P \delta_\eta.
\ee

By \eqref{max diff in B_k} and \eqref{max diff in cov(c^I)},
\begin{align*}
\lefteqn{\delta_{B_k\Sigma_c}:=\left\|\widehat{\cov}(\bd{x}_k,\bd{z}_k^{\mathcal{I}_0})\{\widehat{\cov}(\bd{z}_k^{\mathcal{I}_0})\}^\dag\widehat{\mb{C}}^{\mathcal{I}_0}(\widehat{\mb{C}}^{\mathcal{I}_0})^\top/n
-\cov(\bd{x}_k,\bd{z}_k^{\mathcal{I}_0})\{\cov(\bd{z}_k^{\mathcal{I}_0})\}^\dag\cov(\bd{c}^{\mathcal{I}_0})\right\|_{\max}}\nonumber\\
&\le \|\cov(\bd{x}_k,\bd{z}_k^{\mathcal{I}_0})\{\cov(\bd{z}_k^{\mathcal{I}_0})\}^\dag\|_{\max}
\sqrt{|\mathcal{I}_0|}\left\| \widehat{\mb{C}}^{\mathcal{I}_0}(\widehat{\mb{C}}^{\mathcal{I}_0})^\top/n-\cov(\bd{c}^{\mathcal{I}_0})\right\|_2\\
&\qquad+\|\widehat{\cov}(\bd{x}_k,\bd{z}_k^{\mathcal{I}_0})\{\widehat{\cov}(\bd{z}_k^{\mathcal{I}_0})\}^\dag
-\cov(\bd{x}_k,\bd{z}_k^{\mathcal{I}_0})\{\cov(\bd{z}_k^{\mathcal{I}_0})\}^\dag\|_{\max}\sqrt{|\mathcal{I}_0|}\left\| \widehat{\mb{C}}^{\mathcal{I}_0}(\widehat{\mb{C}}^{\mathcal{I}_0})^\top/n\right\|_2\\
&\lesssim_P  (\lambda_{k,1}/p_k)^{1/2}\delta_\eta
+(\lambda_{k,1}/p_k)^{1/2}\delta_\eta+
\delta_{V_k}\lambda_{k,1}^{1/2}\\
&\lesssim_P  (\lambda_{k,1}/p_k)^{1/2}\delta_\eta
+\delta_{V_k}\lambda_{k,1}^{1/2},
\end{align*}
and thus,
\begin{align}
\lefteqn{\|\widehat{\cov}(\bd{c}_k)-\cov(\bd{c}_k) \|_{\max}}\nonumber\\
&=\|\widehat{\cov}(\bd{x}_k,\bd{z}_k^{\mathcal{I}_0})\{\widehat{\cov}(\bd{z}_k^{\mathcal{I}_0})\}^\dag n^{-1}\widehat{\mb{C}}^{\mathcal{I}_0}(\widehat{\mb{C}}^{\mathcal{I}_0})^\top(\widehat{\cov}(\bd{x}_k,\bd{z}_k^{\mathcal{I}_0})\{\widehat{\cov}(\bd{z}_k^{\mathcal{I}_0})\}^\dag)^\top\nonumber\\
&\qquad -\cov(\bd{x}_k,\bd{z}_k^{\mathcal{I}_0})\{\cov(\bd{z}_k^{\mathcal{I}_0})\}^\dag\cov(\bd{c}^{\mathcal{I}_0})(\cov(\bd{x}_k,\bd{z}_k^{\mathcal{I}_0})\{\cov(\bd{z}_k^{\mathcal{I}_0})\}^\dag)^\top\|_{\max}\nonumber\\
&\le\|\cov(\bd{x}_k,\bd{z}_k^{\mathcal{I}_0})\{\cov(\bd{z}_k^{\mathcal{I}_0})\}^\dag\cov(\bd{c}^{\mathcal{I}_0}) \|_{\max}
|\mathcal{I}_0|\delta_{B_k}
+\delta_{B_k\Sigma_c}
|\mathcal{I}_0|\|\widehat{\cov}(\bd{x}_k,\bd{z}_k^{\mathcal{I}_0})\{\widehat{\cov}(\bd{z}_k^{\mathcal{I}_0})\}^\dag\|_{\max}
\nonumber\\
&\lesssim_P (\lambda_{k,1}/p_k)^{1/2}[(\lambda_{k,1}/p_k)^{1/2}\delta_\eta+
\delta_{V_k}\lambda_{k,1}^{1/2}]\nonumber\\
&\qquad+[(\lambda_{k,1}/p_k)^{1/2}\delta_\eta
+\delta_{V_k}\lambda_{k,1}^{1/2}][ (\lambda_{k,1}/p_k)^{1/2}+(\lambda_{k,1}/p_k)^{1/2}\delta_\eta
+\delta_{V_k}\lambda_{k,1}^{1/2}]\nonumber\\
&\lesssim_P [(\lambda_{k,1}/p_k)^{1/2}\delta_\eta
+\delta_{V_k}\lambda_{k,1}^{1/2}](\lambda_{k,1}/p_k)^{1/2}.
\label{max diff in cov_ck}	
\end{align}

From Assumption~\ref{assump1}\,\ref{assump1(iv)},
\begin{align}
\|\cov(\bd{c}_k)\|_{\max}&\le \max_{1\le i\le p_k}|\var(\bd{c}_k^{[i]}) |\nonumber\\
&=\max_{1\le i\le p_k}|\mb{V}_{xk}^{[i,:]}\mb{\Lambda}_{xk}^{1/2}\mb{H}_k^\top[\cov(\bd{z}_k^{\mathcal{I}_0})]^\dag \mb{A}\mb{N}\cov(\bd{f})(\mb{V}_{xk}^{[i,:]}\mb{\Lambda}_{xk}^{1/2}\mb{H}_k^\top[\cov(\bd{z}_k^{\mathcal{I}_0})]^\dag \mb{A}\mb{N})^\top |\nonumber\\
&\le \max_{1\le i\le p_k}\{
\| \mb{V}_{xk}^{[i,:]}\mb{\Lambda}_{xk}^{1/2}\|_2^2
\|  \mb{H}_k^\top[\cov(\bd{z}_k^{\mathcal{I}_0})]^\dag \mb{A}\mb{N}\cov(\bd{f})  (\mb{H}_k^\top[\cov(\bd{z}_k^{\mathcal{I}_0})]^\dag \mb{A}\mb{N})^\top   \|_2
\nonumber\\
&\lesssim_P  \max_{1\le i\le p_k}\| \mb{V}_{xk}^{[i,:]}\mb{\Lambda}_{xk}^{1/2}\|_2^2\le \max_{1\le i\le p_k}\| \mb{V}_{xk}^{[i,:]}\mb{\Lambda}_{xk}^{1/2}\|_{\max}^2r_k
\nonumber\\
&\lesssim_P \lambda_{k,1}/p_k.
\label{max norm cov_ck}
\end{align}

Denote $\widehat{\var}(\bd{c}_k^{[i]})=\|\widehat{\mb{C}}_k^{[i,:]} \|_F^2/n$ and $\widehat{\var}(\bd{x}_k^{[i]})=\|\widehat{\mb{X}}_k^{[i,:]} \|_F^2/n$.
By the triangle inequality, \eqref{max diff in cov_ck}, \eqref{max norm cov_ck}, $\delta_\eta=o(1)$, and $o(1)=\delta_k\asymp \delta_{V_k}\sqrt{p_k}$ (from 
\eqref{lambda_k=R_k}, \eqref{lambda equal}, and $\lambda_{k,1}\asymp \lambda_{k,r_k}$ in Assumption~\ref{assump1}~\ref{assump1(ii)}),
we have
$\max_{i\le p_k}\widehat{\var}(\bd{c}_k^{[i]})\le \|\widehat{\cov}(\bd{c}_k)\|_{\max}\lesssim_P \lambda_{k,1}/p_k$.
By \eqref{max diff in cov_xk} and $\min_{i\le p_k}\var(\bd{x}_k^{[i]})\ge M_k
\lambda_{r_k}(\cov(\bd{x}_k))/p_k$, we obtain
$\min_{i\le p_k}\widehat{\var}(\bd{x}_k^{[i]})\ge M_k
\lambda_{r_k}(\cov(\bd{x}_k))/p_k-o_P(\lambda_{k,1}/p_k)$.
Then together with \eqref{max diff in cov_xk} and \eqref{max diff in cov_ck}, we have that,
uniformly for all $i=1,\dots, p_k$,
\begin{align*}
\lefteqn{\left|\widehat{\PVE}_c(\bd{x}_k^{[i]})-\PVE_c(\bd{x}_k^{[i]})\right|=\left| \frac{\var(\bd{c}_k^{[i]})}{\var(\bd{x}_k^{[i]})} - \frac{\widehat{\var}(\bd{c}_k^{[i]})}{\widehat{\var}(\bd{x}_k^{[i]})} \right|}\\
&\le
\frac{|\widehat{\var}(\bd{x}_k^{[i]})-\var(\bd{x}_k^{[i]})|}{\widehat{\var}(\bd{x}_k^{[i]})\var(\bd{x}_k^{[i]})}
\widehat{\var}(\bd{c}_k^{[i]})+
\left| \widehat{\var}(\bd{c}_k^{[i]}) -\var(\bd{c}_k^{[i]})   \right| \frac{1}{\var(\bd{x}_k^{[i]})} \\
&\lesssim_P \frac{|\widehat{\var}(\bd{x}_k^{[i]})-\var(\bd{x}_k^{[i]})|}{\var(\bd{x}_k^{[i]})}+
\left| \widehat{\var}(\bd{c}_k^{[i]}) -\var(\bd{c}_k^{[i]})   \right| \frac{1}{\var(\bd{x}_k^{[i]})}\\
&\lesssim_P  \left\{\delta_{V_k}\lambda_{k,1}/\sqrt{p_k}
+[(\lambda_{k,1}/p_k)^{1/2}\delta_\eta
+\delta_{V_k}\lambda_{k,1}^{1/2}](\lambda_{k,1}/p_k)^{1/2}\right\}\frac{1}{\var(\bd{x}_k^{[i]})}\\
&\lesssim_P \delta_\eta+\delta_{V_k}p_k^{1/2}\\
&\lesssim_P \delta_\eta+\delta_k.
\end{align*}
}	
\end{enumerate}	

	The proof is complete.	
\end{proof}

{\color{black}
\begin{proof}{\bf of Corollary~\ref{consistency cor}.}
Let $\cov(\bd{d}_k^{(t)})=\mb{V}_{d_k^{(t)}}\mb{\Lambda}_{d_k^{(t)}}\mb{V}_{d_k^{(t)}}^\top$ be its compact SVD, where $\mb{\Lambda}_{d_k^{(t)}}$ is a diagonal matrix with nonincreasing diagonal elements.
Then, $\bd{f}_k^{(t)}=\mb{\Lambda}_{d_k^{(t)}}^{-1/2}\mb{V}_{d_k^{(t)}}^\top\bd{d}_k^{(t)}$ is an orthonormal basis of $\lspan((\bd{d}_k^{(t)})^\top)$, and 
$\bd{d}_k^{(t)}=\mb{V}_{d_k^{(t)}}\mb{\Lambda}_{d_k^{(t)}}^{1/2}\bd{f}_k^{(t)}$.
Denote
$\mb{F}_k^{(t)}$ and $\mb{D}_k^{(t)}$ to be the sample matrices of $\bd{f}_k^{(t)}$ and $\bd{d}_k^{(t)}$, respectively.
By the central limit theorem,
we have
\be\label{diff in cov fij(1) initial}
\|n^{-1}\mb{F}_j^{(1)}(\mb{F}_k^{(1)} )^\top  -\cov(\bd{f}_j^{(1)}, \bd{f}_k^{(1)}) \|_F=O_P(n^{-1/2}).
\ee
Consequently,
\begin{align}
&\|n^{-1}\mb{D}_k^{(1)}(\mb{D}_k^{(1)} )^\top-\cov(\bd{d}_k^{(1)}) \|_2\le\|n^{-1}\mb{D}_k^{(1)}(\mb{D}_k^{(1)} )^\top-\cov(\bd{d}_k^{(1)}) \|_F \nonumber\\
&\quad=\|\mb{V}_{d_k^{(t)}}\mb{\Lambda}_{d_k^{(t)}}^{1/2}n^{-1}\mb{F}_k^{(1)}(\mb{F}_k^{(1)} )^\top\mb{\Lambda}_{d_k^{(t)}}^{1/2} \mb{V}_{d_k^{(t)}}^\top -\mb{V}_{d_k^{(t)}}\mb{\Lambda}_{d_k^{(t)}}^{1/2}\cov(\bd{f}^{(1)})\mb{\Lambda}_{d_k^{(t)}}^{1/2} \mb{V}_{d_k^{(t)}}^\top  \|_F\nonumber\\
&\quad\lesssim_P \lambda_1(\cov(\bd{d}_k^{(1)}))n^{-1/2}.
\label{DD-covd part 1}
\end{align}
Moreover, 
\begin{align}
\|\mb{D}_k^{(1)}\|_2&=\|\mb{D}_k^{(1)} (\mb{D}_k^{(1)})^{\top}\|_2^{1/2}
\le n^{1/2}(\|n^{-1}\mb{D}_k^{(1)}(\mb{D}_k^{(1)} )^\top-\cov(\bd{d}_k^{(1)}) \|_2+\|\cov(\bd{d}_k^{(1)})\|_2)^{1/2}
\nonumber\\
&\lesssim_P [n\lambda_1(\cov(\bd{d}_k^{(1)}))]^{1/2}
\label{scale of D_k}
\end{align}
On the other hand, it follows from \eqref{scale of norm of X} and \eqref{diff in D in relative norm} that
\be\label{unscaled diff in D_k}
\|\widehat{\mb{D}}_k^{(1)}-\mb{D}_k^{(1)} \|_2
\lesssim_P \delta_\eta [n\lambda_1(\cov(\bd{x}_k))]^{1/2}.
\ee
Then by \eqref{diff in D in relative norm} and \eqref{scale of D_k},
\begin{align*}
\lefteqn{\|\widehat{\mb{D}}_k^{(1)}(\widehat{\mb{D}}_k^{(1)})^\top- \mb{D}_k^{(1)}(\mb{D}_k^{(1)})^\top\|_F}
\\
&\lesssim_P 
\{[n\lambda_1(\cov(\bd{d}_k^{(1)}))]^{1/2}+\delta_\eta [n\lambda_1(\cov(\bd{x}_k))]^{1/2}
\}\delta_\eta [n\lambda_1(\cov(\bd{x}_k))]^{1/2}.
\end{align*}
Using the above inequality,  \eqref{DD-covd part 1} and the triangle inequality yields
\begin{align}
\lefteqn{\|n^{-1}\widehat{\mb{D}}_k^{(1)}(\widehat{\mb{D}}_k^{(1)})^\top-\cov(\bd{d}_k^{(1)})\|_F}\nonumber\\
&\le\sqrt{3r_k}\|n^{-1}\widehat{\mb{D}}_k^{(1)}(\widehat{\mb{D}}_k^{(1)})^\top-\cov(\bd{d}_k^{(1)})\|_2\nonumber\\
&\lesssim_P\{[n\lambda_1(\cov(\bd{d}_k^{(1)}))]^{1/2}+\delta_\eta [n\lambda_1(\cov(\bd{x}_k))]^{1/2}
\}\delta_\eta [n\lambda_1(\cov(\bd{x}_k))]^{1/2}/n  +\lambda_1(\cov(\bd{d}_k^{(1)}))n^{-1/2}\nonumber\\
&\lesssim_P \delta_\eta \lambda_1(\cov(\bd{x}_k)),
\label{unscaled diff in covd}
\end{align}
where we used
\begin{align}
\lambda_1(\cov(\bd{d}_k))& \le \tr(\cov(\bd{d}_k))=\tr(\cov(\bd{x}_k-\bd{c}_k))\nonumber\\
&\le \tr(\cov(\bd{x}_k))+\tr(\cov(\bd{c}_k))+2|\tr(\cov(\bd{x}_k,\bd{c}_k))|\nonumber\\
&\lesssim_P\lambda_1(\cov(\bd{x}_k))
\label{eig(covdk)<=eig(covxk)}
\end{align}
following from
\begin{align*}
\tr(\cov(\bd{c}_k) )&\le |\mathcal{I}_0|\| \cov(\bd{c}_k)   \|_2\\
&=|\mathcal{I}_0|\|\mb{V}_{xk}\mb{\Lambda}_{xk}^{1/2}\mb{H}_k^\top[\cov(\bd{z}_k^{\mathcal{I}_0})]^\dag \mb{A}\mb{N}\cov(\bd{f})(\mb{V}_{xk}\mb{\Lambda}_{xk}^{1/2}\mb{H}_k^\top[\cov(\bd{z}_k^{\mathcal{I}_0})]^\dag \mb{A}\mb{N})^\top \|_2\\
&\lesssim_P\lambda_1(\cov(\bd{x}_k))
\end{align*}
and
\begin{align*}
|\tr(\cov(\bd{x}_k,\bd{c}_k))|&=\sum_{i=1}^{p_k}|\cov(\bd{x}_k^{[i]},\bd{c}_k^{[i]})|\le \sum_{i=1}^{p_k} [\var(\bd{x}_k^{[i]})]^{1/2} [\var((\bd{c}_k^{[i]})]^{1/2}\\
&\le[\sum_{i=1}^{p_k}  \var(\bd{x}_k^{[i]}) ]^{1/2} [\sum_{i=1}^{p_k}  \var(\bd{c}_k^{[i]})]^{1/2}
=[\tr(\cov(\bd{x}_k))\tr(\cov(\bd{c}_k))]^{1/2}\\
&\lesssim_P \lambda_1(\cov(\bd{x}_k)).
\end{align*}

Denote $r_k^{(t)}=\rank(\cov(\bd{d}_k^{(t)}))$.
By Theorem~2 in \citet{Yu15}, \eqref{unscaled diff in covd}, 
and the condition that $\lambda_{k,1}^{(1)}>\kappa^{(1)}\lambda_{r_k}(\cov(\bd{x}_k))$, $\lambda_{k,1}^{(1)}\asymp\lambda_{k,m_k^{(1)}}^{(1)}$, and 
$(\lambda_{k,\ell}^{(1)}-\lambda_{k,\ell+1}^{(1)})/\lambda_{k,\ell}^{(1)}\ge \delta^{(1)}$ for $1\le \ell\le m_k^{(1)}$,
%{\color{black}[$(\lambda_{d_k^{(t)},\ell}-\lambda_{d_k^{(t)},\ell+1})/\lambda_{d_k^{(t)},\ell}\ge \delta_t$, $\lambda_{d_k^{(t)},1}\asymp\lambda_{d_k^{(t)},r}$, $\lambda_{d_k^{(t)},r}>\kappa\lambda_{r_k}(\cov(\bd{x}_k))$]},
we have that
there exists 
$\widehat{\mb{V}}_{d_k^{(1)}}\in \mathbb{R}^{p_k\times r_k^{(1)}}$,
whose columns are the left-singular vectors of $n^{-1}\widehat{\mb{D}}_k^{(1)}(\widehat{\mb{D}}_k^{(1)})^\top$ 
corresponding to its $r_k^{(1)}$ largest singular values,
such that
\be\label{diff in Vdt}
\|\widehat{\mb{V}}_{d_k^{(1)}}-\mb{V}_{d_k^{(1)}}\|_F\lesssim_P 
\delta_\eta .
\ee
From Weyl's inequality and \eqref{unscaled diff in covd}, 
\be\label{diff in Lambda_dt}
\|\widehat{\mb{\Lambda}}_{d_k^{(1)}}-\mb{\Lambda}_{d_k^{(1)}}\|_{\max}\le \|n^{-1}\widehat{\mb{D}}_k^{(1)}(\widehat{\mb{D}}_k^{(1)})^\top-\cov(\bd{d}_k^{(1)})   \|_2\lesssim_P \delta_\eta \lambda_1(\cov(\bd{x}_k)),
\ee
where $\widehat{\mb{\Lambda}}_{d_k^{(1)}}\in \mathbb{R}^{r_k^{(1)}\times r_k^{(1)}}$ is a diagonal matrix
with nonincreasing diagonal elements being
the $r_k^{(1)}$ largest singular values of $n^{-1}\widehat{\mb{D}}_k^{(1)}(\widehat{\mb{D}}_k^{(1)})^\top$.
Then by $\delta_\eta=o(1)$ and $\lambda_{r_k^{(1)}}(\cov(\bd{d}_k^{(1)}))\asymp\lambda_1(\cov(\bd{d}_k^{(1)}))>\kappa^{(1)} \lambda_{r_k}(\cov(\bd{x}_k))\asymp \lambda_1(\cov(\bd{x}_k))$, 
for $1\le \ell\le r_k^{(1)}$
we have
\be\label{lower bound of eig_cov_dk1}
\widehat{\mb{\Lambda}}_{d_k^{(1)}}^{[\ell,\ell]}\ge\mb{\Lambda}_{d_k^{(1)}}^{[\ell,\ell]}-|\widehat{\mb{\Lambda}}_{d_k^{(1)}}^{[\ell,\ell]}-\mb{\Lambda}_{d_k^{(1)}}^{[\ell,\ell]}|\ge \kappa^{(1)}_*(1-o_P(1))\lambda_{r_k}(\cov(\bd{x}_k))
\ee
with a constant $\kappa^{(1)}_*>0$,
and consequently from the mean value theorem, 
\begin{align}\label{diff in half Lambda_dt}
\|\widehat{\mb{\Lambda}}_{d_k^{(1)}}^{1/2}-\mb{\Lambda}_{d_k^{(1)}}^{1/2}\|_{\max}&\le \frac{1}{2}[\kappa^{(1)}_*(1-o_P(1))\lambda_{r_k}(\cov(\bd{x}_k))]^{-1/2}
\|\widehat{\mb{\Lambda}}_{d_k^{(1)}}-\mb{\Lambda}_{d_k^{(1)}}\|_{\max}\nonumber\\
&\lesssim_P \delta_\eta \lambda_1^{1/2}(\cov(\bd{x}_k)).
\end{align}
and
\begin{align}\label{diff in neg half Lambda_dt}
\|(\widehat{\mb{\Lambda}}_{d_k^{(1)}}^{1/2})^\dag-\mb{\Lambda}_{d_k^{(1)}}^{-1/2}\|_{\max}
&\le \frac{1}{2}[\kappa^{(1)}_*(1-o_P(1))\lambda_{r_k}(\cov(\bd{x}_k))]^{-3/2}
\|\widehat{\mb{\Lambda}}_{d_k^{(1)}}-\mb{\Lambda}_{d_k^{(1)}}\|_{\max}\nonumber\\
&\lesssim_P \delta_\eta \lambda_1^{-1/2}(\cov(\bd{x}_k)).
\end{align}
By \eqref{chain ineq}, \eqref{diff in Vdt}, \eqref{diff in half Lambda_dt} and \eqref{diff in neg half Lambda_dt},
\be\label{diff in VLambda_dt_sq}
\| \widehat{\mb{V}}_{d_k^{(1)}}\widehat{\mb{\Lambda}}_{d_k^{(1)}}^{1/2}  -\mb{V}_{d_k^{(1)}}\mb{\Lambda}_{d_k^{(1)}}^{1/2}\|_2
\lesssim_P \delta_\eta \lambda_1^{1/2}(\cov(\bd{x}_k))
\ee
and
\be\label{diff in Lambda_dt_sq-1 V}
\|(\widehat{\mb{\Lambda}}_{d_k^{(1)}}^{1/2} )^\dag \widehat{\mb{V}}_{d_k^{(1)}}^\top -\mb{\Lambda}_{d_k^{(1)}}^{-1/2}\mb{V}_{d_k^{(1)}}^\top\|_2
\lesssim_P \delta_\eta \lambda_1^{-1/2}(\cov(\bd{x}_k)).
\ee
Define $\widehat{\mb{F}}_k^{(t)}=(\widehat{\mb{\Lambda}}_{d_k^{(t)}}^{1/2})^\dag\widehat{\mb{V}}_{d_k^{(t)}}^\top\widehat{\mb{D}}_k^{(t)}$. By \eqref{chain ineq}, \eqref{scale of D_k}, \eqref{unscaled diff in D_k} and \eqref{diff in Lambda_dt_sq-1 V},
\be\label{diff in F_kt}
\| \widehat{\mb{F}}_k^{(1)}-\mb{F}_k^{(1)}     \|_2
=\|(\widehat{\mb{\Lambda}}_{d_k^{(1)}}^{1/2})^\dag\widehat{\mb{V}}_{d_k^{(1)}}^\top\widehat{\mb{D}}_k^{(1)}-\mb{\Lambda}_{d_k^{(1)}}^{-1/2}\mb{V}_{d_k^{(1)}}^\top\mb{D}_k^{(1)} \|_2
\lesssim_P\delta_\eta n^{1/2}.
\ee
Then by \eqref{chain ineq} again,
$
\|  n^{-1}\widehat{\mb{F}}_j^{(1)}(\widehat{\mb{F}}_k^{(1)})^\top-   n^{-1}\mb{F}_j^{(1)}(\mb{F}_k^{(1)})^\top\|_2\lesssim_P \delta_\eta
$.
Using the above inequality, \eqref{diff in cov fij(1) initial} and the triangle inequality yields
\be\label{diff in cov fij(1)}
\|  n^{-1}\widehat{\mb{F}}_j^{(1)}(\widehat{\mb{F}}_k^{(1)})^\top-   \cov(\bd{f}_j^{(1)},\bd{f}_k^{(1)})\|_2\lesssim_P \delta_\eta.
\ee

From the central limit theorem, 
$\|\mb{F}_k^{(1)}(\mb{F}_k^{(1)})^\top/n-\mb{I}_{r_k^{(1)}\times r_k^{(1)}} \|_{\max}=O_P(n^{-1/2})$
and thus
\be\label{scale of F_k1}
\|\mb{F}_k^{(1)}\|_F^2=\tr(\mb{F}_k^{(1)}(\mb{F}_k^{(1)})^\top)
=O_P(\sqrt{n}).
\ee

Following parts 2 and 3 in the proof of Theorem~\ref{consistency thm} with \eqref{unscaled diff in D_k}, \eqref{unscaled diff in covd}, \eqref{eig(covdk)<=eig(covxk)}, \eqref{diff in Lambda_dt}, \eqref{lower bound of eig_cov_dk1}, \eqref{diff in VLambda_dt_sq}, \eqref{diff in F_kt}, \eqref{diff in cov fij(1)} and \eqref{scale of F_k1}, we can obtain the error bounds given in \eqref{level-t: RNE of C and D} and \eqref{level-t: dataset-level PVE} for $T=1$.

Now we consider to prove \eqref{level-t: variable-level PVE} for $T=1$.
By
$\|\cov(\bd{x}_k)\|_{\max}\le r_k \kappa_B^2 \lambda_{k,1}/p_k$ (from Assumption~\ref{assump1}\,\ref{assump1(iv)}) and \eqref{max norm cov_ck}, we have
\begin{align}
\|\cov(\bd{d}_k)\|_{\max}&=\|\cov(\bd{x}_k-\bd{c}_k) \|_{\max}
\le \|\cov(\bd{x}_k) \|_{\max}+\|\cov(\bd{c}_k) \|_{\max}
+2\|\cov(\bd{x}_k,\bd{c}_k)\|_{\max}\nonumber\\
&\le \|\cov(\bd{x}_k) \|_{\max}+\|\cov(\bd{c}_k) \|_{\max}
+2\max_{1\le i,j\le p_k}[\var(\bd{x}_k^{[i]})\var(\bd{c}_k^{[j]})]^{1/2}\nonumber\\
&\lesssim_P \lambda_{k,1}/p_k.
\label{max diff in cov_dk}
\end{align}
Then by \eqref{lambda equal}, 
$\lambda_{k,1}\asymp \lambda_{k,r_k}$, 
%in Assumption~\ref{assump1}~\ref{assump1(ii)}, 
and
$
\kappa^{(1)} \lambda_{r_k}(\cov(\bd{x}_k))\|\mb{V}_{d_k^{(1)}}\|_{\max}^2\le 
\lambda_1(\cov(\bd{d}_k)) \|\mb{V}_{d_k^{(1)}}\|_{\max}^2 
\asymp\lambda_{r_k^{(1)}}(\cov(\bd{d}_k)) \|\mb{V}_{d_k^{(1)}}\|_{\max}^2 
\le \max\limits_{1\le i\le p_k}\sum_{j=1}^{r_k^{(1)}}(\mb{V}_{d_k^{(1)}}^{[i,j]})^2\lambda_j(\cov(\bd{d}_k))=\max\limits_{1\le i\le p_k}\var(\bd{d}_k^{[i]})\lesssim_P\lambda_{k,1}/p_k,
$
we have
\be\label{max norm Vd1}
\|\mb{V}_{d_k^{(1)}}\|_{\max}\lesssim_P 1/\sqrt{p_k}.
\ee
%{\color{black}impose subgaussian on d. And lower bound of for min var($d_k$), distinct eigenvalue of cov($d_k$)}
Then following part 4 in the proof of Theorem~\ref{consistency thm}
with  \eqref{max diff in cov_dk}, \eqref{max norm Vd1}, \eqref{diff in Lambda_dt}, \eqref{diff in half Lambda_dt} and \eqref{eig(covdk)<=eig(covxk)}, 
we can obtain the error bound in \eqref{level-t: variable-level PVE} for $T=1$.

The proof of \eqref{level-t: RNE of C and D}--\eqref{level-t: variable-level PVE} for any fixed $T\ge 1$ follows the same way
as that for $T=1$.
\end{proof}

}

\section{Additional Simulation Results}\label{sec: add simul}
\begin{sidewaysfigure}[p!]
	\begin{subfigure}[b]{0.3\textwidth}
			\centering
		\includegraphics[width=1\textwidth]{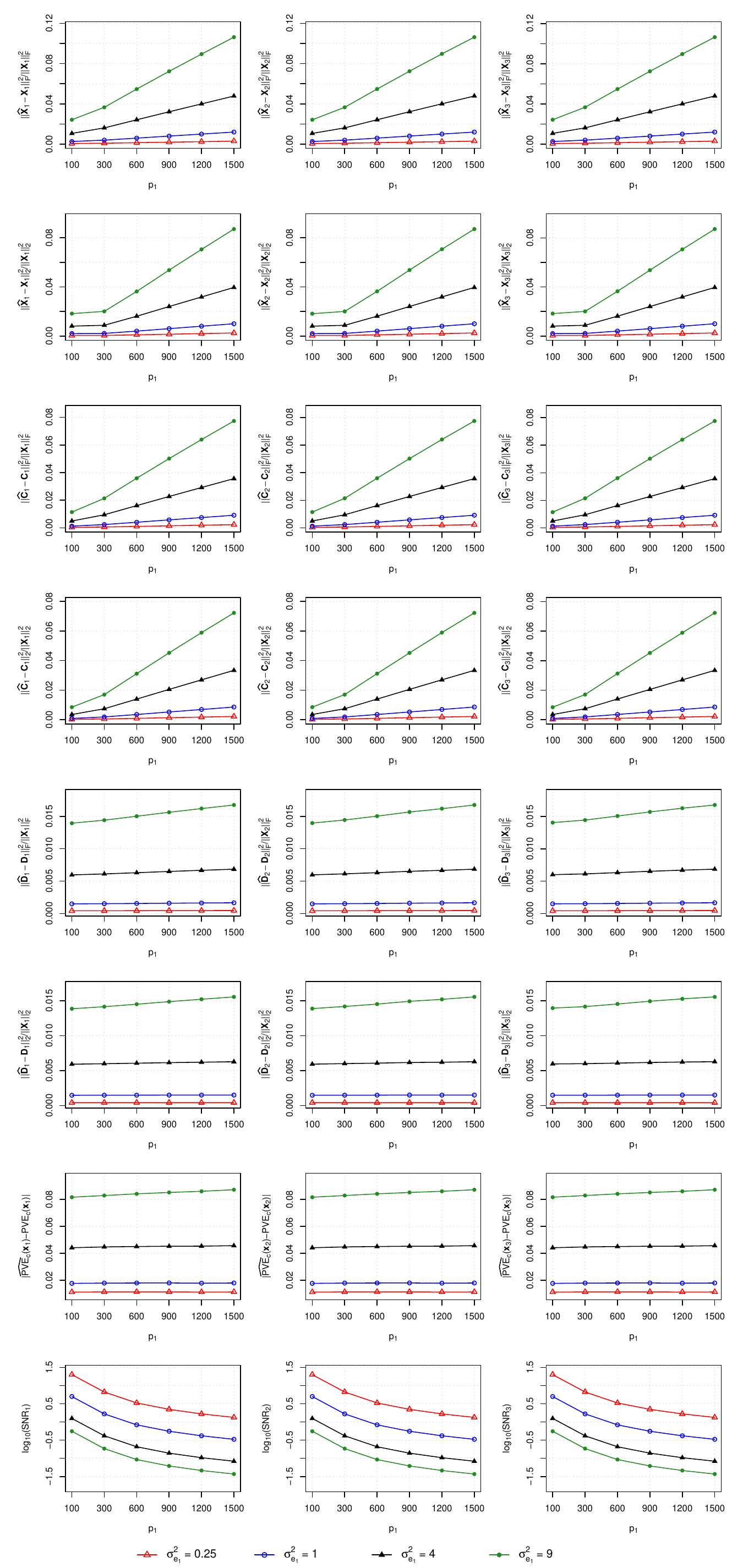}
		\caption{$\theta_z=10^\circ$}
		\label{noise_level}
	\end{subfigure}
	\hspace{0.05\textwidth}
	\begin{subfigure}[b]{0.3\textwidth}
			\centering
		\includegraphics[width=1\textwidth]{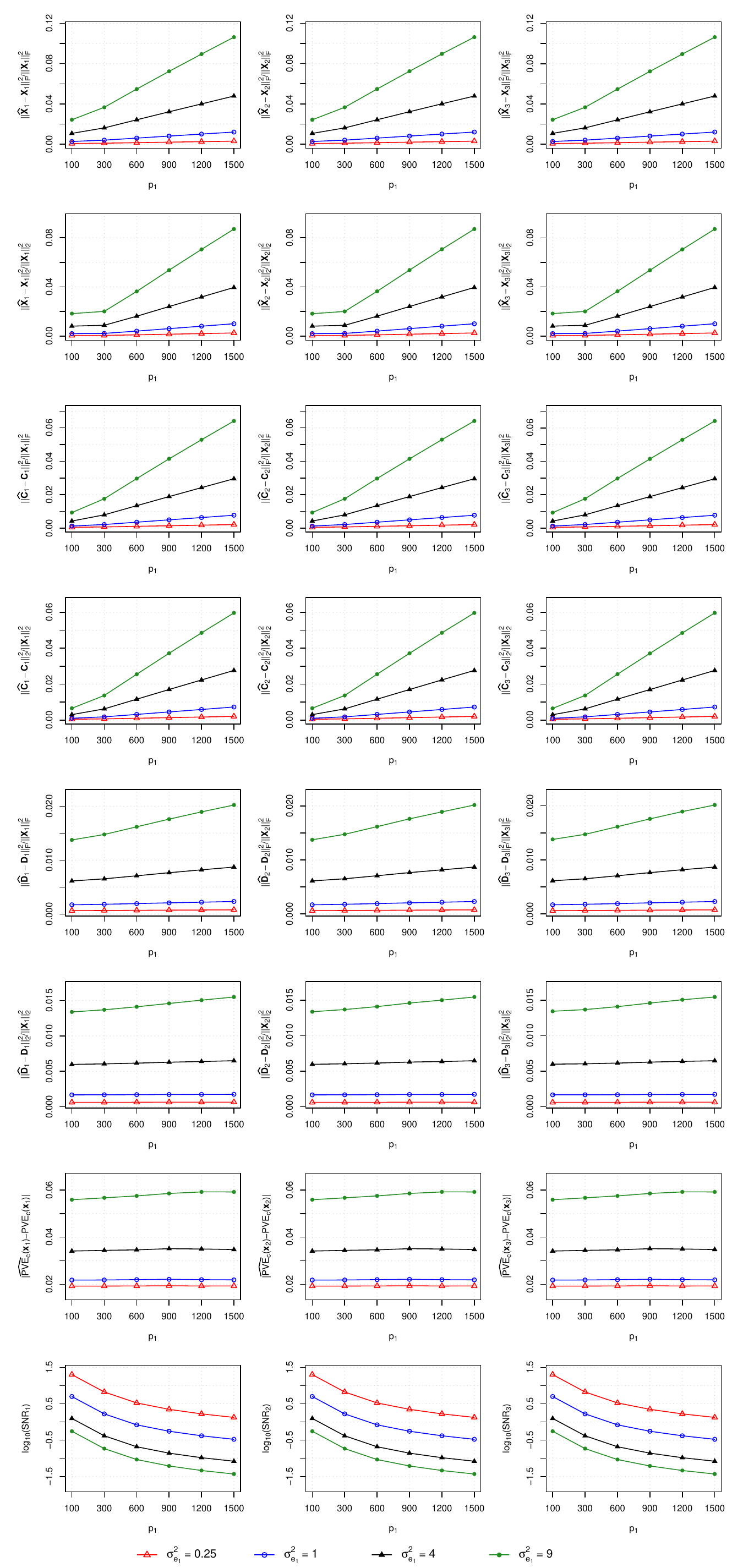}
		\caption{$\theta_z=20^\circ$}
		\label{noise_level}
	\end{subfigure}
	\hspace{0.05\textwidth}
	\begin{subfigure}[b]{0.3\textwidth}
		\centering
	\includegraphics[width=1\textwidth]{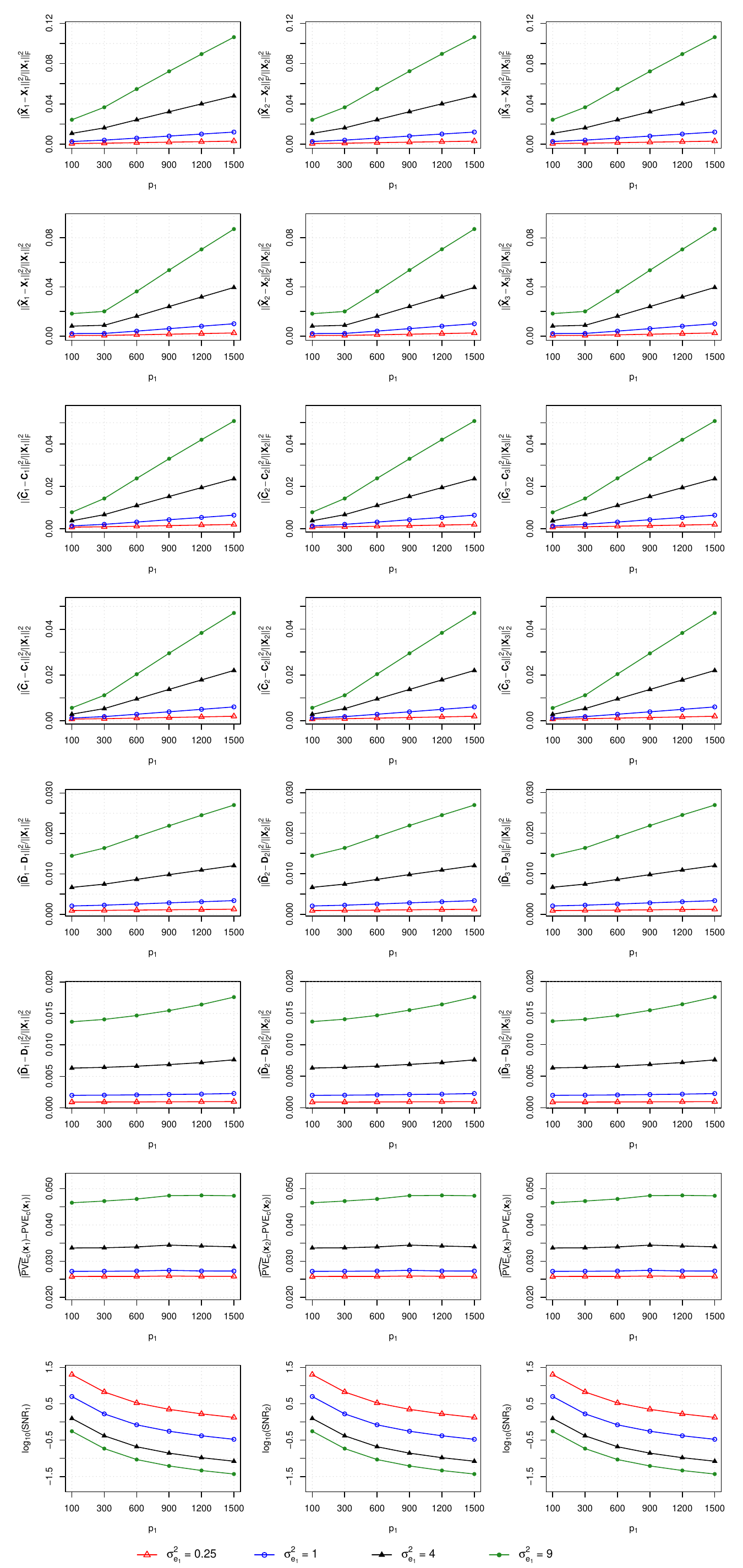}
	\caption{$\theta_z=30^\circ$}
	\label{noise_level}
\end{subfigure}
	\caption{Average errors of D-GCCA estimates over 1000 replications for Setup 1.1 with $\theta_z\in\{10^\circ,20^\circ,30^\circ\}$.}
	\label{figA: Setup1.1 error plot F}
\end{sidewaysfigure}

In Setup 1.1, 
the angle $\theta_z=10^\circ,20^\circ,30^\circ,40^\circ,50^\circ,60^\circ,70^\circ$ corresponds to 
$\PVE_c(\bd{x}_k)=0.853, 0.702, 0.552, 0.409, 0.279, 0.167, 0.079$ for all $k\in \{1,2,3\}$.
In Setup 2.1, the covariance matrix $\cov(\bd{f})\in \mathbb{R}^{15\times 15}$ has blocks 
\vspace{-0.0001cm}
\begin{align*}
&\cov(\bd{f}_1,\bd{f}_2)\\
&=\begin{bsmallmatrix}
0.02498103503160578 & -0.3734791596502449  &
-0.1482674122573037  & -0.3913807076061239  &
-0.05845072081373771 \\
0.1298912403724416  & -0.2915966482089937  &
-0.703223066831662   & -0.286977394728156   &
-0.07037562289439672 \\
-0.4691315902716665  & -0.02216628581934877 &
-0.05789731182102772 & -0.1224434530178697  &
0.7359965879693088  \\
-0.005270967060252731& -0.1916047000827934  &
0.1572469950904809  & -0.1862928969932901  &
0.0648022978041196  \\
0.3309749556233325  &  0.2910731038141944  &
-0.2222302484678626  &  0.4183644600274041  &
-0.09116219316544609 
\end{bsmallmatrix},
\end{align*}
\begin{align*}
&\cov(\bd{f}_1,\bd{f}_3)\\
&=
\begin{bsmallmatrix}
-0.1652455953442644 &  0.07288409202801582&  0.4797927991048995 &
-0.1974810941368655 &  0.2123320697504773 \\
-0.3889488816571995 &  0.05377416249857463&  0.5653871787847853 &
0.03845218160536631& -0.2069628634535125 \\
0.4125592431747815 & -0.7372033575312142 &  0.2721804829221633 &
-0.0862772040030661 & -0.2227478031028198 \\
-0.02345535210198419& -0.1075518721538277 &  0.1394751370539585 &
-0.1625882523272944 &  0.3301641568167817 \\
-0.3328426143159536 & -0.09361178321406048& -0.4483940610130605 &
0.3455811570541347 & -0.09767404221183135
\end{bsmallmatrix},
\end{align*}
\begin{align*}
&\cov(\bd{f}_2,\bd{f}_3)\\
&=
\begin{bsmallmatrix}
-0.1234093117538375  &  0.2223022967058531  &
-0.3593383789512091  &  0.04344070064196999 &
0.2617381817815529  \\
-0.09993460814692552 & -0.008819786526375878&
-0.4039397802979183  &  0.2933537865045707  &
-0.2650032054127345  \\
0.5075563895372593  & -0.1098865559264541  &
-0.4771360952896037  & -0.1119099874049149  &
0.2079731636733454  \\
-0.08232391689469482 & -0.01395485249078317 &
-0.5724368834706903  &  0.3121430368957581  &
-0.1821568224740747  \\
0.3937761144502051  & -0.6998227270213208  &
0.1161733947993463  & -0.04568041770157075 &
-0.1795827017135321  
\end{bsmallmatrix},
\end{align*}
and $\cov(\bd{f}_k)=\mb{I}_{5\times 5}$ for $k=1,2,3$.
Figures \ref{figA: Setup1.1 error plot F}--\ref{figA: Setup 1.2 prop nuiparameter plot} show the additional simulation results for Setups 1.1-2.2. The result analysis described in Section~\ref{Sec: simul finite sample of proposed} also holds here.

\begin{sidewaysfigure}[p!]
\begin{subfigure}[b]{0.3\textwidth}
		\centering
	\includegraphics[width=1\textwidth]{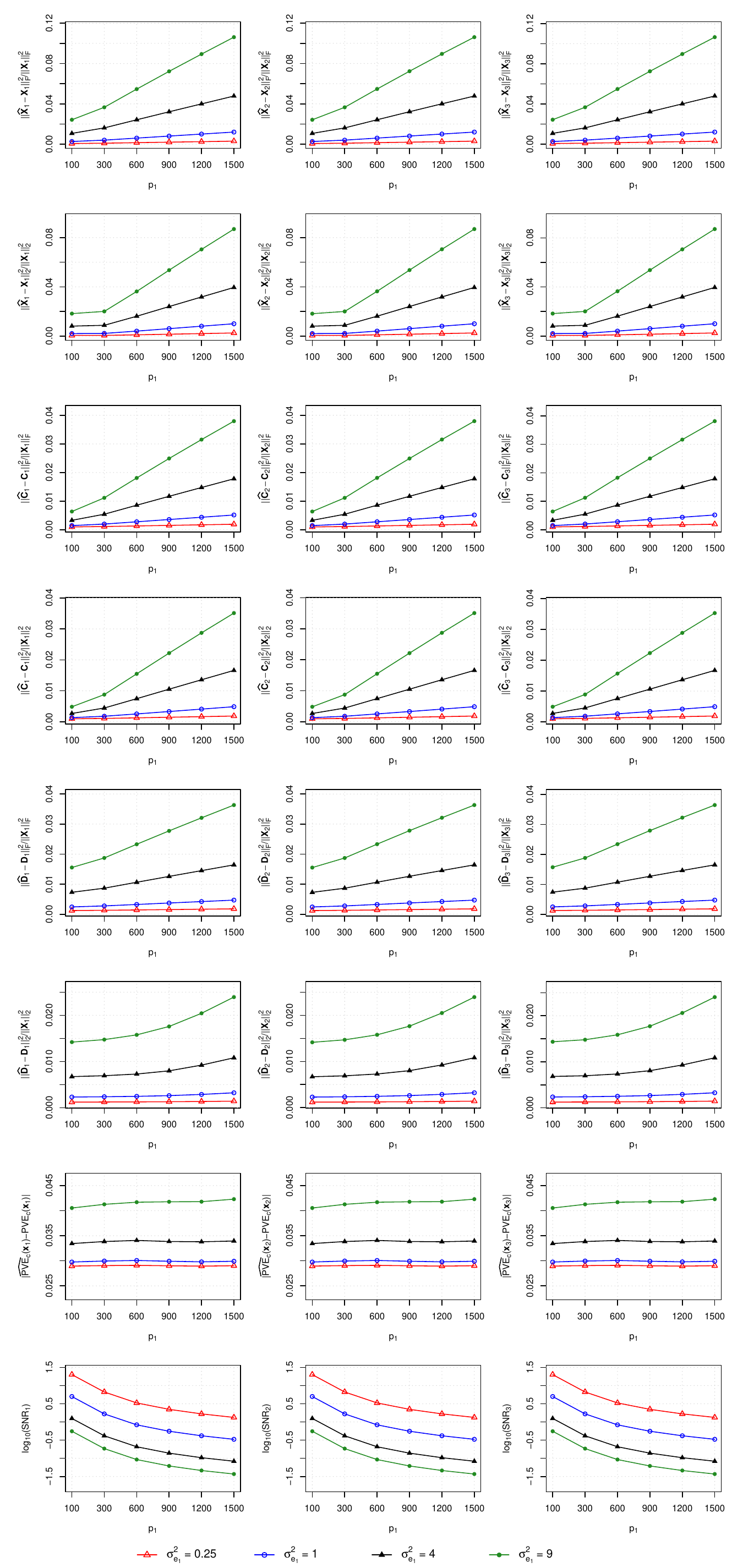}
	\caption{$\theta_z=40^\circ$}
	\label{noise_level}
\end{subfigure}
	\hspace{0.05\textwidth}
\begin{subfigure}[b]{0.3\textwidth}
		\centering
	\includegraphics[width=1\textwidth]{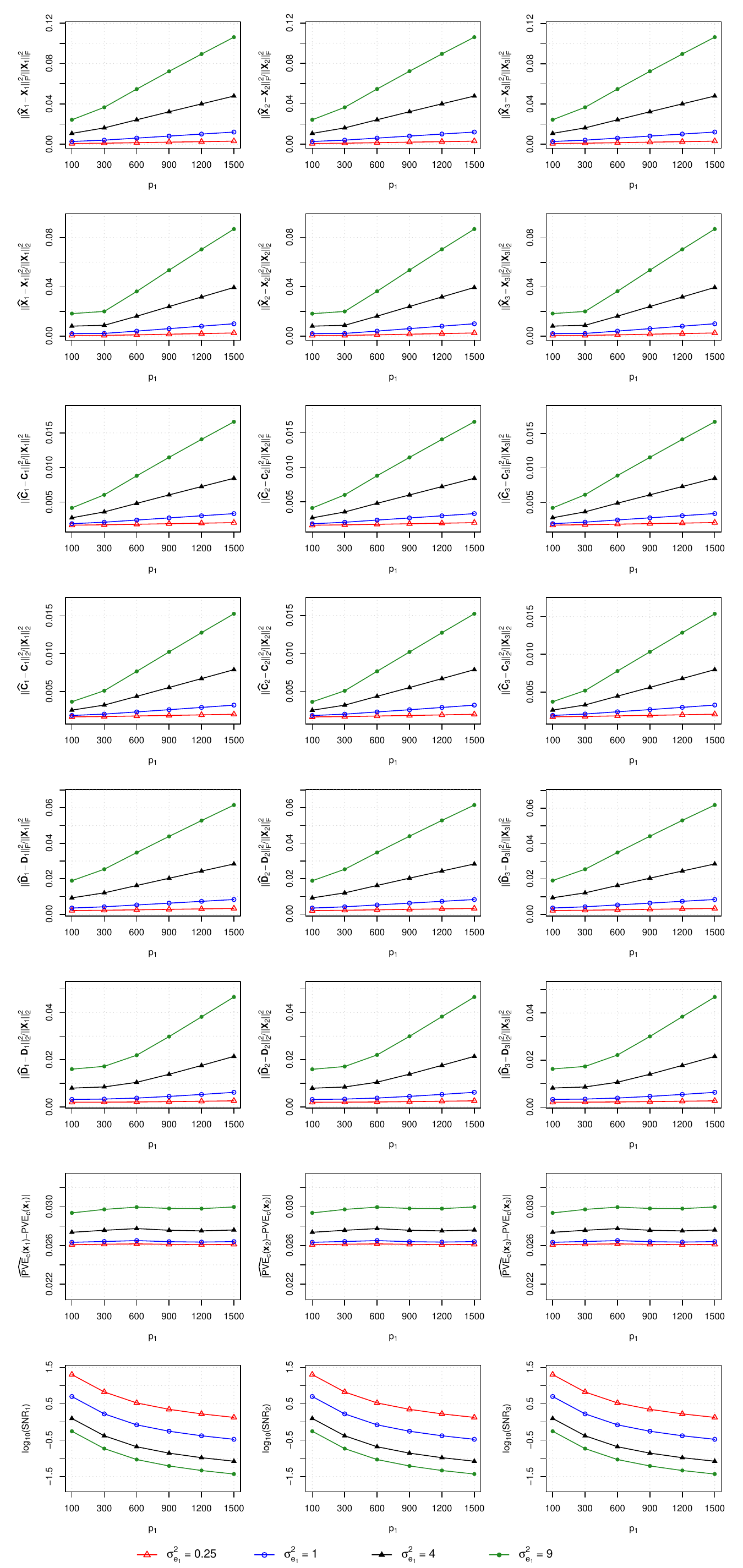}
	\caption{$\theta_z=60^\circ$}
	\label{noise_level}
\end{subfigure}
	\hspace{0.05\textwidth}
\begin{subfigure}[b]{0.3\textwidth}
		\centering
	\includegraphics[width=1\textwidth]{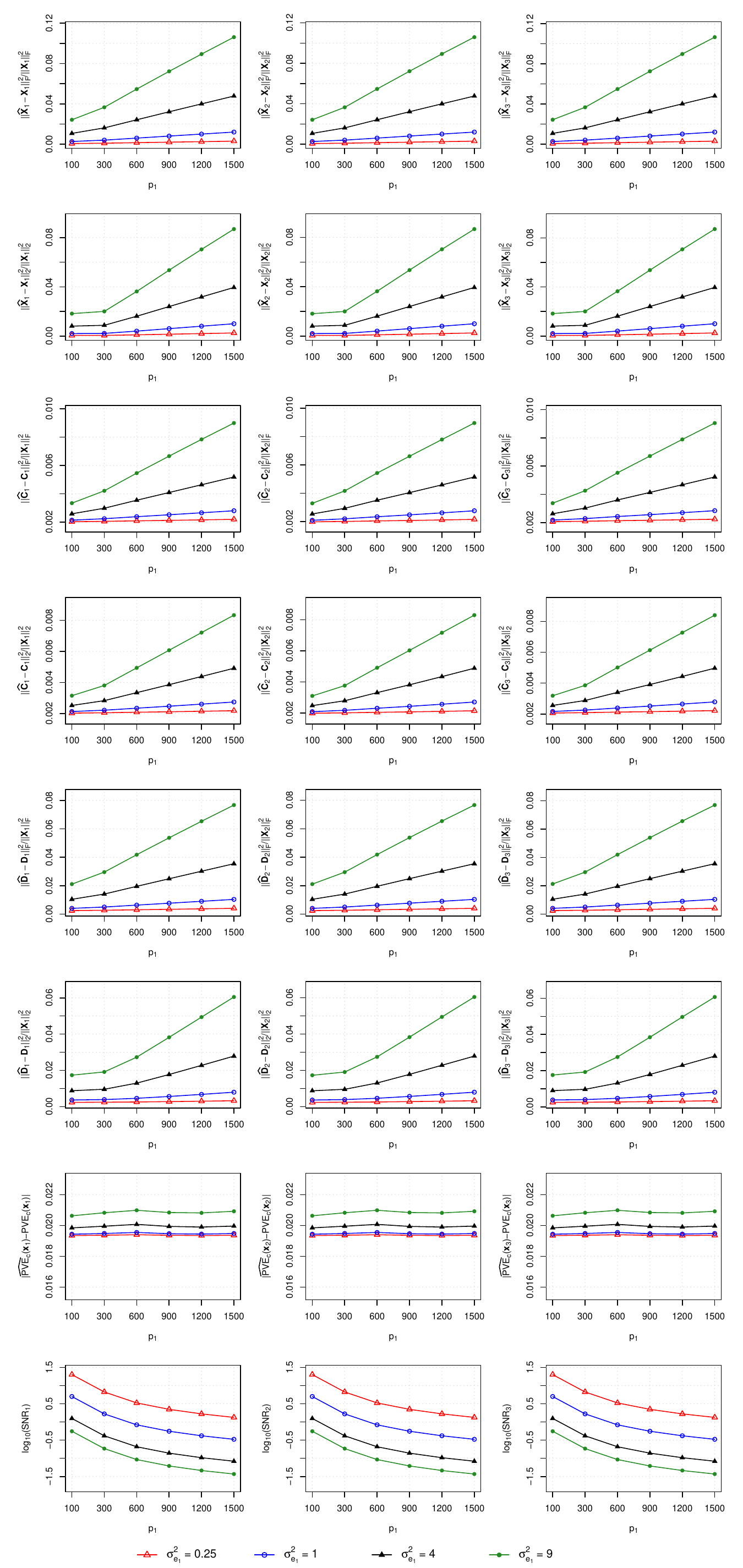}
	\caption{$\theta_z=70^\circ$}
	\label{noise_level}
\end{subfigure}
	\caption{Average errors of D-GCCA estimates over 1000 replications for Setup 1.1 with $\theta_z\in\{40^\circ,60^\circ,70^\circ\}$.}
	%\label{figA: Setup1.1 error plot F}
\end{sidewaysfigure}

\begin{sidewaysfigure}[p!]
	\begin{subfigure}[b]{0.3\textwidth}
			\centering
		\includegraphics[width=1\textwidth]{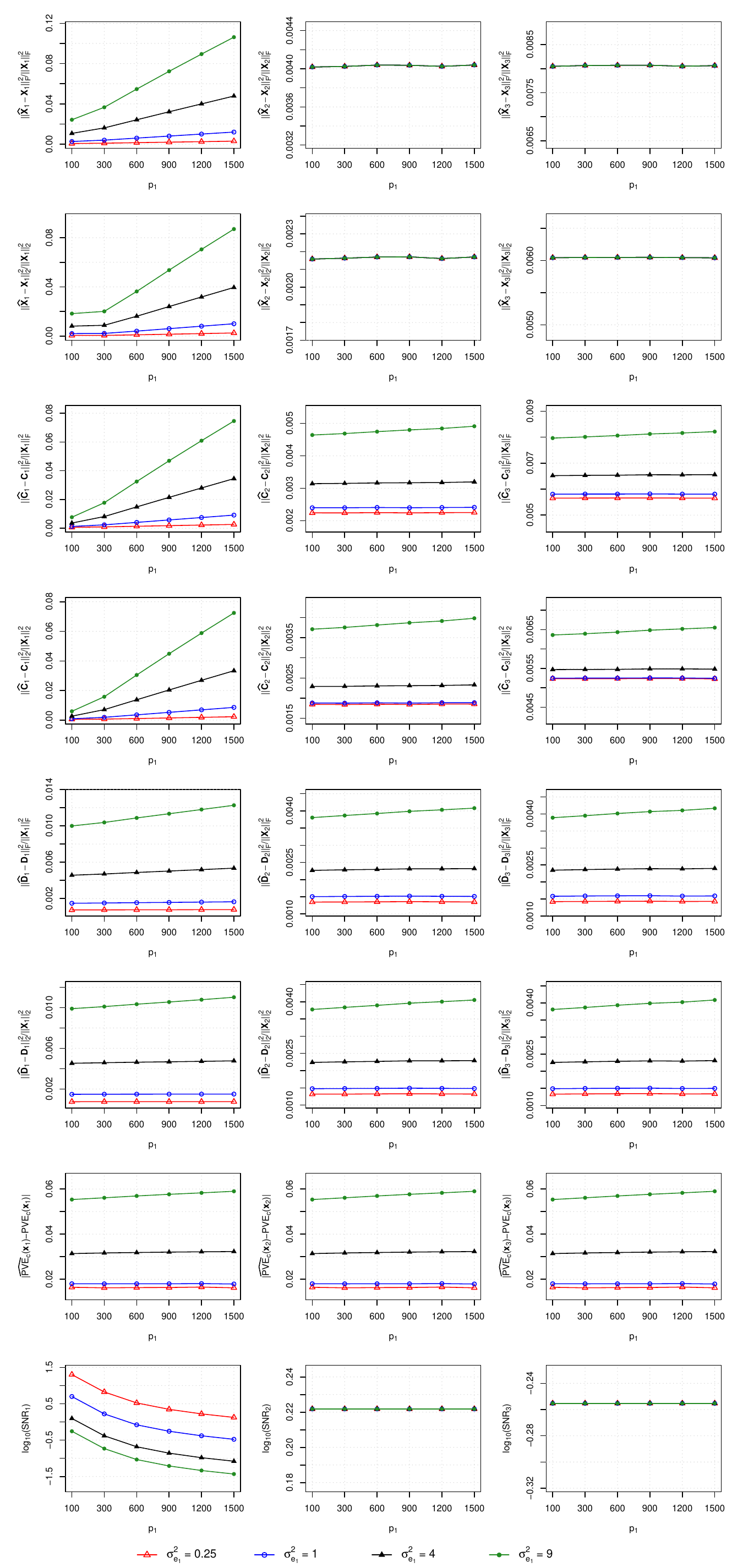}
		\caption{$\theta_z=10^\circ$}
		\label{noise_level}
	\end{subfigure}
		\hspace{0.05\textwidth}
	\begin{subfigure}[b]{0.3\textwidth}
			\centering
		\includegraphics[width=1\textwidth]{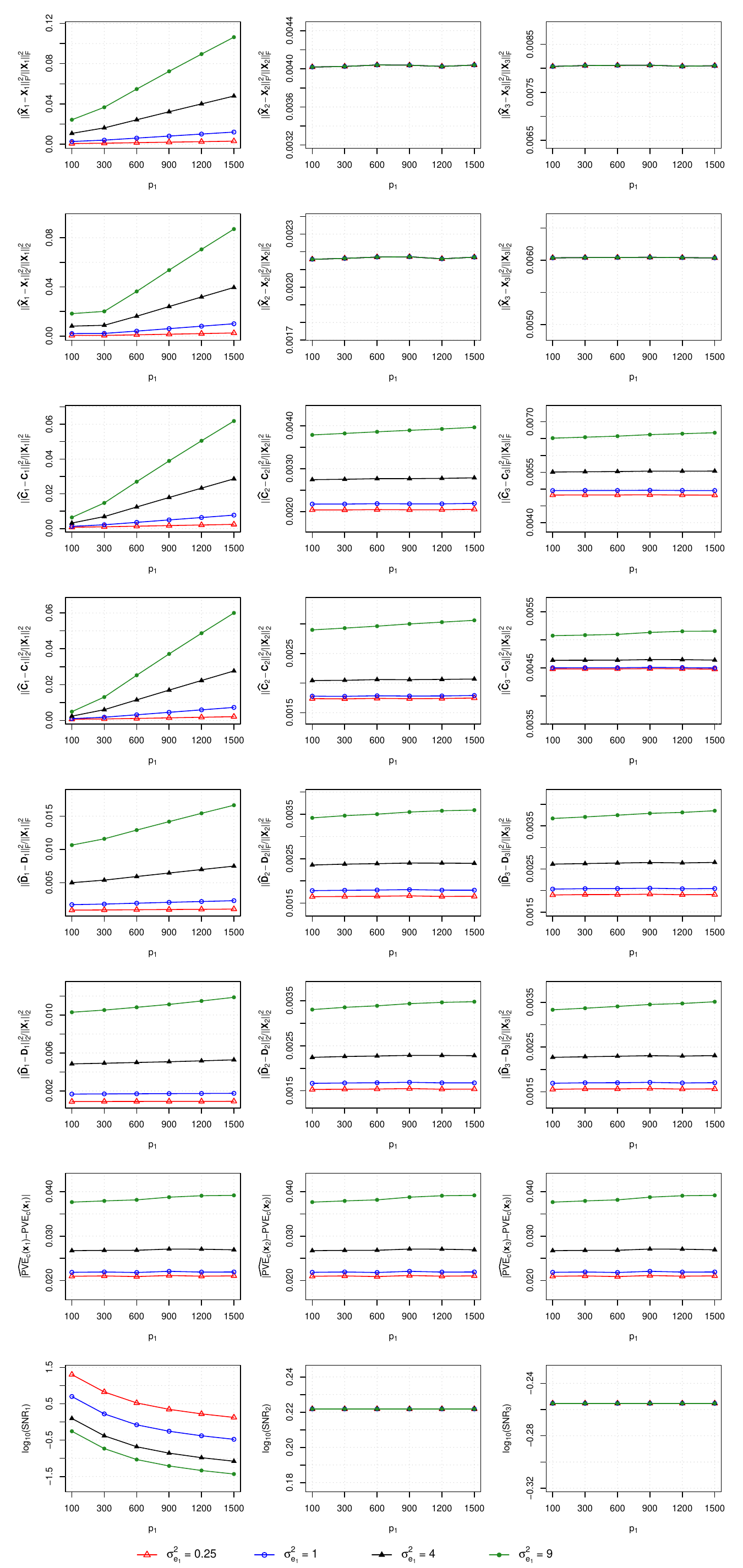}
		\caption{$\theta_z=20^\circ$}
		\label{noise_level}
	\end{subfigure}
		\hspace{0.05\textwidth}
	\begin{subfigure}[b]{0.3\textwidth}
			\centering
		\includegraphics[width=1\textwidth]{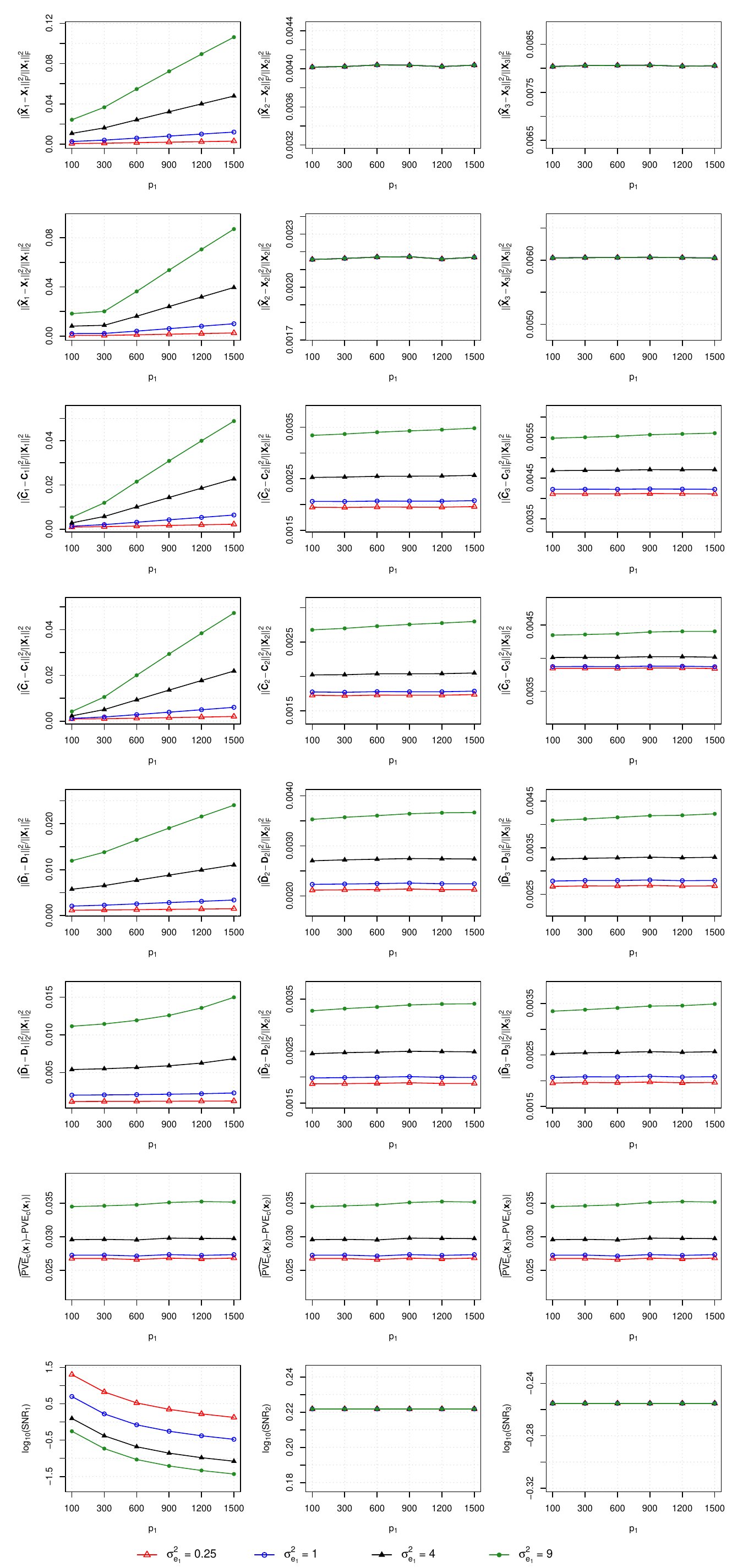}
		\caption{$\theta_z=30^\circ$}
		\label{noise_level}
	\end{subfigure}
	\caption{Average errors of D-GCCA estimates over 1000 replications for Setup 1.2 with $\theta_z\in\{10^\circ,20^\circ,30^\circ\}$.}
\end{sidewaysfigure}

\begin{sidewaysfigure}[p!]
	\begin{subfigure}[b]{0.3\textwidth}
			\centering
		\includegraphics[width=1\textwidth]{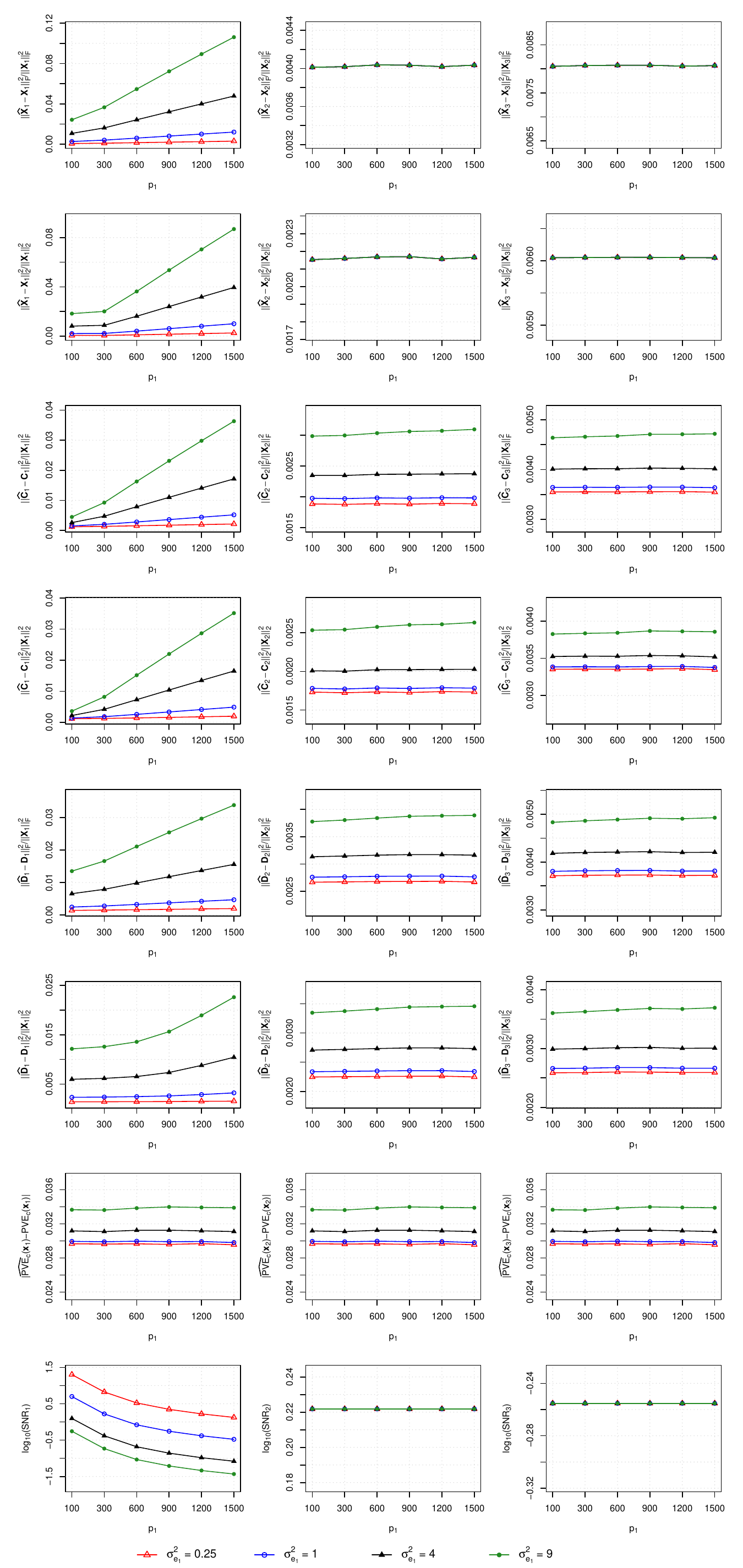}
		\caption{$\theta_z=40^\circ$}
		\label{noise_level}
	\end{subfigure}
	\hspace{0.05\textwidth}
	\begin{subfigure}[b]{0.3\textwidth}
			\centering
		\includegraphics[width=1\textwidth]{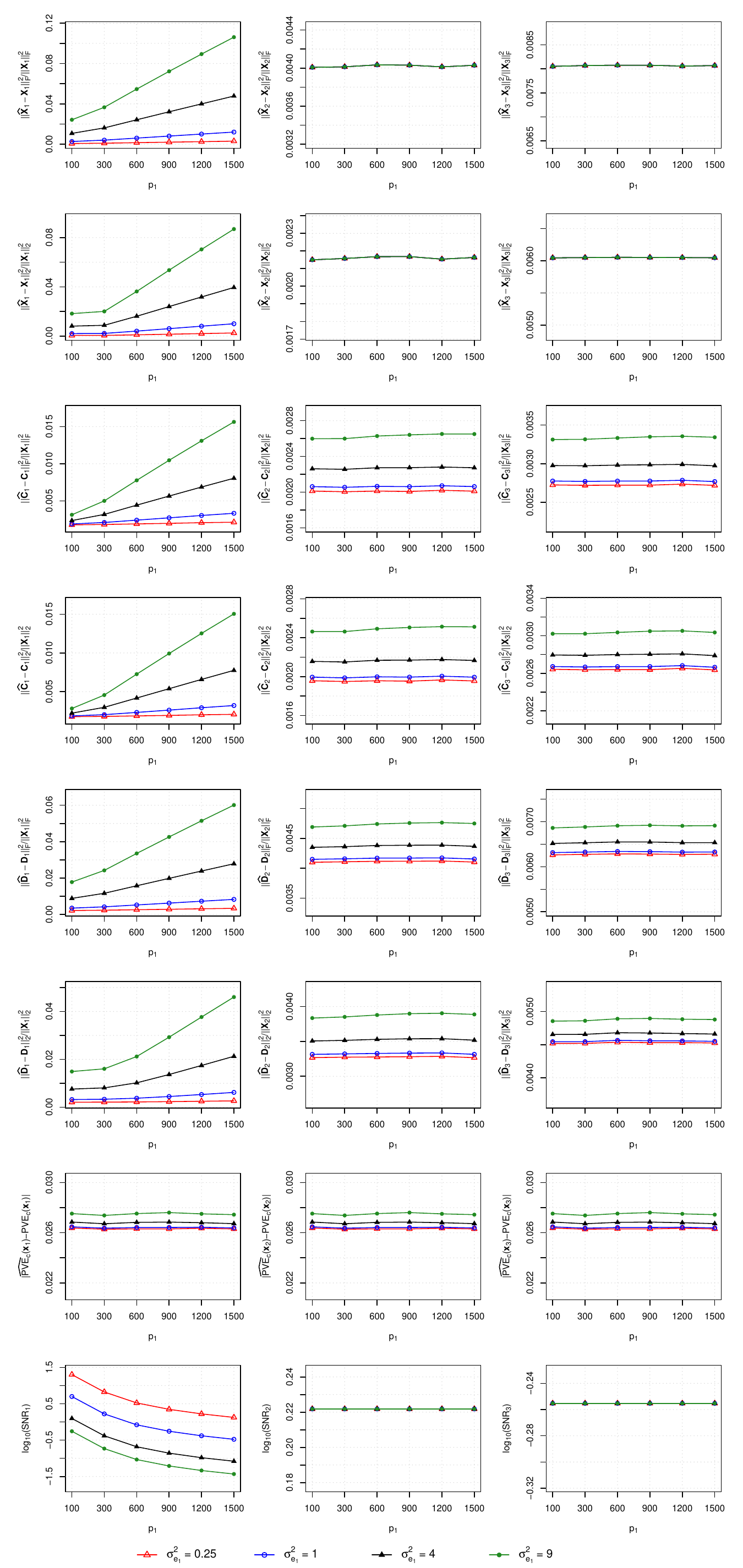}
		\caption{$\theta_z=60^\circ$}
		\label{noise_level}
	\end{subfigure}
		\hspace{0.05\textwidth}
	\begin{subfigure}[b]{0.3\textwidth}
			\centering
		\includegraphics[width=1\textwidth]{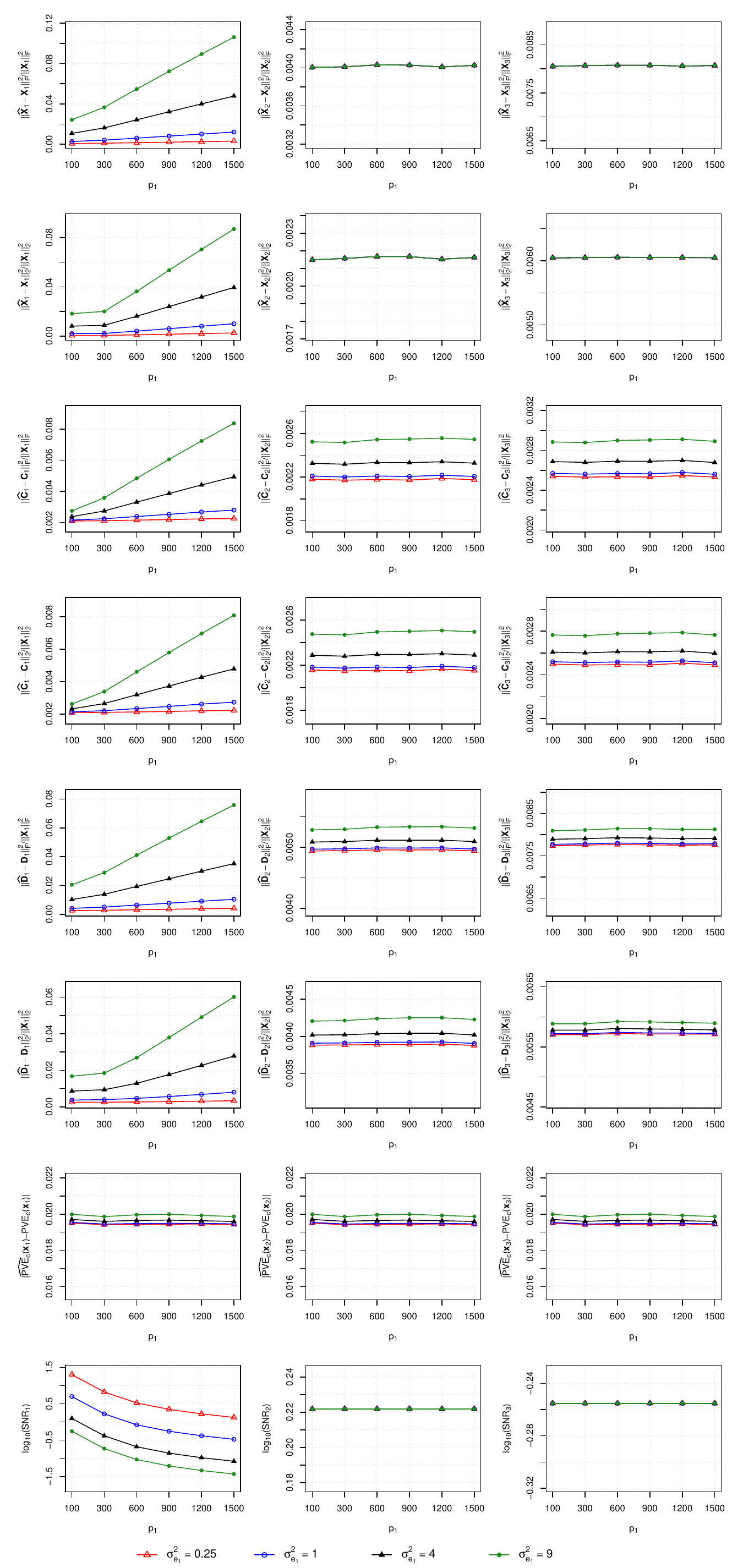}
		\caption{$\theta_z=70^\circ$}
		\label{noise_level}
	\end{subfigure}
	\caption{Average errors of D-GCCA estimates over 1000 replications for Setup 1.2 with $\theta_z\in\{40^\circ,60^\circ,70^\circ\}$.}
\end{sidewaysfigure}

\begin{figure}[p!]
	\begin{subfigure}[b]{1\textwidth}
		\centering
		\includegraphics[width=1\textwidth]{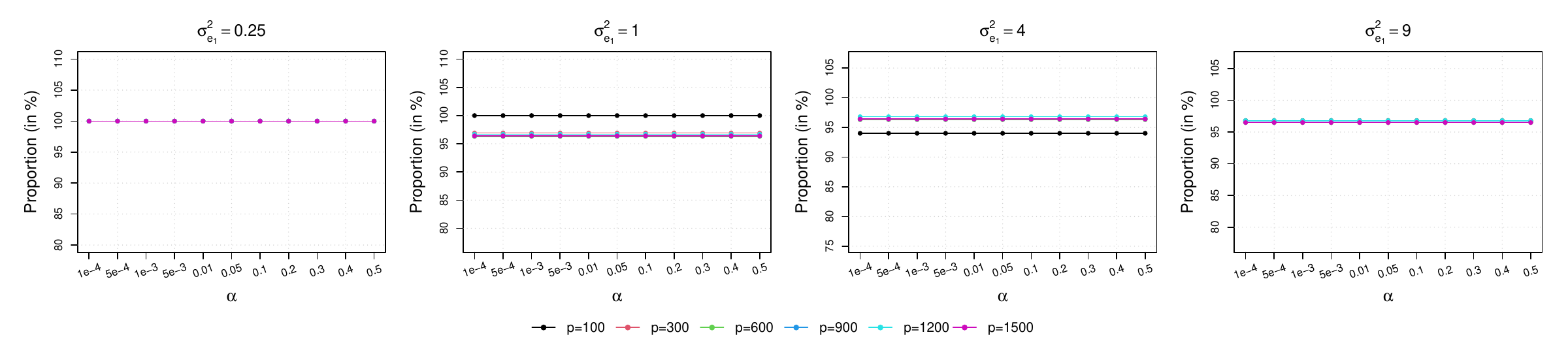}
		\caption{$\theta_z=10^\circ$}
		\label{noise_level}
	\end{subfigure}
\hspace{0.3cm}
	\begin{subfigure}[b]{1\textwidth}
	\centering
	\includegraphics[width=1\textwidth]{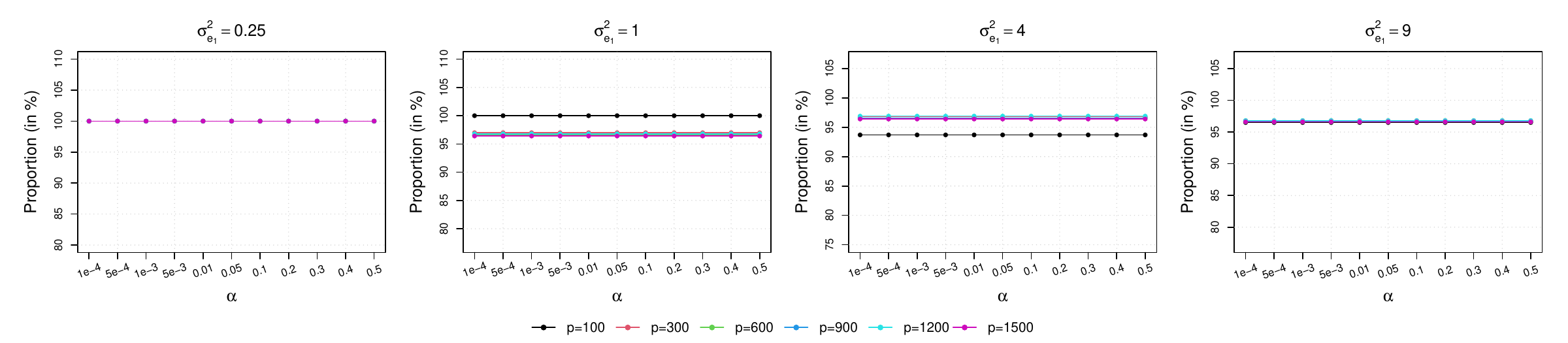}
	\caption{$\theta_z=20^\circ$}
	\label{noise_level}
\end{subfigure}
	\begin{subfigure}[b]{1\textwidth}
	\centering
	\includegraphics[width=1\textwidth]{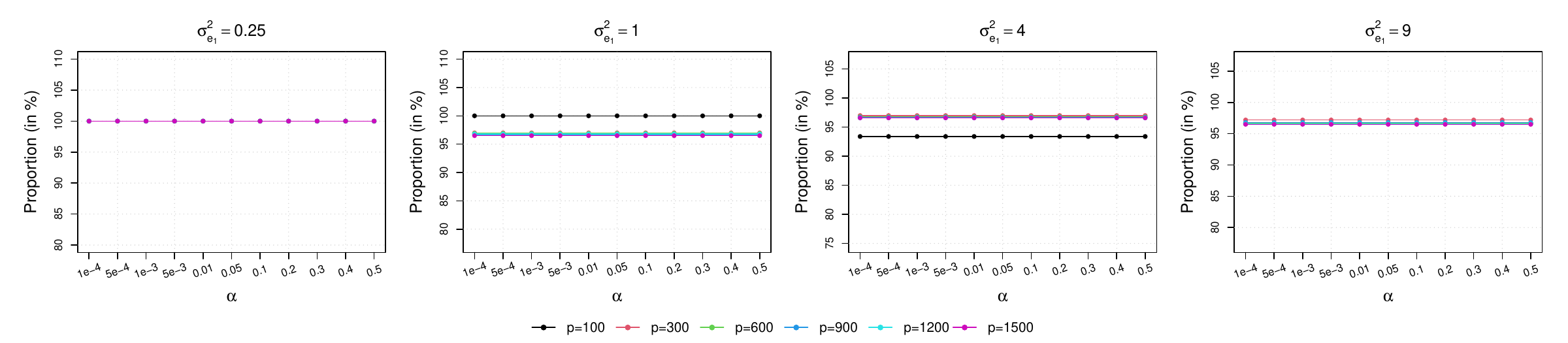}
	\caption{$\theta_z=30^\circ$}
	\label{noise_level}
\end{subfigure}
\hspace{0.3cm}
	\begin{subfigure}[b]{1\textwidth}
	\centering
	\includegraphics[width=1\textwidth]{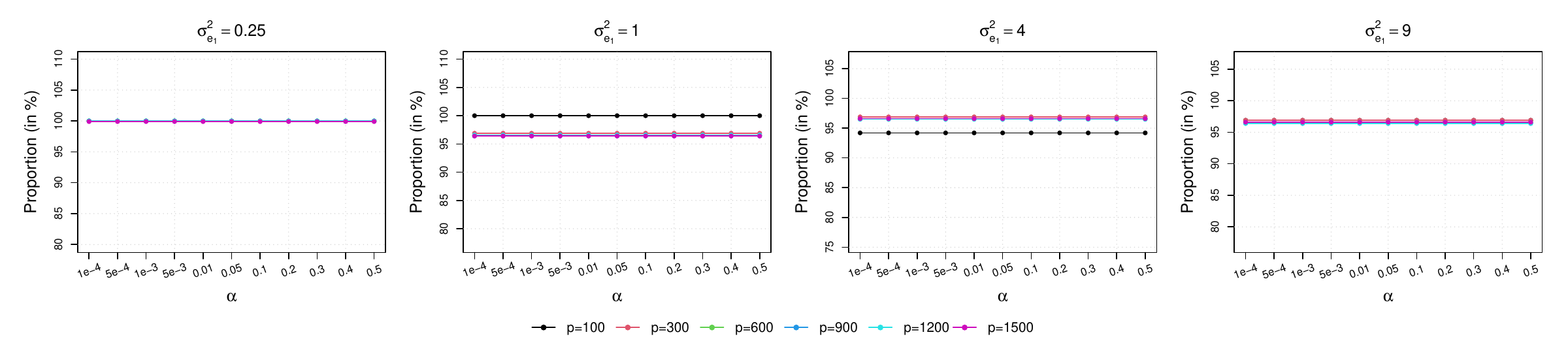}
	\caption{$\theta_z=40^\circ$}
	\label{noise_level}
\end{subfigure}
\hspace{0.3cm}
	\begin{subfigure}[b]{1\textwidth}
	\centering
	\includegraphics[width=1\textwidth]{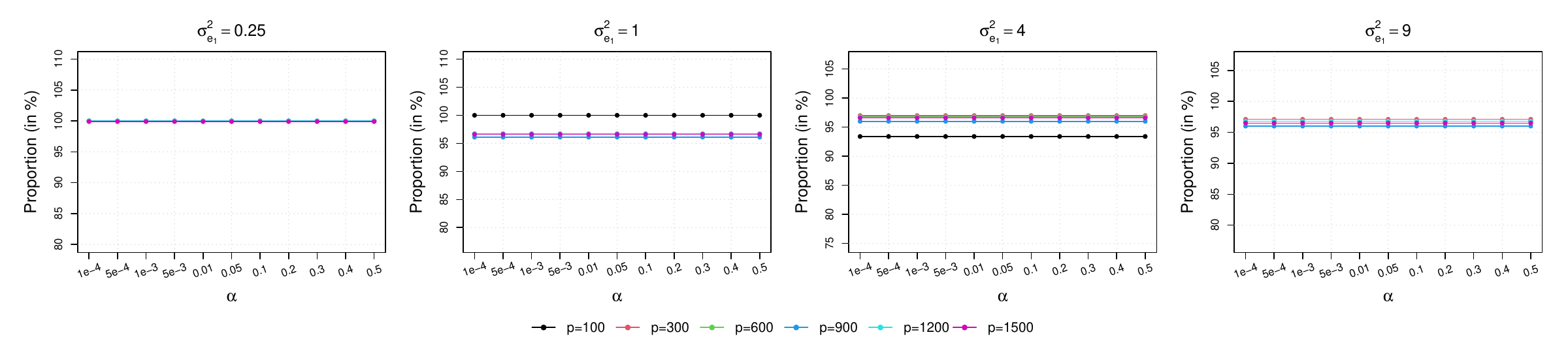}
	\caption{$\theta_z=60^\circ$}
	\label{noise_level}
\end{subfigure}
	\caption{The proportion of 1000 simulation replications of Setup~1.1 where all nuisance parameters of D-GCCA are correctly selected. The nuisance parameters 
		%$\big\{\{r_k,r_k^*\}_{k=1}^K, \mathcal{I}_0, \{\mathcal{I}_\Delta^{(\ell)},\sign(\alpha^{(\ell)})\}_{\ell\in \mathcal{I}_0}\big\}$ 
		are selected using the approach in Section~\ref{subsec: rank and set selection} with a significance level $\alpha$ uniformly applied to all tests.
	}
	\label{Fig1 for nuisance in appendix}
\end{figure}

\begin{figure}[p!]
	\begin{subfigure}[b]{1\textwidth}
		\centering
		\includegraphics[width=1\textwidth]{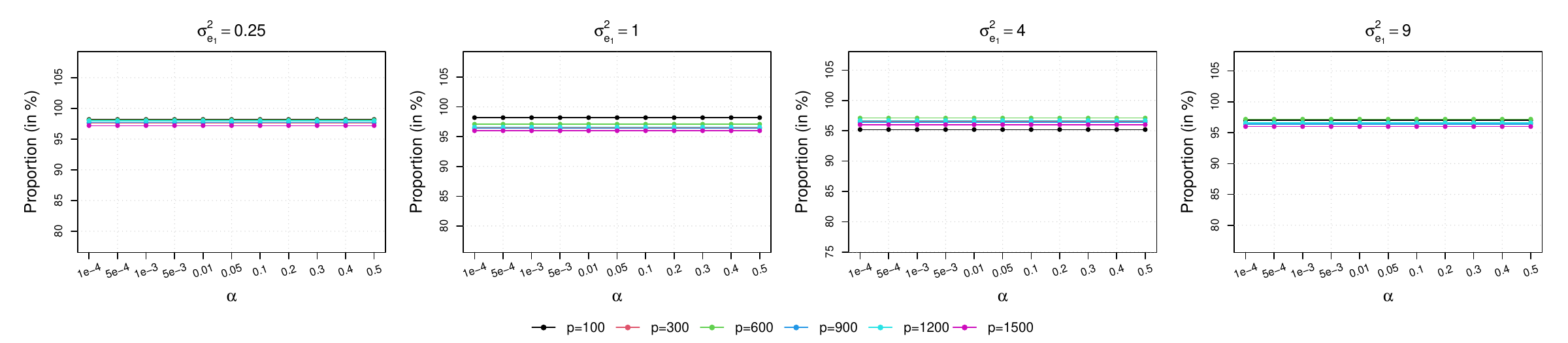}
		\caption{$\theta_z=10^\circ$}
		\label{noise_level}
	\end{subfigure}
	\hspace{0.3cm}
	\begin{subfigure}[b]{1\textwidth}
		\centering
		\includegraphics[width=1\textwidth]{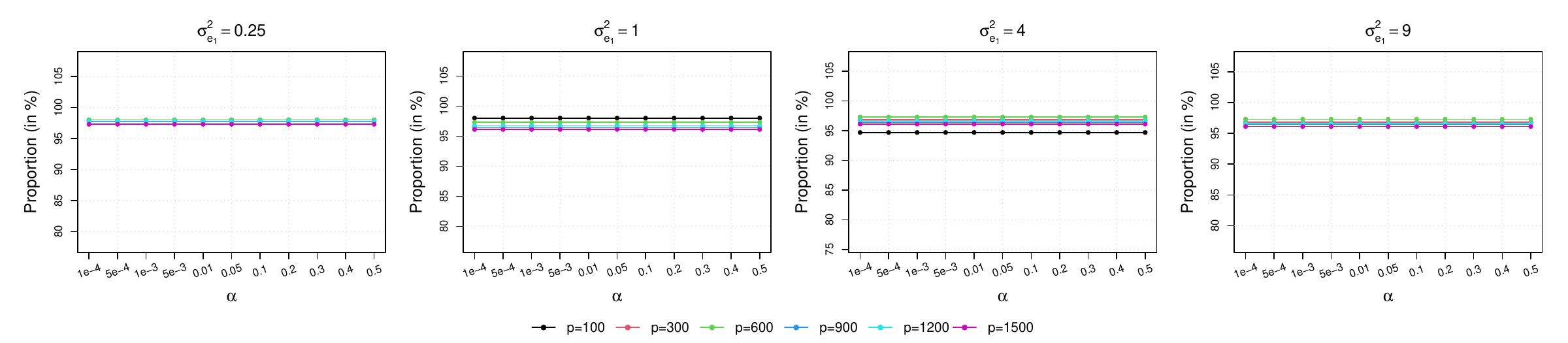}
		\caption{$\theta_z=20^\circ$}
		\label{noise_level}
	\end{subfigure}
	\begin{subfigure}[b]{1\textwidth}
		\centering
		\includegraphics[width=1\textwidth]{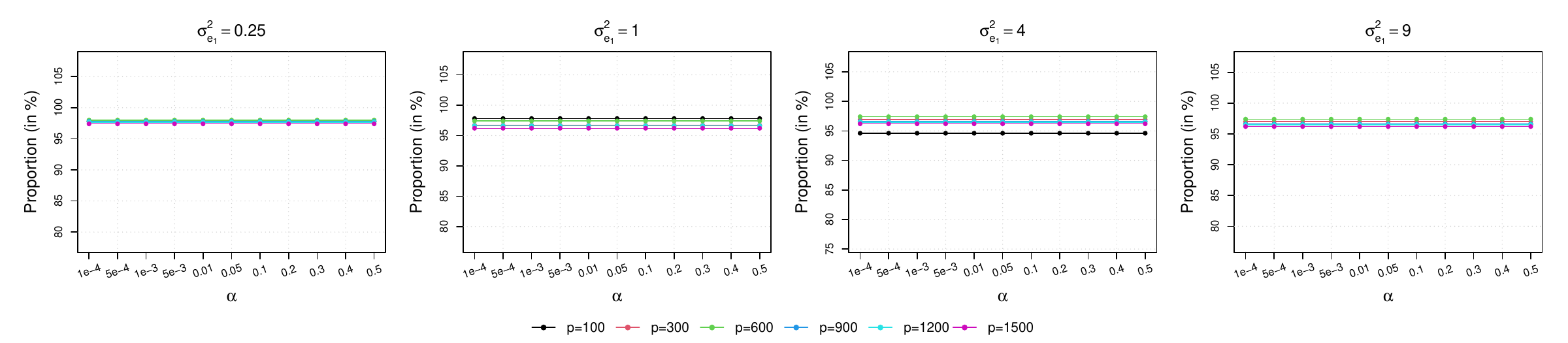}
		\caption{$\theta_z=30^\circ$}
		\label{noise_level}
	\end{subfigure}
	\hspace{0.3cm}
	\begin{subfigure}[b]{1\textwidth}
		\centering
		\includegraphics[width=1\textwidth]{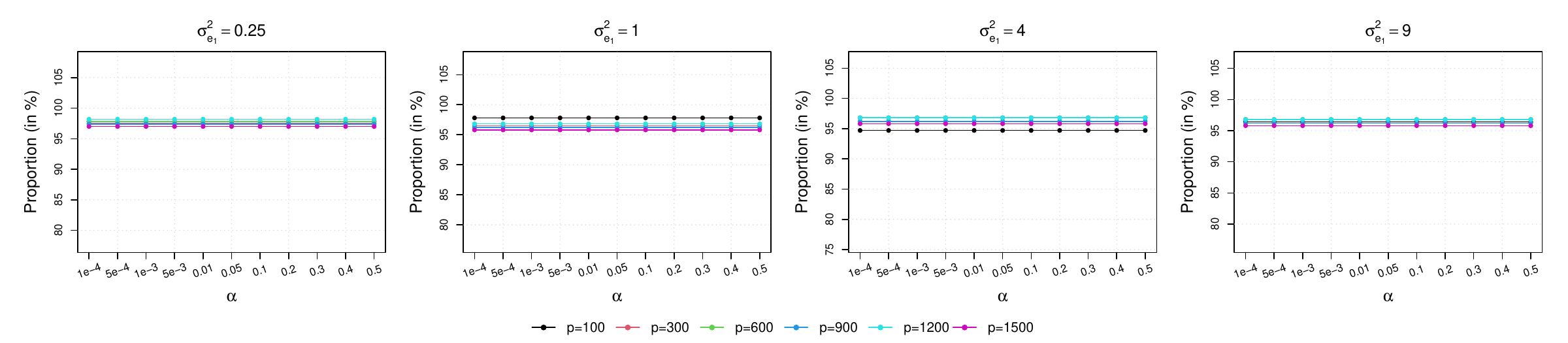}
		\caption{$\theta_z=40^\circ$}
		\label{noise_level}
	\end{subfigure}
	\hspace{0.3cm}
	\begin{subfigure}[b]{1\textwidth}
		\centering
		\includegraphics[width=1\textwidth]{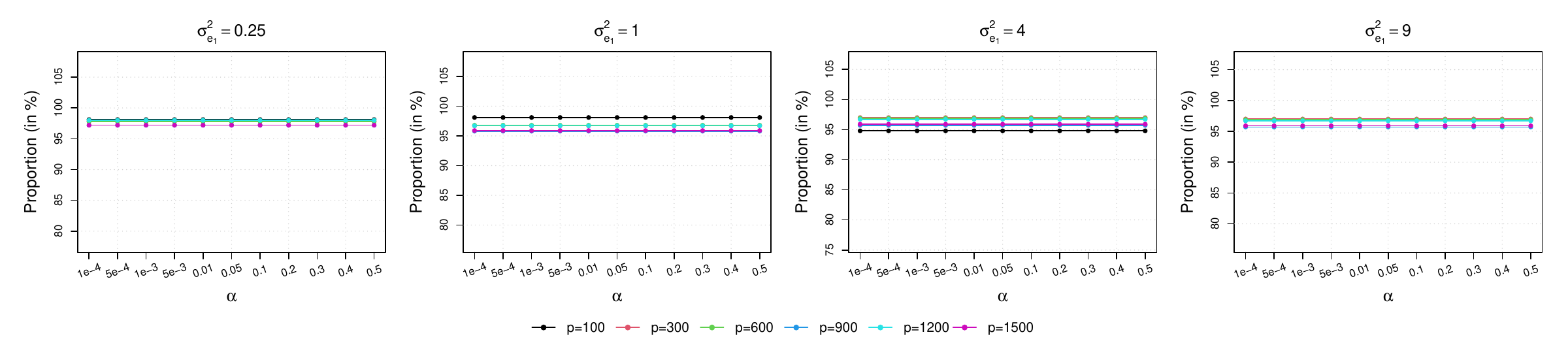}
		\caption{$\theta_z=60^\circ$}
		\label{noise_level}
	\end{subfigure}
	\caption{The proportion of 1000 simulation replications of Setup~1.2 where all nuisance parameters of D-GCCA are correctly selected. The nuisance parameters 
		%$\big\{\{r_k,r_k^*\}_{k=1}^K, \mathcal{I}_0, \{\mathcal{I}_\Delta^{(\ell)},\sign(\alpha^{(\ell)})\}_{\ell\in \mathcal{I}_0}\big\}$ 
		are selected using the approach in Section~\ref{subsec: rank and set selection} with a significance level $\alpha$ uniformly applied to all tests.
	}
\label{figA: Setup 1.2 prop nuiparameter plot}
\end{figure}

%\clearpage

%\vskip 0.2in
\newpage
\bibliography{main}

\end{document}